\documentclass[journal,onecolumn]{IEEEtran}
\usepackage{amsmath,amsfonts,amssymb}
\usepackage{algorithmic}
\usepackage{algorithm}
\usepackage{array,soul}
\usepackage[caption=false,font=normalsize,labelfont=sf,textfont=sf]{subfig}
\usepackage{textcomp}
\usepackage{stfloats,colortbl,xcolor,graphicx}
\usepackage{url,hyperref}
\usepackage{verbatim}
\usepackage{graphicx}
\usepackage{cite}
\usepackage{rsfso}
\usepackage{scalerel}
\usepackage{enumerate}
\newtheorem{theorem}{Theorem}

\newtheorem{lemma}{Lemma}

\newtheorem{definition}{Definition}
\definecolor{Gray}{gray}{0.93}

\definecolor{Gray}{gray}{0.85}

\usepackage{etoolbox}
\makeatletter
\patchcmd{\@makecaption}
  {\scshape}
  {}
  {}
  {}
\makeatother

\begin{document}
\title{Robust Non-adaptive Group Testing under Errors in Group Membership Specifications}

\author{Shuvayan Banerjee, Radhendushka Srivastava, James Saunderson and Ajit Rajwade ~\IEEEmembership{Senior Member,~IEEE}
\thanks{Shuvayan Banerjee is with the IITB-Monash Research Academy.}
\thanks{Radhendushka Srivastava is with the Department of Mathematics, IIT Bombay.}
\thanks{James Saunderson is with the Department of Electrical and Computer Systems Engineering, Monash University.}
\thanks{Ajit Rajwade is with the Department of Computer Science and Engineering, IIT Bombay.}}

%\markboth{Submitted to IEEE Trans. Inf. Th.}

\maketitle

\begin{abstract}
Given $p$ samples, each of which may or may not be defective, group testing aims to determine their defect status indirectly by performing tests on $n < p$ `groups' (also called `pools'), where a group is formed 
by mixing a subset of the $p$ samples. Under the assumption that the number of defective samples is very small compared to $p$, group testing algorithms have provided excellent recovery of the status of all $p$ samples with even a small number of groups. Most existing methods, however, assume that the group memberships are accurately specified. This assumption may not always be true in all applications, due to various resource constraints. For example, such errors could occur when a technician, preparing the groups in a laboratory, unknowingly mixes together an incorrect subset of samples as compared to what was specified. We develop a new group testing method, the Optimal Debiased Robust \textsc{Lasso} Test Method (\textsc{ODrlt}), that handles such group membership specification errors.  
The proposed \textsc{ODrlt} method is based on an approach to debias, i.e., reduce the inherent bias in, estimates produced by \textsc{Lasso}, a popular and effective sparse regression technique. We also provide theoretical upper bounds on the reconstruction error produced by our estimator. Our approach is then combined with two carefully designed hypothesis tests respectively for (\textit{i}) the identification of defective samples in the presence of errors in group membership specifications, and (\textit{ii}) the identification of groups with erroneous membership specifications. The \textsc{ODrlt} approach extends the literature on bias mitigation of statistical estimators such as the \textsc{Lasso}, to handle the important case when some of the measurements contain outliers, due to factors such as group membership specification errors. We present several numerical results which show that our approach outperforms several intuitive baselines and robust regression techniques for identification of defective samples as well as erroneously specified groups.
\end{abstract}


\section{Introduction}
Group testing is a well-studied area of data science, information theory and signal processing, dating back to the classical work of Dorfman in \cite{Dorfman1943}. Consider $p$ samples, one per subject, where each sample is either defective or non-defective. In the case of defective samples, additional quantitative information regarding the extent or severity of the defect in the sample may be available. Group testing typically replaces individual testing of these $p$ samples by testing of $n < p$ `groups' of samples, thereby saving on the number of tests.  Each group (also called a `pool') consists of a mixture of small, equal portions taken from a subset of the $p$ samples. Let the (perhaps noisy) test results on the $n$ groups be arranged in an $n$-dimensional vector $\boldsymbol{z}$. Let the true status of each of the $p$ samples be arranged in a $p$-dimensional vector $\boldsymbol{\beta^*}$. The aim of group testing is to infer $\boldsymbol{\beta^*}$ from $\boldsymbol{z}$ given accurate knowledge of the group memberships. We encode group memberships in an $n\times p$-dimensional binary matrix $\boldsymbol{B}$ (called the  `pooling matrix') where $B_{ij} = 1$ if the $j^{\text{th}}$ sample is a member of the $i^{\text{th}}$ group, and $B_{ij}=0$ otherwise. If the overall status of a group is the sum of the status values of each of the samples that participated in the group, we have:
\begin{equation}
\boldsymbol{z} = \boldsymbol{B\beta^*} + \boldsymbol{\tilde{\eta}}, 
\label{eq:forward_model_gt}
\end{equation}
where $\boldsymbol{\tilde{\eta}} \in \mathbb{R}^n$ is a noise vector.
In a large body of the literature on group testing (e.g., \cite{Atia2012,Chan2011,Dorfman1943}), $\boldsymbol{z}$ and $\boldsymbol{\beta^*}$ are modeled as binary vectors, leading to the forward model $\boldsymbol{z} = \mathfrak{N}(\boldsymbol{B\beta^*})$, 
where the matrix-vector `multiplication' $\boldsymbol{B\beta^*}$ involves binary OR, AND operations instead of sums and products, and $\mathfrak{N}$ is a noise operator that could at random flip some of the bits in $\boldsymbol{z}$. In this work, however, we consider $\boldsymbol{z}$ and $\boldsymbol{\beta^*}$ to be vectors in $\mathbb{R}^n$ and $\mathbb{R}^p$ respectively, as also done in \cite{Ghosh2021,Shental2020,Gilbert2008,bah2024compressed,Heiderzadeh2021,saeedi2022group}, and adopt the linear model~\eqref{eq:forward_model_gt}. This enables encoding of quantitative information in $\boldsymbol{z}$ and $\boldsymbol{\beta^*}$; moreover, $\boldsymbol{B\beta^*}$ now involves the usual matrix-vector multiplication.

In commonly considered situations in group testing, the number of non-zero (i.e., defective) samples $s \triangleq \|\boldsymbol{\beta^*}\|_0$ is much less than $p$, and $\beta^*_j = 0$ indicates that the $j^{\text{th}}$ sample is non-defective where $1 \leq j \leq p$. In such cases, group testing algorithms have shown excellent results for the recovery of $\boldsymbol{\beta^*}$ from $\boldsymbol{z}, \boldsymbol{B}$. These algorithms are surveyed in detail in \cite{Aldridge2021} and can be classified into two broad categories: adaptive and non-adaptive. Adaptive algorithms \cite{Dorfman1943,Heiderzadeh2021,Hwang1972} process the measurements (i.e., the results of pooled tests available in $\boldsymbol{z}$) in two or more stages of testing, where the output of each stage determines the choice of pools in the subsequent testing stage. Non-adaptive algorithms \cite{Shental2020,Ghosh2021,Gilbert2008,Bharadwaja2022}, on the other hand, process the measurements with only a single stage of testing. Non-adaptive algorithms are known to be more efficient in terms of time as well as the number of tests required, at the cost of somewhat higher recovery errors, as compared to adaptive algorithms \cite{Larremore2021, Ghosh2021}. In this work, we focus on non-adaptive algorithms. 
\newline
\newline
\noindent\textbf{Problem Motivation:} In the recent COVID-19 pandemic, RT-PCR (reverse transcription polymerase chain reaction) has been the primary method of testing a person for this disease. Due to widespread shortage of various resources for testing, group testing algorithms were widely employed in many countries \cite{PooledWiki2021}. Many of these approaches used Dorfman testing \cite{Dorfman1943} (an adaptive algorithm), but non-adaptive algorithms have also been recommended or used for this task \cite{Ghosh2021,Shental2020,Bharadwaja2022}. In this application, the vectors $\boldsymbol{\beta^*}$ and $\boldsymbol{z}$ refer to the real-valued \emph{viral loads} in the individual samples and the pools respectively, and $\boldsymbol{B}$ is again a binary pooling matrix. In a pandemic situation, there is heavy demand on testing labs. This leads to practical challenges for the technicians to implement pooling due to factors such as ({\em i}) a heavy workload, ({\em ii}) differences in pooling protocols across different labs, and ({\em iii}) the fact that pooling is inherently more complicated than individual sample testing \cite{Nytimes2020},\cite[`Results']{Fenichel2021}. Due to this, there is the possibility of a small number of inadvertent errors in creating the pools. This causes a difference between {a few entries of} the pre-specified matrix $\boldsymbol{B}$ and the actual matrix {$\boldsymbol{\hat{B}}$} used for pooling. {Note that $\boldsymbol{B}$ is known whereas $\boldsymbol{\hat{B}}$ is unknown in practice. The sparsity of the difference between $\boldsymbol{B}$ and $\boldsymbol{\hat{B}}$ is a reasonable assumption, if the technicians are competent. Hence only a small number of group membership specifications contain errors.} This issue of errors during pool creation is well documented in several independent sources such as \cite{Nytimes2020},\cite[`Results']{Fenichel2021}, \cite[Page 2]{Comess2022}, \cite{Rohde2020}, \cite[Sec. 3.1]{Zabeti2021}, \cite[`Discussion']{Christoff2021}, \cite[`Specific consideration related to SARS-CoV-2']{Grobe2021} and \cite[`Laboratory infrastructure']{Aldridge2022}. However the vast majority of group testing algorithms --- adaptive as well as non-adaptive --- do not account for these errors. To the best of our knowledge, this is the first piece of work on the problem of a mismatched pooling matrix (i.e., a pooling matrix that contains errors in group membership specifications) for non-adaptive group testing with real-valued $\boldsymbol{\beta^*}$ and (possibly) noisy $\boldsymbol{z}$. We emphasize that besides pooled RT-PCR testing, faulty specification of pooling matrices may also naturally occur in group testing in many other scenarios, for example when applied to verification of electronic circuits \cite{Kahng2006}. Another scenario is in epidemiology \cite{Cheragchi2011}, for identifying infected individuals who come in contact with agents who are sent to mix with individuals in the population. The health status of various individuals is inferred from the health status of the agents. However, sometimes an agent may remain uninfected even upon coming in contact with an infected individual, which can be interpreted as an error in the pooling matrix. 
\newline
\newline
\noindent\textbf{Related Work:} We now comment on two related pieces of work which deal with group testing with errors in pooling matrices via non-adaptive techniques. The work in \cite{Cheragchi2011} considers probabilistic and structured errors in the pooling matrix, where an entry $b_{ij}$ with a value of 1 could flip to 0 with a probability $\vartheta \in (0,1)$, but not vice versa, i.e., a genuinely zero-valued $b_{ij}$ never flips to 1. The work in \cite{Mazumdar2014} considers a small number of `pretenders' in the unknown binary vector $\boldsymbol{\beta^*}$, i.e., there exist elements in $\boldsymbol{\beta^*}$ which flip from 1 to 0 with probability $0.5$, but not vice versa. Both these techniques consider \emph{binary valued} vectors $\boldsymbol{z}$ and $\boldsymbol{\beta^*}$, unlike real-valued vectors as considered in this work. They also do not consider noise in $\boldsymbol{z}$ in addition to the errors in $\boldsymbol{B}$. Furthermore, we also present a method to \emph{identify} the errors in $\boldsymbol{B}$, unlike the techniques in \cite{Cheragchi2011,Mazumdar2014}. Due to these differences between our work and \cite{Cheragchi2011,Mazumdar2014}, a direct numerical comparison between our results and theirs will not be meaningful. 
\newline
\newline
\noindent\textbf{Sensing Matrix and Basis Matrix Perturbation in Compressed Sensing:} There exists a nice relationship between the group testing problem and the problem of signal reconstruction in compressed sensing (CS), as explored in \cite{Ghosh2021,Gilbert2008}. Likewise, there is literature in the CS community which deals with perturbations in sensing matrices \cite{Pandotra2019,Parker2011,Zhu2011,Ince2014,Aldroubi2012,Herman2010,Fosson2020,herman2010mixed}. However, these works either consider dense random perturbations (i.e., perturbations in every entry with a bound on the norm of the perturbation matrix) \cite{Zhu2011,Ince2014,Parker2011,Aldroubi2012,Herman2010,Fosson2020,herman2010mixed} or perturbations in specifications of Fourier frequencies \cite{Pandotra2019,Ianni2016}. These perturbation models are vastly different from the sparse set of errors in binary matrices as considered in this work. Furthermore, apart from \cite{Pandotra2019,Ianni2016}, these techniques just perform robust signal estimation, without any attempt to identify rows of the sensing matrix which contained those errors. 
In typical compressed sensing, the measurements follow the model $\boldsymbol{y} = \boldsymbol{B \beta^*} + \boldsymbol{\tilde{\eta}} = \boldsymbol{B \Psi \tilde{\theta}^*} + \boldsymbol{\tilde{\eta}}$, where the signal $\boldsymbol{\beta^*} \in \mathbb{R}^p$ is expressed as a sparse linear combination of the columns of the orthonormal or overcomplete basis matrix $\boldsymbol{\Psi} \in \mathbb{R}^{p \times p}$, where $\boldsymbol{\tilde{\theta}^*} \in \mathbb{R}^p$ is the coefficient vector, and $\boldsymbol{y}$, $\boldsymbol{\tilde{\eta}}$, $\boldsymbol{B}$ have the same meaning as defined earlier. There is a rich literature on compressed sensing where there are errors in the basis matrix $\boldsymbol{\Psi}$ and not in the sensing matrix $\boldsymbol{B}$. These are commonly used in DoA estimation in radar signal processing for localizing off-grid sources \cite{nehorai2014structured,Chi2011,Fannjiang2013,zhang2012robustly}. The concept of a perturbed representation matrix $\boldsymbol{\Psi}$ also has applications in line spectral estimation, as seen in papers such as \cite{Nichols2014}. For the work in our paper, we note that $\boldsymbol{\Psi}$ is the identity matrix. However, even more importantly, the basis mismatch problem is fundamentally different from the approach in this paper which deals with perturbations in the sensing matrix $\boldsymbol{B}$. This difference can be explained in the following manner: A perturbation in the $k$th column of $\boldsymbol{\Psi}$ ($k \in \{1,2,...,p\}$) will affect \emph{all} measurements which use the $k$th coefficient. On the other hand, a perturbation in the $l$th row ($l \in \{1,2,...,n\}$) of $\boldsymbol{B}$, does not affect the other measurements. There also exist papers such as \cite{Aldroubi2012} which consider perturbations in both $\boldsymbol{B}$ and $\boldsymbol{\Psi}$ simultaneously. 
\newline
\newline
\noindent\textbf{Overview of contributions:} In this paper, we present a robust approach for recovering $\boldsymbol{\beta^*} \in \mathbb{R}^p$ from noisy $\boldsymbol{z} \in \mathbb{R}^n$ when $n < p$, given a \emph{known} pre-specified pooling matrix $\boldsymbol{B}$, but where the measurements in $\boldsymbol{z}$  correspond to another \emph{unknown} pooling matrix $\boldsymbol{\hat{B}}$ which contains  errors in group membership specification, i.e., $\boldsymbol{z} = \boldsymbol{\hat{B}\beta^*} + \boldsymbol{\tilde{\eta}}$. 
The well known debiased \textsc{Lasso} estimator in statistics \cite{Javanmard2014,zhang2014confidence,vandegeer2014} is not directly applicable in the presence of errors in group membership specification. We develop a novel debiasing methodology for the Robust \textsc{Lasso} Estimator and establish its asymptotic distribution which leads to a statistical test to determine the non-zero elements of $\boldsymbol{\beta^*}$. We refer to this approach as Optimal Debiased Robust \textsc{Lasso} Test (\textsc{ODrlt}).
In this approach, we present a principled method to identify which measurements in $\boldsymbol{z}$ correspond to rows with errors in $\boldsymbol{B}$, using a statistical hypothesis test. We also present an algorithm for direct estimation of $\boldsymbol{\beta^*}$ and a hypothesis test for identification of the defective samples in $\boldsymbol{\beta^*}$, given errors in $\boldsymbol{B}$. We establish the desirable properties of these statistical tests such as consistency and asymptotic unbiasedness. Though our approach is motivated by pooling errors during preparation of pools of COVID-19 samples, it is not restricted to any one particular modality, and is broadly applicable to any group-testing problem where the pool membership specifications contain a small number of errors.
\newline
\newline
\noindent\textbf{Notations:} Throughout this paper, $\boldsymbol{I_n}$ denotes the identity matrix of size $n \times n$. We use the notation $[n] \triangleq \{1,2,\cdots,n\}$ for $n \in \mathbb{Z}_{+}$. Given a matrix $\boldsymbol{A}$, its $i^\text{th}$ row is denoted as $\boldsymbol{a_{i.}}$, its $j^\text{th}$ column is denoted as $\boldsymbol{a_{.j}}$ and the $(i,j)^{\text{th}}$ element is denoted by $a_{ij}$. The $i^{\text{th}}$ column of the identity matrix will be denoted as $\boldsymbol{e_i}$. For any vector $\boldsymbol{z} \in \mathbb{R}^n$ and index set $S \subseteq [n]$, we define $\boldsymbol{z}_S \in \mathbb{R}^n$ such that $\forall i \in S, (z_S)_i = z_i$ and $\forall i \notin S, (z_S)_i = 0$. $S^c$ denotes the complement of set $S$. We define the entrywise $l_\infty$ norm of a matrix $\boldsymbol{A}$ as $|\boldsymbol{A}|_\infty \triangleq \underset{i,j}{\max} |a_{ij}|$. 
Consider two real-valued random sequences $x_n$ and $r_n$. Then, we say that $x_n$ is $o_P(r_n)$ if $x_n/r_n \rightarrow 0$ in probability, i.e., $\lim_{n \rightarrow \infty} P(|x_n/r_n| \geq \epsilon) = 0$ for any $\epsilon > 0$. Also, we say that $x_n$ is $O_P(r_n)$ if $x_n/r_n$ is bounded in probability, i.e., for any $\epsilon > 0$ there exist $m, n_0>0$, such that $P(|x_n/r_n| < m) \geq 1- \epsilon$ for all $n > n_0$.
\newline
\newline
\noindent\textbf{Organization of the paper:} The noise model involving measurement noise and errors in the pooling matrix is presented in Sec.~\ref{sec:pform}. Our core technique, \textsc{ODrlt}, is presented in Sec.~\ref{sec:DLT}, with essential background literature summarized in Sec.~\ref{subsec:debiased_lasso}, and our key innovations presented in Sec.~\ref{sec:deb_lasso_test}. Detailed experimental results are given in Sec.~\ref{sec:experiments}. We conclude in Sec.~\ref{sec:conclusion}. Proofs of all theorems and lemmas are provided in Appendices \ref{Sec:App_A1},\ref{sec:App_A2} and \ref{sec:App_A3}. 
\section{Problem formulation}
\label{sec:pform}
\subsection{Basic Noise Model} 
We now formally describe the model setup used in this paper.  
Suppose that the defect status vector $\boldsymbol{\beta^*}\in \mathbb{R}^p$ and that the elements of the $n\times p$ pooling matrix $\boldsymbol{B}$ are independently drawn from $\text{Bernoulli}(0.5)$ in \eqref{eq:forward_model_gt}.
The random Bernoulli model (Bernoulli($\theta$) for $\theta \in (0,1)$) has been widely used in group testing \cite{scarlett2016limits,scarlett2017phase,truong2020all} as well as compressed sensing \cite{duarte2008single}. In particular, the $\theta = 0.5$ case has been deemed to be optimal for compressed sensing \cite{hunt2018data}. Hence, we first focus on the $\theta = 0.5$ case and then generalize to any sensing matrix with independent and identically distributed (i.i.d.) bounded entries in Sec.~\ref{Sec:Bounded_matrix}.   
Additionally, let $\boldsymbol{\beta^*}$ be sparse (as commonly assumed in group testing) with at most $s \ll p$ non-zero distinct elements. Assume that the elements of the noise vector $\boldsymbol{\tilde{\eta}} \in \mathbb{R}^n$ in \eqref{eq:forward_model_gt} are i.i.d. Gaussian random variables with mean $0$ and variance $\tilde{\sigma}^2$. 
This additive noise model is a natural choice to represent errors in measurements and has been considered before in group testing and the closely related `pooled data' problem, for example, in\cite{hahn2022distributed}, \cite{scarlett2017phase} (equation 14), \cite{tan2024approximate} (equation 1.3), \cite{iwen2009group} (equation 1), \cite{yi2023error} (equations 1 and 3, and the beginning of section IV). Note that, throughout this work, we assume $\tilde{\sigma}^2$ to be known. 
The \textsc{Lasso} estimator $\boldsymbol{\hat{\beta}}$, used to estimate $\boldsymbol{\beta^*}$, is defined as
\begin{equation}
\boldsymbol{\hat{\beta}} = \arg \min_{\boldsymbol{\beta}} \frac{1}{2n}\|\boldsymbol{z}-\boldsymbol{B\beta}\|^2_2 + \lambda \|\boldsymbol{\beta}\|_1. 
\label{eq:lasso_z}
\end{equation}
Given a sufficient number of measurements, the \textsc{Lasso} is known to be consistent for sparse $\boldsymbol{\beta^*}$ \cite[Chapter 11]{THW2015} if the penalty parameter $\lambda > 0$ is chosen appropriately and if $\boldsymbol{B}$ satisfies the Restricted Eigenvalue Condition (REC)\footnote{Restricted Eigenvalue Condition: For some constant $\psi \geq 1$ and $S\subseteq [p]$, let $C_\psi(S) \triangleq \{\boldsymbol{\Delta}\in \mathbb{R}^n:\|\boldsymbol{\Delta}_{S^c}\|_1 \leq \psi \|\boldsymbol{\Delta}_S\|_1\}$. We say that a $n\times p$ matrix $\boldsymbol{B}$ satisfies the REC with respect to $C_\psi(S)$ if there exists a constant $\gamma>0$ such that $\frac{1}{n}\|\boldsymbol{B\Delta}\|_2^2 \geq \gamma\|\boldsymbol{\Delta}\|_2^2$ for all $\boldsymbol{\Delta}\in C_{\psi}(S)$. Here $\gamma$ is the restricted eigenvalue (RE) constant. The vector $\boldsymbol{\Delta}$ is intended to be the error vector between the true signal vector and its estimate via (say) the \textsc{Lasso}.}. Certain deterministic binary pooling matrices can also be used as in \cite{Ghosh2021, Shental2020} for a consistent estimator of $\boldsymbol{\beta^*}$. However, we focus on the chosen random pooling matrix in this paper.
 
It is more amenable for analysis via the REC, if the elements of the pooling matrix have mean $0$. Since the elements of $\boldsymbol{B}$ are drawn independently from Bernoulli$(0.5)$, it does not obey the mean-zero property.
Hence, we transform the random binary matrix $\boldsymbol{B}$ to a random Rademacher matrix $\boldsymbol{A}$ by means of a simple one-one transformation similar to that adopted in \cite{raginsky2010compressed} for Poisson compressive measurements. We also correspondingly transform the measurements in $\boldsymbol{z}$ to equivalent measurements $\boldsymbol{y}$ associated with Rademacher matrix $\boldsymbol{A}$. This transformation can be accomplished by considering $2n$ measurements instead of $n$, where $\forall i \in [n], \boldsymbol{b}_{n+i,.} := 1-\boldsymbol{b}_{i,.}$. That is, the $(n+i)$th row of $\boldsymbol{B}$ is obtained by toggling the corresponding entries of the $i$th row of $\boldsymbol{B}$. In such a case, we have $\forall i \in [n], y_i = z_i - z_{n+i}$. The elements of $\boldsymbol{y}$ correspond to measurements using a random Rademacher matrix $\boldsymbol{A}$ where $\forall i \in [n], \boldsymbol{a}_{i,.} = \boldsymbol{b}_{i,.}-\boldsymbol{b}_{n+i,.}$.
\footnote{This transformation also can be accomplished by (\textit{i}) acquiring an additional measurement $\bar{z}_s$ with all ones in the corresponding row of the pooling matrix, i.e., $\bar{z}_s \approx \boldsymbol{1_p^T \beta^*}$, and (\textit{ii}) subtracting $\bar{z}_s$ from twice of each $z_i$ to yield $y_i = 2z_i - \bar{z}_s$. Note that $\bar{z}_s$ can be obtained by averaging over some $\tilde{K}$ measurements, each taken with a row consisting of all ones in the pooling matrix.}

The expression for each measurement in $\boldsymbol{y}$ is now given by:
\begin{eqnarray}
\forall i \in [n], y_i=\boldsymbol{a_{i.}} \boldsymbol{\beta^{*}} + \eta_i \implies \boldsymbol{y} = \boldsymbol{A\beta^*} + \boldsymbol{\eta},
\label{eq:forward_model_bitflip}
\end{eqnarray}
where $\eta_i \triangleq \tilde{\eta}_i-\tilde{\eta}_{n+i} \sim \mathcal{N}(0,\sigma^2), \ \sigma^2 \triangleq 2\tilde{\sigma}^2$. 
We will henceforth consider $\boldsymbol{y}, \boldsymbol{A}$ for the \textsc{Lasso} estimates in the following manner:
The \textsc{Lasso} estimator $\boldsymbol{\hat{\beta}}$, used to estimate $\boldsymbol{\beta^*}$, is now defined as
\begin{equation}
\boldsymbol{\hat{\beta}} = \arg \min_{\boldsymbol{\beta}} \frac{1}{2n}\|\boldsymbol{y}-\boldsymbol{A\beta}\|^2_2 + \lambda \|\boldsymbol{\beta}\|_1. 
\label{eq:lasso}
\end{equation}
In this paper, we begin with random Rademacher distribution for $\boldsymbol{A}$, corresponding to the $\text{Bernoulli}(0.5)$ distribution for $\boldsymbol{B}$, for the theoretical development. We then subsequently extend our analysis for a general $\boldsymbol{A}$ (corresponding to different models for $\boldsymbol{B}$) with entries drawn independently from any zero-mean, unit-variance distribution defined over a bounded domain. This is demonstrated in Sec.~\ref{Sec:Bounded_matrix}.

\subsection{Model Mismatch Errors} \label{sec:bitflips}
Consider the measurement model defined in \eqref{eq:forward_model_bitflip}. We now examine the effect of mis-specification of samples in a pool. That is, we consider the case where due to errors in mixing of the samples, the pools are generated using an unknown matrix $\boldsymbol{\hat{A}}$ (say) instead of the pre-specified matrix $\boldsymbol{A}$. Note that $\boldsymbol{A}$ and $\boldsymbol{\hat{A}}$ are respectively obtained from $\boldsymbol{B}$ and $\boldsymbol{\hat{B}}$. The elements of matrix $\boldsymbol{\hat{A}}$ and $\boldsymbol{A}$ are equal everywhere except for the misspecified samples in each pool. 
We refer to these errors in group membership specifications as `\textbf{bit-flips}'. For example, suppose that the $i^{\text{th}}$ pool is specified to consist of samples $j_1,j_2,j_3 \in [p]$. But due to errors during pool creation, the $i^{\text{th}}$ pool is generated using samples $j_1,j_2,j_5$. In this specific instance, $a_{i,j_3} \ne \hat{a}_{i,j_3} $ and $a_{i,j_5} \ne \hat{a}_{i,j_5} $.

Note that $\boldsymbol{A}$ is known whereas $\boldsymbol{\hat{A}}$ is unknown. Moreover, the locations of the bit-flips are unknown. Hence they induce signal-dependent and possibly large `\textbf{model mismatch errors}' 
$\delta_{i}^* \triangleq (\boldsymbol{\hat{a}_{i.}}-\boldsymbol{a_{i.}})\boldsymbol{\beta^{*}}$ in the $i^{\text{th}}$ measurement. In the presence of bit-flips, the model in \eqref{eq:forward_model_bitflip} can be expressed as:  
\begin{eqnarray}
    y_i = \boldsymbol{a_{i.} \beta^{*}} +   \delta_i^*+{\eta}_i, \ \mbox{for} \ i\in [n], \implies
\boldsymbol{y} = \boldsymbol{A \beta^{*}} + \boldsymbol{\delta^*} + \boldsymbol{{\eta}} = (\boldsymbol{A}|\boldsymbol{I_n})\begin{pmatrix}
    \boldsymbol{\beta^*} \\
    \boldsymbol{\delta^*}
\end{pmatrix} + \boldsymbol{\eta}. 
\label{eq:fm_delta} 
\end{eqnarray}
We assume $\boldsymbol{\delta^*}$, which we call the `model mismatch error' (MME) vector in $\mathbb{R}^n$, to be sparse, and  $r \triangleq\|\boldsymbol{\delta^*}\|_0 \ll n$.  The sparsity assumption on $\boldsymbol{\delta^*}$ is reasonable in many applications (e.g., given a competent technician performing pooling). 

Suppose for a fixed  $i\in[n]$, $\boldsymbol{\hat{a}_{i.}}$ contains a bit-flip at index $j$. If  ${\beta}_j^*$ is $0$ then $\delta^*_i$ would remain 0 despite the presence of a bit-flip in $\boldsymbol{\hat{a}_{i.}}$. 
Furthermore, such a bit-flip has no effect on the  measurements and is {\it not identifiable} from the measurements. However, if $\beta_j^*$ is non-zero then $\delta_i^*$ is also non-zero. Such a bit-flip adversely affects the measurement and we henceforth refer to it as an \emph{effective bit-flip}. 
Effective bit-flips lead to non-zero elements in the MME vector $\boldsymbol{\delta^*}$. We refer to the non-zero elements of $\boldsymbol{\delta^*}$ as \emph{effective} MMEs. 
Without loss of generality, we consider the identification of \emph{effective} MMEs in this paper.
\section{Debiasing the Robust Lasso} \label{sec:DLT}
We now present our proposed approach, named the `Optimal Debiased Robust Lasso Test Method' (\textsc{ODrlt}), for recovering the signal $\boldsymbol{\beta^*}$ given measurements $\boldsymbol{y}$ obtained from the erroneous, unknown matrix $\boldsymbol{\hat{A}}$ which is different from the pre-specified, known sensing matrix $\boldsymbol{A}$.  The main objectives of this work are: 
\begin{enumerate}[\text{Aim} (i):]
    \item Estimation of $\boldsymbol{\beta^*}$ under model mismatch and development of a statistical test to determine whether or not the $j^{\text{th}}$ sample ($j \in [p]$) is defective/diseased, i.e., whether or not $\beta^*_j$ is non-zero.
    \item Development of a statistical test to determine whether or not the $i^\text{th}$ measurement ($i\in [n]$) contains an effective MME i.e., whether or not $\delta^*_i$ is non-zero.
\end{enumerate}
A measurement containing an effective MME will appear like an outlier in comparison to other measurements due to the non-zero values in $\boldsymbol{\delta^*}$. Therefore identification of measurements containing effective MMEs is equivalent to determining the non-zero entries of $\boldsymbol{\delta^*}$. This idea is inspired by the concept of `Studentised residuals' which is widely used in the statistics literature to identify outliers in \emph{full-rank} regression models \cite{Montgomery}. Since our model operates in a compressive regime where $n < p$, the distributional property of studentized residuals may not hold. Therefore, we develop our \textsc{Drlt} method which is tailored for the compressive regime.  

Our basic estimator for $\boldsymbol{\beta^*}$ and $\boldsymbol{\delta^*}$ from $\boldsymbol{y}$ and $\boldsymbol{A}$ is given as
\begin{eqnarray}
\begin{pmatrix}
    \boldsymbol{\hat{\beta}_{\lambda_1}} \\
    \boldsymbol{\hat{\delta}_{\lambda_2}}
\end{pmatrix}  &=& \arg\min_{\boldsymbol{\beta},\boldsymbol{\delta}} {\frac{1}{2n}}\left\|\boldsymbol{y}-\boldsymbol{A\beta}-\boldsymbol{\delta}\right\|^2_2 + \lambda_1 \|\boldsymbol{\beta}\|_1 + \lambda_2 \left\|\boldsymbol{\delta}\right\|_1, \label{eq:lasso_delta}
\end{eqnarray}
where $\lambda_1, \lambda_2$ are appropriately chosen regularization parameters. This estimator is a robust version of the \textsc{Lasso} regression \cite{Nguyen2013}. The robust \textsc{Lasso}, just like the \textsc{Lasso}, will incur a bias due to the $\ell_1$ penalty terms. 
 
The work in \cite{Javanmard2014,zhang2014confidence,vandegeer2014} provides a method to mitigate the bias in the \textsc{Lasso} estimate and produces a `debiased' signal estimate whose distribution turns out to be approximately Gaussian with specific observable parameters in the compressive regime (for details, see Sec.~\ref{subsec:debiased_lasso} below). However, their approach does not take into account errors in sensing matrix specification. We non-trivially adapt the technique of \textsc{Lasso} `debiasing' to our specific application which considers bit-flips in the pooling matrix, and we also develop novel procedures to realize Aims~(\textit{i}) and (\textit{ii}) mentioned above.

We now first review important concepts which are used to develop our method for the specified aims. We subsequently develop our method in the rest of this section. However, before that, we present error bounds on the estimates $\boldsymbol{\hat{\beta}_{\lambda_1}}$ and $\boldsymbol{\hat{\delta}_{\lambda_2}}$ from \eqref{eq:lasso_delta}, which are non-trivial extensions of results in \cite{Nguyen2013}. These bounds will be essential in developing hypothesis tests to achieve Aims (i) and (ii).

\subsection{Bounds on the Robust \textsc{Lasso} Estimate}
\label{subsec:bounds_rlasso_estimate}
When the elements of sensing matrix $\boldsymbol{A}$ are i.i.d. Gaussian random variables, upper bounds on $\|\boldsymbol{\beta^*}-\boldsymbol{\hat{\beta}_{\lambda_1}}\|_2$ and $\|\boldsymbol{\delta^*}-\boldsymbol{\hat{\delta}_{\lambda_2}}\|_2$  have been presented in \cite{Nguyen2013}. In our case, $\boldsymbol{A}$ is i.i.d. Rademacher, and hence some modifications to the results from \cite{Nguyen2013} are required.
We now state a theorem for the upper bound on the reconstruction error of both $\boldsymbol{\hat{\beta}_{\lambda_1}}$ and $\boldsymbol{\hat{\delta}_{\lambda_2}}$ for a random Rademacher pooling matrix $\boldsymbol{A}$. We further use the so called `cone constraint' to derive separate bounds on the estimates of both $\boldsymbol{\beta^*}$ and $\boldsymbol{\delta^*}$. These bounds will be very useful in deriving theoretical results for debiasing. 
\begin{theorem}[Robust \textsc{Lasso} Bounds]
\label{th:upper_bound_robustLasso}
    Let  $\boldsymbol{\hat{\beta}}_{\boldsymbol{\lambda_1}},\boldsymbol{\hat{\delta}_{\lambda_2}}$ be as in \eqref{eq:lasso_delta} and set $\lambda_1 \triangleq \frac{4\sigma\sqrt{\log p}}{\sqrt{n}}, \lambda_2 \triangleq \frac{4\sigma\sqrt{\log n}}{{n}}$. Let $n < p$, $\mathcal{S} \triangleq \{j: \beta^*_j \ne 0\}$, $\mathcal{R} \triangleq \{i: \delta^*_i \ne 0\}$, $s \triangleq |\mathcal{S}|$ and $r \triangleq |\mathcal{R}|$.
    If $|\boldsymbol{A}|_\infty \leq 1$ and $\boldsymbol{A}$ satisfies the Extended Restricted Eigenvalue Condition (EREC) from Definition~\ref{def:EREC} with $\kappa>0$ and with respect to the cone $\mathcal{C}(\mathcal{S}, \mathcal{R},(\sqrt{n}\lambda_2)/\lambda_1)$, then we have the following: 
    \begin{enumerate}[(1)]
        \item Error bound on $\boldsymbol{\hat{\beta}_{\lambda_1}}$:
        \begin{equation}
            \label{eq:beta_lasso_bound}
P\left(\left\|\boldsymbol{\hat{\beta}_{\lambda_1}}-\boldsymbol{\beta^*}\right\|_1 \leq 48\kappa^{-2} (s+r)\sigma\sqrt{\frac{{\log(p)}}{{n}}}\right) \geq 1-\left(\frac{1}{p}+\frac{1}{n}\right).
        \end{equation}
        \item Error bound on $\boldsymbol{\hat{\delta}_{\lambda_2}}$: Additionally if $n \log n \geq (48 \kappa^{-2})^2 (s+r)^2 \log p$, 
        \begin{equation}
            \label{eq:delta_lasso_bound}
P\Biggl(\left\|\boldsymbol{\hat{\delta}_{\lambda_2}}-\boldsymbol{\delta^*}\right\|_1 \leq \frac{24\sigma r \sqrt{\log n}}{n}\Biggr) \geq 1-\left(\frac{1}{p}+\frac{2}{n}\right).
        \end{equation}
    \end{enumerate}
\hfill{$\blacksquare$}
\end{theorem}
In Lemma.~\ref{le:Ext_RE} of Appendix~\ref{Sec:App_A1}, we show that the chosen random Rademacher sensing matrix $\boldsymbol{A}$ satisfies the EREC with $\kappa=1/16$ if $\lambda_1$ and $\lambda_2$ are chosen as in Theorem~\ref{th:upper_bound_robustLasso}. Furthermore,  $|\boldsymbol{A}|_{\infty}=1$. Therefore, the sufficient conditions for Theorem \ref{th:upper_bound_robustLasso} are satisfied with high probability for a random Rademacher sensing matrix.
\newline
\newline
\noindent \textbf{Remarks on Theorem~\ref{th:upper_bound_robustLasso}:} 
\begin{enumerate}
\item From Result (1), we see that $\left\|\boldsymbol{\hat{\beta}_{\lambda_1}}-\boldsymbol{\beta^*}\right\|_1$ = $O_P\left((s+r)\sqrt{\frac{\log p}{n}}\right)$.
\item From Result (2), we see that $\left\|\boldsymbol{\hat{\delta}_{\lambda_2}}-\boldsymbol{\delta^*}\right\|_1$ = $O_P\left(\frac{r\sqrt{\log n}}{n}\right)$. 
\item The upper bounds of errors given in Theorem~\ref{th:upper_bound_robustLasso} increase with $\sigma$, as well as $s$ and $r$, which is quite intuitive. They also decrease with $n$.
\item Theorem \ref{th:upper_bound_robustLasso} serves as a useful tool for developing theoretical results for debiasing, as will be seen later in this section.
\item Theorem \ref{th:upper_bound_robustLasso} holds  with a slight modification in constants when $\boldsymbol{\eta}$ is drawn from a zero-mean sub-Gaussian noise. In particular, if sub-Gaussian norm of $\boldsymbol{\eta}$ is  $\sigma^2$, then the constants $48$ and $24$ in \eqref{eq:beta_lasso_bound} and \eqref{eq:delta_lasso_bound} are replaced by $96$ and $48$, respectively. These constants are obtained by using tail bounds of sub-Gaussian random variables (see Theorem~2.6.3 of \cite{Vershynin2018}), and \eqref{eq:lambda_1}, \eqref{eq:lambda_2} in our proof.
\end{enumerate}
\subsection{A note on the Debiased \textsc{Lasso}}
\label{subsec:debiased_lasso}
Let us consider the measurement vector from \eqref{eq:fm_delta}, momentarily setting $\boldsymbol{\delta^*} \triangleq \boldsymbol{0}$, i.e., we have $\boldsymbol{y} = \boldsymbol{A\beta^*} + \boldsymbol{\eta}$. Let $\boldsymbol{\hat{\beta}_{\lambda}}$ be the minimizer of the following \textsc{Lasso} problem 
\begin{equation}
\min_{\boldsymbol{\beta}}{\frac{1}{2n}}\|\boldsymbol{y}-\boldsymbol{A\beta}\|^2_2 + \lambda \|\boldsymbol{\beta}\|_1, 
\label{eq:basic_lasso}
\end{equation}
for a given value of $\lambda$. 
Though \textsc{Lasso} provides excellent theoretical guarantees \cite[Chapter 11]{THW2015}, it is well known that it produces biased estimates, i.e., $E(\boldsymbol{\hat{\beta}_{\lambda}}) \neq \boldsymbol{\beta^*}$, where the expectation is taken over different instances of $\boldsymbol{\eta}$. There has been considerable effort in mitigating this bias, as reported in  \cite{Javanmard2014,zhang2014confidence,vandegeer2014}. These works essentially replace $\boldsymbol{\hat{\beta}_{\lambda}}$ by a `debiased' estimate $\boldsymbol{\hat{\beta}_{d}}$ given by:
\begin{equation}
\boldsymbol{\hat{\beta}_{d}} = \boldsymbol{\hat{\beta}_{\lambda}} + \dfrac{1}{n} \boldsymbol{M} \boldsymbol{A}^{\top} (\boldsymbol{y} - \boldsymbol{A \hat{\beta}_{\lambda}}),
\label{eq:debiased_beta1}
\end{equation}
where $\boldsymbol{M}$ is an approximate inverse %(defined as in Alg.~\ref{alg:approx_inverse}) 
of  $\boldsymbol{\hat{\Sigma}} \triangleq \boldsymbol{A}^{\top} \boldsymbol{A}/n$. The second term on the RHS of \eqref{eq:debiased_beta1} can be viewed as an adjustment to $\boldsymbol{\hat{\beta}_{\lambda}}$ with a weighted sum of the entries of the residual vector ($\boldsymbol{y} - \boldsymbol{A \hat{\beta}_{\lambda}}$).
Substituting $\boldsymbol{y} = \boldsymbol{A\beta^*} + \boldsymbol{\eta}$ into \eqref{eq:debiased_beta1} and treating $\frac{1}{n} \boldsymbol{MA}^{\top} \boldsymbol{A}$ as approximately equal to the identity matrix, yields:
\begin{equation}\label{eq:basic_deb}
\boldsymbol{\hat{\beta}_{d}} = \boldsymbol{\hat{\beta}_{\lambda}} + \dfrac{1}{n} \boldsymbol{M} \boldsymbol{A}^{\top} (\boldsymbol{A\beta^*} + \boldsymbol{\eta} - \boldsymbol{A \hat{\beta}_{\lambda}}) \approx  \boldsymbol{\beta^*} + \dfrac{1}{n}\boldsymbol{ \boldsymbol{M} \boldsymbol{A}^{\top}\eta},
\end{equation}
which is referred to as a debiased estimate, because $E(\boldsymbol{\hat{\beta}_d}) \approx \boldsymbol{\beta^*}$. In fact, the debiased estimate $\boldsymbol{\hat{\beta}_d}$ is decomposed as: 
\begin{eqnarray}\label{eq:Jm_debiasing}
\boldsymbol{\hat{\beta}_d}-\boldsymbol{\beta^*} = \frac1n\boldsymbol{MA}^{\top}\boldsymbol{\eta} + (\boldsymbol{M\hat{\Sigma}}-\boldsymbol{I_n})(\boldsymbol{\beta^*}-\boldsymbol{\hat{\beta}_d}).
\end{eqnarray}
The second term on the right hand side of \eqref{eq:Jm_debiasing} is the bias of $\boldsymbol{\hat{\beta}_d}$, whereas the variance-covariance matrix of the first term is $\frac1n\boldsymbol{M \hat{\Sigma} M}^{\top}$.   
Reference~\cite{Javanmard2014} provides an algorithm to construct the matrix $\boldsymbol{M}$ which minimizes the variances of the elements of the first term \eqref{eq:Jm_debiasing} together with reducing the bias effect due to the second term. This is accomplished by solving the following optimization problem: 
\begin{equation}
\forall j \in [p], \text{minimize}_{\boldsymbol{m}_j} \;\;\boldsymbol{m_j}^{\top}\boldsymbol{\hat{\Sigma}m_j}\;\; %\text{ w. r. t. } \boldsymbol{m}_j 
\text{ subject to }\;\;\|\boldsymbol{\hat{\Sigma}m_j}-\boldsymbol{e_j}\|_{\infty} \leq \mu.
\label{eq:design_m_jav}
\end{equation}
The constraints ensure that the asymptotic bias of $\boldsymbol{\hat{\beta}_d}$ is negligible under some suitable conditions on $\boldsymbol{\Sigma}$, $n,\ p, s$ and $\mu$.

When $\boldsymbol{\delta^*}\ne \boldsymbol{0}$, then the theoretical guarantees of the debiased \textsc{Lasso} estimator $\boldsymbol{\hat{\beta}_d}$ do not hold. Moreover, we observe in a simulation study that $\boldsymbol{\hat{\beta}_d}$ has larger estimation errors in comparison to the proposed debiased robust \textsc{Lasso} estimator (see the \textsc{Baseline-1} and \textsc{ODrlt} columns of Table~\ref{tb:prop_deb} in Sec.~\ref{subsec:Debiasing_baselines}). In the present setup (i.e., $\boldsymbol{\delta^*}\ne \boldsymbol{0}$), we develop a debiased approach which adjusts the Robust \textsc{Lasso} estimator with a carefully chosen weighted sum of the corresponding residual vectors. The presence of MMEs leads to significant differences in the debiasing procedure, as we show in the next section.

\subsection{Debiasing in the Presence of MMEs}
\label{sec:deb_lasso_test}
In the presence of MMEs, the design matrix $(\boldsymbol{A}|\boldsymbol{I_n})$ from \eqref{eq:fm_delta} plays the role of $\boldsymbol{A}$ in \eqref{eq:design_m_jav}. However 
$(\boldsymbol{A}|\boldsymbol{I_n})$
is partly random and partly deterministic, whereas the theory in \cite{Javanmard2014} applies to either purely random or purely deterministic (Theorem~8 and Theorem~6 of \cite{Javanmard2014}, respectively) matrices, but not a combination of both. Additionally, Theorem~8 of \cite{Javanmard2014} requires the rows of the sensing matrix to have zero mean, which is not satisfied by $(\boldsymbol{A}|\boldsymbol{I_n})$. A simple mean correction will not work because it would alter the structure of $\boldsymbol{\delta^*}$ which itself arises from bit-flips. 
Hence, their theoretical results  do not apply for the approximate inverse of 
$\frac{1}{n}(\boldsymbol{A}|\boldsymbol{I_n})^{\top}(\boldsymbol{A}|\boldsymbol{I_n})$
obtained using \eqref{eq:design_m_jav}. Numerical results for the weaker performance (with respect to sensitivity and specificity, defined in Sec.~\ref{sec:experiments}) of `debiasing' (as in \eqref{eq:debiased_beta1}) with such an approximate inverse are demonstrated in Sec.~\ref{subsec:Debiasing_baselines} (see the \textsc{Baseline-2} column of Table~\ref{tb:prop_deb}).

To produce a debiased estimate of $\boldsymbol{\beta^*}$ in the presence of MMEs in the pooling matrix, we adopt a different approach. 
We define a linear combination of the residual error vector 
(i.e., $\boldsymbol{y}-\boldsymbol{A\hat{\beta}_{\lambda_1}} - \boldsymbol{\hat{\delta}_{\lambda_2}}$)
produced by the robust \textsc{Lasso} estimator from \eqref{eq:lasso_delta} {via a carefully chosen set of weights}, in order to debias the robust \textsc{Lasso} estimates $\boldsymbol{\hat{\beta}_{\lambda_1}}$, $\boldsymbol{\hat{\delta}_{\lambda_2}}$. These weights are represented in the form of an appropriately designed matrix $\boldsymbol{W} \in \mathbb{R}^{n \times p}$ for debiasing $\boldsymbol{\hat{\beta}_{\lambda_1}}$ and produce an estimate with minimal asymptotic variance. The matrix $\boldsymbol{W}$ also leads to a derived weights matrix 
$\left(\boldsymbol{I_n}-\frac{1}{n}\boldsymbol{WA}^{\top}\right)$ for  debiasing $\boldsymbol{\hat{\delta}_{\lambda_2}}$. Our procedure to design an ``optimal'' $\boldsymbol{W}$ is given in Alg.~\ref{alg:design_W}. 

Given weight matrix $\boldsymbol{W}$, we define debiased robust \textsc{Lasso} estimates for $\boldsymbol{\beta^*}$ and $\boldsymbol{\delta^*}$ as follows:
\begin{eqnarray} 
\boldsymbol{\hat{\beta}_{W}}&\triangleq&\boldsymbol{\hat{\beta}_{\lambda_1}}+\frac1n\boldsymbol{W}^{\top} (\boldsymbol{y}-\boldsymbol{A\hat{\beta}_{\lambda_1}}-\boldsymbol{\hat{\delta}_{\lambda_2}})\label{eq:deb_beta_est},\\
\boldsymbol{\hat{\delta}_{W}}&\triangleq&\boldsymbol{y}-\boldsymbol{A}\boldsymbol{\hat{\beta}_W} = \boldsymbol{\hat{\delta}_{\lambda_2}}+\left(\boldsymbol{I_n}-\frac1n \boldsymbol{AW}^{\top} \right)(\boldsymbol{y}-\boldsymbol{A\hat{\beta}_{\lambda_1}}-\boldsymbol{\hat{\delta}_{\lambda_2}}). \label{eq:deb_delta_est}
\end{eqnarray}
Note that, in our work, the matrix $\boldsymbol{W}$ %does not play the role of the approximate inverse $\boldsymbol{M}$ from Alg.~\ref{alg:M}, but instead 
plays the role of $\boldsymbol{AM}^{\top}$ (comparing \eqref{eq:deb_beta_est} and \eqref{eq:debiased_beta1}).
From the forward model in \eqref{eq:fm_delta}, and using \eqref{eq:deb_beta_est} and \eqref{eq:deb_delta_est}, we obtain the following expressions:
\begin{eqnarray}
 \boldsymbol{\hat{\beta}_W}-\boldsymbol{\beta^*}&=&\frac{1}{n}\boldsymbol{W}^{\top}\boldsymbol{{\eta}}-\underbrace{\left(\boldsymbol{I_p}-\frac{1}{n}\boldsymbol{W}^{\top}\boldsymbol{A}\right)  (\boldsymbol{\beta^*-\hat{\beta}_{\lambda_1}})+
\frac{1}{n}\boldsymbol{W}^{\top}
\left(\boldsymbol{\delta^{*}}-\boldsymbol{\hat{\delta}_{\lambda_2}}\right)}_{\text{bias terms}}, \label{eq:beta_d_unscaled}
\\
 \boldsymbol{\hat{\delta}_W}-\boldsymbol{\delta^*}&=&
\left(\boldsymbol{I_n}-\frac{1}{n}\boldsymbol{A}\boldsymbol{W}^{\top}\right)
\boldsymbol{{\eta}}+
\underbrace{\left(\boldsymbol{I_n}-\frac{1}{n}\boldsymbol{A}\boldsymbol{W}^{\top}\right) \boldsymbol{A} (\boldsymbol{\beta^*-\hat{\beta}_{\lambda_1}})-
\frac{1}{n}\boldsymbol{AW}^{\top}
\left(\boldsymbol{\delta^{*}}-\boldsymbol{\hat{\delta}_{\lambda_2}}\right)}_{\text{bias terms}}. \label{eq:delta_d_unscaled}
\end{eqnarray}
Note that the first terms on the RHS of both \eqref{eq:beta_d_unscaled} and \eqref{eq:delta_d_unscaled} are zero-mean Gaussian (given $\boldsymbol{W}$). The remaining two terms in both equations are referred to as `bias terms'. We intend to design $\boldsymbol{W}$ in such a way that the two bias terms in each equation are negligible in comparison to the standard deviation of the elements of the first term. In such a scenario, the sum of the asymptotic variance of the elements of $\boldsymbol{\hat{\beta}_W}$ will be $\frac{\sigma^2}{n^2}\sum_{j=1}^p \boldsymbol{w_{.j}}^{\top} \boldsymbol{w_{.j}}$ (which is also referred to as asymptotic total variance of $\boldsymbol{\hat{\beta}_W}$). It is well known that a statistical test based on a statistic with smaller variance is generally more powerful than that based on a statistic with higher variance \cite{Berger}. Hence, we design the weights matrix $\boldsymbol{W}$ which \emph{minimizes} this sum total variance, and also satisfies the condition of negligible bias. We develop a procedure for the design of $\boldsymbol{W}$ as presented in Alg.~\ref{alg:design_W} which ensures negligible bias. In particular, the constraints given in Alg.~\ref{alg:design_W} ensure the following asymptotic results regarding the bias and standard deviation (formally presented in Theorems~\ref{th:2_3term_beta_decompose_W} and \ref{th:dist_conv_W}):
\begin{enumerate}[(i)]
\item The two bias terms on the RHS of \eqref{eq:beta_d_unscaled} are $o_P\left(\frac{1}{\sqrt{n}}\right)$, whereas the standard deviation of the elements of the first term is $O\left(\frac{1}{\sqrt{n}}\right)$.
\item The two bias terms on the RHS of \eqref{eq:delta_d_unscaled} are $o_P\left(\frac{p\sqrt{1-n/p}}{n}\right)$, whereas the standard deviation of the elements of the first term is $O\left(\frac{p\sqrt{1-n/p}}{n}\right)$. 
\end{enumerate}
Thus for both $\boldsymbol{\hat{\beta}_W}$ and $\boldsymbol{\hat{\delta}_W}$, the bias terms are negligible in comparison to the first term. 
\begin{algorithm} 
\caption{Design of $\boldsymbol{W}$ for Rademacher matrix $\boldsymbol{A}$}
\begin{algorithmic}[1]
\REQUIRE 
  $\boldsymbol{A}$, $\mu_1$, $\mu_2$ and $\mu_3$ 
\ENSURE
$\boldsymbol{W}$ 
\STATE We solve the following optimisation problem :
\begin{eqnarray}
\nonumber \underset{\boldsymbol{W}}{\textrm{minimize}}& &\quad 
\sum_{j=1}^p\boldsymbol{w_{.j}}^{\top}\boldsymbol{w_{.j}} \\ \nonumber
\textrm{subject to}& & 
\mathsf{C0} : \boldsymbol{w_{.j}}^{\top}\boldsymbol{w_{.j}}/n \leq 1 \ \forall \ j \in [p] 
\\ & & \nonumber \mathsf{C1} : \left| \left(\boldsymbol{I_p}-\frac1n\boldsymbol{W^{\top}A}\right)\right|_\infty \leq \mu_1 , \\ \nonumber & & \mathsf{C2} : \left|\frac{1}{p}\left(\boldsymbol{I_n}-\frac{1}{n}\boldsymbol{A}\boldsymbol{W}^{\top}\right) \boldsymbol{A}\right|_\infty \leq \mu_2, \\ \nonumber & & \mathsf{C3} : \left|\left(\frac{\boldsymbol{AW^{\top}}}{p}-\boldsymbol{I_n}\right)\right|_{\infty} \leq \mu_3,
\end{eqnarray}
where $\mu_1 \triangleq 2\sqrt{\frac{2\log(p)}{n}}$, $\mu_2 \triangleq 2\sqrt{\frac{\log({2np})}{np}}+\frac{1}{n}$ and $\mu_3\triangleq 2\sqrt{\frac{2\log(n)}{p}}$.
\STATE If the above problem is not feasible, then set $\boldsymbol{W=A}$.
\end{algorithmic}
\label{alg:design_W}
\end{algorithm}

We now describe the procedure to design $\boldsymbol{W}$ which minimizes  $\sum_{j=1}^p \boldsymbol{w_{.j}^{\top}  w_{.j}}$, subject to the constraints $\mathsf{C0}, \mathsf{C1}, \mathsf{C2}, \mathsf{C3}$ on $\boldsymbol{W}$, as summarized in Alg.~\ref{alg:design_W}. This is a key contribution of our work. The constraint $\mathsf{C0}$ is required to upper bound the asymptotic variance of the elements of $\boldsymbol{\hat{\beta}_W}$. The constraint $\mathsf{C1}$ (via $\mu_1$) controls the rate of convergence of the first bias term on the RHS of \eqref{eq:beta_d_unscaled}, as shown in Theorem~\ref{th:2_3term_beta_decompose_W}. We later show within the proof of this theorem(see \eqref{eq:rate_B2_orig_W}) that due to this, the second bias term on the RHS of \eqref{eq:beta_d_unscaled} also converges at the same rate. The constraint $\mathsf{C2}$ (via $\mu_2$) controls the rate of convergence of bias terms on the RHS of \eqref{eq:delta_d_unscaled}, again shown in Theorem~\ref{th:2_3term_beta_decompose_W}. Likewise, we later show in the proof of this theorem (see \eqref{eq:rate_D2_W}) that the second bias term on the RHS of \eqref{eq:delta_d_unscaled} consequently also converges at the same rate. 
Furthermore, the constraint $\mathsf{C3}$ (via $\mu_3$) allows us to control the variance of the first term on the RHS of \eqref{eq:delta_d_unscaled}, i.e. the asymptotic variance of $\boldsymbol{\hat{\delta}_W}$, as will be shown in Theorem~\ref{th:dist_conv_W}. %\textcolor{blue}{The asymptotic variance of $\boldsymbol{\hat{\delta}_W}$ scales with the order $O(p(\sqrt{1-n/p})/n)$. Therefore, to ensure that the asymptotic variance of $\boldsymbol{\hat{\delta}_W}$ converges, we scale \eqref{eq:delta_d_unscaled} by $n/p(\sqrt{1-n/p})$.} \textcolor{red}{Not clear. Need to discuss.}

We note that Alg.~\ref{alg:design_W} is a convex optimization problem, as it has a convex cost function and convex constraints. The values of $\mu_1, \mu_2, \mu_3$ are selected in such a way that each of the constraints $\mathsf{C1}, \mathsf{C2}, \mathsf{C3}$ in Alg.~\ref{alg:design_W} holds with high probability for the choice $\boldsymbol{W} \triangleq \boldsymbol{A}$, as will be formally established in Lemma~\ref{le:Rad_mat_prop} for any Rademacher matrix $\boldsymbol{A}$. Note that $\mu_1, \mu_2, \mu_3$ are all $o(1)$ asymptotically, as would be required for the bias terms to be negligible. These constraints are derived from Theorem~\ref{th:2_3term_beta_decompose_W} and ensure that the bias terms are negligible. The choice of $\boldsymbol{W} \triangleq \boldsymbol{A}$ is partly motivated by the fact that in case the optimization problem in Alg.~\ref{alg:design_W} is infeasible, we set $\boldsymbol{W} \triangleq \boldsymbol{A}$. Furthermore, the choice $\boldsymbol{W} \triangleq \boldsymbol{A}$ helps us establish that the set of all possible $\boldsymbol{W}$ matrices which satisfy the constraints in Alg.~\ref{alg:design_W} is non-empty with high probability. 

We now formally state Theorems~\ref{th:2_3term_beta_decompose_W} and \ref{th:dist_conv_W}. These theorems play a vital role in deriving  Theorem~\ref{th:distribution_beta_delta_opt} that leads to developing the optimal debiased robust \textsc{Lasso} (\textsc{ODrlt}) tests. These theorems constitute the main theoretical contributions of this work.
\newline
\begin{theorem}[\textsc{ODrlt}-Bias]
\label{th:2_3term_beta_decompose_W} 
 Let  $\boldsymbol{\hat{\beta}}_{\boldsymbol{\lambda_1}},\boldsymbol{\hat{\delta}_{\lambda_2}}$ be as in \eqref{eq:lasso_delta}, $\boldsymbol{\hat{\beta}_{W}}$, $\boldsymbol{\hat{\delta}_{W}}$ be as in \eqref{eq:deb_beta_est}, \eqref{eq:deb_delta_est} respectively and set $\lambda_1 \triangleq \frac{4\sigma\sqrt{\log p}}{\sqrt{n}}, \lambda_2 \triangleq \frac{4\sigma\sqrt{\log n}}{{n}}$. Let $\boldsymbol{A}$ be a random Rademacher matrix and let $\boldsymbol{W}$ be obtained from Alg.~\ref{alg:design_W}. Then if $n$ is $o(p)$  and $n$ is $\omega[((s+r)\log p)^2]$, as $p,n\to \infty$, we have:
\begin{enumerate}
\item    
\begin{equation}
        \left\|\sqrt{n}\left(\boldsymbol{I_p}-\frac{1}{n}\boldsymbol{W^{\top}}\boldsymbol{A}\right)  (\boldsymbol{\beta^*-\hat{\beta}_{\lambda_1}})\right\|_\infty=o_P(1). \label{eq:bias_beta_1}
        \end{equation}
\item 
\begin{equation}
\left\|\sqrt{n} \ \frac{1}{n}\boldsymbol{W}^{\top}
\left(\boldsymbol{\delta^{*}}-\boldsymbol{\hat{\delta}_{\lambda_2}}\right)\right\|_\infty=o_P(1).\label{eq:bias_beta_2}
\end{equation}
\item 
\begin{equation}
\left\|\frac{n}{p\sqrt{1-n/p}}\left(\boldsymbol{I_n}-\frac{1}{n}\boldsymbol{A}\boldsymbol{W}^{\top}\right) \boldsymbol{A} (\boldsymbol{\beta^*-\hat{\beta}_{\lambda_1}})\right\|_\infty=o_P(1).\label{eq:bias_delta_1}
\end{equation}
\item
\begin{equation}
        \left\|\frac{n}{p\sqrt{1-n/p}}\frac{1}{n}\boldsymbol{AW}^{\top}
\big(\boldsymbol{\delta^{*}}-\boldsymbol{\hat{\delta}_{\lambda_2}}\big)\right\|_\infty = o_P(1).\label{eq:bias_delta_2}
    \end{equation}
\end{enumerate}
\hfill{$\blacksquare$}
\end{theorem}

\medskip
Define the following matrices:
\begin{eqnarray}
\boldsymbol{\Sigma_{\beta}}&\triangleq& {\text{Var}\left(\frac{1}{\sqrt{n}}\boldsymbol{W}^{\top}\boldsymbol{{\eta}}\right)}= \sigma^2\frac{1}{n}\boldsymbol{W^{\top}W},\label{eq:Sig_beta}\\
\boldsymbol{\Sigma_{\delta}} &\triangleq& \text{Var}\left(\left(\boldsymbol{I_n}-\frac{1}{n}\boldsymbol{A}\boldsymbol{W}^{\top}\right)
\boldsymbol{{\eta}}\right)=\sigma^2\left(\boldsymbol{I_n}-\frac{1}{n}\boldsymbol{AW}^{\top}\right)\left(\boldsymbol{I_n}-\frac{1}{n}\boldsymbol{AW}^{\top}\right)^{\top}. \label{eq:Sig_delta}
\end{eqnarray}
Note that $\boldsymbol{\Sigma_\beta}/n$ and $\boldsymbol{\Sigma_\delta}$ are the variance-covariance matrix of the first terms of the RHS of \eqref{eq:beta_d_unscaled} and \eqref{eq:delta_d_unscaled}, respectively.

Theorem~\ref{th:dist_conv_W} shows that when $\boldsymbol{W}$ is chosen as per Alg.~\ref{alg:design_W}, the element-wise variances of the first term of the RHS of \eqref{eq:beta_d_unscaled} (diagonal elements of $\boldsymbol{\Sigma_{\beta}}$) approach $1$ in probability. The constraints $\mathsf{C0}$ and $\mathsf{C1}$ of Alg.~\ref{alg:design_W} are mainly used to establish this theorem.  
Further, for the optimal choice of $\boldsymbol{W}$ as in Alg.~\ref{alg:design_W}, we show that the element-wise variances of the first term of the RHS of \eqref{eq:delta_d_unscaled} (diagonal elements of $\boldsymbol{\Sigma_{\delta}}$) tend to $\sigma^2$ in probability. To establish this, we use the constraint $\mathsf{C3}$ of Alg.~\ref{alg:design_W}. 
\bigskip
\begin{theorem}[\textsc{ODrlt}-Variance]
\label{th:dist_conv_W}
Let $\boldsymbol{A}$ be a Rademacher matrix. Suppose  $\boldsymbol{W}$ is obtained from Alg.~\ref{alg:design_W} and $\boldsymbol{\Sigma_{\beta}}$ and $\boldsymbol{\Sigma_{\delta}}$ are defined as in \eqref{eq:Sig_beta} and \eqref{eq:Sig_delta}, respectively. If $n \log n$ is $o(p)$ and $n$ is $\omega[((s+r)\log p)^2]$, as $n,p \to \infty$, we have the following:
    \begin{enumerate}[(1)]
        \item For $j \in [p]$,
        \begin{equation}
            \Sigma_{\beta_{jj}} \overset{P}{\to} \sigma^2.
        \end{equation}
        \item For $i \in [n]$,
        \begin{equation}
            \frac{n^2}{p^2(1-n/p)}\Sigma_{\delta_{ii}} \overset{P}{\to} \sigma^2. 
        \end{equation}
        \hfill $\blacksquare$
    \end{enumerate}
\end{theorem}
When we choose an optimal $\boldsymbol{W}$ as per the Alg.~\ref{alg:design_W}, the equations \eqref{eq:beta_d_unscaled} and \eqref{eq:delta_d_unscaled} along with Theorem~\ref{th:2_3term_beta_decompose_W} and Theorem~\ref{th:dist_conv_W} can be used to derive the asymptotic distribution of $\boldsymbol{\hat{\beta}_W}$
 and $\boldsymbol{\hat{\delta}_W}$. This is accomplished in Theorem~\ref{th:distribution_beta_delta_opt}.
\bigskip
\begin{theorem}[\textsc{ODrlt}-Distribution]
\label{th:distribution_beta_delta_opt}  
Let  $\boldsymbol{\hat{\beta}}_{\boldsymbol{\lambda_1}},\boldsymbol{\hat{\delta}_{\lambda_2}}$ be as in \eqref{eq:lasso_delta}, $\boldsymbol{\hat{\beta}_{W}}$, $\boldsymbol{\hat{\delta}_{W}}$ be as in \eqref{eq:deb_beta_est}, \eqref{eq:deb_delta_est} respectively and set $\lambda_1 \triangleq \frac{4\sigma\sqrt{\log p}}{\sqrt{n}}, \lambda_2 \triangleq \frac{4\sigma\sqrt{\log n}}{{n}}$. Let $\boldsymbol{A}$ be a random Rademacher matrix and  $\boldsymbol{W}$ be the debiasing matrix obtained from Alg.~\ref{alg:design_W}. If  $n$ is $\omega[((s+r)\log p)^2]$ and $n \log n$ is $o(p)$, then we have:
\begin{enumerate}[(1)]
    \item For fixed $j\in[p]$, 
\begin{equation}\label{eq:beta_marginal_dist}
   \frac{\sqrt{n}(\hat{\beta}_{{\scaleto {Wj}{5pt}}}-{\beta}^*_j)}{\sqrt{\Sigma_{\beta_{jj}}}}\xrightarrow{\mathcal{L}} \mathcal{N}(0,1) \mbox { as } p,n\to\infty.
\end{equation}
\item For fixed $i\in[n]$
\begin{eqnarray}\label{eq:delta_marginal_dist}
   \frac{\hat{\delta}_{{\scaleto {Wi}{5pt}}}-\delta^*_i}{\sqrt{\Sigma_{\delta_{ii}}}}\xrightarrow{\mathcal{L}} \mathcal{N}(0,1) \mbox { as } p,n\to\infty,
\end{eqnarray}
where $\Sigma_{\beta_{jj}}$ and $\Sigma_{\delta_{ii}}$ are the $j^{\text{th}}$ and $i^{\text{th}}$  diagonal elements of matrices $ \boldsymbol{\Sigma_{\beta}}$ (as in \eqref{eq:Sig_beta}) and 
$ \boldsymbol{\Sigma_{\delta}}$ (as in \eqref{eq:Sig_delta}), respectively.
\end{enumerate}
\hfill{$\blacksquare$}
\end{theorem}

\medskip
Theorem \ref{th:distribution_beta_delta_opt} paves the way to develop the following optimal debiased robust \textsc{Lasso} test or \textsc{ODrlt} for Aim (i) and (ii) of this work.
\newline
\noindent\textbf{Optimal \textsc{Drlt} for $\boldsymbol{\beta^*}$:} We now present a hypothesis testing procedure for an optimally designed $\boldsymbol{W}$ to determine defective samples based on Theorem~\ref{th:distribution_beta_delta_opt}. Given $\alpha>0$, we reject the null hypothesis $\mathsf{G_{0,j}}: \beta^*_j=0$ in favor of $\mathsf{G_{1,j}}: \beta^*_j \neq 0$, for each
$j \in [p]$ when
\begin{equation}\label{eq:beta_ODRLT}
\sqrt{n}|\hat{\beta}_{Wj}|\big/\sqrt{ \Sigma_{\beta_{jj}}} > z_{\alpha/2},
\end{equation}
where $z_{\alpha/2}$ is the upper $(\alpha/2)^{\text{th}}$ quantile of a standard normal random variable.

\noindent\textbf{Optimal \textsc{Drlt} for $\boldsymbol{\delta^*}$:} We develop a hypothesis testing procedure based on Theorem~\ref{th:distribution_beta_delta_opt} corresponding to optimal $\boldsymbol{W}$ to determine whether or not a measurement in $\boldsymbol{y}$ is affected by an effective MME. As before, given $\alpha>0$, for $i \in  [n]$, we reject the null hypothesis $\mathsf{H_{0,i}}:\delta^*_i = 0$ in favor of  $\mathsf{H_{1,i}} : \delta^*_i \ne 0$ when
\begin{equation}\label{eq:delta_ODRLT}
 |\hat{\delta}_{Wi}| \big/ \sqrt{\Sigma_{\delta_{ii}}} > z_{\alpha/2}. 
\end{equation}
\\
\noindent\textbf{Remarks on Theorem~\ref{th:distribution_beta_delta_opt}:}
\begin{enumerate}
\item A desirable property of a statistical test is that the probability of rejecting the null hypothesis when the alternate is true converges to $1$ as $n \rightarrow \infty$ (referred to as a consistent test). Theorem~\ref{th:distribution_beta_delta_opt} ensures that the proposed \textsc{ODrlt} is consistent for $\boldsymbol{\beta^*}$. This implies that the sensitivity (defined in Sec.~\ref{sec:experiments}) of the test for $\boldsymbol{\beta^*}$ approaches $1$ as $n,p \rightarrow \infty$. On the other hand, its specificity (defined in Sec.~\ref{sec:experiments}) approaches $1-\alpha$ as $n,p \rightarrow \infty$. 
\item Additionally, Theorem \ref{th:distribution_beta_delta_opt}  shows that probability of rejecting the null hypothesis when the null is true, converges to $\alpha$ (referred to as an asymptotically unbiased test).
\item The asymptotic distributions of the LHS terms in \eqref{eq:beta_marginal_dist} and \eqref{eq:delta_marginal_dist} do not depend on $\boldsymbol{A}$. These distributions are asymptotically Gaussian because the noise vector $\boldsymbol{\eta}$ is normally distributed. 

\item The condition $n < p$ in Result (1) emerges from \eqref{eq:min_sing_A} and \eqref{eq:psuedo_sing}, which are based on probabilistic bounds on the singular values of random Rademacher matrices \cite{Rudelson2009}. For the special case where $n = p$ (which is no longer a compressive regime), these bounds are no longer applicable, and instead results such as \cite[Thm. 1.2]{rudelson2008littlewood} can be used. 

\item Theorem~\ref{th:2_3term_beta_decompose_W} and~\ref{th:dist_conv_W} also apply when noise $\boldsymbol{\eta}$ is drawn from a known zero-mean sub-Gaussian distribution 
with sub-Gaussian norm $\sigma^2$. Thus, asymptotic bias of $\boldsymbol{\hat{\beta}_{W}}$ and $\boldsymbol{\hat{\delta}_{W}}$ is negligible even when the additive noise is sub-Gaussian. Moreover, the expression for the asymptotic variances of the debiased robust \textsc{Lasso} estimators for Gausian and sub-Gaussian noise are the same. In view of this, the asymptotic distribution of $\boldsymbol{\hat{\beta}_{W}}$ and $\boldsymbol{\hat{\delta}_{W}}$ can be obtained by using the central limit theorem, applied respectively to the first terms on the RHS of \eqref{beta_marginal_W} and \eqref{delta_W_dist_W}. However for the purpose of a statistical test, we propose an empirical version \textsc{ODrlt} below. This empirical \textsc{ODrlt} determines the thresholds for rejection of the test based on the \emph{empirical} distribution of the first terms on the RHS of \eqref{beta_marginal_W} and \eqref{delta_W_dist_W}.\\ 
\noindent\textbf{Empirical \textsc{ODrlt}:} For a known zero-mean sub-Gaussian $\boldsymbol{\eta}$, we now present the following version of \textsc{ODrlt}:
\begin{enumerate}
    \item For $\boldsymbol{\beta^*}$: Reject $\mathsf{G_{0,j}}: \beta^*_j=0$ in favor of $\mathsf{G_{1,j}}: \beta^*_j \neq 0$, if
\begin{equation}\label{eq:beta_EODRLT}
\sqrt{n}|\hat{\beta}_{Wj}|\big/\sqrt{ \Sigma_{\beta_{jj}}} > \zeta_{j,\alpha/2},
\end{equation}
where $j\in[p]$, and $\zeta_{j,\alpha/2}$ is the upper $(\alpha/2)^{\text{th}}$ quantile of $\frac1n \boldsymbol{w}_{\cdot j}^\top \boldsymbol{\eta}/\sqrt{ \Sigma_{\beta_{jj}}}$.
\item For $\boldsymbol{\delta^*}$: Reject $\mathsf{H_{0,i}}:\delta^*_i = 0$ in favor of  $\mathsf{H_{1,i}} : \delta^*_i \ne 0$ if
\begin{equation}\label{eq:delta_EODRLT}
 |\hat{\delta}_{Wi}| \big/ \sqrt{\Sigma_{\delta_{ii}}} > \xi_{i,\alpha/2}, 
\end{equation}
where $i\in[n]$, and  $\xi_{i,\alpha/2}$ is the upper $(\alpha/2)^{\text{th}}$ quantile of $(\eta_i-\frac1n \boldsymbol{a}_{i\cdot}\boldsymbol{W}^\top\boldsymbol{\eta})/\sqrt{\Sigma_{\delta_{ii}}}$.
\end{enumerate}
The cut-offs $\zeta_{j,\alpha/2}$ and $\xi_{i,\alpha/2}$ are obtained using simulated $\boldsymbol{\eta}$ and weight matrix $\boldsymbol{W}$.
%\textcolor{blue}{Consider the case that the additive noise $\boldsymbol{\eta}$ obeys some known zero-mean sub-Gaussian distribution. The noise terms for the debiased robust \textsc{Lasso} estimates $\boldsymbol{\hat{\beta}_W}$ and $\boldsymbol{\hat{\delta}_W}$ are $\frac{1}{n}\boldsymbol{W^\top \eta}$ and $\left(\boldsymbol{I_n}-\frac{1}{n}\boldsymbol{W}\boldsymbol{A}^{\top}\right)
%\boldsymbol{{\eta}}$ respectively. For a given $\boldsymbol{A}$ and $\boldsymbol{W}$, the distribution of these terms converges to a Gaussian distribution using the Central Limit Theorem (since the values of $\boldsymbol{W}$ are bounded as per constraint $\mathsf{C0}$ in Alg.~\ref{alg:design_W}).  
%Alternatively, we can obtain the cut-off for the hypothesis test empirically as follows.  Since we assume that $\boldsymbol{A}$ and the distribution of $\boldsymbol{\eta}$ are known, we can sample from $\frac{1}{n}\boldsymbol{W^\top \eta}$ and $\left(\boldsymbol{I_n}-\frac{1}{n}\boldsymbol{W}\boldsymbol{A}^{\top}\right)
%\boldsymbol{{\eta}}$ multiple times and then obtain the $2.5\%$ quantile and $97.5\%$ quantile of the distributions respectively. Then we reject the null-hypothesis when the scaled test statistics $\sqrt{n}|\hat{\beta}_{Wj}|\big/\sqrt{ \Sigma_{\beta_{jj}}}$ and $|\hat{\delta}_{Wi}| \big/ \sqrt{\Sigma_{\delta_{ii}}}$ lie outside these quantiles.}

\item The values in $\boldsymbol{\delta^*}$ depend on $\boldsymbol{\beta^*}$ in the form $\delta^*_i = (\boldsymbol{\hat{a}}_{i,.} - \boldsymbol{a}_{i,.}) \boldsymbol{\beta^*}$. However, $\boldsymbol{\hat{a}}_{i,.}$ is completely unknown to us. Hence, we estimate $\boldsymbol{\delta^*}$ in an independent manner so that we have a computationally tractable estimator with provable performance bounds, because searching for $\boldsymbol{\hat{a}}_{i,.}$ directly is infeasible. Note that our primary aim is to find out the defective samples in $\boldsymbol{\beta^*}$. Determining which pools could possibly contain errors during preparation, aids this process. The erroneous pools are determined by the hypothesis test in this theorem. Hence, very accurate values of $\boldsymbol{\delta^*}$ are not necessary. Once, the erroneous pools are determined, the corresponding measurements can be either discarded, or else the technician can be asked to repeat those particular measurements. There is a chance of false positives and false negatives for detecting erroneous pools, especially at high noise levels. However, the experimental results in Figure 2 show that the sensitivity and specificity for the tests for MMEs is more than 90\% for a very wide range of experimental parameters such as number of measurements, measurement noise level, proportion of effective MMEs and signal sparsity.
\end{enumerate}

We now state an additional theorem, i.e., Theorem~\ref{th:dist_debiased_beta_delta_A} below, which shows that the estimates $\boldsymbol{\hat{\beta}_W}, \boldsymbol{\hat{\delta}_W}$ are debiased even for the choice $\boldsymbol{W} \triangleq \boldsymbol{A}$. For clarity, we denote such debiased estimates by $\boldsymbol{\hat{\beta}_A}, \boldsymbol{\hat{\delta}_A}$. Note that, Theorem \ref{th:dist_debiased_beta_delta_A} is based on the (additive Gaussian) noise model given in \eqref{eq:forward_model_bitflip}.

\begin{theorem}[\textsc{Drlt}-Distribution]
\label{th:dist_debiased_beta_delta_A} 
Let  $\boldsymbol{\hat{\beta}}_{\boldsymbol{\lambda_1}},\boldsymbol{\hat{\delta}_{\lambda_2}}$ be as in \eqref{eq:lasso_delta}, $\boldsymbol{\hat{\beta}_{A}}$, $\boldsymbol{\hat{\delta}_{A}}$ be as in \eqref{eq:deb_beta_est}, \eqref{eq:deb_delta_est} respectively corresponding to $\boldsymbol{W} = \boldsymbol{A}$ and set $\lambda_1 \triangleq \frac{4\sigma\sqrt{\log p}}{\sqrt{n}}, \lambda_2 \triangleq \frac{4\sigma\sqrt{\log n}}{{n}}$. Suppose that $n$ is $\omega[((s+r)\log p)^2]$ \footnote{Given functions $f(n)$ and $g(n)$ of $n \in \mathbb{R}$, we say that $f(n)$ is $\omega(g(n))$ if $\lim_{n \rightarrow \infty} \dfrac{f(n)}{g(n)} = \infty$, i.e. $f(n)$ asymptotically `dominates' $g(n)$.}, and that $\boldsymbol{A}$ is a Rademacher matrix. \begin{enumerate}[(1)]
    \item If $n < p$, then for any $j\in[p]$,     \begin{equation}\label{eq:beta_marginal_dist_A}
   \sqrt{n}(\hat{\beta}_{{\scaleto {Aj}{5pt}}}-{\beta}^*_j)\xrightarrow{\mathcal{L}} \mathcal{N}\left(0, \sigma^2\right) \mbox { as } p,n\to\infty.
\end{equation}
\item If $n\log n$ is $o(p)$, then for any $i\in[n]$,
\begin{eqnarray}\label{eq:delta_marginal_dist_A}
   \frac{\hat{\delta}_{{\scaleto {Ai}{5pt}}}-\delta^*_i}{\sqrt{1-\frac{2p}{n}
     +\frac{1}{n^2}\boldsymbol{a_{i.}A^{\top}A}\boldsymbol{a_{i.}}^{\top}}}\xrightarrow{\mathcal{L}} \mathcal{N}\left(0, \sigma^2\right) \mbox { as } p,n\to\infty.
\end{eqnarray}
%where $\boldsymbol{\Sigma_{A}}$ is defined as follows:
%\begin{equation} \label{eq:Sigma_A}
%   \boldsymbol{\Sigma_{A}} \triangleq \left(\boldsymbol{I_n}-\frac{1}{n}\boldsymbol{AA}^{\top}\right)\left(\boldsymbol{I_n}-\frac{1}{n}\boldsymbol{AA}^{\top}\right)^{\top} .
%\end{equation}
\end{enumerate}
Here $\xrightarrow{\mathcal{L}}$ denotes the convergence in law/distribution.
\hfill{$\blacksquare$}
\end{theorem}
\noindent\textbf{Remarks on Theorem~\ref{th:dist_debiased_beta_delta_A}}
\begin{enumerate}
\item With $\boldsymbol{W} = \boldsymbol{A}$, the diagonal elements of the covariance matrices $\boldsymbol{\Sigma_{\beta}}$ and $\boldsymbol{\Sigma_{\delta}}$ from \eqref{eq:Sig_beta} and \eqref{eq:Sig_delta} simplify to the following: $\forall j \in [p], \Sigma_{\beta_{jj}} = \sigma^2$, and $\forall i \in [n], \Sigma_{\delta_{ii}} = \left(1-\frac{2p}{n}
     +\frac{1}{n^2}\boldsymbol{a_{i.}A^{\top}A}\boldsymbol{a_{i.}}^{\top}\right) \sigma^2$.
\item Note that the \emph{asymptotic variance} of $\boldsymbol{\hat{\beta}_W}$ with the optimal $\boldsymbol{W}$ is lower than that of $\boldsymbol{\hat{\beta}_A}$. This follows since $\boldsymbol{W}$ designed by Alg.~\ref{alg:design_W} minimizes the asymptotic variance. The probabilistic bounds for $\Sigma_{\beta_{jj}}$, which is proportional to the variance of the $j$th element of $\boldsymbol{\hat{\beta}_W}$, are provided in \eqref{eq:sigma_beta_jj_bound} in the proof of Theorem~\ref{th:dist_conv_W}, which indicate that $\Sigma_{\beta_{jj}} \leq \sigma^2$ with high probability. Indeed, 
we show numerically in Sec.~\ref{subsec:var_comp} that the hypothesis tests based on $\boldsymbol{W}$ designed via Alg.~\ref{alg:design_W} perform better in comparison to those using $\boldsymbol{W} = \boldsymbol{A}$. 
\item \textsc{Drlt} is statistically inefficient in comparison to \textsc{ODrlt}. However it is computationally more efficient as it does not require execution of Alg.~\ref{alg:design_W}.
\end{enumerate}

The hypothesis tests for the case where $\boldsymbol{W} = \boldsymbol{A}$ are referred to as \textsc{Drlt}. They are defined as follows:
\\
\noindent\textbf{\textsc{Drlt} for $\boldsymbol{\beta^*}$:} %In Aim (i), we intended to develop a statistical test to determine whether a sample is defective or not. 
Given the significance level $\alpha \in [0,1]$, for each $j \in [p]$, we reject the null hypothesis $\mathsf{G_{0,j}}: \beta^*_j=0$ in favor of $\mathsf{G_{1,j}}: \beta^*_j \neq 0$ when 
\begin{equation}
\dfrac{\sqrt{n}|\hat{\beta}_{Aj}|}{\sigma}>z_{\alpha/2}, 
\label{eq:drlt_beta}
\end{equation}
where $z_{\alpha/2}$ is the upper $(\alpha/2)^{\text{th}}$ quantile of a standard normal random variable.  
\newline
\noindent\textbf{\textsc{Drlt} for $\boldsymbol{\delta^*}$:} %In Aim (ii), we intended to develop a statistical test to determine whether or not a pooled measurement is affected by MMEs.
Given the significance level $\alpha \in [0,1]$, for each $i \in  [n]$, we reject the null hypothesis $\mathsf{H_{0,i}}:\delta^*_i = 0$ in favor of  $\mathsf{H_{1,i}} : \delta^*_i \ne 0$ when
\begin{equation}
\dfrac{|\hat{\delta}_{Ai}|}{\sigma \sqrt{1-\frac{2p}{n}
     +\frac{1}{n^2}\boldsymbol{a_{i.}A^{\top}A}\boldsymbol{a_{i.}}^{\top}}}>z_{\alpha/2}. 
\label{eq:drlt_delta}
\end{equation}

\subsection{\textsc{ODrlt} for Centered Bounded Pooling Matrices}\label{Sec:Bounded_matrix}
In this subsection, we discuss the extension of our results for the Debiased Robust \textsc{Lasso} to handle a randomly generated, `centered' pooling matrix
$\boldsymbol{A}$ with bounded values for all entries. A \emph{centered} pooling matrix is one with a mean value of 0 in every row. Let the elements of $\boldsymbol{A}$ be drawn independently and identically from a distribution with mean $0$ and variance $1$, and defined on a bounded domain, i.e. we have $a_{ij} \in [-h,h], h>0$ for all $i \in [n], j \in [p]$. Hence henceforth refer to such a matrix as a \emph{centered bounded pooling matrix}. 
\\
Consider the same measurement model as given in \eqref{eq:fm_delta}. We now obtain $\boldsymbol{\hat{\beta}_{\lambda_1}}$ and $\boldsymbol{\hat{\delta}_{\lambda_2}}$ using the Robust \textsc{Lasso} Estimator given in \eqref{eq:lasso_delta}. 
For this new model, we show in Lemma~\ref{le:EREC_bound_mat} that $\boldsymbol{A}$ satisfies the EREC. 
Hence, the upper bounds on the reconstruction error for the robust \textsc{Lasso} from Theorem~\ref{th:upper_bound_robustLasso} remain the same. 
\\
In order to establish theoretical properties of the debiased robust \textsc{Lasso} Estimators from \eqref{eq:deb_beta_est} and \eqref{eq:deb_delta_est} for the centered bounded matrix, the procedure for designing the weights matrix $\boldsymbol{W}$ changes and requires a modified set of constraints. This procedure is presented in the form of Alg.~\ref{alg:design_W_bnd}. 
We again show that $\boldsymbol{W}=\boldsymbol{A}$ is a feasible solution for the optimization problem in Alg.~\ref{alg:design_W_bnd}. In order to show this, we present 
Lemmas~\ref{le:Bnd_mat_prop} and \ref{le:W_mat_prop_bnd}, which are analogues of Lemmas \ref{le:W_mat_prop} and \ref{le:Rad_mat_prop} respectively, for the case  of centered, bounded pooling matrices.
\begin{algorithm} 
\caption{{Design of $\boldsymbol{W}$ for centered bounded matrix $\boldsymbol{A}$}}
\begin{algorithmic}[1]
{\REQUIRE 
  $\boldsymbol{A}$ with elements in $[-h,h]$, $\mu_1(h)$, $\mu_2(h)$ and $\mu_3(h)$ 
\ENSURE
$\boldsymbol{W}$ 
\STATE We solve the following optimization problem :
\begin{eqnarray}
\nonumber \underset{\boldsymbol{W}}{\textrm{minimize}}& &\quad 
\sum_{j=1}^p\boldsymbol{w_{.j}}^{\top}\boldsymbol{w_{.j}} \\ \nonumber
\textrm{subject to}& & 
\mathsf{C0} : \boldsymbol{w_{.j}}^{\top}\boldsymbol{w_{.j}}/n \leq 1+h^2\sqrt{\frac{\log p}{n}} \ \forall \ j \in [p] 
\\ & & \nonumber \mathsf{C1} : \left| \left(\boldsymbol{I_p}-\frac1n\boldsymbol{W^{\top}A}\right)\right|_\infty \leq \mu_1(h)\triangleq 2h^2 \sqrt{\frac{2 \log p}{n}} , \\ \nonumber & & \mathsf{C2} : \left|\frac{1}{p}\left(\boldsymbol{I_n}-\frac{1}{n}\boldsymbol{A}\boldsymbol{W}^{\top}\right) \boldsymbol{A}\right|_\infty \leq \mu_2(h)\triangleq 4h^3\sqrt{\frac{\log 2np}{np}}+\frac{h}{n}, \\ \nonumber & & \mathsf{C3} : \left|\left(\frac{\boldsymbol{AW^{\top}}}{p}-\boldsymbol{I_n}\right)\right|_{\infty} \leq \mu_3(h)\triangleq 2h^2 \sqrt{\frac{2 \log n}{p}},
\end{eqnarray}
\STATE If the above problem is not feasible, then set $\boldsymbol{W}=\boldsymbol{A}$.}
\end{algorithmic}
\label{alg:design_W_bnd}
\end{algorithm}
{By using Lemmas~\ref{le:Bnd_mat_prop} and \ref{le:W_mat_prop_bnd} and a similar line of argument as in earlier proofs, we establish that Theorems~\ref{th:2_3term_beta_decompose_W}, \ref{th:dist_conv_W} and \ref{th:distribution_beta_delta_opt} also apply for the centered bounded matrix $\boldsymbol{A}$. Therefore, the  \textsc{ODrlt}s from \eqref{eq:beta_ODRLT} and \eqref{eq:delta_ODRLT} also apply for the case of a centered bounded matrix $\boldsymbol{A}$. Furthermore, these  \textsc{ODrlt}s depend on $h$ only via the optimal debiasing matrix $\boldsymbol{W}$.

The key theoretical results from this subsection, i.e., Lemmas ~\ref{le:EREC_bound_mat}, \ref{le:Bnd_mat_prop} and \ref{le:W_mat_prop_bnd}, are stated and proved in Appendix~\ref{Sec:App_A4}.

As a special case of the centered bounded matrix, we now show that our results on the Debiased Robust \textsc{Lasso} hold for the commonly used random Bernoulli($\theta$) pooling matrix $\boldsymbol{B}$ with known $\theta \in (0,1)$. We start by centering the pooling matrix $\boldsymbol{B}$ by considering $2n$ rows. Consider the linear model in \eqref{eq:forward_model_gt} given as $\boldsymbol{z} = \boldsymbol{B\beta^*} + \boldsymbol{\tilde{\eta}}$, where $\boldsymbol{z} \in \mathbb{R}^{2n}$. Here, the elements of $\boldsymbol{B}$ are obtained from the following distribution: 
\begin{equation}\label{eq:Bernoulli_theta_distn}
    P(b_{ij}=1)=\theta, \quad \text{and}\  P(b_{ij}=0)=1-\theta.
\end{equation}
We now both center and scale our measurements as follows:
\begin{equation}
\forall i \in [n], \ y_i=\frac{1}{2\theta(1-\theta)}(z_i- z_{n+i}). 
\end{equation}
The elements of $\boldsymbol{y}$ correspond to measurements with a $n \times p$ matrix $\boldsymbol{A}$ where $\boldsymbol{a}_{i,.} := \frac{1}{2\theta(1-\theta)}(\boldsymbol{b}_{i,.}-\boldsymbol{b}_{n+i,.})$. Here $2\theta(1-\theta)$ is the variance of the elements of $\boldsymbol{b}_{i,.}-\boldsymbol{b}_{n+i,.}$. Without loss of generality, we consider $\theta \in (0,0.5]$ since the centering equation is $y_i=\frac{1}{2\theta(1-\theta)}(z_i- z_{n+i})$
and the same analysis follows by using $\theta'$ where $\theta' := (1-\theta)$. Therefore, the transformed linear model becomes
\begin{eqnarray}
\forall i \in [n], y_i=\boldsymbol{a_{i.}} \boldsymbol{\beta^{*}} + \eta_i \implies \boldsymbol{y} = \boldsymbol{A\beta^*} + \boldsymbol{\eta},
\label{eq:forward_model_bitflip_Btheta}
\end{eqnarray}
where $\eta_i \triangleq \frac{1}{2\theta(1-\theta)}(\tilde{\eta}_i-\tilde{\eta}_{n+i}) \sim \mathcal{N}\left(0,\frac{1}{2\theta^2(1-\theta)^2}\tilde{\sigma}^2\right)$. 
Here, $\boldsymbol{A}$ is a random matrix whose elements acquire values from $\{-1, 0, 1\}$ with the following probability mass function:
\begin{equation}\label{eq:Bernoulli_theta_pmf}
    P(a_{ij}=-1)=\theta(1-\theta), \quad  P(a_{ij}=0)=\theta^2 + (1-\theta)^2 \quad \text{and}\  P(a_{ij}=1)=\theta(1-\theta).    
    \end{equation}
We refer to such a matrix as a `centered Bernoulli($\theta$)' matrix. Since $\mathbb{E}(a_{ij})=0$ and $Var(a_{ij})=1$ with $a_{ij}$ being bounded, all the theoretical analysis mentioned earlier in this subsection follows directly for this model. Therefore the  \textsc{ODrlt}s given in \eqref{eq:beta_ODRLT} and \eqref{eq:delta_ODRLT} directly apply to the case where $\boldsymbol{A}$ is a centered Bernoulli($\theta$) matrix. Note that a matrix whose elements are i.i.d. Centered Bernoulli(0.5) is not equivalent to a matrix whose elements are i.i.d. Rademacher. In the former case,  the elements of the matrix $\boldsymbol{A}$ can be $-1,0,1$, as each row of $\boldsymbol{A}$ is obtained from the difference between any two different rows of $\boldsymbol{B}$. The probability mass function of individual elements in this case is as defined in \eqref{eq:Bernoulli_theta_pmf} with $\theta=0.5$. On the other hand, a Rademacher matrix $\boldsymbol{A}$ is obtained by subtracting a row of $\boldsymbol{B}$ \emph{specifically} from its toggled counterpart.}
%\section{Optimal Debiased Lasso Test} \label{sec:opt_deb_test}
\section{Experimental Results}\label{sec:experiments}
\noindent\textbf{Data Generation:} We now describe the method of data generation for our simulation study. We synthetically generated signals (i.e., $\boldsymbol{\beta^*}$) with $p=500$ elements in each. For the non-zero values of $\boldsymbol{\beta^*}$, 40$\%$ were drawn i.i.d. from $U(50,100)$ and the remaining 60$\%$ were drawn i.i.d. from $U(500,10^3)$, and were placed at randomly chosen indices. The elements of the matrix $\boldsymbol{A}$ were drawn from the Rademacher distribution. In order to generate \emph{effective} MMEs, sign changes were induced in an adversarial manner in randomly chosen rows of $\boldsymbol{A}$ and at column indices corresponding to the non-zero locations of $\boldsymbol{\beta^*}$. This yielded the perturbed matrix $\boldsymbol{\hat{A}}$, produced via an adversarial form of the model mismatch error (MMEs) for bit-flips which will be described in the following paragraph. 
Define the fractions $f_{sp} \triangleq s/p$, $f_{adv} \triangleq r/n$ to express signal sparsity and a fraction of the number of measurements with effective MMEs respectively. We chose the noise standard deviation $\sigma$ to be a fraction 
of the mean absolute value of the noiseless measurements, i.e., we set $\sigma \triangleq f_{\sigma}\sum_{i=1}^n |\boldsymbol{a_{i.} \beta^*}|/n$ where $0<f_{\sigma}<1$. For different simulation scenarios, different values of $s=\|\boldsymbol{\beta^*}\|_0$ (via $f_{sp}$), $r=\|\boldsymbol{\delta^*}\|_0$ (via $f_{adv}$), noise standard deviation $\sigma$ (via $f_\sigma$) and number of measurements $n$ were chosen, as will be described in the following paragraphs.
\newline
\newline
\noindent\textbf{Choice of Model Mismatch Error:}
In our work, all effective MMEs were generated in the following manner: In our convention, a bit-flipped pool (measurement as described in \eqref{eq:fm_delta}) contains exactly one bit-flip at a randomly chosen index. Suppose that the $i^{\text{th}}$ pool (measurement) contains a bit-flip.  Then exactly one of the following two can happen: (1) some $j^{\text{th}}$ sample that was intended to be in the pool (as defined in $\boldsymbol{A}$) is excluded, or (2) some $j^{\text{th}}$ sample that was not intended to be part of the pool (as defined in $\boldsymbol{A}$) is included. These two cases lead to the following changes in the $i^{\text{th}}$ row of $\boldsymbol{\hat{A}}$ (as compared to the $i^{\text{th}}$ row of $\boldsymbol{A}$), and in both cases the choice of $j \in [p]$ is random uniform: Case 1: $\hat{a}_{ij} = -1$ but $a_{ij} = 1$, Case 2: $\hat{a}_{ij} = 1$ but $a_{ij}=-1$. Note that under this scheme, the generated MMEs may not be effective. Hence MMEs were applied in an adversarial setting by inducing bit-flips only at those entries in any row of $\boldsymbol{\hat{A}}$ corresponding to indices with \emph{non-zero} values of $\boldsymbol{\beta^*}$.
\newline
\newline
\noindent\textbf{Choice of Regularization Parameters:} The regularization parameters $\lambda_1$ and $\lambda_2$ were chosen such that both $\log(\lambda_1) \in [1:0.25:7], \log(\lambda_2) \in [1:0.25:7]$, in the following manner: We first identified values of $\lambda_1$ and $\lambda_2$ from this range such that the Lilliefors test \cite{lilliefors1967kolmogorov} confirmed the Gaussian distribution for both $ \sqrt{n}\hat{\beta}_{Wj}/\left(\sqrt{ \Sigma_{\beta_{jj}}}\right)$ and $\hat{\delta}_{Wi} /\left( \sqrt{\Sigma_{\delta_{ii}}}\right)$ (see \textsc{ODrlt} as in \eqref{eq:beta_ODRLT} and \eqref{eq:delta_ODRLT}) at the 1\% significance level, for at least 70\% of $j \in [p]$ (coordinates of $\boldsymbol{\beta^*}$) and $i \in [n]$ (coordinates of $\boldsymbol{\delta^*}$). Out of these chosen values, we determined the value of $\lambda_1, \lambda_2$ that minimized the average cross-validation error over 10 folds. In each fold, 90\% of the $n$ measurements (denoted by a sub-vector $\boldsymbol{y_r}$ corresponding to sub-matrix $\boldsymbol{A_r}$) were used to obtain ($\boldsymbol{\hat{\beta}_{\lambda_1}}, \boldsymbol{\hat{\delta}_{\lambda_2}}$) via the robust \textsc{Lasso}, and the remaining 10\% of the measurements (denoted by a sub-vector $\boldsymbol{y_{cv}}$ corresponding to measurements generated by the sub-matrix $\boldsymbol{A_{cv}}$) were used to estimate the cross-validation error $\|\boldsymbol{y_{cv}}-\boldsymbol{A_{cv}} \boldsymbol{\hat{\beta}_{\lambda_1}}-\boldsymbol{I_{cv}}\boldsymbol{\hat{\delta}_{\lambda_2}}\|^2_2$. Note that $\boldsymbol{I_{cv}}$ is a sub-matrix of the identity matrix which samples only some elements of $\boldsymbol{y}$ (and hence $\boldsymbol{\hat{\delta}_{\lambda_2}}$) to yield smaller vectors $\boldsymbol{y_{cv}}$ (and $\boldsymbol{\hat{\delta}_{cv}}$). The cross-validation error is known to be an observable, data-driven proxy for the mean-squared error \cite{Zhang2014}, which justifies its choice as a method for parameter selection.
\newline
\newline
\noindent\textbf{Evaluation Measures of Hypothesis Tests:} Many different variants of the \textsc{Lasso} estimator were compared empirically against each other as will be described in subsequent subsections. Each of them were implemented using the \texttt{CVX} (SDPT3) package in \texttt{MATLAB}. 
Results for the hypothesis tests (given in \eqref{eq:drlt_beta},\eqref{eq:drlt_delta},\eqref{eq:beta_ODRLT} and \eqref{eq:delta_ODRLT}) are reported in terms of sensitivity and specificity (defined below). The significance level of these tests was chosen to be $1\%$. Consider a binary signal $\boldsymbol{\hat{b}_{\beta}}$ with $p$ elements. In our simulations, a sample at index $j$ in $\boldsymbol{\hat{\beta}_W}$ was declared to be defective if the hypothesis test $\mathsf{G_{0,j}}$ is rejected, in which case we set $\hat{b}_{\beta,j} = 1$. In all other cases, we set $\hat{b}_{\beta,j} = 0$. We declared an element to be a \textit{true defective} if $\beta^*_j \ne 0$ and $\hat{b}_{\beta,j} \ne 0$, and a \textit{false defective} if $\beta^*_j = 0$ but $\hat{b}_{\beta,j} \ne 0$. We declared it to be a \textit{false non-defective} if $\beta^*_j \neq 0$ but $\hat{b}_{\beta,j} = 0$, and a \textit{true non-defective} if $\beta^*_j = 0$ and $\hat{b}_{\beta,j} = 0$. The \textbf{sensitivity} for $\boldsymbol{\beta^*}$ is defined as ($\#$ true defectives)/($\#$ true defectives + $\#$ false non-defectives) and
\textbf{specificity} for $\boldsymbol{\beta^*}$ is defined as ($\#$ true non-defectives)/($\#$ true non-defectives + $\#$ false defectives). We reported the results of testing for the debiased tests using: (\textit{i}) $\boldsymbol{W} \triangleq \boldsymbol{A}$ corresponding to \textsc{Drlt} (see \eqref{eq:drlt_beta} and \eqref{eq:drlt_delta}), and (\textit{ii}) the optimal  $\boldsymbol{W}$ using  Alg.~\ref{alg:design_W} corresponding to \textsc{ODrlt} (see \eqref{eq:beta_ODRLT} and \eqref{eq:delta_ODRLT}).
\\
\\
\noindent\textbf{Experiment Settings:} 
We now describe the experimental parameter setups used in this section to illustrate the numerical performances. 
\begin{itemize}
\item Setup \textsf{EA}: Varying $f_{adv} \in [0.01:0.01:0.10]$ with $n = 400, f_{sp} = 0.01, f_{\sigma} = 0.1$. 
\item Setup \textsf{EB}: Varying $n\in[200:50:500]$ with $f_{adv} = 0.01, f_{sp} = 0.01, f_{\sigma} = 0.1$. 
\item Setup \textsf{EC}: Varying $f_{\sigma} \in [0: 0.05: 0.5]$ with $n = 400, f_{adv} = 0.01, f_{sp} = 0.01$. 
\item Setup \textsf{ED}: Varying $f_{sp} \in [0.01:0.01:0.10]$ with $n = 400, f_{adv} = 0.01, f_{\sigma} = 0.1$. 
\end{itemize}
The experiments were run 100 times across different noise instances in $\boldsymbol{\eta}$, for the same signal $\boldsymbol{\beta^*}$ (in \textsf{EA}, \textsf{EB} and \textsf{EC}) and sensing matrix $\boldsymbol{A}$ (in \textsf{EA}, \textsf{EC} and \textsf{ED}). In \textsf{ED}, the sparsity of the signal varies; therefore the signal vector $\boldsymbol{\beta^*}$ also varies. Similarly, in \textsf{EB} as $n$ varies, the sensing matrix $\boldsymbol{A}$ also varies. 
\\
\noindent\textbf{Implementation of Alg.~\ref{alg:design_W}:} We implemented Alg.~\ref{alg:design_W} using the \texttt{CVX} package (SDPT3) in \texttt{MATLAB}. The time complexity of the method depends on various algorithmic and implementation details. Instead, we report execution times: computing a $\boldsymbol{W}$ matrix of size $300 \times 500$ required only about 30 minutes on a standard desktop machine (these timings can be further reduced with faster solvers). A detailed 
study of algorithmic complexity and optimization is left for future work. However, we have worked on a closed-form formula for $\boldsymbol{W}$ for a slightly different problem in another separate piece of work \cite{banerjee2025fast}. Furthermore, we note that for a fixed pooling matrix $\boldsymbol{A}$, the weights matrix $\boldsymbol{W}$ has to be obtained only once. It does not need to be obtained afresh for every execution of the robust \textsc{Lasso}. A \texttt{MATLAB} implementation of the algorithms in this paper can be found at \url{https://github.com/Shuvayan21/DRLT-for-MMEs}.

\subsection{Total variance of debiased robust \textsc{Lasso} estimators}\label{subsec:var_comp}
{In this subsection, we show that the variance of the debiased \textsc{Lasso} estimator $\boldsymbol{\hat{\beta}_W}$ is significantly smaller than that of $\boldsymbol{\hat{\beta}_A}$. We also compare the empirical variance of the debiased \textsc{Lasso} estimators from \eqref{eq:deb_beta_est} and \eqref{eq:deb_delta_est} 
 with their respective asymptotic variance, for both cases: where $\boldsymbol{W}=\boldsymbol{A}$, and where the optimal $\boldsymbol{W}$ obtained using Alg.\ref{alg:design_W}.
Note that the total variance (TV) of a random vector is defined as the sum of the variances of its components. By using \eqref{eq:Sig_beta} and \eqref{eq:Sig_delta}, the asymptotic total variance (ATV) of $\boldsymbol{\hat{\beta}_W}$ and $\boldsymbol{\hat{\delta}_W}$ are $\frac{1}{n}\sum_{j=1}^p \Sigma_{\boldsymbol{\beta}_{jj}}$ and $\sum_{i=1}^n \Sigma_{\boldsymbol{\delta}_{ii}}$, respectively. Similarly, using \eqref{eq:beta_marginal_dist_A} and \eqref{eq:delta_marginal_dist_A},  the asymptotic total variance of $\boldsymbol{\hat{\beta}_A}$ and $\boldsymbol{\hat{\delta}_A}$ are $\frac{p}{n} \sigma^2$ and $\sigma^2\sum_{i=1}^n\left(1-\frac{2p}{n}
     +\frac{1}{n^2}\boldsymbol{a_{i.}A^{\top}A}\boldsymbol{a_{i.}}^{\top}\right)$, respectively.
The empirical total variance (ETV) of the debiased \textsc{Lasso} estimators is obtained using $100$ simulation runs over different instances of $\boldsymbol{\eta}$ with varying $n\in[100:100:500]$, $f_{\sigma}=0.01$, $p = 500$, $s = 5$ and $r = 4$ where the signal $\boldsymbol{\beta^*}$ was generated in the same manner as described in the beginning of this section. In these simulations, the elements of $\boldsymbol{A}$ were generated from the Rademacher distribution. The ratio between empirical total variances of $\boldsymbol{\hat{\beta}_W}$ and $\boldsymbol{\hat{\beta}_A}$ as well as the ratio between asymptotic total variances of $\boldsymbol{\hat{\beta}_W}$ and $\boldsymbol{\hat{\beta}_A}$ are shown in second and third column of Table~\ref{tab:mean_sigma_final_A}, respectively. For larger $n$, it indicates that the total variance of $\boldsymbol{\hat{\beta}_W}$ is around $2/3$ that of $\boldsymbol{\hat{\beta}_A}$. This illustrates the superiority of the debiasing matrix $\boldsymbol{W}$ constructed in Alg.~\ref{alg:design_W}. Interestingly, we observe a similar phenomenon for the performance of $\boldsymbol{\hat{\delta}_W}$ and $\boldsymbol{\hat{\delta}_A}$ from Table~\ref{tab:mean_sigma_final_A} (fourth and fifth column). Table \ref{tab:mean_sigma_final_B} compares the ratio of the empirical total variance to the asymptotic total variance of $\boldsymbol{\hat{\beta}_W}$, $\boldsymbol{\hat{\beta}_A}$, $\boldsymbol{\hat{\delta}_W}$ and $\boldsymbol{\hat{\delta}_A}$ in the columns~2,~3,~4 and~5, respectively. In most regimes, the ratio is close to $1$, which indicates that the theoretical asymptotic variances of the debiased robust \textsc{Lasso} estimates are close to their empirical counterparts in a large regime.}
\begin{table}[ht]
{\centering
\begin{minipage}{0.48\textwidth}
\centering
\begin{tabular}{|c|c|c|c|c|}
\hline
\rowcolor{Gray}
$n$ & $\frac{ETV(\boldsymbol{\hat{\beta}_W})}{ETV(\boldsymbol{\hat{\beta}_A})}$ & $\frac{ATV(\boldsymbol{\hat{\beta}_W})}{ATV(\boldsymbol{\hat{\beta}_A})}$ & $\frac{ETV(\boldsymbol{\hat{\delta}_W})}{ETV(\boldsymbol{\hat{\delta}_A})}$ & $\frac{ATV(\boldsymbol{\hat{\delta}_W})}{ATV(\boldsymbol{\hat{\delta}_A})}$ \\
\hline
100 & 0.564 & 0.729 & 0.626 & 0.678 \\
200 & 0.628 & 0.709 & 0.534 & 0.624 \\
300 & 0.652 & 0.698 & 0.678 & 0.629 \\
400 & 0.639 & 0.690 & 0.728 & 0.684 \\
500 & 0.628 & 0.687 & 0.727 & 0.719 \\
\hline
\end{tabular}
\caption{Ratio of the empirical total variances of $\boldsymbol{\hat{\beta}_W}$ to those of $\boldsymbol{\hat{\beta}_A}$ (second column); ratio of asymptotic total variances of $\boldsymbol{\hat{\beta}_W}$ to those of $\boldsymbol{\hat{\beta}_A}$ (third column); the same ratios for $\boldsymbol{\hat{\delta}_W}$ and $\boldsymbol{\hat{\delta}_A}$ (fourth and fifth columns respectively). %The empirical total variances are obtained over 100 different instances of $\boldsymbol{\eta}$ with $f_{\sigma} = 0.01$, $p = 500$, $s = 5$ and $r = 4$.
}

\label{tab:mean_sigma_final_A}
\end{minipage}
\hfill
\begin{minipage}{0.48\textwidth}
\centering
\begin{tabular}{|c|c|c|c|c|}
\hline
\rowcolor{Gray}
$n$ & $\frac{ETV(\boldsymbol{\hat{\beta}_W})}{ATV(\boldsymbol{\hat{\beta}_W})}$ & $\frac{ETV(\boldsymbol{\hat{\beta}_A})}{ATV(\boldsymbol{\hat{\beta}_A})}$ & $\frac{ETV(\boldsymbol{\hat{\delta}_W})}{ATV(\boldsymbol{\hat{\delta}_W})}$ & $\frac{ETV(\boldsymbol{\hat{\delta}_A})}{ATV(\boldsymbol{\hat{\delta}_A})}$ \\
\hline
100 & 1.221 & 1.576 & 1.791 & 1.937 \\
200 & 1.061 & 1.197 & 1.462 & 1.711 \\
300 & 1.033 & 1.106 & 1.365 & 1.265 \\
400 & 1.02 & 1.101 & 1.105 & 1.039 \\
500 & 1.007 & 1.102 & 1.047 & 1.036 \\
\hline
\end{tabular}
\caption{ Ratio of empirical to asymptotic total variances of $\boldsymbol{\hat{\beta}_W}$ (second column); the same ratios for $\boldsymbol{\hat{\beta}_A}$, $\boldsymbol{\hat{\delta}_W}$ and $\boldsymbol{\hat{\delta}_A}$ (third, fourth and fifth columns respectively). %The empirical total variances are obtained over 100 different instances of $\boldsymbol{\eta}$ with $f_{\sigma} = 0.01$, $p = 500$, $s = 5$ and $r = 4$.
}
\label{tab:mean_sigma_final_B}
\end{minipage}}
\end{table}
\subsection{Results with Baseline Debiasing Techniques in the Presence of Effective MMEs}\label{subsec:Debiasing_baselines}
We now describe the results of an experiment to show the impact of 
\textsc{ODrlt} \eqref{eq:beta_ODRLT}  in the presence of effective MMEs induced in $\boldsymbol{A}$. We compare \textsc{ODrlt} with the baseline hypothesis test for $\boldsymbol{\beta^*}$ as defined by \cite{Javanmard2014}, which is equivalent to ignoring MMEs (i.e., setting $\boldsymbol{\delta^*}= \boldsymbol{0}$ in \eqref{eq:fm_delta}). Considering the presence of effective MMEs, we further compare \textsc{ODrlt} with the baseline test defined in \cite{Javanmard2014} which would use the approximate inverse of the augmented sensing matrix $(\boldsymbol{A} | \boldsymbol{I_n})$ (as would be obtained from \eqref{eq:design_m_jav}) with $\mu=2\sqrt{\frac{\log(n+p)}{n}}$.  
We now describe these two chosen baseline hypothesis tests for $\boldsymbol{\beta^*}$ in more detail.
\begin{enumerate}
\item \textbf{Baseline ignoring MMEs:} (\textsc{Baseline-1}) This approach computes the following `debiased' estimate of $\boldsymbol{\beta^*}$ as given in Equation (5) of \cite{Javanmard2014}: 
\begin{equation}
\boldsymbol{\hat{\beta}_b} \triangleq \boldsymbol{\hat{\beta}_{\lambda,b}} + \frac{1}{n}\boldsymbol{MA}^{\top}(\boldsymbol{y}-\boldsymbol{A \hat{\beta}_{\lambda,b}}),
\end{equation}
where  $\boldsymbol{\hat{\beta}_{\lambda,b}} \triangleq \text{argmin}_{\boldsymbol{\beta}} \|\boldsymbol{y}-\boldsymbol{A\beta}\|^2_2 + \lambda \|\boldsymbol{\beta}\|_1$, and $\boldsymbol{M}$ is the approximate inverse of $\boldsymbol{A}$ obtained from \eqref{eq:design_m_jav}. 
In this baseline approach, we reject the null hypothesis $\mathsf{G_{0,j}}: \beta^*_j=0$ in favor of $\mathsf{G_{1,j}}: \beta^*_j \neq 0$, for each
$j \in [p]$ when $\sqrt{n}\boldsymbol{\hat{\beta}_{bj}}/\sqrt{\sigma^2[\boldsymbol{MA^{\top}A M^{\top}}]_{jj}/n} > z_{\alpha/2}$.
\item \textbf{Baseline considering MMEs:} (\textsc{Baseline-2}) In this approach, we consider the MMEs which is equivalent to the sensing matrix as $(\boldsymbol{A}|\boldsymbol{I_n})$ and signal vector $\boldsymbol{x^*}=\left(\boldsymbol{\beta^*}^\top,\boldsymbol{\delta^*}^\top\right)^\top$. The `debiased' estimate of $\boldsymbol{x^*}$ in this approach is given as:
\begin{equation}\boldsymbol{\tilde{x}_b} \triangleq \boldsymbol{\tilde{x}_{\lambda}} + \frac{1}{n}\boldsymbol{\tilde{M}} (\boldsymbol{A}|\boldsymbol{I_n})^{\top}(\boldsymbol{y}-(\boldsymbol{A}|\boldsymbol{I_n})\boldsymbol{\tilde{x}_{\lambda}} ),
\end{equation}
where $\boldsymbol{\tilde{x}_{\lambda}} \triangleq \text{argmin}_{\boldsymbol{\beta}} \|\boldsymbol{y}-(\boldsymbol{A}|\boldsymbol{I_n})\boldsymbol{x}\|^2_2 + \lambda \|\boldsymbol{x}\|_1$ and $\boldsymbol{\tilde{M}}$ is the approximate inverse of $(\boldsymbol{A}|\boldsymbol{I_n})$ obtained using \eqref{eq:design_m_jav}.
Then $\boldsymbol{\tilde{\beta}_{b}}$ is obtained by extracting the first $p$ elements of $\boldsymbol{\tilde{x}_{b}}$. In this approach, we reject the null hypothesis $\mathsf{G_{0,j}}: \beta^*_j=0$ in favor of $\mathsf{G_{1,j}}: \beta^*_j \neq 0$, for each
$j \in [p]$ when $\sqrt{n}\boldsymbol{\tilde{\beta}_{bj}}/\sqrt{\sigma^2[\boldsymbol{\tilde{M} (A|I_n)^{\top}(A|I_n) \tilde{M}^{\top}}]_{jj}/n} > z_{\alpha/2}$.
\end{enumerate}
Note that the theoretical results established in \cite{Javanmard2014} hold for completely random or purely deterministic sensing matrices, whereas the sensing matrix corresponding to the MME model, i.e., $(\boldsymbol{A} | \boldsymbol{I_n})$,  is partly random and partly deterministic. Nonetheless, the second baseline test, i.e. \textsc{Baseline-2} with the augmented matrix, is useful as a numerical benchmark. 
For both baseline approaches, the regularization parameter $\lambda$ was chosen using cross validation. We chose the $\lambda$ value which minimized the validation error with $90\%$ of the measurements used for reconstruction and the remaining $10\%$ used for cross-validation. 
\begin{table}
\centering
    \begin{tabular}{|c|c|c|c|c|c|c|}
    \hline
        \rowcolor{Gray}
        &\multicolumn{3}{c|}{Sensitivity } &\multicolumn{3}{c|}{Specificity } \\
        \hline
        \rowcolor{Gray}
        $n$ & \textsc{Baseline-1} &\textsc{Baseline-2} & \textsc{ODrlt} & \textsc{Baseline-1} &  \textsc{Baseline-2} & \textsc{ODrlt}\\
    \hline
      100 & 0.522 & 0.602 & 0.647 & 0.678 & 0.703 & 0.771 \\
       \hline 
      200 & 0.597 & 0.682 & 0.704 & 0.832 & 0.895 & 0.931 \\
       \hline
        300 & 0.698 & 0.803 & 0.879 & 0.884 & 0.915 & 0.963 \\
       \hline
       400 & 0.791 & 0.835 & 0.951 & 0.902 & 0.927 & 0.999 \\
       \hline
       500 & 0.858 & 0.894 & 0.984 & 0.923 & 0.956 & 1 \\
       \hline 
    \end{tabular}
    \\
    \caption{Comparison of average Sensitivity (Sen.) and Specificity (Spec.), each based on 100 independent noise runs (keeping $\boldsymbol{\beta^*}$, $\boldsymbol{A}$ and $\boldsymbol{\delta^*}$ fixed), for the tests \textsc{Baseline-1}, \textsc{Baseline-2} and \textsc{ODrlt} for determining defectives in $\boldsymbol{\beta^*}$ from their respective debiased estimates in the presence of MMEs induced in $\boldsymbol{{A}}$  (See Sec.~\ref{subsec:Debiasing_baselines} for detailed definitions). }
    \label{tb:prop_deb}
\end{table}
In Table ~\ref{tb:prop_deb}, we compare 
the average values (over $100$ instances of measurement noise $\boldsymbol{\eta}$ (keeping $\boldsymbol{\beta^*}$, $\boldsymbol{A}$ and $\boldsymbol{\delta^*}$ fixed) of Sensitivity and Specificity of \textsc{Baseline-1}, \textsc{Baseline-2} and \textsc{ODrlt} for different values of $n$ varying in $\{100,200,300,400,500\}$ and $p=500$.
It is clear from Table~\ref{tb:prop_deb},
that for all the values of $n$, the Sensitivity and Specificity value of \textsc{ODrlt} is higher as compared to that of \textsc{Baseline-1} and \textsc{Baseline-2}. The performance of \textsc{Baseline-2} dominates \textsc{Baseline-1} which indicates that ignoring MMEs may lead to misleading inferences in small sample scenarios. Furthermore, the Sensitivity and Specificity of \textsc{ODrlt} approaches 1 as $n$ increases. This highlights the superiority of our proposed technique and its associated hypothesis tests over two carefully chosen baselines. Note that there is no prior literature on debiasing in the presence of MMEs, and hence these two baselines are the only possible competitors for our technique. 

 \subsection{Empirical verification of asymptotic results of Theorem~\ref{th:distribution_beta_delta_opt}}
\label{subsec:distr_test_stats}
In this subsection, we compare the empirical distribution of $T_{G,j} \triangleq \sqrt{n}(\hat{\beta}_{Wj}-\beta^*_j)/\sqrt{[ \boldsymbol{\Sigma_{\beta}}]_{jj}}$ and $T_{H,i} \triangleq (\hat{\delta}_{Wi}-\delta^*_i) /\sqrt{[\boldsymbol{\Sigma_{\delta}}]_{{ii}}}$, for the optimal weight matrix $\boldsymbol{W}$, with its asymptotic distribution $\mathcal{N}(0,1)$ as derived in Theorem~\ref{th:distribution_beta_delta_opt}.
We chose $p = 500, n=400$, $f_{adv}=0.01$, $f_{sp}=0.01$ and $f_{\sigma} = 0.01$. The measurement vector $\boldsymbol{y}$ was generated with a perturbed matrix $\boldsymbol{\hat{A}}$ containing effective MMEs using the procedure  described earlier.  
Here, $T_{G,j}$ and $T_{H,i}$ were computed for $100$ runs across different noise instances in $\boldsymbol{\eta}$.

\begin{figure}
\centering
    \includegraphics[scale=0.26]{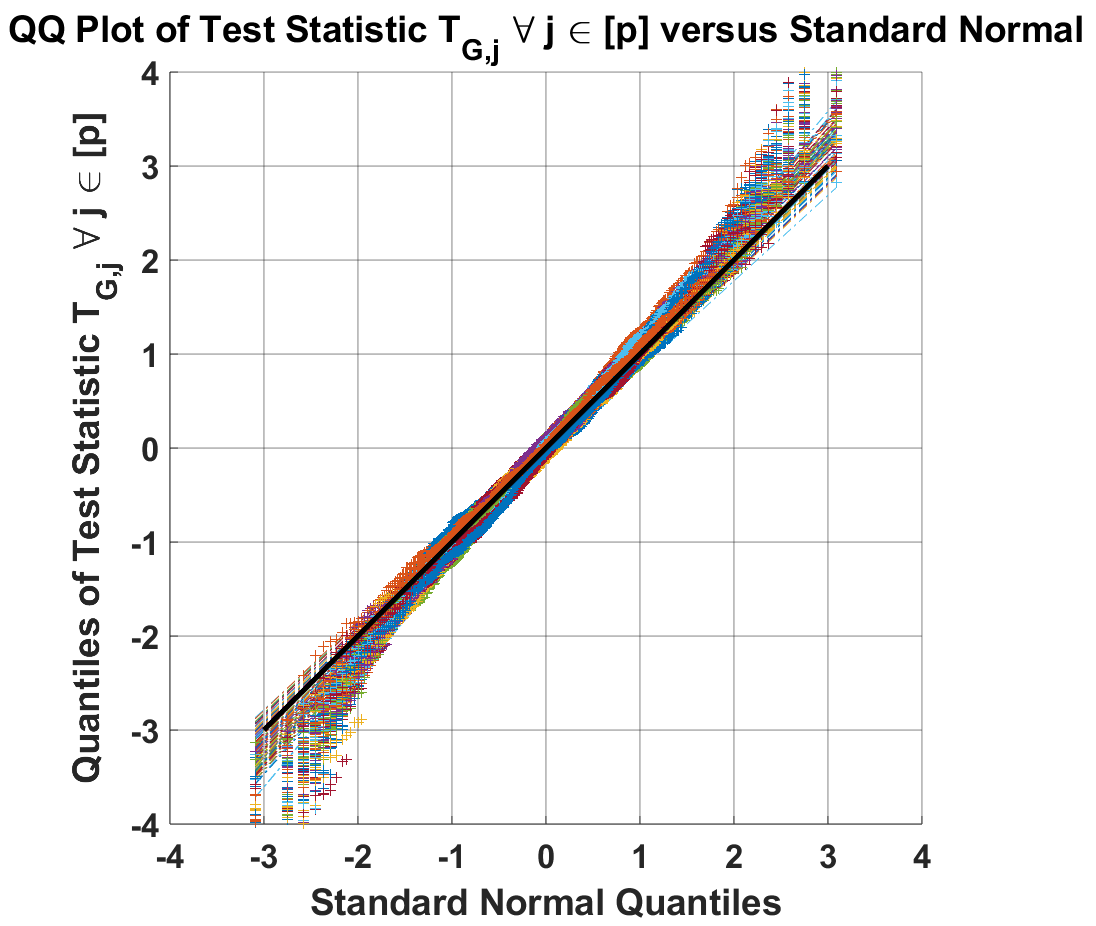}
    \includegraphics[scale=0.26]{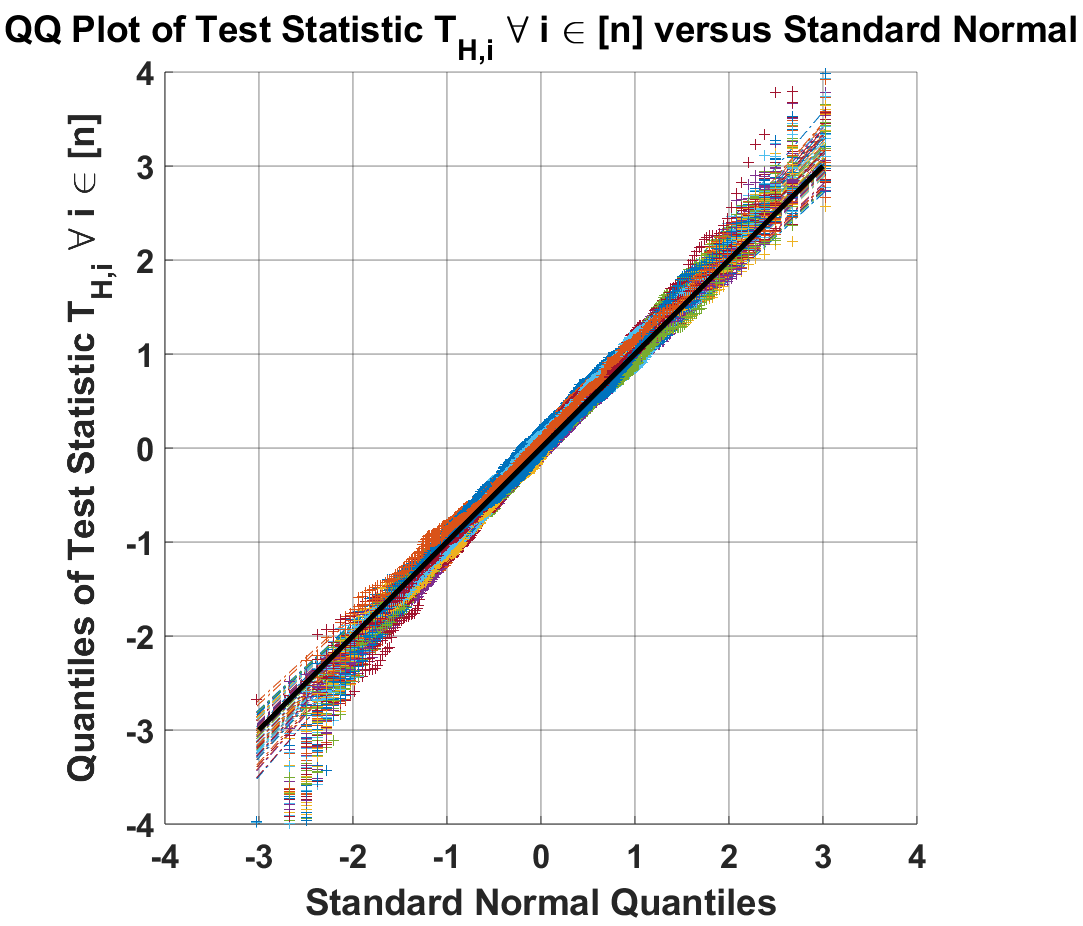}
    \caption{Left: Quantile-Quantile plots of $\mathcal{N}(0,1)$ vs. $T_{G,j}$ (defined in the beginning of Sec.~\ref{subsec:distr_test_stats}) using 100 independent noise runs for all $j \in [p]$ (one plot per index $j$ with different colors). Right: Quantile-Quantile plots of $\mathcal{N}(0,1)$ vs. $T_{H,i}$ (defined in the beginning of Sec.~\ref{subsec:distr_test_stats}) using 100 independent noise runs for all $i \in [n]$ (one plot per index $i$ with different colors). For both plots, the pooling matrix contained effective MMEs.}
    \label{fig:qqplot_test_stat}   
\end{figure}

The left sub-figure of Fig.~\ref{fig:qqplot_test_stat} shows plots of the quantiles of a standard normal random variable versus the quantiles of $T_{G,j}$ computed over $100$ runs for each $j\in[p]$. For the quantiles, each plot is presented in a different color. A $45^{\circ}$ straight line passing through the origin is also plotted (black solid line) as a reference. These $p$ different quantile-quantile (QQ) plots corresponding to $j\in[p]$, all super-imposed on one another, indicate that the quantiles of the $\{T_{G,j}\}_{j=1}^p$ are close to that of a standard normal distribution in the range of $[-2,2]$ (thus covering $95\%$ range of the area under the standard bell curve) for defective as well as non-defective samples. This confirms that the distribution of the $T_{G,j}$ values is each approximately $\mathcal{N}(0,1)$, even in this chosen finite sample scenario. Similarly, the right sub-figure of Fig.~\ref{fig:qqplot_test_stat} shows the QQ-plot corresponding to $T_{H,i}$
for each $i\in[n]$ in different colors. As before, these $n$ different QQ-plots, one for each $i \in [n]$, all super-imposed on one another, indicate that the $\{T_{H,i}\}_{i=1}^n$ values are also each approximately standard normal, with or without MMEs. 
\subsection{Sensitivity and Specificity of \textsc{ODrlt} and \textsc{Drlt} for \texorpdfstring{$\boldsymbol{\delta^*}$}{delta*}}\label{sec:sensitivity_delta}

\begin{figure*}
\centering
    \includegraphics[scale=0.195]{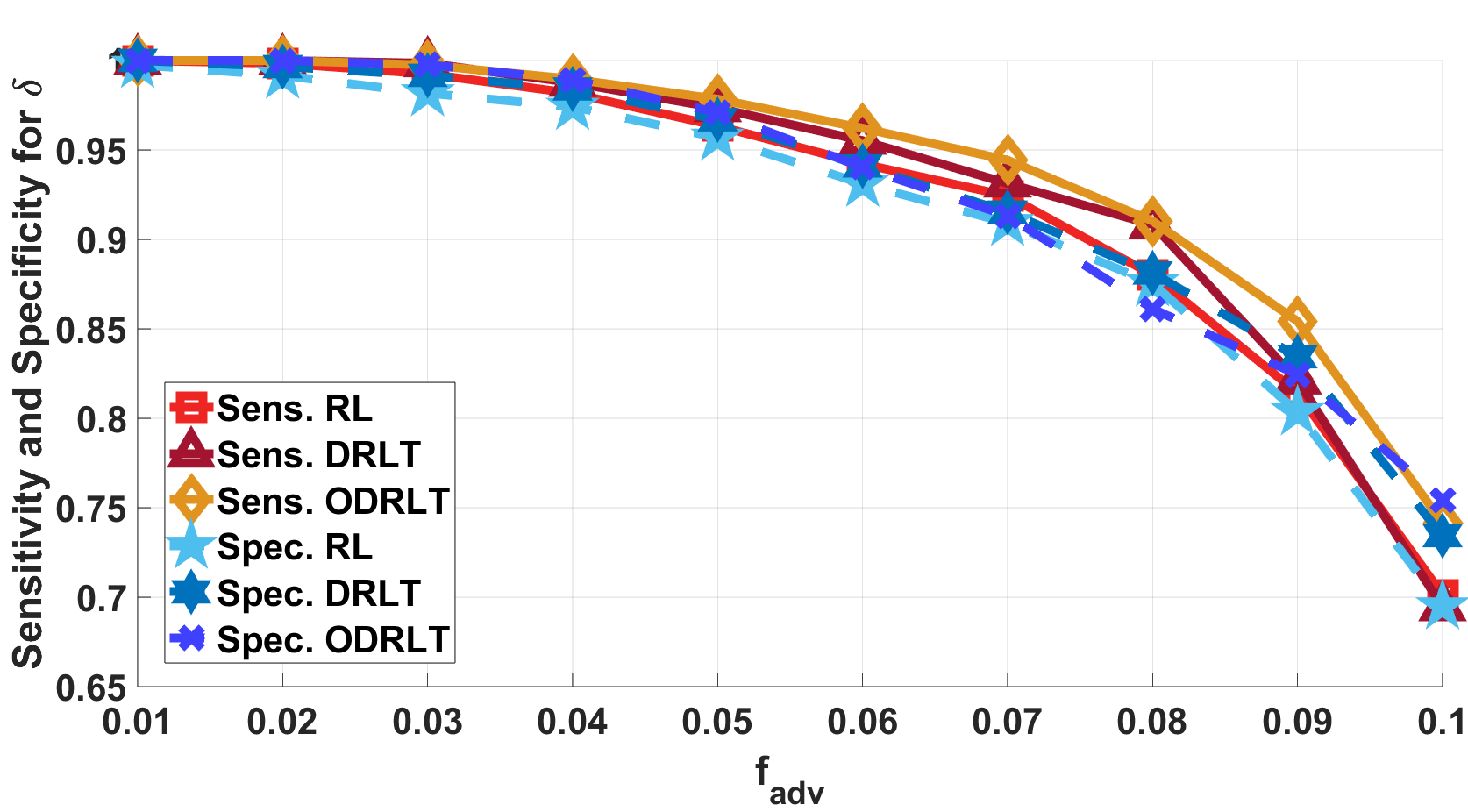}      
    \includegraphics[scale=0.195]{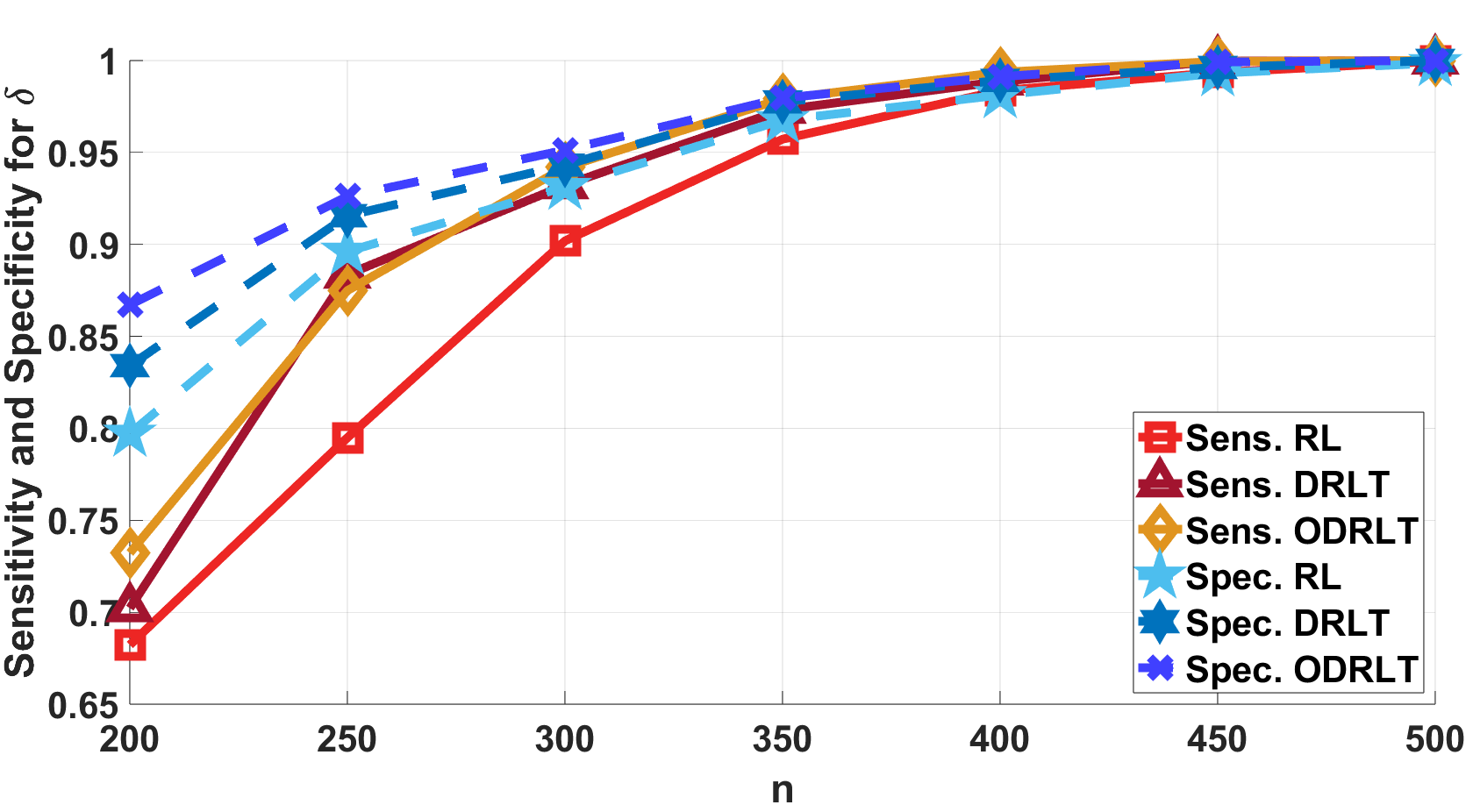}\\
    \includegraphics[scale=0.195]{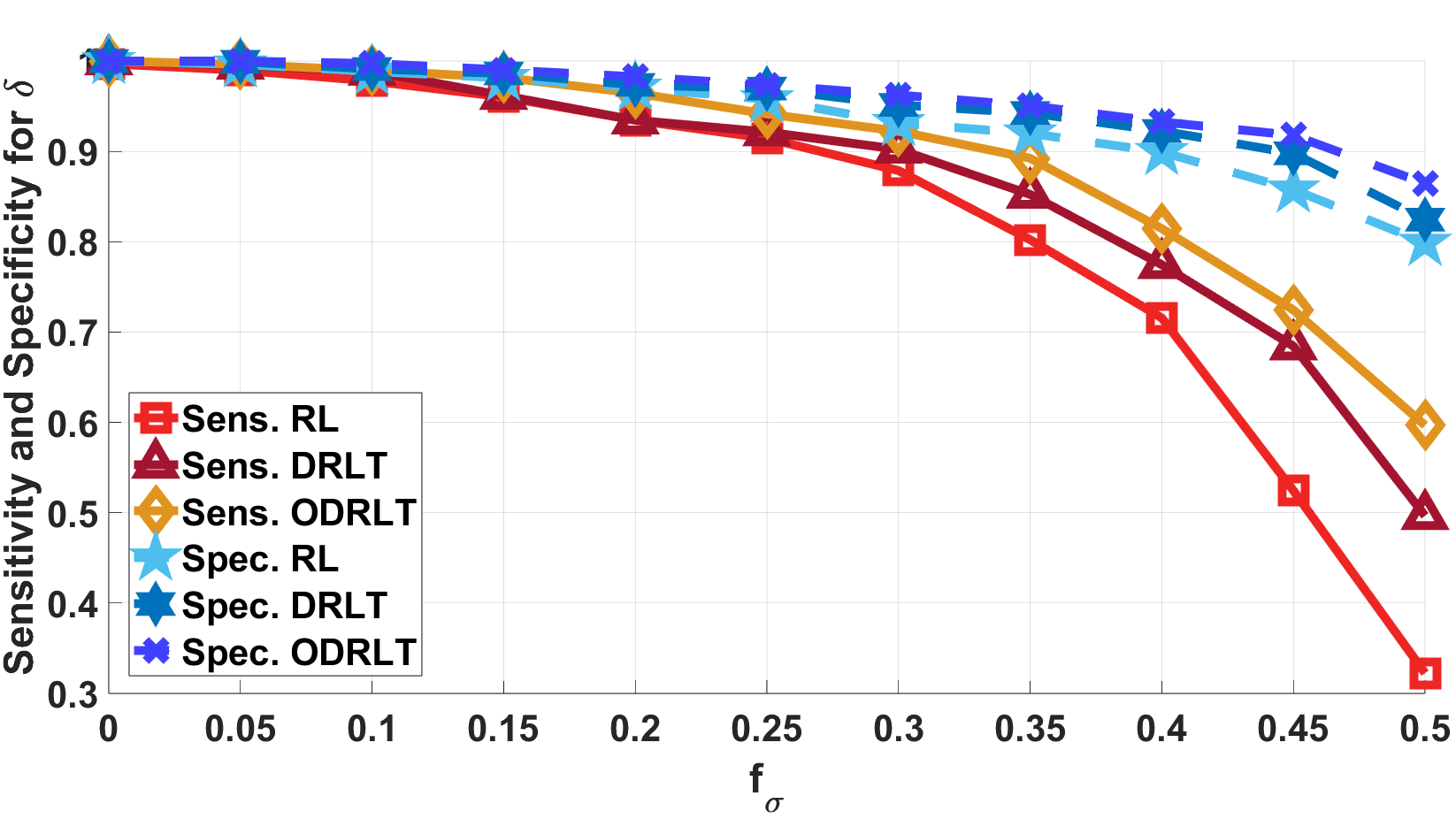}      
    \includegraphics[scale=0.195]{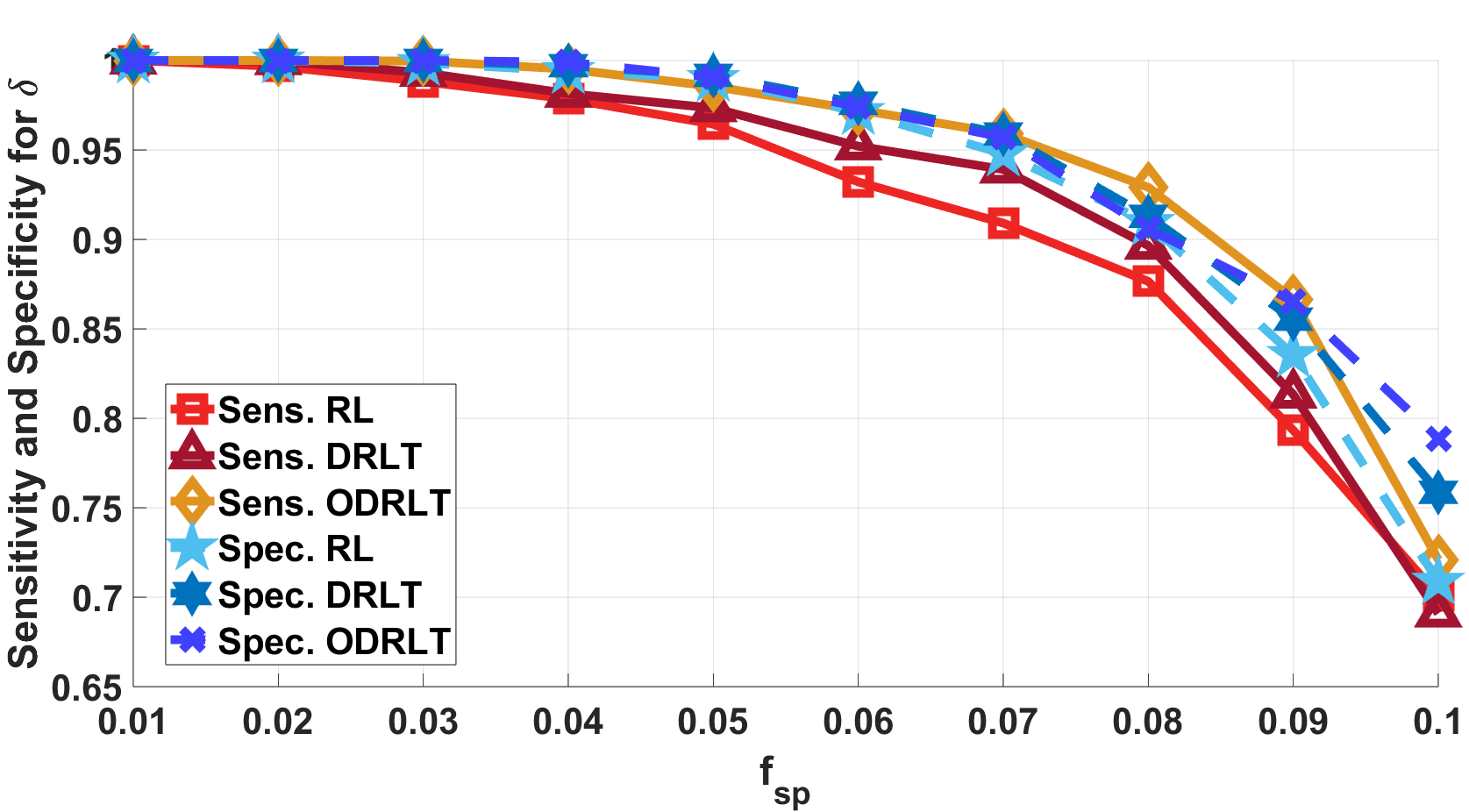}
    \caption{Average Sensitivity and Specificity plots (over 100 independent noise runs keeping $\boldsymbol{\beta^*}$, $\boldsymbol{\delta^*}$ and $\boldsymbol{A}$ fixed) for detecting measurements containing MMEs (i.e. detecting non-zero values of $\boldsymbol{\delta^*}$) using \textsc{Drlt}, \textsc{ODrlt} and Robust \textsc{Lasso} (\textsc{Rl}). The experimental parameters are $p = 500, f_{\sigma}=0.1, f_{adv}=0.01,f_{sp}=0.1,n = 400$. Left to right, top to bottom: results for experiments \textsf{EA}, \textsf{EB}, \textsf{EC}, \textsf{ED} (see ``Experiment Settings'' in the beginning of this section, for details).}   
    \label{fig:sens_spec_E1234}    
\end{figure*}

The empirical sensitivity and specificity of a test was computed as follows. The estimate $\boldsymbol{\hat{\delta}_W}$ was binarized to create a vector $\boldsymbol{\hat{b}_{W,\delta}}$ such that for all $i \in [n]$, the value of $\hat{b}_{W,\delta}(i)$ was set to 1 if \textsc{Drlt} or \textsc{ODrlt} rejected the hypothesis $\mathsf{H_{0,i}}$, and $\hat{b}_{W,\delta}(i)$ was set to 0 otherwise. Likewise, a ground truth binary vector $\boldsymbol{b_{\delta}^*}$ was created which satisfied $b^*_{\delta}(i) = 1$ at all locations $i$ where $\delta^*_i \ne 0$ and $b^*_{\delta}(i) = 0$ otherwise. Sensitivity and specificity values were computed by comparing corresponding entries of $\boldsymbol{b^*_{\delta}}$ and $\boldsymbol{\hat{b}_{W,\delta}}$.  The sensitivity of \textsc{Drlt} and \textsc{ODrlt} test for $\boldsymbol{\delta^*}$ averaged over 100 runs of different $\boldsymbol{\eta}$ instances is reported in Fig.~\ref{fig:sens_spec_E1234} for the different experimental settings \textsf{EA}, \textsf{EB}, \textsf{EC}, \textsf{ED}. Under setup \textsf{EB}, the sensitivity plot indicates that the sensitivity of \textsc{Drlt} and \textsc{ODrlt} increases as $n$ increases. Under setups \textsf{EA}, \textsf{EC}, and \textsf{ED}, the sensitivity of both \textsc{Drlt} and \textsc{ODrlt} is reasonable even with larger values of $f_{adv}$, $f_\sigma$, and $f_{sp}$ (which are difficult regimes). In Fig.~\ref{fig:sens_spec_E1234}, we compare the sensitivity of \textsc{Drlt} and \textsc{ODrlt} to that of Robust \textsc{Lasso} from \eqref{eq:lasso_delta} without any debiasing step, which is abbreviated as \textsc{Rl}. To determine defectives and non-defectives for the \textsc{Rl} method, we adopted a thresholding strategy where an estimated element was considered defective (resp. non-defective) if its value was greater than or equal to (resp. less than)
a threshold $\tau_{ss}$. The optimal value of $\tau_{ss}$ was chosen clairvoyantly (i.e., assuming knowledge of the ground truth signal vector $\boldsymbol{\beta^*}$) on a training set so as to maximize Youden's index defined as $\text{Sensitivity} + \text{Specificity} - 1$. Furthermore, Fig.~\ref{fig:sens_spec_E1234} indicates that the sensitivity of \textsc{ODrlt} is superior to that of \textsc{Rl} and \textsc{Drlt} with \textsc{Drlt} also slightly better than \textsc{Rl}. Note that, in practice, a choice of the threshold $\tau_{ss}$ for \textsc{Rl} would be challenging and require a \emph{representative} training set, whereas \textsc{Drlt} and \textsc{ODrlt} do not require any training set for the choice of such a threshold.

\subsection{Identification of Defective Samples in \texorpdfstring{$\boldsymbol{\beta^*}$}{beta*}}
\label{subsec:exp_beta_hypothesis}

In the next set of experimental results, we first examined the effectiveness of \textsc{Drlt} and \textsc{ODrlt} to detect defective samples in $\boldsymbol{\beta^*}
$ in the presence of bit-flips in $\boldsymbol{A}$ induced as per adversarial MMEs. We compared the performance of \textsc{Drlt} and \textsc{ODrlt} to two other closely related algorithms to enable performance calibration:
(1) Robust \textsc{Lasso} (\textsc{Rl}) from \eqref{eq:lasso_delta} without debiasing; (2) A hypothesis testing mechanism on a pooling matrix without model mismatch, which we refer to as \textsc{Baseline-3}. In \textsc{Baseline-3}, we generated measurements with the \emph{correct} pooling matrix $\boldsymbol{A}$ (i.e., $\boldsymbol{\delta^*}=\boldsymbol{0}$) and obtained a debiased \textsc{Lasso} estimate as given by \eqref{eq:basic_deb}. 
 (Note that \textsc{Baseline-3} is very different from \textsc{Baseline-1} and \textsc{Baseline-2} from Sec.~\ref{subsec:Debiasing_baselines} as in this approach $\boldsymbol{\delta^*}=\boldsymbol{0}$.) Using this debiased estimate, we obtained a hypothesis test similar to Equation (5) of \cite{Javanmard2014}. In the case of \textsc{Rl}, the decision regarding whether a sample is defective or not was taken based on a threshold $\tau_{ss}$ that was chosen to maximize the Youden's index on a \emph{training set} of signals from the \emph{same} distribution. The regularization parameters $\lambda_1,\lambda_2$ were chosen separately for every choice of parameters $f_{adv}, f_{\sigma}, f_{sp}$ and $n$.

%We examined the variation in sensitivity and specificity with regard to change in the following parameters, keeping all other parameters fixed: (\textsf{EA}) the number of bit-flips in the matrix $\boldsymbol{A}$ expressed as a fraction $f_{adv} \in [0,1]$ of $n$; (\textsf{EB}) number of pools $n$; (\textsf{EC}) noise standard deviation $\sigma$ expressed as a fraction $f_{\sigma} \in [0,1]$ of the quantity $\bar{y}$ defined in Sec.~\ref{subsec:distr_test_stats}; (\textsf{ED}) sparsity ($\ell_0$ norm) $s$ of vector $\boldsymbol{\beta^*}$ expressed as fraction $f_{sp} \in [0,1]$ of signal dimension $p$. For the bit-flips experiment i.e., (\textsf{EA}), $f_{adv}$ was varied in $\{0.01,0.02,\ldots,0.1\}$ with $n=400,f_{sp}=0.01,f_{\sigma}=0.1$. For the measurements experiment (\textsf{EB}), $n$ was varied over $\{200,150,\ldots,500\}$ with $f_{sp} = 0.01 , f_{adv} = 0.01, f_{\sigma} = 0.1$. For the noise experiment (i.e., (\textsf{EC}), we varied $f_{\sigma}$ in $\{0,0.05,\ldots,0.5\}$ with $n=400,f_{sp}=0.01,f_{adv}=0.01$. For the sparsity experiment (i.e., (\textsf{ED}), $f_{sp}$ was varied in $\{0.01,0.02,\ldots,0.1\}$ with $n = 400, f_{adv} = 0.01, f_{\sigma} = 0.1$.  

\begin{figure*}
   \centering
    \includegraphics[scale=0.195]{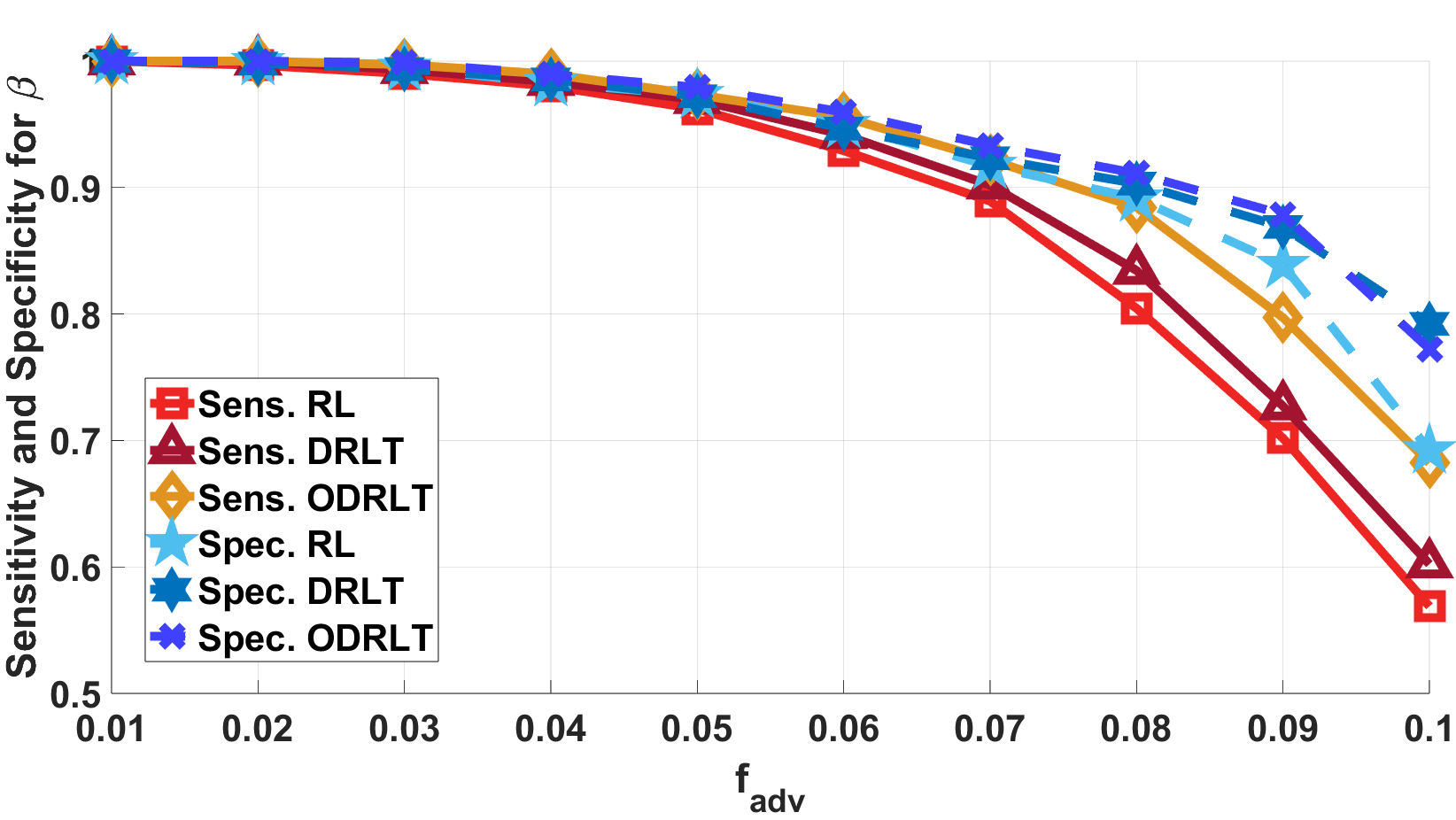}
    \includegraphics[scale=0.195]{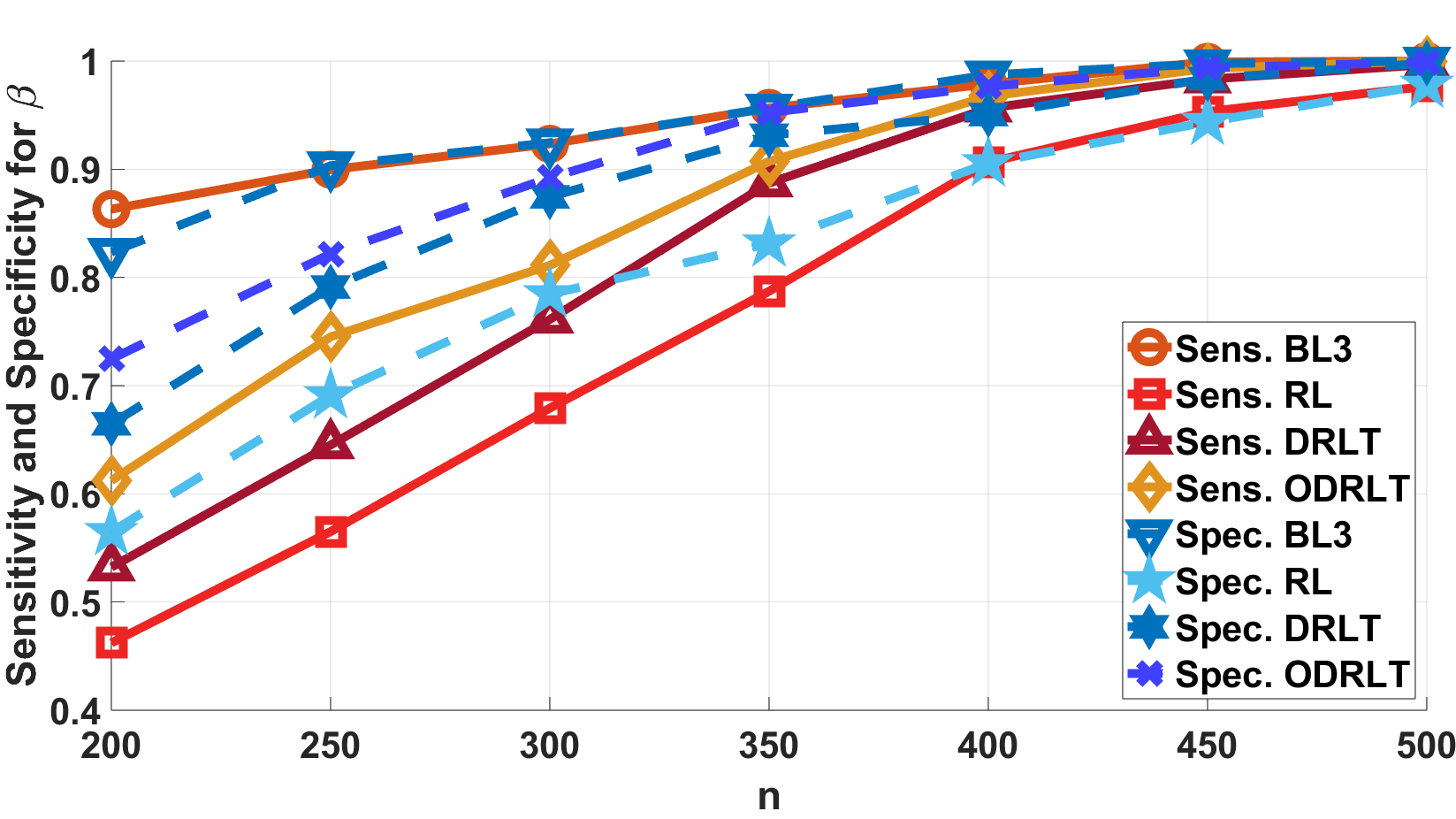}\\
    \includegraphics[scale=0.195]{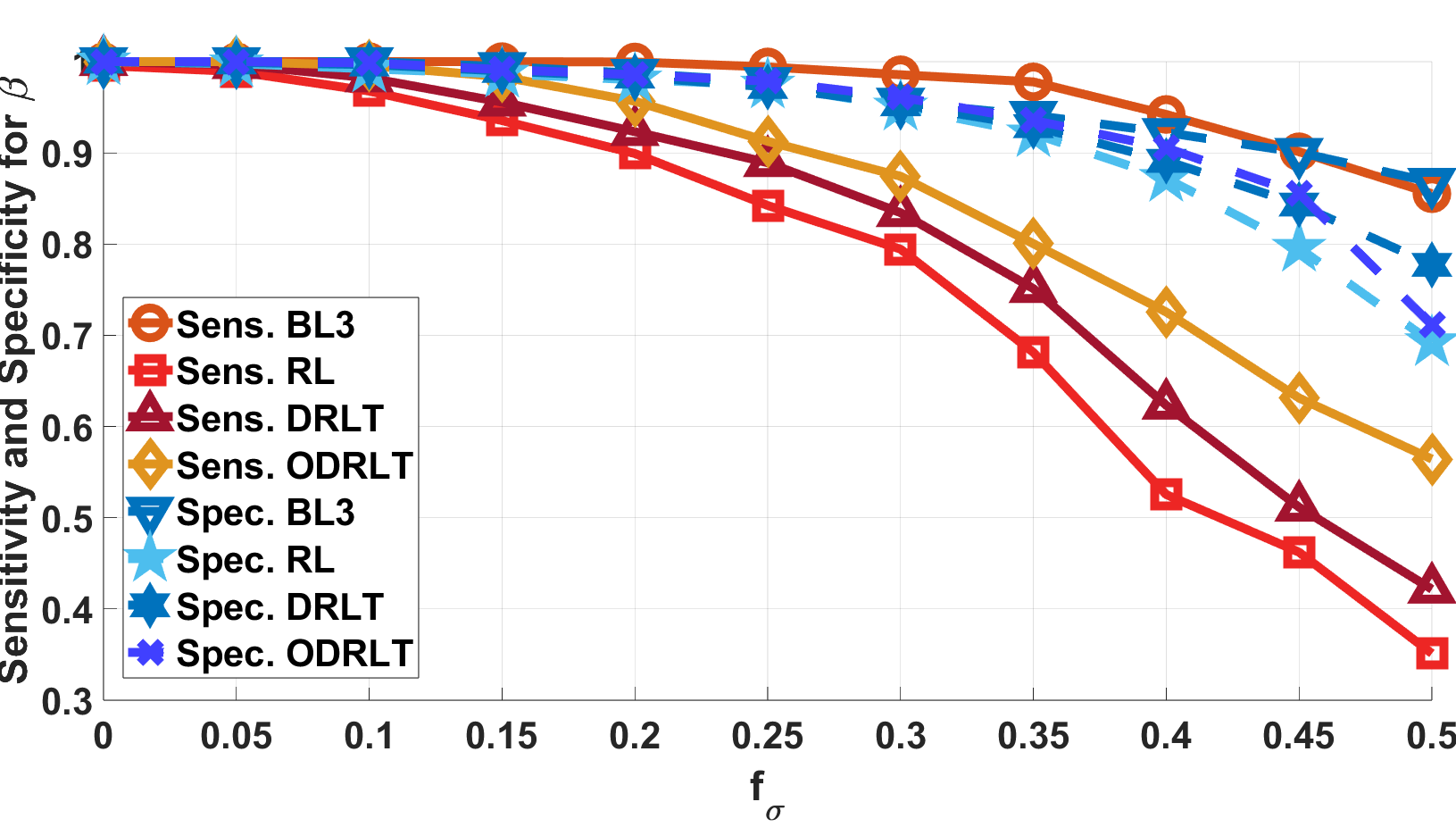}
    \includegraphics[scale=0.195]{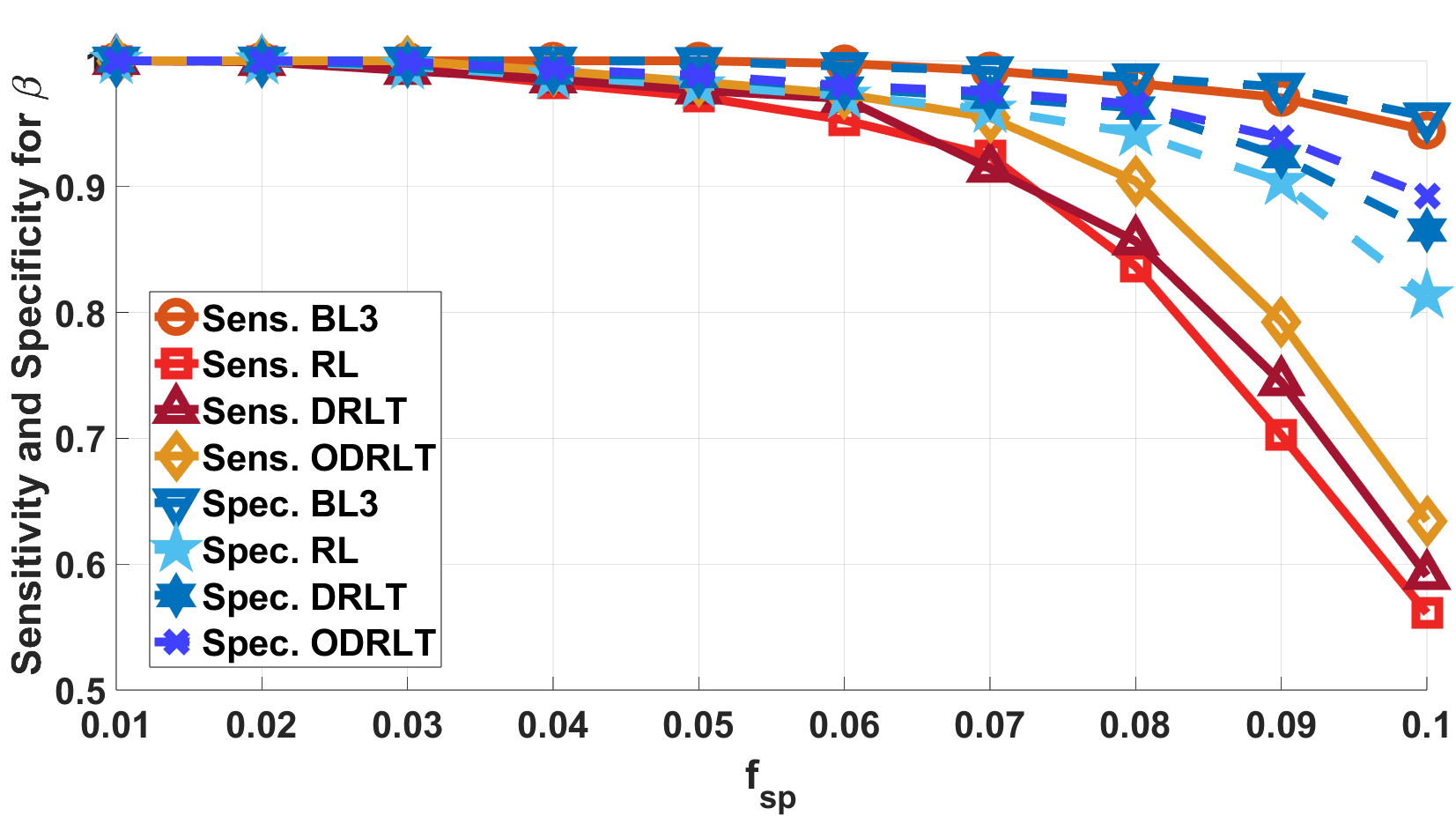}
    \caption{Average Sensitivity and Specificity plots (over 100 independent noise runs keeping $\boldsymbol{\beta^*}$, $\boldsymbol{A}$ and $\boldsymbol{\delta^*}$ fixed) plots for detecting defective samples (i.e., non-zero values of $\boldsymbol{\beta^*}$) using
    \textsc{Drlt}, \textsc{ODrlt}, Robust \textsc{Lasso} (\textsc{Rl}) and \textsc{Baseline~3}. Left to right, top to bottom: results for experiments (\textsf{EA}), (\textsf{EB}), (\textsf{EC}), (\textsf{ED}) --see ``Experiment Settings'' in the beginning of this section for more details.}  
    \label{fig:Sens_spec}    
\end{figure*}

The sensitivity and specificity values, averaged over 100 noise instances, for all four setups \textsf{EA}, \textsf{EB}, \textsf{EC} and \textsf{ED} are plotted in Fig.~\ref{fig:Sens_spec}. The plots demonstrate the superior performance of \textsc{ODrlt} over \textsc{Rl} and \textsc{Drlt}. Furthermore, the performance of \textsc{Drlt} is also superior to \textsc{Rl}. In all regimes, \textsc{Baseline-3} performs best as it is an oracular method which uses an error-free sensing matrix. We also see that for higher $n$, lower $f_{\sigma}$ and lower $f_{sp}$, the sensitivity and specificity of \textsc{ODrlt} come very close to those of \textsc{Baseline-3}. 
\subsection{RRMSE Comparison of Debiased Robust Lasso Techniques to Baseline Algorithms} \label{sec:rmse comparison}
\begin{figure*}
\centering
    \includegraphics[scale=0.195]{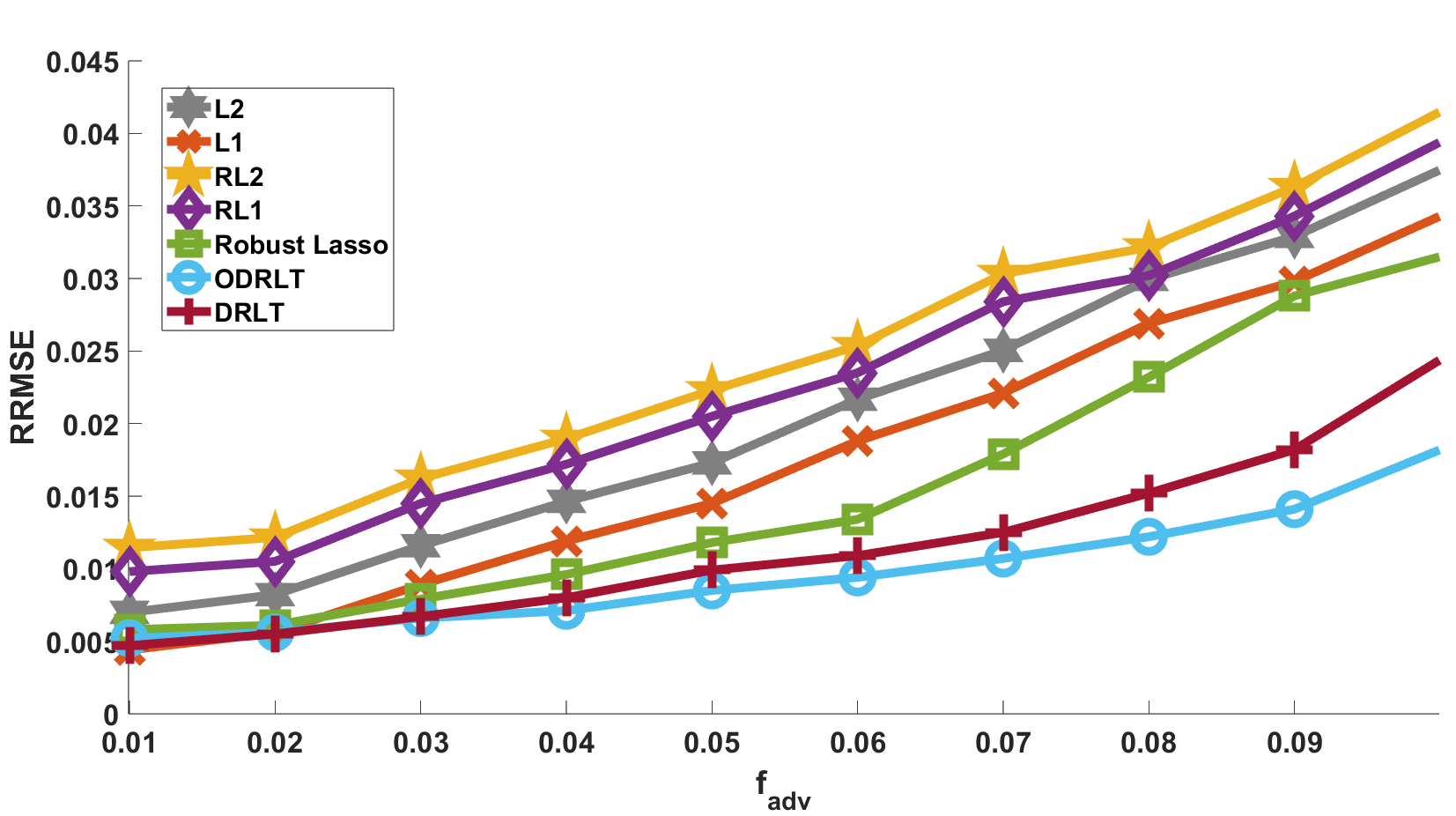}
    \includegraphics[scale=0.195]{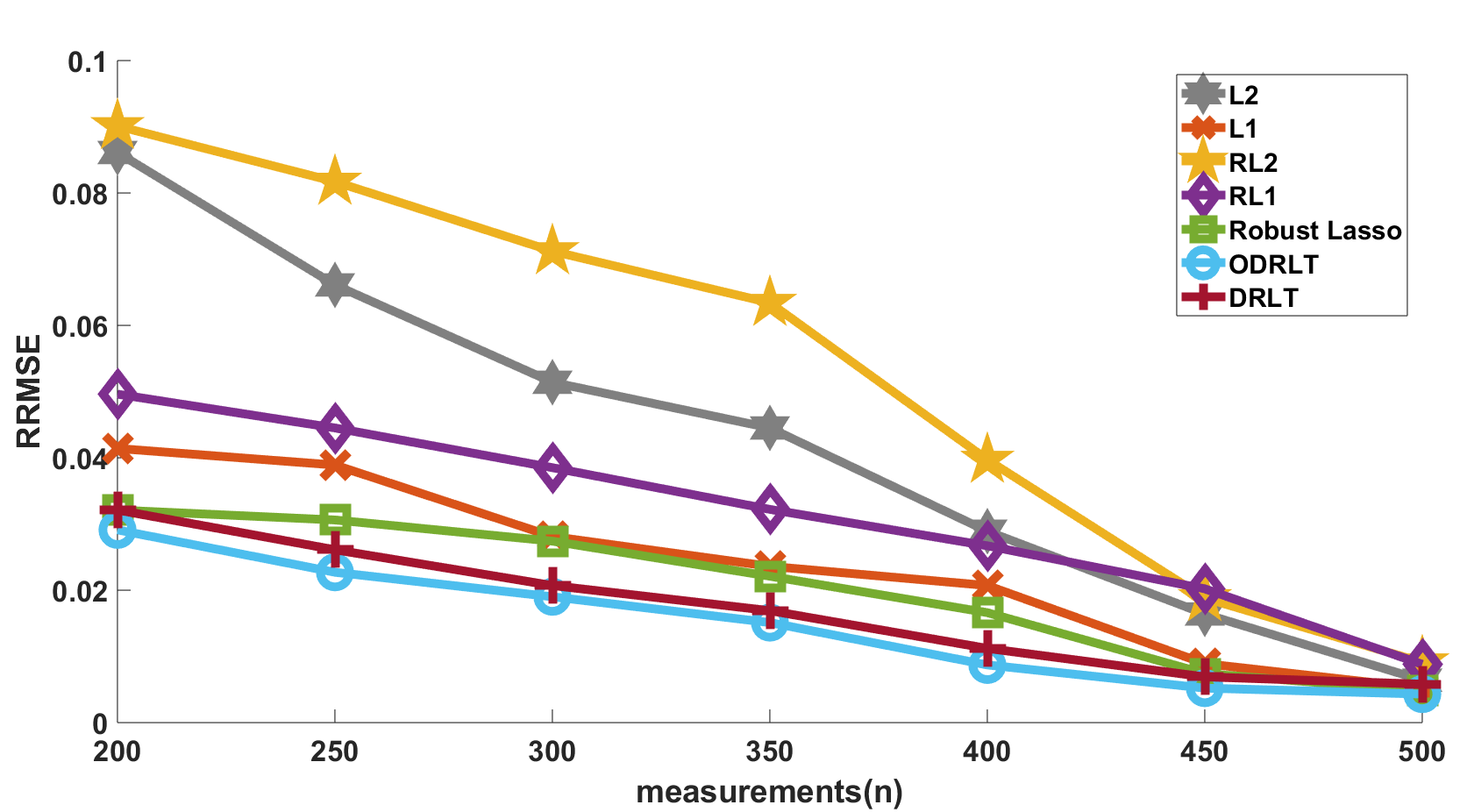}\\
    \includegraphics[scale=0.195]{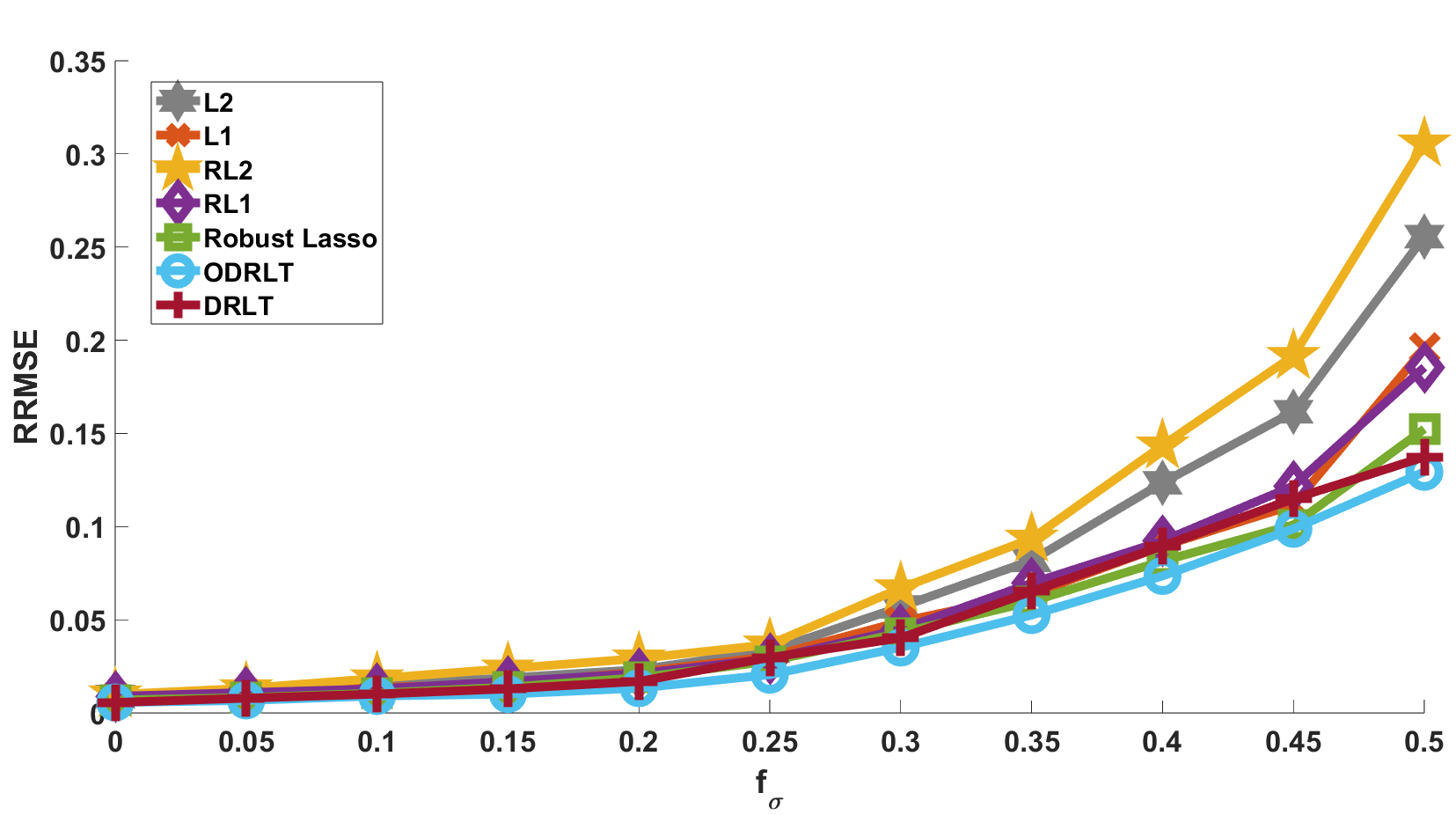}
    \includegraphics[scale=0.195]{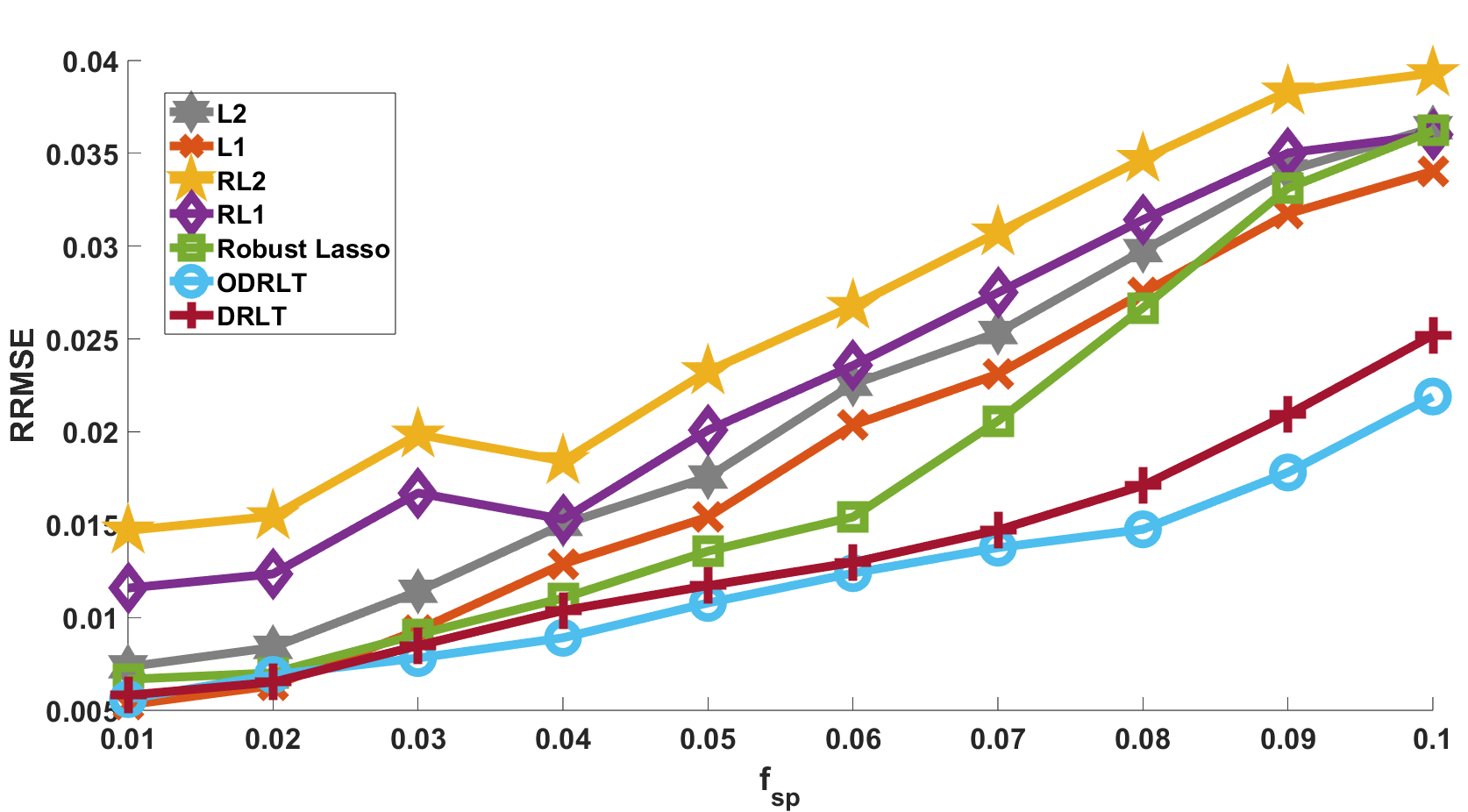}
    \caption{Average RRMSE comparison (over 100 independent noise runs keeping $\boldsymbol{\beta^*}$, $\boldsymbol{A}$ and $\boldsymbol{\delta^*}$ fixed) using  \textsc{ODrlt}, \textsc{Drlt}, \textsc{L1} (L1 \textsc{Lasso}), \textsc{L2} (L2 \textsc{Lasso}), \textsc{RL1} (L1 \textsc{Lasso} with \textsc{Ransac}), the \textsc{RL2} (L2 \textsc{Lasso} with \textsc{Ransac}), and robust \textsc{Lasso} (\textsc{Rl}). Left to right, top bottom: results for settings \textsc{EA}, \textsf{EB}, \textsf{EC}, \textsf{ED} -- see ``Experiment Settings'' in the beginning of this section for more details.}
    \label{fig:SSM_RMSE}    
\end{figure*}
We computed estimates of $\boldsymbol{\beta^*}$ using the debiased robust \textsc{Lasso} technique in two ways: (\textit{i}) with the weights matrix $\boldsymbol{W} \triangleq \boldsymbol{A}$, and (\textit{ii}) the optimal $\boldsymbol{W}$ as obtained using Alg.~\ref{alg:design_W}. We henceforth refer to these estimators as Debiased Robust Lasso (\textsc{Drl}) and Optimal Debiased Robust Lasso (\textsc{Odrl}) respectively. 
 
We computed the relative root mean squared error (RRMSE) for \textsc{Drl} and \textsc{Odrl} as follows: First, the pooled measurements with MMEs were identified as described in Sec.~\ref{sec:sensitivity_delta} and then discarded. From the remaining measurements, an estimate of $\boldsymbol{\beta^*}$ was obtained using robust \textsc{Lasso} with the optimal $\lambda_1, \lambda_2$ chosen by cross-validation. Given the resultant estimate $\boldsymbol{\hat{\beta}}$, the RRMSE was computed as $\|\boldsymbol{\beta^*}-\boldsymbol{\hat{\beta}}\|_2/\|\boldsymbol{\beta^*}\|_2$.
 
We compared the RRMSE of \textsc{Drl} and \textsc{Odrl} to that of the following algorithms:
\begin{enumerate}
\item Robust \textsc{Lasso} or \textsc{Rl} from \eqref{eq:lasso_delta}.
\item \textsc{Lasso} (referred to as \textsc{L2}) based on minimizing $\|\boldsymbol{y}-\boldsymbol{A \beta}\|^2_2 + \lambda \|\boldsymbol{\beta}\|_1$ with respect to $\boldsymbol{\beta}$. Note that this ignores MMEs.
\item An inherently outlier-resistant version of \textsc{Lasso} which uses the $\ell_1$ data fidelity (referred to as \textsc{L1}), based on minimizing $\|\boldsymbol{y}-\boldsymbol{A\beta}\|_1 + \lambda \|\boldsymbol{\beta}\|_1$ with respect to $\boldsymbol{\beta}$. 
\item Variants of \textsc{L1} and \textsc{L2} combined with the well-known \textsc{Ransac} (Random Sample Consensus) framework \cite{Fischler1981} (described below in more detail). The combined estimators are referred to as \textsc{Rl1} and \textsc{Rl2} respectively.
\end{enumerate}

\textsc{Ransac} is a popular randomized robust regression algorithm, widely used in computer vision \cite[Chap. 10]{Forsyth2012}. We apply it here to the signal reconstruction problem considered in this paper. In \textsc{Ransac}, multiple small subsets of measurements from $\boldsymbol{y}$ are randomly chosen. Let the total number of subsets be $N_S$. Let the set of the chosen subsets be denoted by $\{\mathcal{Z}_i\}_{i=1}^{N_S}$. From each subset $\mathcal{Z}_i$, the vector $\boldsymbol{\hat{\beta}}^{(i)}$ is estimated, using either \textsc{L2} or \textsc{L1}. Every measurement is made to `cast a vote' for one of the models from the set $\{\boldsymbol{\hat{\beta}}^{(i)}\}_{i=1}^{N_S}$. We say that measurement $y_l$ (where $l \in [n]$) casts a vote for model $\boldsymbol{\hat{\beta}}^{(j)}$ (where $j \in [N_S]$) if $|y_l - \boldsymbol{a_{l.}} \boldsymbol{\hat{\beta}}^{(j)}| \leq |y_l - \boldsymbol{a_{l.}} \boldsymbol{\hat{\beta}}^{(k)}|$ for $k \in [N_S], k \neq j$. Let the model which garners the largest number of votes be denoted by $\boldsymbol{\hat{\beta}}^{j_s}$, where $j_s \in [N_S]$. The set of measurements which voted for this model is called the consensus set. \textsc{Ransac} when combined with \textsc{L2} and \textsc{L1} is respectively called \textsc{Rl2} and \textsc{Rl1}. In \textsc{Rl2}, the estimator \textsc{L2} is used to determine $\boldsymbol{\beta^*}$ using measurements only from the consensus set. Likewise, in \textsc{Rl1}, the estimator \textsc{L1} is used to determine $\boldsymbol{\beta^*}$ using measurements only from the consensus set.

Our experiments in this section were performed for signal and sensing matrix settings identical to those described in Sec.~\ref{subsec:exp_beta_hypothesis}. The performance in all experiments was measured using RRMSE, averaged over reconstructions from 100 independent noise runs. For all techniques, the regularization parameters were chosen using cross-validation following the procedure in \cite{Zhang2014}. The maximum number of subsets for finding the consensus set in \textsc{Ransac} was set to $N_S = 500$ with  $0.9n$ measurements in each subset. RRMSE plots for all competing algorithms are presented in Fig.~\ref{fig:SSM_RMSE}, where we see that \textsc{Odrl} and \textsc{Drl} outperformed all other algorithms for all parameter ranges considered here. We also observe that \textsc{Odrl} produces lower RRMSE than \textsc{Drl}, particularly in the regime involving higher $f_{adv}$. 

\subsection{Comparison with Different Sensing Matrices}
\begin{figure*}[ht]
\centering      
    \includegraphics[height=1.25in]{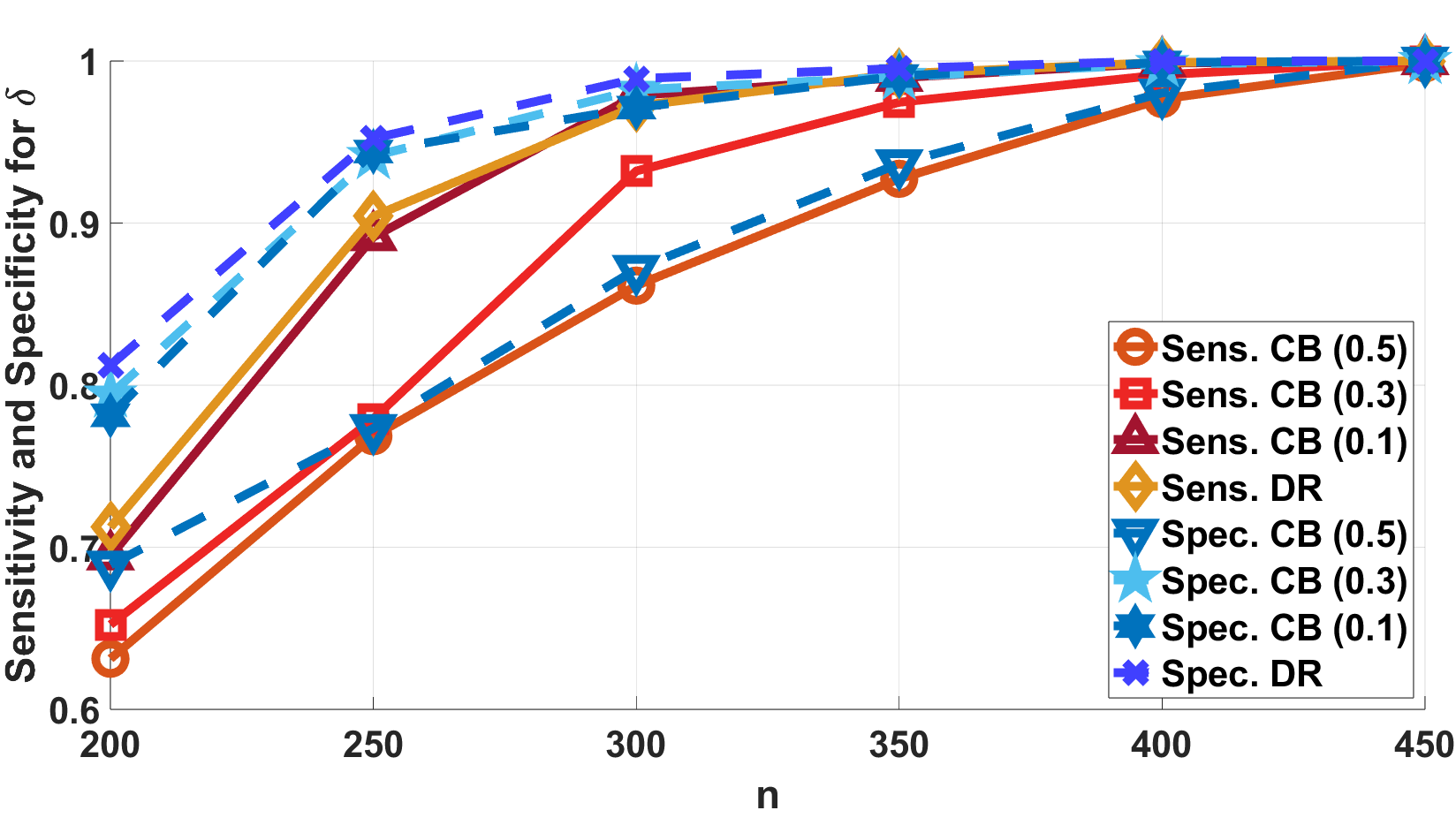}
    \includegraphics[height=1.25in]{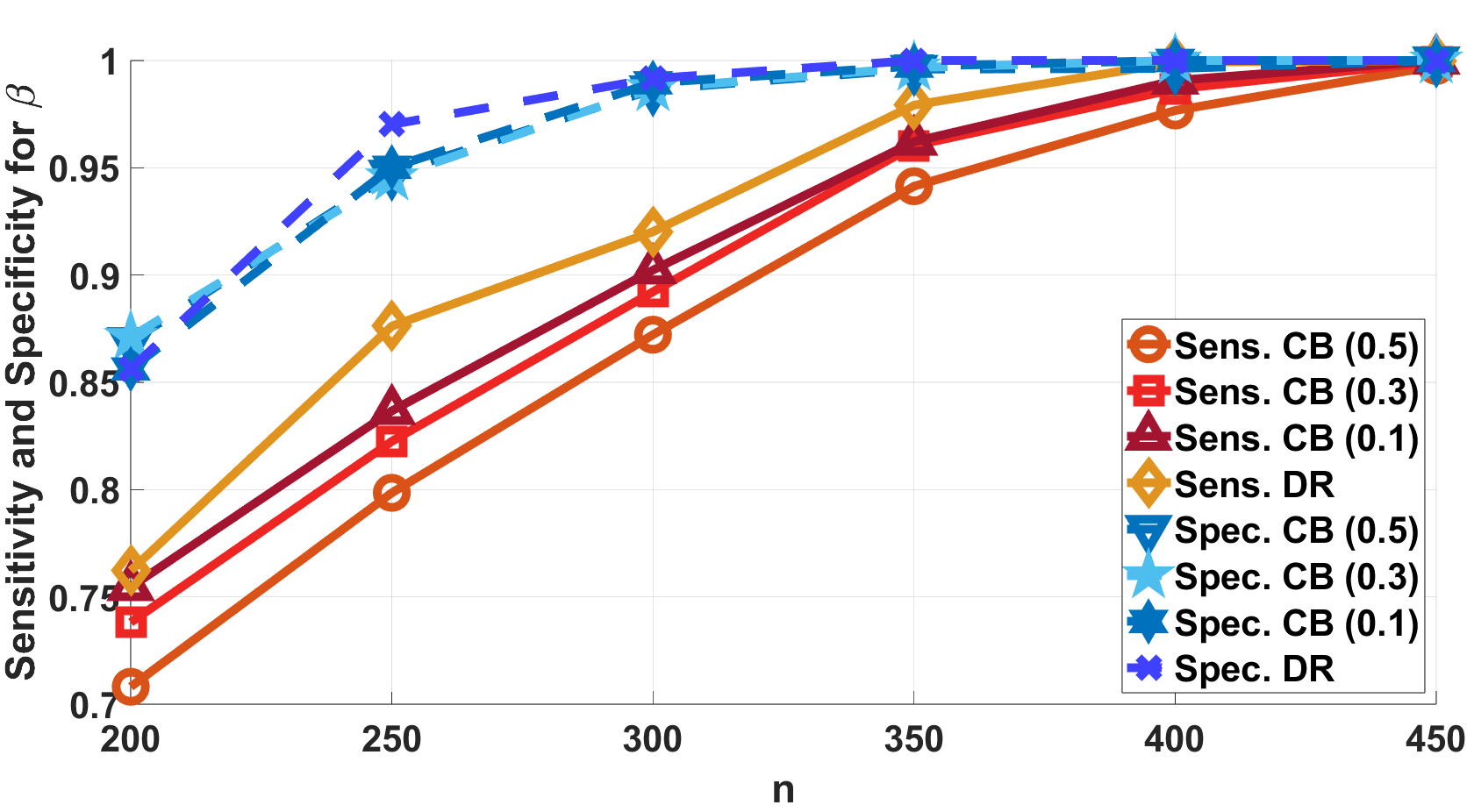}      
    \includegraphics[height=1.25in]{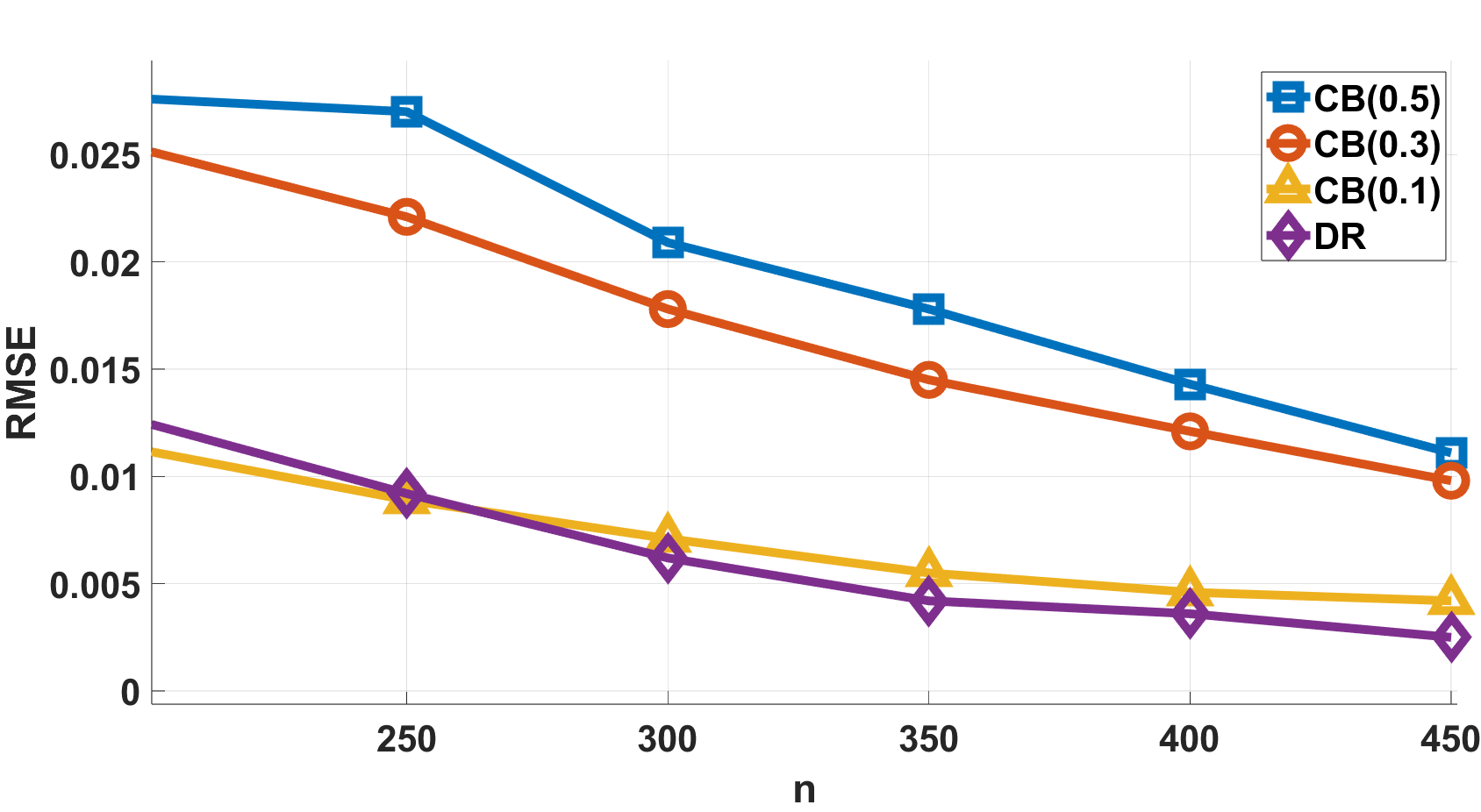}
    \caption{Plots for the setting \textsf{EA} for \textsc{ODrlt} for $\boldsymbol{\delta^*}$ (left) , \textsc{ODrlt} for $\boldsymbol{\beta^*}$ (center) and RRMSE for $\boldsymbol{\beta^*}$ (right). These are for Centered Bernoulli (CB) matrices with $\theta \in \{0.5,0.3,0.1\}$ and Doubly-regular (DR) matrices. The experimental parameters are $p = 500, f_{\sigma}=0.05, f_{adv}=0.01,f_{sp}=0.1$, and $n$ varying from 200 to 450.}   
    \label{fig:sens_spec_matrix}    
\end{figure*}

{In this subsection, we compare the performance of \textsc{ODrlt} with different sensing matrices:
(\textit{i}) Centered Bernoulli($0.1$), (\textit{ii}) Centered Bernoulli($0.3$), (\textit{iii}) Centered Bernoulli($0.5$), and (\textit{iv}) Centered \textbf{Doubly-regular}. The elements of the sensing matrices in (\textit{i}), (\textit{ii}) and (\textit{iii}) have a distribution given in \eqref{eq:Bernoulli_theta_pmf}. Doubly-regular matrices, i.e. matrices with equal number of $1$'s and $0$'s in each row and column ($n/50$ ones per column and $p/50$ ones per row) with the locations of the $1$'s randomly chosen in each row, are a common model in group testing \cite{tan2022performance}. The centering for doubly-regular matrices was done as in Sec.~\ref{Sec:Bounded_matrix} by choosing $\theta=1/50$. We compared the performances of these matrices in terms of RRMSE, Sensitivity and Specificity for the \textsc{ODrlt} estimates of $\boldsymbol{\beta^*}$ and $\boldsymbol{\delta^*}$.

For \textsc{ODrlt} estimates of $\boldsymbol{\beta^*}$ and $\boldsymbol{\delta^*}$ using all four types of sensing matrices, 
%In experimental setup \textsf{EA}, we varied $f_{adv} \in \{0.01,0.03,\ldots,0.09\}$ with fixed values $n = 400, f_{sp} = 0.01, f_{\sigma} = 0.05$. In \textsf{EB}, we varied $n$ from 200 to 450 in steps of 50 with $f_{adv} = 0.01, f_{sp} = 0.01, f_{\sigma} = 0.05$. In \textsf{EC}, we varied $f_{\sigma} \in \{0.01,0.03,\ldots,0.09\}$ with $n = 400, f_{adv} = 0.01, f_{sp} = 0.1$. In  \textsf{ED}, we varied $f_{sp} \in \{0.01,0.03,\ldots,0.09\}$ with $n = 400, f_{adv} = 0.01, f_{\sigma} = 0.05$. The experiments were run 25 times across different noise instances in $\boldsymbol{\eta}$, for the same signal $\boldsymbol{\beta^*}$.
experiments were performed using setups \textsf{EA}, \textsf{EB}, \textsf{EC} and \textsf{ED} described in the beginning of Sec.~\ref{sec:experiments} under `Experiment Settings'. The set of experimental results for setups \textsf{EB}, \textsf{EC}, \textsf{ED} are shown in the \underline{supplemental material}. Here, we show the plots only for setup \textsf{EA} for the \textsc{ODrlt} for $\boldsymbol{\delta^*}$, the \textsc{ODrlt} for $\boldsymbol{\beta^*}$, and for RRMSE for $\boldsymbol{\beta^*}$, all for varying $n$. In Fig.~\ref{fig:sens_spec_matrix}, we see that the \textsc{ODrlt} for $\boldsymbol{\delta^*}$ and $\boldsymbol{\beta^*}$ for doubly-regular designs performs the best, followed by Centered Bernoulli($0.1$),  Centered Bernoulli($0.3$) and lastly Centered Bernoulli($0.5$) matrices. Similar trends are observed for the RRMSE, as shown in Fig.~\ref{fig:sens_spec_matrix}.}

\subsection{Comparison with Different Noise Models}
\begin{figure*}[ht]
\centering      
    \includegraphics[height=1.25in]{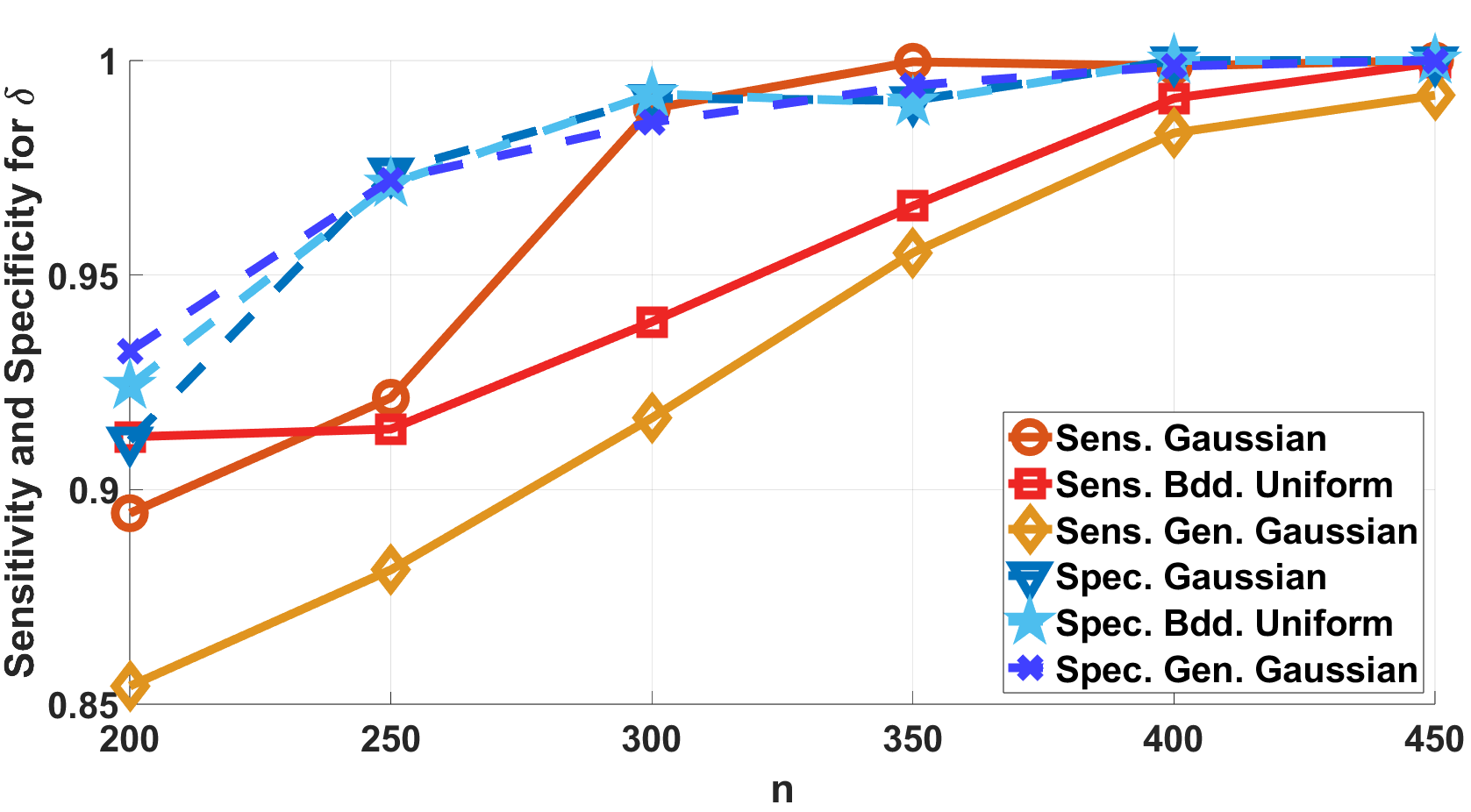}
    \includegraphics[height=1.25in]{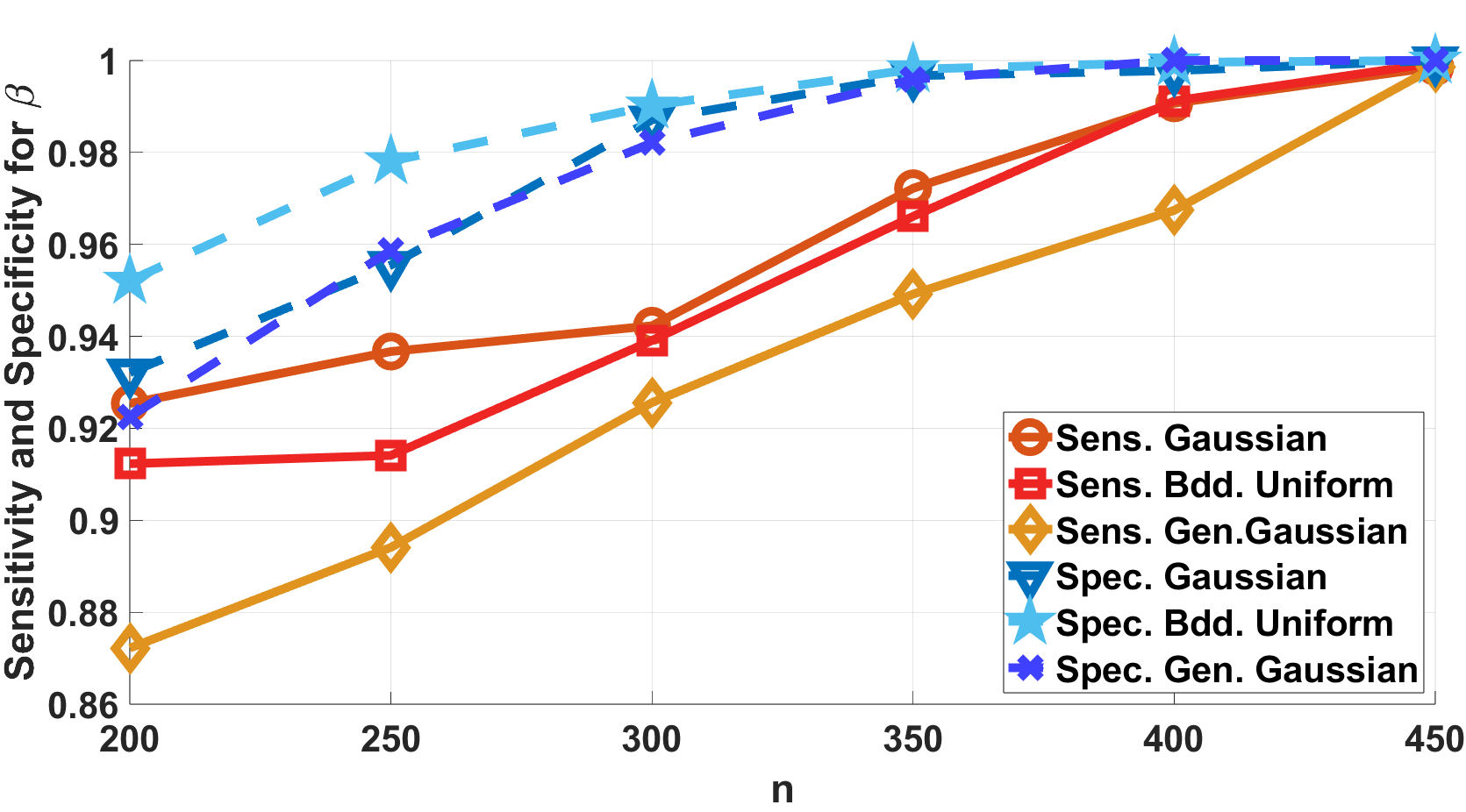}      
    \includegraphics[height=1.25in]{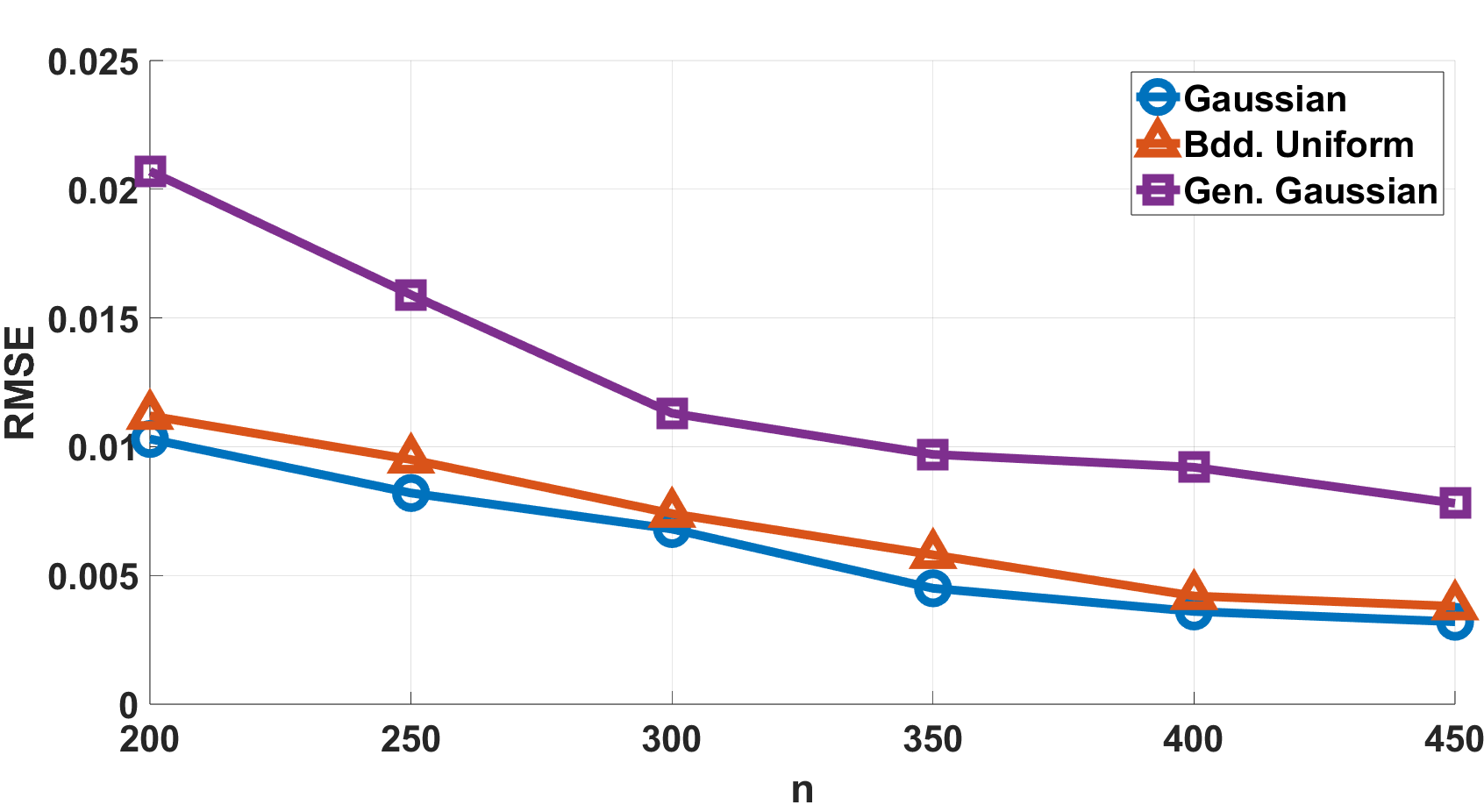}
    \caption{Plots for \textsf{EA} for \textsc{ODrlt} for $\boldsymbol{\delta^*}$ (left), \textsc{ODrlt} for $\boldsymbol{\beta^*}$ (center) and RRMSE (right) for additive noise models: $N(0,\sigma^2)$, Bounded uniform $[-\sqrt{3}\sigma,\sqrt{3}\sigma]$ and Generalized Gaussian with shape parameter $1.5$. The experimental parameters are $p = 500, f_{\sigma}=0.05, f_{adv}=0.01,f_{sp}=0.1$ and $n$ varying from 200 to 450.}   
    \label{fig:sens_spec_noise}    
\end{figure*}

{In this set of experiments, we compared the performance of \textsc{ODrlt} in the presence of additive noise $\boldsymbol{\eta}$ obtained from two different distributions in addition to $\mathcal{N}(0,\sigma^2)$: (\textit{i}) Bounded Uniform$[-\sqrt{3}\sigma,\sqrt{3}\sigma]$, and (\textit{ii}) a Generalized Gaussian (GG) distribution with shape parameter $1.5$ and scale $\sigma^2$, having the following probability density function $$f(x) = \frac{3}{4 \sigma \, \Gamma\left( \frac{2}{3} \right)} \exp\left( -\left| \frac{x}{\sigma} \right|^{1.5} \right).
$$
For the bounded uniform and GG noise models, we used the empirical \textsc{ODrlt} defined in the remarks on Theorem~\ref{th:distribution_beta_delta_opt}. For the different noise models, we observe the sensitivity and specificity of the \textsc{ODrlt} estimates for $\boldsymbol{\delta^*}$ and $\boldsymbol{\beta^*}$, and RRMSE for the estimates of $\boldsymbol{\beta^*}$ in Fig.~\ref{fig:sens_spec_noise} for varying $n$ (setup \textsf{EA} as defined in the beginning of Sec.~\ref{sec:experiments}). For other setups (\textsf{EB}, \textsf{EC}, \textsf{ED}), the results are shown in the \underline{supplemental material}.

We observe that the performance under Gaussian and Uniform noise is approximately the same. Since the GG has heavier tails than the Gaussian, it leads to some deterioration in performance. However, for large $n$ or small $f_{sp}, f_{adv}$ and $f_{\sigma}$, the \textsc{ODrlt} performs well even in the presence of the chosen GG noise model. }

\subsection{Comparison with the Probabilistic Group Testing Technique from \cite{Cheragchi2011}} 
{The work in \cite{Cheragchi2011} considers probabilistic and structured errors in the pooling matrix $\boldsymbol{B}$, where an entry $b_{ij}$ with a value of 1 could flip to 0 with  probability $\vartheta' := 1-\vartheta$ where $\vartheta \in (0,1)$, but not vice versa, i.e., a genuinely zero-valued $b_{ij}$ never flips to 1. They develop a combinatorial method called the Distance Decoder (\textsc{Dist-D}) to estimate the true (binary-valued) signal vector in the presence of such errors from binary measurement vectors. Since such errors in the pooling matrix can be considered as a form of MME's, we apply our algorithm based on the \textsc{ODrlt} to detect these errors and compare its performance to that of \textsc{Dist-D}.

For a fair comparison between \textsc{ODrlt} and \textsc{Dist-D} \cite{Cheragchi2011}, we converted the real-valued signal $\boldsymbol{\beta^*}$ (containing values of infection levels) to a binary-valued signal $\boldsymbol{x^*}$ where $x^*_j=1 \ \text{if} \ \beta_j^*>0$ and $x^*_j=0$ otherwise. Then we obtained the pooled measurement vector $\boldsymbol{u} := \boldsymbol{Bx^*}$ using binary AND/OR operations for the method in \cite{Cheragchi2011} where $\boldsymbol{B}$ obeys the $\textrm{Bernoulli}(0.5)$ model. For \textsc{ODrlt}, we obtained a pooled measurement vector $\boldsymbol{y} = \boldsymbol{Ax^*} + \boldsymbol{\eta}$ using the usual matrix-vector product for real numbers, where $\boldsymbol{A}$ is obtained by a centering operation on the \emph{same} $\boldsymbol{B}$ matrix. Here, the elements of $\boldsymbol{\eta}$ are drawn from $\mathcal{N}(0,\sigma^2)$ with $\sigma$ calculated using $f_{\sigma} = 0.01$. For the method in \cite{Cheragchi2011}, the aim was to estimate $\boldsymbol{x^*}$ given $\boldsymbol{u}, \boldsymbol{B}$. For \textsc{ODrlt}, the aim was to estimate $\boldsymbol{x^*}$ given $\boldsymbol{y}, \boldsymbol{A}$.  

We compared the performance of \textsc{ODrlt} (at 5\% significance level) and \textsc{Dist-D} in terms of Sensitivity and Specificity for (\textit{i}) varying $\vartheta \in [0.75:0.05:0.95]$, and (\textit{ii}) varying number of measurements $n \in [250:50:450]$. The other parameters were fixed to $p=500$, $s=10$. We chose the error parameter $e$ for \textsc{Dist-D} as per the formula given in Sec.~VI of \cite{Cheragchi2011}. Note that the model in \cite{Cheragchi2011} does not account for any noise in $\boldsymbol{u}$ other than that due to bit-flips in $\boldsymbol{B}$.}
\begin{table}[ht]
\centering
\begin{minipage}{0.48\textwidth}
    {\begin{tabular}{|c|c|c|c|c|}
\hline
\rowcolor{Gray}
$\vartheta$ & Sens.\textsc{ODrlt} & Sens.\textsc{Dist-D} & Spec.\textsc{ODrlt} & Spec.\textsc{Dist-D}  \\
\hline
0.75 & 0.892 & 0.901 & 0.967 & 0.972 \\
0.80  & 0.954 & 0.941 & 0.978 & 0.984 \\
0.85 & 0.987 & 0.984 & 0.981 & 0.995 \\
0.90  & 0.995 & 0.992 & 0.989 & 0.997 \\
0.95 & 1      & 0.998 & 0.997 & 1      \\
\hline
\end{tabular}
\caption{Sensitivity (Sens.) and Specificity (Spec.) values for \textsc{ODrlt} and \textsc{Dist-D} across different $\vartheta$ values where $\vartheta' := 1-\vartheta$ is the probability of a bitflip in $\boldsymbol{B}$. The other parameters are fixed at $p=500,n=400, s=10$.}
\label{tab:sens_spec_DD}}
\end{minipage}
\begin{minipage}{0.48\textwidth}
   \vskip -10pt
    \centering
{\begin{tabular}{|c|c|c|c|c|}
\hline
\rowcolor{Gray}
$n$ & Sens.\textsc{ODrlt} & Sens.\textsc{Dist-D} & Spec.\textsc{ODrlt} & Spec.\textsc{Dist-D}  \\
\hline
250 & 0.804 & 0.781 & 0.889 & 0.891 \\
300 & 0.892 & 0.875 & 0.934 & 0.955 \\
350 & 0.969 & 0.936 & 0.975 & 0.986 \\
400 & 0.995 & 0.989 & 0.999 & 1      \\
450 & 1      & 1      & 1      & 1      \\
\hline
\end{tabular}
\caption{Sensitivity and Specificity values for \textsc{ODrlt} and \textsc{Dist-D} across different $n$ values. The other parameters are fixed at $p=500, s=10, \vartheta=0.9$.}
\label{tab:sens_spec_n}}
\end{minipage}
\end{table}
{In Tables~\ref{tab:sens_spec_DD} and \ref{tab:sens_spec_n}, we see that the performance of \textsc{ODrlt} and \textsc{Dist-D} are quite similar for varying $\vartheta$ and $n$, despite the presence of some additive noise in $\boldsymbol{y}$ for \textsc{ODrlt}. 

\noindent \textbf{Comparison in the presence of two-way probabilistic errors:} We now consider a two-way bit-flip model where the entries of $\boldsymbol{B}$ can flip from 1 to 0 or from 0 to 1, with probability $\vartheta' := 1-\vartheta$. We performed the same experiments as before with the same parameter settings.}
\begin{table}[ht]
    \centering
\begin{minipage}{0.48\textwidth}
    {\centering
\begin{tabular}{|c|c|c|c|c|}
\hline
\rowcolor{Gray}
$\vartheta$ & Sens.\textsc{ODrlt} & Sens.\textsc{Dist-D} & Spec.\textsc{ODrlt} & Spec.\textsc{Dist-D}  \\
\hline
0.75 & 0.692 & 0.601 & 0.769 & 0.672 \\
0.80  & 0.812 & 0.691 & 0.879 & 0.780 \\
0.85 & 0.902 & 0.785 & 0.951 & 0.897 \\
0.90  & 0.943 & 0.861 & 0.983 & 0.953 \\
0.95 & 0.988 & 0.921 & 0.9994 & 0.977  \\
\hline
\end{tabular}
\caption{Sensitivity and Specificity values for \textsc{ODrlt} and \textsc{Dist-D} across different activation probability $\vartheta$ values for two-way probabilistic errors in $\boldsymbol{B}$. The other parameters are fixed at $p=500,n=400, s=10$.}
\label{tab:sens_spec_DD_both}}
\end{minipage}
\begin{minipage}{0.48\textwidth}
    {\centering
\begin{tabular}{|c|c|c|c|c|}
\hline
\rowcolor{Gray}
$n$ & Sens.\textsc{ODrlt} & Sens.\textsc{Dist-D} & Spec.\textsc{ODrlt} & Spec.\textsc{Dist-D}  \\
\hline
250 & 0.627 & 0.552 & 0.782 & 0.672 \\
300 & 0.782 & 0.667 & 0.895 & 0.794 \\
350 & 0.896 & 0.756 & 0.952 & 0.8798 \\
400 & 0.959 & 0.889 & 0.992 & 0.957      \\
450 & 0.983 & 0.9496 & 1      & 0.989      \\
\hline
\end{tabular}
\caption{Sensitivity and Specificity values for \textsc{ODrlt} and \textsc{Dist-D} across different $n$ values for two-way probabilistic errors in $\boldsymbol{B}$. The other parameters are fixed at $p=500, s=10, \vartheta=0.9$.}
\label{tab:sens_spec_n_both}}
\end{minipage}
\end{table}
{In Tables \ref{tab:sens_spec_DD_both} and \ref{tab:sens_spec_n_both}, we see that \textsc{ODrlt} clearly outperforms \textsc{Dist-D} in terms of both sensitivity and specificity across many values of $\vartheta$ and $n$. Since the algorithm \textsc{Dist-D} is not designed to handle the two-way errors, this experiment showcases the robustness of our algorithm \textsc{ODrlt} for different models of MME's. Besides this, we also note that the \textsc{ODrlt} approach is also designed to determine defect/infection levels and identify which pools contained errors. }

\section{Conclusion}
\label{sec:conclusion}
We have presented a technique for determining the sparse vector $\boldsymbol{\beta^*}$ of health status values from noisy pooled measurements in $\boldsymbol{y}$, with the additional feature that our technique is designed to handle bit-flip errors in the pooling matrix. These bit-flip errors can occur at a small number of unknown locations, due to which the pre-specified matrix $\boldsymbol{A}$ (known) and the actual pooling matrix $\boldsymbol{\hat{A}}$ (unknown) via which pooled measurements are acquired, differ from each other. We use the theory of \textsc{Lasso} debiasing as our basic scaffolding to identify the defective samples in $\boldsymbol{\beta^*}$, but with extensive and non-trivial theoretical and algorithmic innovations to (\textit{i}) make the debiasing robust to model mismatch errors (MMEs), and also to (\textit{ii}) enable identification of the pooled measurements that were affected by the MMEs. Our approach is also validated by an extensive set of simulation results, where the proposed method outperforms intuitive baseline techniques. To our best knowledge, there is no prior literature on using \textsc{Lasso} debiasing to identify measurements with MMEs. Our weights matrix $\boldsymbol{W}$ was designed to minimize the variance of the elements of $\boldsymbol{\hat{\beta}_W}$. One could have alternatively designed $\boldsymbol{W}$ to minimize the variance of $\boldsymbol{\hat{\delta}_W}$. Likewise, another approach for debiasing could have involved designing an approximate inverse for $(\boldsymbol{A}|\boldsymbol{I_n})$, which would correspond to an approach that minimizes the total variance of the debiased estimates of the elements of $\boldsymbol{\beta^*}$ 
and $\boldsymbol{\delta^*}$. But our primary goal is to accurately estimate $\boldsymbol{\beta^*}$, hence we chose to minimize the variance of $\boldsymbol{\hat{\beta}_W}$. 

There are several interesting avenues for future work:
\begin{enumerate}
\item Currently the optimal weights matrix $\boldsymbol{W}$ is designed to minimize the variance of the debiased estimates of $\boldsymbol{\beta^*}$ and not necessarily those of $\boldsymbol{\delta^*}$. Our technique could in principle be extended to derive another weights matrix to minimize the variance of the debiased estimates of $\boldsymbol{\delta^*}$. 
\item A specific form of MMEs consists of unknown permutations in the pooling matrix (also called `permutation noise' \cite{Zabeti2021}) where the pooled results are swapped with one another. The techniques in this paper can be extended to identify pooled measurements that suffer from permutation noise, and potentially correct them. 
\item Our technique, which involves debiasing, requires more stringent constraints than $\|\boldsymbol{\beta^*}\|_0 < n/\log p$ used for standard sparse regression including \textsc{Lasso}. However in the literature on \textsc{Lasso} debiasing, there exist techniques such as \cite{cai2017confidence} which relax the condition on $\|\boldsymbol{\beta^*}\|_0$ to allow for $\sqrt{n}/\log p < \|\boldsymbol{\beta^*}\|_0 \leq  n/\log p$, with the caveat that \textit{a priori} knowledge of $\|\boldsymbol{\beta^*}\|_0$ is (provably) essential. Incorporating these results within the current framework is another avenue for future research. It is also of interest to combine our results with those on \textit{in situ} estimation of $\|\boldsymbol{\beta^*}\|_0$ from pooled or compressed measurements as in \cite{ravazzi2018sparsity,lopes2016unknown}.
\item Another interesting direction of research is to explore applications of the developed method to handle signal-dependent noise models such as Poisson, explore model mismatch errors in newly emerged group testing modalities such as tropical group testing \cite{wang2023tropical,paligadu2024small}, or explore specific modalities such as digital drop PCR \cite{vasudevan2021digital,mao2019principles,suo2020ddpcr}.
\end{enumerate}

\appendices
\section{Overview of Proofs of Theoretical Results}
We first establish all theoretical results pertaining to the robust \textsc{Lasso} in Appendix~\ref{Sec:App_A1}. Next, we establish all the theoretical results pertaining to the Debiased Robust \textsc{Lasso} Estimator in Appendix.~\ref{sec:App_A2}. In Appendix \ref{sec:App_A3}, we derive some properties of the random Rademacher sensing matrix $\boldsymbol{A}$ used in the proof of the Debiased \textsc{Lasso} Estimator. Lastly, in Appendix~\ref{Sec:App_A4}, we derive properties of a matrix with entries drawn from a centered $\text{Bernoulli}(\theta)$ distribution. The results are useful for proving key theoretical results in \ref{sec:App_A2}. Within Appendix~\ref{sec:App_A2}, we first establish theoretical properties of the Debiased Robust \textsc{Lasso} (\textsc{Drlt}) estimator given in \eqref{eq:beta_marginal_dist_A} and \eqref{eq:delta_marginal_dist_A} in Theorem \ref{th:dist_debiased_beta_delta_A} by taking $\boldsymbol{W}=\boldsymbol{A}$. Next, we describe the proof of the debiased \textsc{Lasso|} estimator with the optimal $\boldsymbol{W}$ obtained from Alg.~\ref{alg:design_W} given in Theorem \ref{th:distribution_beta_delta_opt}. The proof follows the same pattern as that of Theorem \ref{th:dist_debiased_beta_delta_A}.

\section{Proofs of Theorems and Lemmas on Robust Lasso}\label{Sec:App_A1}
%We extend Theorem 1 and Lemma 1 of \cite{Nguyen2013} to prove our Theorem \ref{th:upper_bound_robustLasso}. 
In order to prove the upper bound on the reconstruction error for the Robust \textsc{Lasso} in Theorem~\ref{th:upper_bound_robustLasso}, we first extend Lemma 1 of \cite{Nguyen2013} to show that the Extended Restricted Eigenvalue Condition (EREC) also holds for the Rademacher sensing matrix $\boldsymbol{A}$. This is established in Lemma ~\ref{le:Ext_RE} of our work. Next, we use Theorem 1 of \cite{Nguyen2013} and the tail bounds of Gaussian random variables in \eqref{eq:lambda_1} and \eqref{eq:lambda_2} to complete the proof of Theorem \ref{th:upper_bound_robustLasso}.

We re-parameterize model \eqref{eq:fm_delta} and  the robust \textsc{Lasso} optimization problem \eqref{eq:lasso_delta} to match those in \cite{Nguyen2013}, i.e.,
\begin{equation}
    \boldsymbol{y} = \left(\boldsymbol{A} \ | \ \sqrt{n}\boldsymbol{I_n}\right) \begin{pmatrix}
    \boldsymbol{\beta^*} \\ \boldsymbol{\delta^*}/\sqrt{n}
\end{pmatrix} + \boldsymbol{{\eta}} = \boldsymbol{A\beta^*}+\sqrt{n}\boldsymbol{e^*}+\boldsymbol{\eta},
\label{eq:noisevector2}    
\end{equation}
where $\boldsymbol{e^*}\triangleq \boldsymbol{\delta^*}/\sqrt{n}$.
Note that the optimization problem \eqref{eq:lasso_delta} 
is
\begin{eqnarray}
   \nonumber \begin{pmatrix}
    \boldsymbol{\hat{\beta}_{\lambda_1}} \\
    \boldsymbol{\hat{\delta}_{\tilde{\lambda}_2}}
\end{pmatrix} &=& \arg\min_{\boldsymbol{\beta},\boldsymbol{\delta}} {\frac{1}{2n}}\left\|\boldsymbol{y}-(\boldsymbol{A}|\sqrt{n}\boldsymbol{I_n}) \begin{pmatrix}
    \boldsymbol{\beta} \\
    \boldsymbol{\delta}/\sqrt{n}
\end{pmatrix}\right\|^2_2 + \lambda_1 \|\boldsymbol{\beta}\|_1 + {\tilde{\lambda}_2} \left\|\frac{\boldsymbol{\delta}}{\sqrt{n}}\right\|_1, 
\end{eqnarray}
where $\tilde{\lambda}_2=\sqrt{n}\lambda_2$. 
The equivalent robust \textsc{Lasso} optimization problem for the model \eqref{eq:noisevector2} is given by:
\begin{eqnarray}
 \begin{pmatrix}
    \boldsymbol{\hat{\beta}_{\lambda_1}} \\
    \boldsymbol{\hat{e}_{\tilde{\lambda}_2}}
\end{pmatrix} &=& \arg\min_{\boldsymbol{\beta},\boldsymbol{e}} {\frac{1}{2n}}\left\|\boldsymbol{y}-(\boldsymbol{A}|\sqrt{n}\boldsymbol{I_n}) \begin{pmatrix}
    \boldsymbol{\beta} \\
    \boldsymbol{e}
\end{pmatrix}\right\|^2_2 + \lambda_1 \|\boldsymbol{\beta}\|_1 + {\tilde{\lambda}_2} \left\|\boldsymbol{e}\right\|_1,
\label{eq:lasso_e} 
\end{eqnarray}
where $\boldsymbol{\hat{e}_{\lambda_2}}= \boldsymbol{\hat{\delta}_{\lambda_2}}/\sqrt{n} $.
In order to prove Theorem \ref{th:upper_bound_robustLasso}, we first recall the Extended Restricted Eigenvalue Condition (EREC) for a sensing matrix from \cite{Nguyen2013}.
Given $\boldsymbol{\beta}^*$ and $\boldsymbol{\delta}^*$, let us define sets 
\begin{equation}\label{eq:sets_S_R}
    \mathcal{S} \triangleq \{j:\beta^*_j \neq 0\} \ , \ \mathcal{R} \triangleq \{i:\delta^*_i \neq 0\}.
\end{equation}
 Note that $s \triangleq |\mathcal{S}|, r \triangleq |\mathcal{R}|$.

\begin{definition}
\textbf{Extended Restricted Eigenvalue Condition (EREC)} \cite{Nguyen2013}: Given $\mathcal{S},\mathcal{R}$ as defined in \eqref{eq:sets_S_R}, and $\lambda_1,\tilde{\lambda}_2>0$, 
an $n\times p$ matrix $\boldsymbol{A}$ is said to satisfy the EREC if there exists a $\kappa>0$ such that
\begin{equation}\label{eq:Extended_RE}
    \frac{1}{\sqrt{n}} \|\boldsymbol{Ah_{\beta}}+\sqrt{n}\boldsymbol{h_{\delta}}\|_2 \geq \kappa(\|\boldsymbol{h_{\beta}}\|_2+\|\boldsymbol{h_{\delta}}\|_2),
\end{equation}
for all $(\boldsymbol{h_{\beta}},\boldsymbol{h_{\delta}}) \in \mathcal{C}(\mathcal{S},\mathcal{R},\lambda)$ with $\lambda := \tilde{\lambda}_2/\lambda_1$ and where $\mathcal{C}$ is defined as follows:
\begin{eqnarray}\label{eq:Coneconstraint}
 \mathcal{C}(\mathcal{S},\mathcal{R},\lambda) &\triangleq& \{(\boldsymbol{h_{\beta}},\boldsymbol{h_{\delta}})\in \mathbb{R}^p\times \mathbb{R}^n:\|(\boldsymbol{h_{\beta}})_{\mathcal{S}^c}\|_1+\lambda \|(\boldsymbol{h_{\delta}})_{\mathcal{R}^c}\|_1 \leq  3(\|(\boldsymbol{h_{\beta}})_{\mathcal{S}}\|_1+\lambda \|(\boldsymbol{h_{\delta}})_{\mathcal{R}}\|_1)\}. 
\end{eqnarray}
Here, $(\boldsymbol{h_{\beta}})_\mathcal{S}$ and $(\boldsymbol{h_{\delta}})_\mathcal{R}$ are $s$ and $r$ dimensional vectors constructed from $\boldsymbol{h_{\beta}}$ and $\boldsymbol{h_{\delta}}$ respectively, with supports restricted to the set $\mathcal{S}$ and $\mathcal{R}$ as defined in \eqref{eq:sets_S_R}. In other words, $\forall j \in \mathcal{S}, (h_{\beta})_{{\mathcal{S}}}(j) = h_{\beta}(j); 
\forall j \notin \mathcal{S}, (h_{\beta})_{\mathcal{S}}(j) = 0$. Note that $\boldsymbol{h_{\beta}}$ and $\boldsymbol{h_{\delta}}$ are intended to be error vectors, i.e., $\boldsymbol{h_{\beta}} := \boldsymbol{\beta^*} - \boldsymbol{\hat{\beta}_{\lambda_1}}$ and $\boldsymbol{h_{\delta}} := \boldsymbol{\delta^*} - \boldsymbol{\hat{\delta}_{\tilde{\lambda}_2}}$, where 
$\boldsymbol{\hat{\beta}_{\lambda_1}}, \boldsymbol{\hat{\delta}_{\tilde{\lambda}_2}}$ are estimates of $\boldsymbol{\beta^*}$ and $\boldsymbol{\delta^*}$ via the robust \textsc{Lasso}. 
$\blacksquare$
\label{def:EREC}
\end{definition}
In Lemma.~\ref{le:Ext_RE}, we extend Lemma 1 from \cite{Nguyen2013} to random Rademacher matrices. In this lemma we show that a random Rademacher matrix $\boldsymbol{A}$ satisfies EREC with high probability for $\kappa=1/16$.

\begin{lemma}\label{le:Ext_RE} 
Let $\boldsymbol{A}$ be an $n \times p$ matrix  with i.i.d. Rademacher entries. There exist positive constants $C_1,C_2,c_3,c_4$ such that if $n \geq C_1 s \log p$ and $r \leq \textrm{min}\{C_2 \frac{n}{ \log n},\frac{s\log p}{\log n}\}$ then
     \begin{equation*}
        P\left(\forall \ (\boldsymbol{h_{\beta}},\boldsymbol{h_{\delta}}) \in \mathcal{C}\left(\mathcal{S},\mathcal{R},\lambda\right), \ \frac{1}{\sqrt{n}} \|\boldsymbol{Ah_{\beta}}+\sqrt{n}\boldsymbol{h_{\delta}}\|_2 \geq \frac{1}{16}(\|\boldsymbol{h_{\beta}}\|_2+\|\boldsymbol{h_{\delta}}\|_2)\right) \geq 1-c_3 \exp{\{-c_4n\}},
    \end{equation*}
      where $\lambda := \sqrt{\frac{\log n}{\log p}}$ and $\mathcal{C}$ as in \eqref{eq:Coneconstraint}.
\hfill{$\blacksquare$}
\end{lemma}
\noindent \textbf{Proof of Lemma~\ref{le:Ext_RE}:}
Using a similar line of argument as in the proof of Lemma 1 of \cite{Nguyen2013}, it is enough to show the following two properties of the sensing matrix $\boldsymbol{A}$ to complete the proof. 
\begin{enumerate}
    \item \textbf{Lower bound on $\frac{1}{n} \|\boldsymbol{Ah_{\beta}}\|_2^2 +\|\boldsymbol{h_{\delta}}\|_2^2$}. For some $\kappa_1>0$ with high probability,
    \begin{equation}
        \frac{1}{n} \|\boldsymbol{Ah_{\beta}}\|_2^2 +\|\boldsymbol{h_{\delta}}\|_2^2 \geq \kappa_1 \left(\|\boldsymbol{h_{\beta}}\|_2+\|\boldsymbol{h_{\delta}}\|_2\right)^2 \ \qquad\forall \ (\boldsymbol{h_{\beta}},\boldsymbol{h_{\delta}}) \in \mathcal{C}\left(\mathcal{S},\mathcal{R},\sqrt{\frac{\log n}{\log p}}\right).
    \label{eq:lower_bound_Ah}
    \end{equation}
    \item \textbf{Mutual Incoherence:} The column space of the matrix $\boldsymbol{A}$ is incoherent with the column space of the identity matrix. For some $\kappa_2>0$ with high probability,
    \begin{equation}\label{eq:upper_bound_mut_coh}
        \frac{1}{\sqrt{n}} |\langle\boldsymbol{Ah_{\beta}},\boldsymbol{h_{\delta}}\rangle| \leq \kappa_2 (\|\boldsymbol{h_{\beta}}\|_2+\|\boldsymbol{h_{\delta}}\|_2)^2 \ \qquad\forall \ (\boldsymbol{h_{\beta}},\boldsymbol{h_{\delta}}) \in \mathcal{C}\left(\mathcal{S},\mathcal{R},\sqrt{\frac{\log n}{\log p}}\right).
    \end{equation}
\end{enumerate}
 By using \eqref{eq:lower_bound_Ah} and \eqref{eq:upper_bound_mut_coh}, we have, with high probability,
 \begin{eqnarray*}
     \frac{1}{n}\|\boldsymbol{Ah_\beta+\sqrt{n}h_\delta}\|_2^2 &=& \frac{1}{n}\left\|\boldsymbol{Ah_{\beta}}\right\|_2^2+\|\boldsymbol{h_{\delta}}\|_2^2+\frac{2}{\sqrt{n}}\langle\boldsymbol{A h_{\beta}},\boldsymbol{h_{\delta}}\rangle \geq \kappa_1\left(\|\boldsymbol{h_{\beta}}\|_2+\|\boldsymbol{h_{\delta}}\|_2\right)^2 - 2\kappa_2(\|\boldsymbol{h_{\beta}}\|_2+\|\boldsymbol{h_{\delta}}\|_2)^2\\ &=& (\kappa_1-2\kappa_2)(\|\boldsymbol{h_{\beta}}\|_2+\|\boldsymbol{h_{\delta}}\|_2)^2
 \end{eqnarray*}
The proof is completed if $\kappa_1>2\kappa_2$. We now show that \eqref{eq:lower_bound_Ah} and \eqref{eq:upper_bound_mut_coh} hold together with $\kappa_1>2\kappa_2$ for a Rademacher sensing matrix $\boldsymbol{A}$.
\newline
We now state two important facts on the Rademacher matrix $\boldsymbol{A}$ which will be used in proving \eqref{eq:lower_bound_Ah} and \eqref{eq:upper_bound_mut_coh} respectively.
\begin{enumerate}[(1)]
    \item We use a result following Lemma 1  \cite{LiRaskutti} (see the equation immediately following Lemma 1 in \cite{LiRaskutti}, and set $\bar{D}$ in that equation to the identity matrix, since we are concerned with signals that are sparse in the canonical basis). Using this result, there exist positive constants $c_2,c'_3,c'_4$, such that with probability
at least $1-c'_3 \exp{\{-c'_4n\}}$:
\begin{equation}\label{eq:lwr_bnd_Rade}
    \frac{1}{\sqrt{n}}\|\boldsymbol{Ah_{\beta}}\|_2 \geq \frac{\|\boldsymbol{h_{\beta}}\|_2}{4}-c_2\sqrt{\frac{\log p}{n}}\|\boldsymbol{h_{\beta}}\|_1 \ \forall \ \boldsymbol{h_{\beta}} \in \mathbb{R}^p. 
\end{equation}
\item From Theorem 4.4.5 of \cite{Vershynin2018}, for a $s \times r'$ dimensional Rademacher matrix $\boldsymbol{A_{\mathcal{R}_i \mathcal{S}_j}}$, there exists a constant $c_1>0$ such that, for any $\tau'>0$, with probability at least $1-2\exp{\{-n\tau'^2\}}$ we have 
\begin{equation}\label{eq:max_sing_val}
   \frac{1}{\sqrt{n}} \|\boldsymbol{A_{\mathcal{R}_i \mathcal{S}_j}}\|_2 = \frac{1}{\sqrt{n}} \sigma_{max}({\boldsymbol{A_{\mathcal{R}_i \mathcal{S}_j}}}) \leq c_1 \left(\sqrt{\frac{s}{n}}+\sqrt{\frac{r'}{n}}+\tau'\right).
\end{equation}
\end{enumerate}
Throughout this proof, we take the constants $C_1 \triangleq \frac{7^2}{(24c_2)^2}$ and $C_2 \triangleq \max\{32^2 c_2^2,4 (51200c_1)^2\}$, where $c_1,c_2$ are as defined in \eqref{eq:lwr_bnd_Rade} and \eqref{eq:max_sing_val} respectively.

\noindent\textbf{Proof of \eqref{eq:lower_bound_Ah}:}
We first obtain a lower bound on $\frac{1}{n}\|\boldsymbol{Ah_{\beta}}\|_2^2$ using \eqref{eq:lwr_bnd_Rade}.
For every $(\boldsymbol{h_{\beta}},\boldsymbol{h_{\delta}}) \in \mathcal{C}\left(\mathcal{S},\mathcal{R},\sqrt{\frac{\log n}{\log p}}\right)$, we have:
\begin{equation}\label{eq:cone_cons_use_1}
    \|\boldsymbol{h_{\beta}}\|_1 \leq 4 \|\boldsymbol{(h_\beta)_{\mathcal{S}}}\|_1 + 3 \sqrt{\frac{\log n}{\log p}} \|\boldsymbol{(h_{\delta})_\mathcal{S}}\|_1 \leq 4 \sqrt{s} \|\boldsymbol{h_{\beta}}\|_2 +3 \sqrt{\frac{\log n}{\log p}} \sqrt{r} \|\boldsymbol{h_{\delta}}\|_2.
\end{equation}
The first inequality above follows from the definition of the set $\mathcal{C}$ in \eqref{eq:Coneconstraint} in the EREC using $\lambda := \sqrt{\frac{\log n}{\log p}}$. Substituting \eqref{eq:cone_cons_use_1} in \eqref{eq:lwr_bnd_Rade}, we obtain that, with probability at least $1-c'_3 \exp{\{-c'_4n\}}$, for every $(\boldsymbol{h_{\beta}},\boldsymbol{h_{\delta}}) \in \mathcal{C}\left(\mathcal{S},\mathcal{R},\sqrt{\frac{\log n}{\log p}}\right)$:
\begin{eqnarray}
   \nonumber \frac{1}{\sqrt{n}}\|\boldsymbol{Ah_{\beta}}\|_2 &\geq& \left(\frac{1}{4}- 4c_2\sqrt{\frac{s \log p}{n}}\right)\|\boldsymbol{h_{\beta}}\|_2 - 3c_2\sqrt{\frac{\log n}{\log p}} \sqrt{\frac{r \log p}{n}} \|\boldsymbol{h_{\delta}}\|_2, \\ \therefore \frac{1}{\sqrt{n}}\|\boldsymbol{Ah_{\beta}}\|_2 + \|\boldsymbol{h_{\delta}}\|_2 &\geq & \left(\frac{1}{4}- 4c_2\sqrt{\frac{s \log p}{n}}\right)\|\boldsymbol{h_{\beta}}\|_2 + \left(1-3c_2\sqrt{\frac{\log n}{\log p}} \sqrt{\frac{r \log p}{n}}\right) \|\boldsymbol{h_{\delta}}\|_2. \label{eq:both_lower}
\end{eqnarray}
Under the assumption $n \geq C_1 s \log p$, the first term in the brackets of \eqref{eq:both_lower} is greater than $\frac{1}{8}$. Again, under the assumption  $r \leq C_2 \frac{n}{ \log n}$, the second term is greater than $\frac{1}{8}$. Thus we have, $\frac{1}{\sqrt{n}}\|\boldsymbol{Ah_{\beta}}\|_2 + \|\boldsymbol{h_{\delta}}\|_2 \geq \frac{1}{8} (\|\boldsymbol{h_{\beta}}\|_2+\|\boldsymbol{h_{\delta}}\|_2)$. Squaring both sides, we have, $\frac{1}{n} \|\boldsymbol{Ah_{\beta}}\|_2^2 +\|\boldsymbol{h_{\delta}}\|_2^2 + \frac{2}{\sqrt{n}}\|\boldsymbol{Ah_{\beta}}\|_2\|\boldsymbol{h_{\delta}}\|_2 \geq \frac{1}{64} \left(\|\boldsymbol{h_{\beta}}\|_2+\|\boldsymbol{h_{\delta}}\|_2\right)^2$. Using the fact that $\|\boldsymbol{a}\|^2_2 + \|\boldsymbol{b}\|^2_2 \geq 2 \|\boldsymbol{a}\|_2 \|\boldsymbol{b}\|_2$ for any vectors $\boldsymbol{a}, \boldsymbol{b}$, we have, 
$2\left(\frac{1}{n} \|\boldsymbol{Ah_{\beta}}\|_2^2 +\|\boldsymbol{h_{\delta}}\|_2^2\right) \geq \frac{1}{n} \|\boldsymbol{Ah_{\beta}}\|_2^2 +\|\boldsymbol{h_{\delta}}\|_2^2 + \frac{2}{\sqrt{n}}\|\boldsymbol{Ah_{\beta}}\|_2\|\boldsymbol{h_{\delta}}\|_2 \geq \frac{1}{64} \left(\|\boldsymbol{h_{\beta}}\|_2+\|\boldsymbol{h_{\delta}}\|_2\right)^2$.
Hence we have with probability at least  $1-c'_3 \exp{\{-c'_4n\}}$, for every $(\boldsymbol{h_{\beta}},\boldsymbol{h_{\delta}}) \in \mathcal{C}\left(\mathcal{S},\mathcal{R},\sqrt{\frac{\log n}{\log p}}\right)$
\begin{equation}\label{eq:RE_A}
    \frac{1}{n} \|\boldsymbol{Ah_{\beta}}\|_2^2 +\|\boldsymbol{h_{\delta}}\|_2^2 \geq \frac{1}{128}(\|\boldsymbol{h_{\beta}}\|_2+\|\boldsymbol{h_{\delta}}\|_2)^2.
\end{equation}
Therefore, we have $\kappa_1=1/128$ completing the proof of \eqref{eq:lower_bound_Ah}.

\noindent\textbf{Proof of \eqref{eq:upper_bound_mut_coh}:} 
This part of the proof directly follows the proof of Lemma 2 in \cite{Nguyen2013}, with a few minor differences in constant factors. Nevertheless, we are including it here to make the paper self-contained. 

Divide the set $\{1,2,\ldots,p\}$ into subsets $\mathcal{S}_1,\mathcal{S}_2,\ldots,\mathcal{S}_q$
of size $s$ each, such that the first set $\mathcal{S}_1$ contains $s$ largest absolute value
entries of $\boldsymbol{h_{\beta}}$ indexed by $\mathcal{S}$, the set $\mathcal{S}_2$ contains $s$ largest absolute value
entries of the vector $\boldsymbol{({h_\beta})_{\mathcal{S}^c}}$, $\mathcal{S}_2$ contains the second largest $s$
absolute value entries of $\boldsymbol{({h_\beta})_{\mathcal{S}^c}}$, and so on. By the same strategy,
we also divide the set $\{1,2,\ldots,n\}$ into subsets $\mathcal{R}_1, \mathcal{R}_2 ,\ldots, \mathcal{R}_k$
such that the first set $\mathcal{R}_1$ contains $r$ entries of $\boldsymbol{h_{\delta}}$ indexed by $\mathcal{R}$
and sets $\mathcal{R}_2,\mathcal{R}_3,\ldots$ are of size $r'\geq r$.
We have for every $(\boldsymbol{h_{\beta}},\boldsymbol{h_{\delta}}) \in \mathcal{C}\left(\mathcal{S},\mathcal{R},\sqrt{\frac{\log n}{\log p}}\right)$,
\begin{align}\label{eq:sing_val}
    \frac{1}{\sqrt{n}}|\langle\boldsymbol{Ah_{\beta}},\boldsymbol{h_{\delta}}\rangle| \leq \sum_{i,j} \frac{1}{\sqrt{n}} |\langle\boldsymbol{A_{\mathcal{R}_i\mathcal{S}_j}}(\boldsymbol{h_\beta})_{\mathcal{S}_j},(\boldsymbol{h_\delta})_{\mathcal{R}_i}\rangle| &\leq \max_{i,j}  \frac{1}{\sqrt{n}} \|\boldsymbol{A_{\mathcal{R}_i\mathcal{S}_j}}\|_2 \sum_{i',j'} \|(\boldsymbol{h_\beta})_{\mathcal{S}_{j'}}\|_2 \|(\boldsymbol{h_\delta})_{\mathcal{R}_{i'}}\|_2\\
    & = \max_{i,j}  \frac{1}{\sqrt{n}} \|\boldsymbol{A_{\mathcal{R}_i\mathcal{S}_j}}\|_2 \left(\sum_{j'} \|(\boldsymbol{h_\beta})_{\mathcal{S}_{j'}}\|_2\right)\left(\sum_{i'} \|(\boldsymbol{h_\delta})_{\mathcal{R}_{i'}}\|_2\right).
\end{align}
Note that $\boldsymbol{A_{\mathcal{R}_i\mathcal{S}_j}}$ (a submatrix of $\boldsymbol{A}$ containing rows belonging to $\mathcal{R}_i$ and columns belonging to $\mathcal{S}_j$) is itself a Rademacher matrix with i.i.d. entries. 
Taking the union bound over all possible values of $\mathcal{S}_j$ and $\mathcal{R}_i$, we have that the inequality in \eqref{eq:max_sing_val} holds with probability at least $1-2\binom{n}{r'}\binom{p}{s}\exp{(-n\tau'^2)}$. If  $n \geq 4 {\tau'}^{-2}s \log(p)$ we obtain, $\binom{p}{s} \leq p^{s} \leq \exp({\tau'}^2n/4)$. Furthermore, if we assume, $n \geq 4 {\tau'}^{-2}r'\log(n)$, we have $\binom{n}{r'} \leq n^{r'} \leq \exp({\tau'}^2n/4)$. Later we will give a choice of $\tau'$ which ensures that these conditions are satisfied.
Therefore, we obtain with probability at least $1-2\exp{\{-n\tau'^2/2\}}$,
\begin{equation}\label{eq:union_max_sing}
     \textrm{max}_{i,j}  \frac{1}{\sqrt{n}} \|\boldsymbol{A_{\mathcal{R}_i \mathcal{S}_j}}\|_2 \leq c_1 \left(\sqrt{\frac{s}{n}}+\sqrt{\frac{r'}{n}}+\tau'\right).
\end{equation}
Using the first inequality in the last equation of Section 2.1 of \cite{Candes2006} we obtain $\sum_{i=3}^q \|\boldsymbol{(h_\beta)_{\mathcal{S}_i}}\|_2 \leq \frac{1}{\sqrt{s}} \|\boldsymbol{(h_\beta)_{\mathcal{S}^c}}\|_1$. Furthermore, for every $(\boldsymbol{h_{\beta}},\boldsymbol{h_{\delta}}) \in \mathcal{C}\left(\mathcal{S},\mathcal{R},\sqrt{\frac{\log n}{\log p}}\right)$, we have $  \|\boldsymbol{(h_\beta)_{\mathcal{S}^c}}\|_1 \leq 3 \sqrt{s} \|\boldsymbol{h_{\beta}}\|_2+ 3 \sqrt{\frac{\log n}{\log p}} \sqrt{r} \|\boldsymbol{h_{\delta}}\|_2$. Hence,
\begin{equation*}
    \sum_{i=1}^q \|\boldsymbol{(h_\beta)_{\mathcal{S}_i}}\|_2 = \|\boldsymbol{(h_\beta)_{\mathcal{S}_1}}\|_2+ \|\boldsymbol{(h_\beta)_{\mathcal{S}_2}}\|_2 + \sum_{i=3}^q \|\boldsymbol{(h_\beta)_{\mathcal{S}_i}}\|_2 \leq 2 \|\boldsymbol{h_{\beta}}\|_2 + \sum_{i=3}^q \|\boldsymbol{(h_\beta)_{\mathcal{S}_i}}\|_2 \leq 5 \|\boldsymbol{h_{\beta}}\|_2 + 3 \sqrt{\frac{\log n}{\log p}} \sqrt{\frac{r}{s}} \|\boldsymbol{h_{\delta}}\|_2.
\end{equation*}
Following a similar process we obtain $\sum_{i=3}^k \|\boldsymbol{(h_\delta)_{\mathcal{R}_i}}\|_2 \leq \frac{1}{\sqrt{r'}} \|\boldsymbol{(h_\delta)_{\mathcal{R}^c}}\|_1$. Furthermore, for every $(\boldsymbol{h_{\beta}},\boldsymbol{h_{\delta}}) \in \mathcal{C}\left(\mathcal{S},\mathcal{R},\sqrt{\frac{\log n}{\log p}}\right)$, we have $  \frac{1}{\sqrt{r'}}\|\boldsymbol{(h_\delta)_{\mathcal{R}^c}}\|_1 \leq 3 \sqrt{\frac{s}{r'}}\frac{1}{\sqrt{\frac{\log n}{\log p}}} \|\boldsymbol{h_{\beta}}\|_2+ 3 \sqrt{\frac{r}{r'}} \|\boldsymbol{h_{\delta}}\|_2$. Since $r'\geq r$,
\begin{equation*}
    \sum_{i=1}^k \|\boldsymbol{(h_\delta)_{\mathcal{R}_i}}\|_2 = \|\boldsymbol{(h_\delta)_{\mathcal{R}_1}}\|_2+ \|\boldsymbol{(h_\delta)_{\mathcal{R}_2}}\|_2 + \sum_{i=3}^k \|\boldsymbol{(h_\delta)_{\mathcal{R}_i}}\|_2 \leq 2 \|\boldsymbol{h_{\delta}}\|_2 + \sum_{i=3}^k \|\boldsymbol{(h_\delta)_{\mathcal{R}_i}}\|_2 \leq 5 \|\boldsymbol{h_{\delta}}\|_2 + \frac{3}{\sqrt{\frac{\log n}{\log p}}} \sqrt{\frac{s}{r'}} \|\boldsymbol{h_{\beta}}\|_2.
\end{equation*}
Hence, joining \eqref{eq:union_max_sing} , \eqref{eq:max_sing_val} into \eqref{eq:sing_val}, we obtain with probability at least $1-2\exp{\{-n\tau'^2/2\}}$, for every $(\boldsymbol{h_{\beta}},\boldsymbol{h_{\delta}}) \in \mathcal{C}\left(\mathcal{S},\mathcal{R},\sqrt{\frac{\log n}{\log p}}\right)$ 
\begin{eqnarray}
    \frac{1}{\sqrt{n}}|\langle\boldsymbol{Ah_{\beta}},\boldsymbol{h_{\delta}}\rangle| \leq c_1 \left(\sqrt{\frac{s}{n}}+\sqrt{\frac{r'}{n}}+\tau'\right) \times \left(5 \|\boldsymbol{h_{\beta}}\|_2 + 3 \sqrt{\frac{\log n}{\log p}} \sqrt{\frac{r}{s}} \|\boldsymbol{h_{\delta}}\|_2\right) \times \left(5 \|\boldsymbol{h_{\delta}}\|_2 + \frac{3}{\sqrt{\frac{\log n}{\log p}}} \sqrt{\frac{s}{r'}} \|\boldsymbol{h_{\beta}}\|_2\right).
\end{eqnarray}
Recall that $r\leq \frac{s\log p}{\log n}$, by assumption. Taking $r'=\frac{s \log p}{ \log n}$  leads to $\sqrt{\frac{\log n}{\log p}} \sqrt{\frac{r}{s}}\leq \sqrt{\frac{\log n}{\log p}} \sqrt{\frac{r'}{s}} = 1$ and $\frac{1}{\sqrt{\frac{\log n}{\log p}}} \sqrt{\frac{s}{r'}}=1$. Thus, we obtain with probability at least $1-2\exp({-n{\tau'^2}/2})$ for every $(\boldsymbol{h_{\beta}},\boldsymbol{h_{\delta}}) \in \mathcal{C}\left(\mathcal{S},\mathcal{R},\sqrt{\frac{\log n}{\log p}}\right)$,
\begin{equation}
    \frac{1}{\sqrt{n}}|\langle\boldsymbol{Ah_{\beta}},\boldsymbol{h_{\delta}}\rangle| \leq 25c_1 \left(\sqrt{\frac{s}{n}}+\sqrt{\frac{r'}{n}}+\tau'\right) \times (\|\boldsymbol{h_{\beta}}\|_2+\|\boldsymbol{h_{\delta}}\|_2)^2
\end{equation}
Let $\tau' \triangleq 1/(51200c_1)$. Recall that, $C_1 \triangleq \max\{32^2 c_2^2,4 (51200c_1)^2\}$. Then $n \geq C_1 s \log p$ implies $n \geq 4 {\tau'}^{-2} s \log p = 4 {\tau'}^{-2} r' \log n$. Furthermore, \begin{equation}
    \sqrt{\frac{s}{n}} \leq \sqrt{\frac{r'}{n}}  =\sqrt{\frac{s\log p}{n \log n}} \leq \tau'/2.
\end{equation}
Therefore, we have with probability at least $1-2\exp({-n{\tau'^2}/2})$, for every $(\boldsymbol{h_{\beta}},\boldsymbol{h_{\delta}}) \in \mathcal{C}\left(\mathcal{S},\mathcal{R},\sqrt{\frac{\log n}{\log p}}\right)$
\begin{equation} \label{eq:mutual_incoherence}
    \frac{1}{\sqrt{n}}|\langle\boldsymbol{Ah_{\beta}},\boldsymbol{h_{\delta}}\rangle| \leq 25c_1 \times 2\tau' (\|\boldsymbol{h_{\beta}}\|_2+\|\boldsymbol{h_{\delta}}\|_2)^2 \leq \frac{1}{512} (\|\boldsymbol{h_{\beta}}\|_2+\|\boldsymbol{h_{\delta}}\|_2)^2
\end{equation}
This completes the proof of \eqref{eq:upper_bound_mut_coh}.

Now, from \eqref{eq:RE_A} and \eqref{eq:mutual_incoherence}, using a union bound, we obtain with probability at least $1-(c'_3\exp(-c'_4n)+2\exp(-n{\tau'^2}/2))$, 
\begin{equation} \label{eq:squared_EREC}
    \frac{1}{n}\|\boldsymbol{Ah_\beta+\sqrt{n}h_\delta}\|_2^2 \geq (\kappa_1-2\kappa_2) (\|\boldsymbol{h_{\beta}}\|_2+\|\boldsymbol{h_{\delta}}\|_2)^2 = \kappa^2 (\|\boldsymbol{h_{\beta}}\|_2+\|\boldsymbol{h_{\delta}}\|_2)^2
\end{equation}
Taking $c_3=c'_3+2$ and $c_4=\min\{c'_4,{\tau'^2}/2\}$, we have $ 1-(c'_3\exp(-c'_4n)+2\exp(-{\tau'^2}n/2)\geq 1-c_3\exp(-c_4n)$. 

Note that, we have, $\kappa=\sqrt{\kappa_1-2\kappa_2} =1/16$.
Taking the root over \eqref{eq:squared_EREC}, we obtain with probability at least $1-c_3\exp(-c_4n)$,
\begin{equation*}
    \frac{1}{\sqrt{n}}\|\boldsymbol{Ah_\beta+\sqrt{n}h_\delta}\|_2 \geq \frac{1}{16} (\|\boldsymbol{h_{\beta}}\|_2+\|\boldsymbol{h_{\delta}}\|_2) \ \forall \ (\boldsymbol{h_{\beta}},\boldsymbol{h_{\delta}}) \in \mathcal{C}\left(\mathcal{S},\mathcal{R},\sqrt{\frac{\log n}{\log p}}\right) .
\end{equation*}
This completes the proof of the lemma.\hfill{$\blacksquare$}
\subsection{Proof of Theorem~\ref{th:upper_bound_robustLasso}}
\noindent\textbf{Overview:} We first establish bounds on $\|\hat{\boldsymbol{\beta}}_{\lambda_1} - \boldsymbol{\beta^*}\|_1$ and $\|\hat{\boldsymbol{\delta}}_{\lambda_2} - \boldsymbol{\delta^*}\|_1$. Defining $\boldsymbol{h_\beta} = \boldsymbol{\hat{\beta}}_{\lambda_1} - \boldsymbol{\beta^*}$ and $\boldsymbol{h_\delta} = \frac{1}{n}(\boldsymbol{\hat{\delta}}_{\lambda_2} - \boldsymbol{\delta^*})$, they satisfy the cone constraint:
\begin{equation*}
    \|(\boldsymbol{h_\beta})_{S^c}\|_1 + \frac{\tilde{\lambda}_2}{\lambda_1} \|(\boldsymbol{h_\delta})_{R^c}\|_1 \leq 3\Big(\|(\boldsymbol{h_\beta})_S\|_1 + \frac{\tilde{\lambda}_2}{\lambda_1} \|(\boldsymbol{h_\delta})_R\|_1\Big),
\end{equation*}
where $S$ and $R$ are the support sets of $\boldsymbol{\beta^*}$ and $\boldsymbol{\delta^*}$, respectively. Using norm inequalities and known bounds, we derive
\begin{equation*}
    \|\boldsymbol{h_\beta}\|_1 \leq 12 \kappa^{-2} (s + r) \lambda_1.
\end{equation*}
Applying Gaussian tail bounds, we obtain the high-probability bound:
\begin{equation*}
P\Bigg(\|\boldsymbol{\hat{\beta}_{\lambda_1}} - \boldsymbol{\beta^*}\|_1 \leq 48 \kappa^{-2} (s + r) \sigma \sqrt{\frac{\log p}{n}}\Bigg) \geq 1 - \frac{1}{n} - \frac{1}{p}.
\end{equation*}
For $\|\boldsymbol{\hat{\delta}_{\lambda_2}} - \boldsymbol{\delta^*}\|_1$, we consider the \textsc{Lasso} estimator for $\boldsymbol{\delta^*}$ with noise vector $\boldsymbol{\varrho} := \boldsymbol{A}(\boldsymbol{\beta^*} - \boldsymbol{\hat{\beta}_{\lambda_1}}) + \boldsymbol{\eta}$. Given $\lambda_2 \geq 2 \|\boldsymbol{\varrho}\|_\infty / n$, results from \cite{THW2015} yield:
\begin{equation*}
\|\boldsymbol{\hat{\delta}_{\lambda_2}} - \boldsymbol{\delta^*}\|_1 \leq 12 r \lambda_2.
\end{equation*}
Using high-probability bounds on $\|\boldsymbol{\rho}\|_\infty$, we conclude:
\begin{equation*}    P\Bigg(\|\boldsymbol{\hat{\delta}_{\lambda_2}} - \boldsymbol{\delta^*}\|_1 \leq 48 r \sigma \frac{\sqrt{\log n}}{n}\Bigg) \geq 1 - \frac{2}{n} - \frac{1}{p}.
\end{equation*}

\noindent \textbf{Proof of \eqref{eq:beta_lasso_bound}:}
 We now derive the bound for the $l_1$ norm of the robust \textsc{Lasso} estimate of the error $\boldsymbol{\hat{\beta}_{\lambda_1}}-\boldsymbol{\beta^*}$ given by the optimization problem \eqref{eq:lasso_delta}. {Recall that we have $\lambda_1=\frac{4\sigma\sqrt{\log p}}{\sqrt{n}}$ and $\lambda_2=\frac{4\sigma\sqrt{\log n}}{{n}}$. We choose $\tilde{\lambda}_2\triangleq \sqrt{n}\lambda_2=\frac{4\sigma\sqrt{\log n}}{\sqrt{n}}$. We use $\tilde{\lambda}_2$ to define the cone constraint in \eqref{eq:Coneconstraint}.} Note that, in the proof of Theorem 1 of \cite{Nguyen2013}, it is shown that $\boldsymbol{h_\beta}\triangleq \boldsymbol{\hat{\beta}_{\lambda_1}-\beta^*}$ and $\boldsymbol{h_\delta}\triangleq\frac{1}{\sqrt{n}}(\boldsymbol{\hat{\delta}_{\lambda_2}-\delta^*})$ satisfies the cone constraint given in \eqref{eq:Coneconstraint}. Therefore, we have 
\begin{equation}\label{eq:cone_cons}
\|(\boldsymbol{h_{\beta}})_{\mathcal{S}^{c}}\|_1+\frac{\tilde{\lambda}_2}{\lambda_1} \|(\boldsymbol{h_{\delta}})_{\mathcal{R}^{c}}\|_1 \leq  3(\|(\boldsymbol{h_{\beta}})_{\mathcal{S}}\|_1+\frac{\tilde{\lambda}_2}{\lambda_1} \|(\boldsymbol{h_{\delta}})_{\mathcal{R}}\|_1).
\end{equation}
Now by using Eqn.~\eqref{eq:cone_cons}, we have
\begin{eqnarray} \label{eq:l1_beta_cone}
    \|\boldsymbol{h_{\beta}}\|_1 &=& \|\boldsymbol{(h_\beta)_{\mathcal{S}}}\|_1+\|\boldsymbol{(h_\beta)_{\mathcal{S}^c}}\|_1 \leq 4\|\boldsymbol{(h_\beta)_{\mathcal{S}}}\|_1 +3\frac{\tilde{\lambda}_2}{\lambda_1} \|\boldsymbol{(h_{\delta})_{\mathcal{R}}}\|_1 \leq 4\sqrt{s}\|\boldsymbol{h_{\beta}}\|_2 +3\sqrt{r}\frac{\tilde{\lambda}_2}{\lambda_1} \|\boldsymbol{h_{\delta}}\|_2.
\end{eqnarray}
Here, the last inequality of Eqn.\eqref{eq:l1_beta_cone} holds since $\left\|\boldsymbol{(h_\beta)_{\mathcal{S}}}\right\|_1 \leq \sqrt{s}\|\boldsymbol{h_{\beta}}\|_2$ and $\left\|\boldsymbol{(h_\delta)_{\mathcal{R}}}\right\|_1 \leq \sqrt{r}\|\boldsymbol{h_{\delta}}\|_2$.
Note that, $\max\{\sqrt{s},\sqrt{r}\}\leq \sqrt{s+r}$. {Based on the values of $\lambda_1,\tilde{\lambda}_2$, we have $\tilde{\lambda}_2<\lambda_1$ since $n<p$.} Hence, by using  Eqn.\eqref{eq:l1_beta_cone}, we have
\begin{eqnarray}
    \|\boldsymbol{h_{\beta}}\|_1  \leq 4\sqrt{s}\|\boldsymbol{h_{\beta}}\|_2 +3\sqrt{r} \|\boldsymbol{h_{\delta}}\|_2 \leq 4\sqrt{s+r}\left(\|\boldsymbol{h_{\beta}}\|_2+\|\boldsymbol{h_{\delta}}\|_2\right). \label{eq:l1_l2_sparsity}
\end{eqnarray}
Recall that, $\boldsymbol{e^*}=\frac{\boldsymbol{\delta^*}}{\sqrt{n}}$ and $\boldsymbol{\hat{e}}=\frac{\boldsymbol{\hat{\delta}_{\tilde{\lambda}_2}}}{\sqrt{n}}$ in Theorem 1 of \cite{Nguyen2013}. Therefore, by the equivalence of the model given in \eqref{eq:noisevector2} and the optimisation problem in \eqref{eq:lasso_delta} with that of \cite{Nguyen2013}, we have 
\begin{equation}\label{eq:upper_bnd}      \|\boldsymbol{\hat{\beta}_{\lambda_1}-\beta^*}\|_2 +\left\|\frac{1}{\sqrt{n}}(\boldsymbol{\hat{\delta}_{\tilde{\lambda}_2}-\delta^*})\right\|_2 \leq 3\kappa^{-2} \textrm{max}\{\lambda_1 \sqrt{s},\tilde{\lambda}_2 \sqrt{r} \} ,
        \end{equation}
as long as 
\begin{equation}\label{eq:cond_lambda}
     \dfrac{2\|\boldsymbol{A}^{\top} \boldsymbol{\eta}\|_{\infty}}{n} \leq \lambda_1 , \ \text{and} \ \dfrac{2\|\boldsymbol{\eta}\|_{\infty}}{\sqrt{n}} \leq \tilde{\lambda}_2.
\end{equation}

Therefore when 
\eqref{eq:cond_lambda} holds, then 
by using \eqref{eq:l1_l2_sparsity} (recall $\boldsymbol{h_{\beta}}=\boldsymbol{\hat{\beta}_{\lambda_1}-\beta^*}$) and \eqref{eq:upper_bnd}, we have 
\begin{eqnarray}\label{eq:bound_with_lambda}
    \|\boldsymbol{\hat{\beta}_{\lambda_1}-\beta^*}\|_1 
\leq 4\sqrt{s+r} \left(3\kappa^{-2} \textrm{max}\{\lambda_1 \sqrt{s},\tilde{\lambda}_2 \sqrt{r} \}\right) \leq 12 \kappa^{-2} (s+r) \textrm{max}\{\lambda_1,\tilde{\lambda}_2\} \leq 12 \kappa^{-2} (s+r) \lambda_1.
\end{eqnarray} 
{We will now bound the probability that $\dfrac{2\|\boldsymbol{A}^{\top} \boldsymbol{\eta}\|_{\infty}}{n} \leq \lambda_1$ and $\dfrac{2\|\boldsymbol{\eta}\|_{\infty}}{\sqrt{n}} \leq \tilde{\lambda}_2$ using the fact that $\boldsymbol{\eta}$ is Gaussian.}
By using Lemma~\ref{le:ma_Gaussian} in Appendix \ref{sec:App_A3} which describes the tail bounds of the Gaussian vector, we have
\begin{eqnarray}
    P(2\|\boldsymbol{\eta}\|_{\infty}/\sqrt{n} \leq 4\sigma\sqrt{\log(n)/n}) \geq 1-\frac{1}{n} \label{eq:lambda_2} , \\
    P\left(2\left\|\frac{1}{\sqrt{n}}\boldsymbol{A}^{\top}\boldsymbol{\eta}\right\|_{\infty}/\sqrt{n} \leq 4\sigma\sqrt{\log(p)/n}\right) \geq 1-\frac{1}{p} \label{eq:lambda_1}.
\end{eqnarray}
Using \eqref{eq:lambda_2},\eqref{eq:lambda_1}  with Bonferroni's inequality in \eqref{eq:bound_with_lambda}, we have:
\begin{eqnarray}\label{eq:beta_bnd_l1}
    P\left(\|\boldsymbol{\hat{\beta}_{\lambda_1}}-\boldsymbol{\beta^*}\|_1 \leq 48\kappa^{-2} (s+r)\sigma\sqrt{\frac{{\log(p)}}{{n}}}\right) \geq 1 - \frac{1}{n}-\frac{1}{p}. 
\end{eqnarray}
This completes the proof of \eqref{eq:beta_lasso_bound}.
\newline
\textbf{Proof of \eqref{eq:delta_lasso_bound}:}
We now derive an upper bound of $\|\boldsymbol{\hat{\delta}_{\lambda_2}}-\boldsymbol{\delta^{*}}\|_{1}$. We approach this by showing that given the optimal estimate of $\boldsymbol{\hat{\beta}}_{\lambda_1}$, we can obtain a unique estimate of $\boldsymbol{\hat{\delta}}_{\lambda_2}$ using \eqref{eq:delta_lasso_est}. We then derive the upper bound on $\|\boldsymbol{\hat{\delta}_{\lambda_2}}-\boldsymbol{\delta^{*}}\|_{1}$ using the \textsc{Lasso} bounds given in \cite{THW2015}. 
Expanding the terms in \eqref{eq:fm_delta}, we obtain:
\begin{eqnarray}
{\min}_{\boldsymbol{\beta},\boldsymbol{\delta}} {\frac{1}{2n}}\left\|\boldsymbol{y}-\boldsymbol{A\beta}-\boldsymbol{\delta}\right\|^2_2 + \lambda_1 \|\boldsymbol{\beta}\|_1 + \lambda_2 \left\|\boldsymbol{\delta}\right\|_1 = {\min}_{\boldsymbol{\beta}}\left\{\lambda_1\|\boldsymbol{\beta}\|_1 +\sum_{i=1}^{n} {\min}_{\delta_{i}}\left\{\frac{1}{2n}((y_{i}-\boldsymbol{a_{i.}\beta})-\delta_{i})^2+{\lambda_2}|\delta_{i}|\right\}\right\} \label{eq:step-wise_estimate}.
\end{eqnarray}
Given the optimal solutions $\boldsymbol{\hat{\beta}_{\lambda_1}}$ and $\boldsymbol{\hat{\delta}_{\lambda_2}}$ of \eqref{eq:lasso_delta}, $\boldsymbol{\hat{\delta}_{\lambda_2}}$ can also be viewed as 
\begin{eqnarray}\label{eq:delta_lasso_est}
& &\boldsymbol{\hat{\delta}_{\lambda_2}} = \textrm{argmin}_{\boldsymbol{\delta}}\frac{1}{2n}\sum_{i=1}^{n}\{(y_{i}-\boldsymbol{a_{i.}}\boldsymbol{\hat{\beta}_{\lambda_1}}-\delta_{i})^2\}+\lambda_2\|\boldsymbol{\delta}\|_{1} 
\end{eqnarray}
Thus \eqref{eq:delta_lasso_est} can also be viewed as \textsc{Lasso} estimator for 
$\boldsymbol{z}=\boldsymbol{I_n\delta^{*}}+\boldsymbol{\varrho}$, where $\boldsymbol{z}\triangleq \boldsymbol{y} -\boldsymbol{A}^{\top}\boldsymbol{\hat{\beta}_{\lambda_1}}$ and $\boldsymbol{\varrho}\triangleq \boldsymbol{A}(\boldsymbol{\beta^{*}}-\boldsymbol{\hat{\beta}_{\lambda_1})+\boldsymbol{\eta}}$ with $\boldsymbol{\delta^*}$ being $r$-sparse.
By using Theorem 11.1(b) of \cite{THW2015} , we have {if $\lambda_2 \geq 2\frac{\|\boldsymbol{\varrho}\|_{\infty}}{n}$, } 
\begin{equation}\label{eq:tib_las_bnd}
\left\|\boldsymbol{\hat{\delta}_{\lambda_2}}-\boldsymbol{\delta^{*}}\right\|_2 \leq \dfrac{3\sqrt{r}\lambda_2}{\gamma_{r}} ,   
\end{equation}
where $\gamma_{r}$ is the Restricted eigenvalue constant of order $r$ which equals one for $\boldsymbol{I_n}$. 
Now using the result in Lemma 11.1 of \cite{THW2015}, when $\lambda_2 \geq 2\frac{\|\boldsymbol{\varrho}\|_{\infty}}{n}$, then 
\begin{equation*}
   \left\|(\boldsymbol{\hat{\delta}_{\lambda_2}}-\boldsymbol{\delta^{*}})_{\mathcal{R}^c}\right\|_1 \leq 3\left\|(\boldsymbol{\hat{\delta}_{\lambda_2}}-\boldsymbol{\delta^{*}})_{\mathcal{R}}\right\|_1. 
\end{equation*}
Therefore by using \eqref{eq:tib_las_bnd} when $\lambda_2 \geq 2\frac{\|\boldsymbol{\varrho}\|_{\infty}}{n}$, we have 
\begin{eqnarray}\label{eq:delta_l1_l2}
    \|\boldsymbol{\hat{\delta}_{\lambda_2}}-\boldsymbol{\delta^{*}}\|_1 &=&\left\|(\boldsymbol{\hat{\delta}_{\lambda_2}}-\boldsymbol{\delta^{*}})_{\mathcal{R}^c}\right\|_1 + \left\|(\boldsymbol{\hat{\delta}_{\lambda_2}}-\boldsymbol{\delta^{*}})_{\mathcal{R}}\right\|_1 \leq 4\left\|(\boldsymbol{\hat{\delta}_{\lambda_2}}-\boldsymbol{\delta^{*}})_{\mathcal{R}}\right\|_1 \leq 4\sqrt{r}\left\|(\boldsymbol{\hat{\delta}_{\lambda_2}}-\boldsymbol{\delta^{*}})_{\mathcal{R}}\right\|_2 \leq 4\sqrt{r}\|\boldsymbol{\hat{\delta}_{\lambda_2}}-\boldsymbol{\delta^{*}}\|_2\nonumber\\
    &\le& 12r \lambda_2.
\end{eqnarray}
Therefore we now show that $\lambda_2\left(=4\sigma\frac{\sqrt{\log n}}{n}\right) \geq 2\frac{\|\boldsymbol{\varrho}\|_{\infty}}{n}$ (i.e., $\|\boldsymbol{\varrho}\|_{\infty}\le 2\sigma\sqrt{\log n}$) holds with high probability.
Now, by Lemma~\ref{le:mat_Op_properties} and the triangle inequality, we have
\begin{equation*}
\label{eq:8a}
\|\boldsymbol{\varrho}\|_{\infty}=\|\boldsymbol{A}(\boldsymbol{\beta^{*}}-\boldsymbol{\hat{\beta}_{\lambda_1}})+\boldsymbol{\eta}\|_{\infty} 
    \leq |\boldsymbol{A}|_{\infty}\|\boldsymbol{\beta^{*}}-\boldsymbol{\hat{\beta}_{\lambda_1}}\|_{1}+\|\boldsymbol{\eta}\|_{\infty}.
\end{equation*}
By using Lemma~\ref{le:ma_Gaussian} in Appendix \ref{sec:App_A3}, we have the following:
\begin{equation} 
    P(\|\boldsymbol{\eta}\|_{\infty} \geq \sigma\sqrt{\log(n)}) \leq \frac{1}{n}.
\end{equation}
Since  $|\boldsymbol{A}|_{\infty}\leq 1$,  by using \eqref{eq:beta_bnd_l1}, we have 
{\begin{eqnarray}\label{eq:varrho}
   P\left(\|\boldsymbol{\varrho}\|_{\infty}\leq 48\kappa^{-2} (s+r) \sigma \sqrt{\frac{\log(p)}{n}}+\sigma\sqrt{\log(n)}\right) \geq 1-\left(\frac{2}{n}+\frac{1}{p}\right).
\end{eqnarray}
 Since $n \log n \geq (48 \kappa^{-2})^2 (s+r)^2 \log p$, we have $48\kappa^{-2} (s+r) \sigma \sqrt{\frac{\log(p)}{n}}<\sigma\sqrt{\log(n)}$. Thus 
\begin{equation}\label{eq:varrho_bnd}
P(\|\boldsymbol{\varrho}\|_{\infty}\le 2\sigma\sqrt{\log n})\ge 1-\left(\frac{2}{n}+\frac{1}{p}\right).
\end{equation}
We now put \eqref{eq:varrho_bnd} in \eqref{eq:delta_l1_l2} to obtain:
\begin{eqnarray}\label{eq:delta_bnd_l1}
 P\left(\|\boldsymbol{\hat{\delta}_{\lambda_2}}-\boldsymbol{\delta^{*}}\|_1 \leq \frac{24\sigma r \sqrt{\log n}}{n}\right) \geq 1-\left(\frac{2}{n}+\frac{1}{p}\right).
\end{eqnarray} 
This completes the proof.
 \hfill{$\blacksquare$}
 
\section{Proofs of Theorems and Lemmas on Debiased Lasso}\label{sec:App_A2}
\subsection{Proof of Theorem~\ref{th:dist_debiased_beta_delta_A}} 
\textbf{Overview:} {The proof proceeds by leveraging the debiased estimator formulation and analyzing its asymptotic behavior. First, setting $\boldsymbol{W} = \boldsymbol{A}$, we express the deviation of $\boldsymbol{\hat{\beta}_A}$ from $\boldsymbol{\beta^*}$ as a sum of three terms: a Gaussian noise term $\frac{1}{n} \boldsymbol{A}^\top \boldsymbol{\eta}$, a bias term involving $\boldsymbol{\hat{\beta}_{\lambda_1}} - \boldsymbol{\beta^*}$, and an additional bias term involving $\boldsymbol{\delta^*} - \boldsymbol{\hat{\delta}_{\lambda_2}}$ as in \eqref{eq:beta_d_unscaled} and \eqref{eq:delta_d_unscaled}. Results from Lemma~\ref{le:2_3term_beta_decompose} (this is similar to Theorem~\ref{th:2_3term_beta_decompose_W} with the important difference that $\boldsymbol{W} = \boldsymbol{A}$) show that the latter two terms vanish in probability as $n, p$ grow large, leaving only the Gaussian term, which establishes that $\sqrt{n}(\hat{\beta}_{Aj} - \beta^*_j) \sim \mathcal{N}(0, \sigma^2)$. The second part of the proof follows a similar decomposition for $\boldsymbol{\hat{\delta}_A} - \boldsymbol{\delta^*}$, showing that the bias terms are negligible and that the dominant term is $(\boldsymbol{I_n} - \frac{1}{n} \boldsymbol{AA}^\top) \boldsymbol{\eta}$, which is Gaussian with covariance matrix $\boldsymbol{\Sigma_{A}}= \left(\boldsymbol{I_n}-\frac{1}{n}\boldsymbol{AA}^{\top}\right)\left(\boldsymbol{I_n}-\frac{1}{n}\boldsymbol{AA}^{\top}\right)^{\top}$. The negligible bias terms vanish, and an argument using Lemma~\ref{le:G_conv_prob} shows that the variance of the $i^{\text{th}}$ term $\frac{n^2}{p(p-n)} \Sigma_{A_{ii}}$ (note the correction factor in this expression) converges to 1 in probability, leading to the final asymptotic normality result for $\boldsymbol{\hat{\delta}_A}$.
Given this overview, we work through the two main steps here below, in detail.}

Note that we have chosen $\boldsymbol{W}=\boldsymbol{A}$. Recalling the expression for $\boldsymbol{\hat{\beta}_W}$ from \eqref{eq:deb_beta_est} and model as given in \eqref{eq:fm_delta} and setting $\boldsymbol{W} = \boldsymbol{A}$, we have
\begin{eqnarray}
\boldsymbol{\hat{\beta}_A}-\boldsymbol{\beta^*}&=&\frac{1}{n}\boldsymbol{A}^{\top}\boldsymbol{{\eta}}+ \left(\boldsymbol{I_p}-\frac{1}{n}\boldsymbol{A^{\top}}\boldsymbol{A}\right)  (\boldsymbol{\hat{\beta}_{\lambda_1}}-\boldsymbol{\beta^*})+
\frac{1}{n}\boldsymbol{A}^{\top}
\left(\boldsymbol{\delta^{*}}-\boldsymbol{\hat{\delta}_{\lambda_2}}\right).\label{eq:beta_d_decompose}
\end{eqnarray}
In Lemma~\ref{le:2_3term_beta_decompose}, \eqref{eq:bias_A_beta1} and \eqref{eq:bias_A_beta2} show that the second and third term on the RHS of \eqref{eq:beta_d_decompose} are negligible as $n$, $p$ increases in probability. Therefore, in view of Lemma~\ref{le:2_3term_beta_decompose},  we have \begin{eqnarray}\label{beta_marginal}
\sqrt{n}(\hat{\beta}_{Aj}-\beta^*_j)&=&\frac{1}{\sqrt{n}}\boldsymbol{a}_{. j}^{\top}\boldsymbol{{\eta}}+ o_P(1),
\end{eqnarray}
where $\boldsymbol{a}_{.j}$ denotes the $j^{\text{th}}$ column of matrix $\boldsymbol{A}$. Given $\boldsymbol{a_{.j}}$, by using the Gaussianity of $\boldsymbol{\eta}$, the first term on the RHS of \eqref{beta_marginal} is a Gaussian random variable with mean $0$ and variance $\sigma^2\frac{ \boldsymbol{a}_{.j}^{\top} \boldsymbol{a}_{.j}}{n}$. Since $\frac{ \boldsymbol{a}_{.j}^{\top} \boldsymbol{a}_{.j}}{n} = 1$, the first term on the RHS is $\mathcal{N}(0,\sigma^2)$. This completes the proof of result (1) of the theorem.   

We now turn to result (2) of the theorem. By using a similar decomposition argument as in the case of $\boldsymbol{\hat{\beta}_A}$ in \eqref{eq:beta_d_decompose} and using the expression of $\boldsymbol{\hat{\delta}_A}$ in \eqref{eq:deb_delta_est}, we have 
\begin{eqnarray}
 \boldsymbol{\hat{\delta}_A}-\boldsymbol{\delta^*}=
\left(\boldsymbol{I_n}-\frac{1}{n}\boldsymbol{A}\boldsymbol{A}^{\top}\right)
\boldsymbol{{\eta}}+
\left(\boldsymbol{I_n}-\frac{1}{n}\boldsymbol{A}\boldsymbol{A}^{\top}\right) \boldsymbol{A} (\boldsymbol{\beta^*-\hat{\beta}_{\lambda_1}})-
\frac{1}{n}\boldsymbol{AA}^{\top}
\big(\boldsymbol{\delta^{*}}-\boldsymbol{\hat{\delta}_{\lambda_2}}\big). \label{eq:delta_d_decompose}
\end{eqnarray}
We define $\boldsymbol{\Sigma_{A}}= \left(\boldsymbol{I_n}-\frac{1}{n}\boldsymbol{AA}^{\top}\right)\left(\boldsymbol{I_n}-\frac{1}{n}\boldsymbol{AA}^{\top}\right)^{\top}$. We note that $\forall i \in [n], \Sigma_{A_{ii}} = 1-\frac{2p}{n}
     +\frac{1}{n^2}\boldsymbol{a_{i.}A^{\top}A}\boldsymbol{a_{i.}}^{\top}$. From \eqref{eq:bias_A_delta1} and \eqref{eq:bias_A_delta2} of Lemma~\ref{le:2_3term_beta_decompose}, the second and third terms on the RHS of \eqref{eq:delta_d_decompose} are both $o_P\left(\frac{p\sqrt{1-n/p}}{n}\right)$. Therefore, using Lemma~\ref{le:2_3term_beta_decompose}, we have  for any $i \in [n]$
\begin{eqnarray}
 \frac{\left({\hat{\delta}_{Ai}}-{\delta^*_i}\right)}{\sqrt{\Sigma_{A_{ii}}}}=
\frac{\left(\boldsymbol{I_n}-\frac{1}{n}\boldsymbol{AA}^{\top}\right)^\top_{i.}
\boldsymbol{{\eta}}}{\sqrt{\Sigma_{A_{ii}}}}+o_P\left(\frac{1}{\sqrt{\Sigma_{A_{ii}}}}\frac{p\sqrt{1-n/p}}{n}\right).\label{delta_W_dist}
\end{eqnarray}
As $\boldsymbol{\eta}$ is Gaussian, the  first term on the RHS of \eqref{delta_W_dist} is a Gaussian random variable with mean $0$ and variance $\sigma^2$.
In Lemma~\ref{le:G_conv_prob}, we show that $\frac{n^2}{p(p-n)} \Sigma_{A_{ii}}$ converges to $1$ in probability if $\boldsymbol{A}$ is a Rademacher matrix. This implies that the second term on the RHS of \eqref{delta_W_dist} is $o_P(1)$. This completes the proof of result (2).   
\hfill{$\blacksquare$}
\newline
\begin{lemma} \label{le:2_3term_beta_decompose} Let  $\boldsymbol{\hat{\beta}}_{\boldsymbol{\lambda_1}},\boldsymbol{\hat{\delta}_{\lambda_2}}$ be as in \eqref{eq:lasso_delta} and set $\lambda_1 \triangleq \frac{4\sigma\sqrt{\log p}}{\sqrt{n}}, \lambda_2 \triangleq \frac{4\sigma\sqrt{\log n}}{{n}}$. Given $\boldsymbol{A}$ is a Rademacher matrix, if $n$ is $o(p)$ and $n$ is $\omega[((s+r)\log p)^2]$, then as $n,p \to \infty$ we have following:
\begin{eqnarray}
     \left\|\sqrt{n}\left(\boldsymbol{I_p}-\frac{1}{n}\boldsymbol{A^{\top}}\boldsymbol{A}\right)  (\boldsymbol{\beta^*-\hat{\beta}_{\lambda_1}})\right\|_\infty&=&o_P(1) \label{eq:bias_A_beta1} 
     \\
     \left\|\frac{1}{\sqrt{n}}\boldsymbol{A}^{\top}
\left(\boldsymbol{\delta^{*}}-\boldsymbol{\hat{\delta}_{\lambda_2}}\right)\right\|_\infty&=&o_P(1) \label{eq:bias_A_beta2}
\\
\left\|\frac{n}{p\sqrt{1-n/p}}\left(\boldsymbol{I_n}-\frac{1}{n}\boldsymbol{A}\boldsymbol{A}^{\top}\right) \boldsymbol{A} (\boldsymbol{\beta^*-\hat{\beta}_{\lambda_1}})\right\|_\infty&=&o_P(1) \label{eq:bias_A_delta1} 
\\
\left\|\frac{n}{p\sqrt{1-n/p}}\frac{1}{n}\boldsymbol{AA}^{\top}
\big(\boldsymbol{\delta^{*}}-\boldsymbol{\hat{\delta}_{\lambda_2}}\big)\right\|_\infty&=&o_P(1) \label{eq:bias_A_delta2}
\end{eqnarray}
\hfill{$\blacksquare$}
\end{lemma}
\textbf{Proof of Lemma \ref{le:2_3term_beta_decompose}:}\\
When $n$ is $\omega[((s+r)\log p)^2]$, the assumptions of Lemma \ref{le:Ext_RE} are satisfied. Hence, the Rademacher matrix $\boldsymbol{A}$ satisfies the assumptions of Theorem \ref{th:upper_bound_robustLasso} with probability that goes to $1$ as $n,p \rightarrow \infty$. Therefore, to prove the results, it suffices to condition on the event that the conclusion of Theorem \ref{th:upper_bound_robustLasso} holds.

\noindent \textbf{Proof of \eqref{eq:bias_A_beta1}:}
Using result (4) of Lemma~\ref{le:mat_Op_properties}, we have:
\begin{eqnarray}\label{eq:l1_decompose}
  \left\|\sqrt{n}\left(\boldsymbol{I_p}-\frac{1}{n}\boldsymbol{A^{\top}}\boldsymbol{A}\right)  (\boldsymbol{\beta^*-\hat{\beta}_{\lambda_1}})\right\|_{\infty} 
  \leq \sqrt{n} |\boldsymbol{A^{\top}A}/n-\boldsymbol{I_p}|_{\infty} \|\boldsymbol{\beta^*}-\boldsymbol{\hat{\beta}_{\lambda_1}}\|_1.
\end{eqnarray}
From result (1) of Lemma \ref{le:Rad_mat_prop}, result (1) of Theorem~\ref{th:upper_bound_robustLasso}, and result (5) of Lemma \ref{le:mat_Op_properties} , we have,
\begin{equation}\label{eq:rate_B1_orig}
    \left\|\sqrt{n}\left(\boldsymbol{I_p}-\frac{1}{n}\boldsymbol{A^{\top}}\boldsymbol{A}\right)  (\boldsymbol{\beta^*-\hat{\beta}_{\lambda_1}})\right\|_{\infty} = O_{P}\left((s+r)\frac{\log p}{\sqrt{n}} \right).
\end{equation}
Under the assumption {$n$ is $\omega[((s+r)\log p)^2]$}, we have:
\begin{equation}\label{eq:rate_B1}
    \left\|\sqrt{n}\left(\boldsymbol{I_p}-\frac{1}{n}\boldsymbol{A^{\top}}\boldsymbol{A}\right)  (\boldsymbol{\beta^*-\hat{\beta}_{\lambda_1}})\right\|_{\infty} = o_{P}(1).
\end{equation}
\textbf{Proof of \eqref{eq:bias_A_beta2}:}
Again by using result (4) of Lemma~\ref{le:mat_Op_properties}, we have
\begin{equation*}
    \left\|\frac{1}{\sqrt{n}}\boldsymbol{A^{\top}}(\boldsymbol{\delta^*}-\boldsymbol{\hat{\delta}_{\lambda_2}})\right\|_{\infty}  \leq  \left|\frac{1}{\sqrt{n}}\boldsymbol{A^{\top}}\right|_{\infty} \|\boldsymbol{\delta^*}-\boldsymbol{\hat{\delta}_{\lambda_2}}\|_1.
\end{equation*}
Since $\boldsymbol{A}$ is a Rademacher matrix, we have, $\left|\frac{1}{\sqrt{n}}\boldsymbol{A^{\top}}\right|_{\infty}=\frac{1}{\sqrt{n}}$. From result (2) of  Theorem~\ref{th:upper_bound_robustLasso} and result (5) of  
Lemma \ref{le:mat_Op_properties}, we have
\begin{equation}\label{eq:rate_B2_orig}
    \left\|\frac{1}{\sqrt{n}}\boldsymbol{A^{\top}}(\boldsymbol{\delta^*}-\boldsymbol{\hat{\delta}_{\lambda_2}})\right\|_{\infty} = O_{P}\left(\frac{r\sqrt{\log n}}{n^{3/2}}\right).
\end{equation}
As $n, p \to \infty$, we have
\begin{equation}\label{eq:rate_B2}
     \left\|\frac{1}{\sqrt{n}}\boldsymbol{A^{\top}}(\boldsymbol{\delta^*}-\boldsymbol{\hat{\delta}_{\lambda_2}})\right\|_{\infty} = o_{P}(1).
\end{equation}

\noindent    \textbf{Proof of \eqref{eq:bias_A_delta1}:}
    Again using result (4) of Lemma~\ref{le:mat_Op_properties}, we have,
    \begin{eqnarray}\label{eq:bound_D1}
  \left\|\frac{n}{p\sqrt{1-n/p}}\left(\boldsymbol{I_n}-\frac{1}{n}\boldsymbol{A}\boldsymbol{A}^{\top}\right) \boldsymbol{A} (\boldsymbol{\beta^*-\hat{\beta}_{\lambda_1}})\right\|_{\infty} \leq \frac{n}{\sqrt{1-n/p}} \times \left|\frac1p\left(\boldsymbol{I_n}-\frac{1}{n}\boldsymbol{A}\boldsymbol{A}^{\top}\right) \boldsymbol{A}\right|_{\infty} \|\boldsymbol{\beta^*}-\boldsymbol{\hat{\beta}_{\lambda_1}}\|_1.
\end{eqnarray}
By using result (5) of Lemma \ref{le:mat_Op_properties}, result (1) of Theorem \ref{th:upper_bound_robustLasso} and result (2) of Lemma \ref{le:Rad_mat_prop}, we have
{\begin{eqnarray}\label{eq:rate_D1_orig}
   \nonumber  \left\|\frac{n}{p\sqrt{1-n/p}}\left(\boldsymbol{I_n}-\frac{1}{n}\boldsymbol{A}\boldsymbol{A}^{\top}\right) \boldsymbol{A} (\boldsymbol{\beta^*-\hat{\beta}_{\lambda_1}})\right\|_{\infty} &\leq& O_P\left((s+r)\frac{n}{\sqrt{1-n/p}}\left(\sqrt{\frac{\log(pn)}{pn}}+\frac{1}{n}\right)\sqrt{\frac{\log p}{n}}\right) \\ \nonumber &=& O_P\left(\frac{(s+r)n}{\sqrt{1-n/p}}\sqrt{\frac{\log(pn)}{pn}}\sqrt{\frac{\log p}{n}}+\frac{(s+r)n}{\sqrt{1-n/p}}\frac{1}{n}\sqrt{\frac{\log p}{n}}\right)\\ &=& O_P\left(\frac{(s+r)}{\sqrt{1-n/p}}\sqrt{\frac{\log(np)\log(p)}{p}}+\frac{(s+r)}{\sqrt{1-n/p}}\sqrt{\frac{\log p}{n}}\right) .
\end{eqnarray}}
{Since $n$ is $o(p)$ and $n$ is $\omega[((s+r)\log p)^2]$}, \eqref{eq:rate_D1_orig} becomes as $n,p \to \infty$,
\begin{equation}\label{eq:rate_D1}
    \left\|\frac{n}{p\sqrt{1-n/p}}\left(\boldsymbol{I_n}-\frac{1}{n}\boldsymbol{A}\boldsymbol{A}^{\top}\right) \boldsymbol{A} (\boldsymbol{\beta^*-\hat{\beta}_{\lambda_1}})\right\|_{\infty} =o_P(1).
\end{equation}
\textbf{Proof of \eqref{eq:bias_A_delta2}:} Using result (4) of Lemma~\ref{le:mat_Op_properties}, we have,
\begin{eqnarray}
 \left\|\frac{n}{p\sqrt{1-n/p}}\frac{1}{n}\boldsymbol{AA}^{\top}
\big(\boldsymbol{\delta^{*}}-\boldsymbol{\hat{\delta}_{\lambda_2}}\big)\right\|_\infty \leq \frac{1}{\sqrt{1-n/p}} \times \left|\frac{1}{p}\boldsymbol{AA}^{\top}\right|_{\infty} \left\|\big(\boldsymbol{\delta^{*}}-\boldsymbol{\hat{\delta}_{\lambda_2}}\big)\right\|_1.
\end{eqnarray}
Since $\boldsymbol{A}$ is a Rademacher matrix, the elements of $\frac{1}{p}\boldsymbol{AA^\top}$ lie between $-1$ and $1$. Therefore, $\left|\frac{1}{p}\boldsymbol{AA^\top}\right|_{\infty}=1$. By using part (5) of Lemma \ref{le:mat_Op_properties} and result (2) of Theorem \ref{th:upper_bound_robustLasso}, we have
\begin{eqnarray}\label{eq:rate_D2_orig}
    \left\|\frac{n}{p\sqrt{1-n/p}}\frac{1}{n}\boldsymbol{AA}^{\top}
\big(\boldsymbol{\delta^{*}}-\boldsymbol{\hat{\delta}_{\lambda_2}}\big)\right\|_\infty =O_P\left(\frac{r}{\sqrt{1-n/p}}\frac{\sqrt{\log n}}{\sqrt{n}}\right). \end{eqnarray}
Since $n$ is $o(p)$, we have
\begin{equation}\label{eq:rate_D2}
    \left\|\frac{n}{p\sqrt{1-n/p}}\frac{1}{n}\boldsymbol{AA}^{\top}
\big(\boldsymbol{\delta^{*}}-\boldsymbol{\hat{\delta}_{\lambda_2}}\big)\right\|_\infty =o_P(1).
\end{equation}
This completes the proof.
\hfill{$\blacksquare$}
\begin{lemma}\label{le:G_conv_prob}
    Let $\boldsymbol{A}$ be a Rademacher matrix and $\boldsymbol{\Sigma_{A}}= \left(\boldsymbol{I_n}-\frac{1}{n}\boldsymbol{AA}^{\top}\right)\left(\boldsymbol{I_n}-\frac{1}{n}\boldsymbol{AA}^{\top}\right)^{\top}$. If $n \log n$ is $o(p)$, we have as $n,p \to \infty$ for any $i \in [n]$,
    \begin{equation}\label{eq:G_conv_prob}
        \frac{n^2}{p^2(1-n/p)}\Sigma_{A_{ii}} \overset{P}{\to} 1.
    \end{equation}
\end{lemma}
\textbf{Proof of Lemma \ref{le:G_conv_prob}:} 
Recall that we consider $\boldsymbol{\Sigma_{A}}= \left(\boldsymbol{I_n}-\frac{1}{n}\boldsymbol{AA}^{\top}\right)\left(\boldsymbol{I_n}-\frac{1}{n}\boldsymbol{AA}^{\top}\right)^{\top}$. Note that for $i \in [n]$, we have
\begin{eqnarray}
   \nonumber \frac{n^2}{p^2(1-n/p)}\Sigma_{A_{ii}} &=& \frac{n^2}{p^2(1-\frac{n}{p})}\left(1-\frac{2}{n}\boldsymbol{a_{i.}}\boldsymbol{a_{i.}}^{\top}
     +\frac{1}{n^2}\boldsymbol{a_{i.}A^{\top}A}\boldsymbol{a_{i.}}^{\top}\right)\\ \nonumber  &=& \left(\frac{n}{p\sqrt{({1-\frac{n}{p}})}}\left(1-\frac{\boldsymbol{a_{i.}a_{i.}^{\top}}}{n}\right)\right)^2 + \underset{k \ne i}{\sum_{k=1}^n} \left(\frac{n}{p\sqrt{({1-\frac{n}{p}})}}\left(\frac{\boldsymbol{a_{i.}a_{k.}^{\top}}}{n}\right)\right)^2\\ &=& {1-\frac{n}{p}} +\underset{k \ne i}{\sum_{k=1}^n} \left(\frac{n}{p\sqrt{({1-\frac{n}{p}})}}\left(\frac{\boldsymbol{a_{i.}a_{k.}^{\top}}}{n}\right)\right)^2 .\label{eq:Gi_struct}
\end{eqnarray}
In \eqref{eq:Gi_struct}, since the second term is positive, we have
\begin{equation}\label{eq:gigiT_lw_bnd}
    \frac{n^2}{p^2(1-n/p)}\Sigma_{A_{ii}} \geq 1-\frac{n}{p}.
\end{equation}
We have from Result (3) of Lemma \ref{le:Rad_mat_prop}, for $k \in [n], k\ne i$,
\begin{equation}\label{eq:off_diag_entries_G}
    P\left(\frac{1}{\sqrt{1-n/p}}\left|\boldsymbol{a_{i.}a_{k.}^{\top}}/p\right|\leq \frac{2}{\sqrt{1-n/p}}\sqrt{\frac{2\log(n)}{p}}\right)\geq 1-\frac{2}{n^2}.
\end{equation}
By using \eqref{eq:off_diag_entries_G}, we have 
\begin{eqnarray}
   \nonumber 
   && \hskip -20pt P\left(\underset{k \ne i}{\sum_{k=1}^n} \left(\frac{n}{p\sqrt{({1-\frac{n}{p}})}}\left(\frac{\boldsymbol{a_{i.}a_{k.}^{\top}}}{n}\right)\right)^2 \leq \frac{4(n-1)}{{1-n/p}}{\frac{2\log(n)}{p}} \right)\\ \nonumber  &\geq&  P\left(\underset{k \ne i}{ \cap_{k=1}^n} \left\{\left(\frac{n}{p\sqrt{({1-\frac{n}{p}})}}\left(\frac{\boldsymbol{a_{i.}a_{k.}^{\top}}}{n}\right)\right)^2 \leq  \frac{4}{{1-n/p}}{\frac{2\log(n)}{p}} \right\}\right) \\ \nonumber &=&  P\left(\underset{k \ne i}{ \cap_{k=1}^n} \left\{\frac{n}{p\sqrt{({1-\frac{n}{p}})}}\left|\frac{\boldsymbol{a_{i.}a_{k.}^{\top}}}{n}\right| \leq  \frac{2}{\sqrt{1-n/p}}{\sqrt{\frac{2\log(n)}{p}}} \right\}\right) \\ &\geq& 1-\frac{2(n-1)}{n^2}.  \label{eq:uppr_bnd_gigit_ij}
\end{eqnarray}
The last inequality comes using Bonferroni's inequality which states that $P(\cap_{i=1}^n U_i) \geq 1-\sum_{i=1}^n P(U_i)$ for any events $U_1, U_2,...,U_n$. Therefore by using \eqref{eq:Gi_struct} and \eqref{eq:uppr_bnd_gigit_ij}, we have
\begin{equation}\label{eq:gigit_uppr_bnd}
   P \left( \frac{n^2}{p^2(1-n/p)}\Sigma_{A_{ii}} \leq 1-\frac{n}{p} + \frac{4(n-1)}{{1-n/p}}{\frac{2\log(n)}{p}} \right) \geq 1-\frac{2}{n}
\end{equation}
By using \eqref{eq:gigiT_lw_bnd} and \eqref{eq:gigit_uppr_bnd}, we have 
\begin{equation}\label{eq:gigit_bnd_full}
     P \left( 1-\frac{n}{p} \leq \frac{n^2}{p^2(1-n/p)}\Sigma_{A_{ii}} \leq 1-\frac{n}{p} + \frac{4(n-1)}{{1-n/p}}{\frac{2\log(n)}{p}} \right) \geq 1-\frac{2}{n}
\end{equation}
Since $n \log n$ is $o(p)$, as $n,p \to \infty$, $1-\frac{n}{p} \to 1$ and $\frac{4(n-1)}{{1-n/p}}{\frac{2\log(n)}{p}} \to 0$. This completes the proof.
\hfill{$\blacksquare$}

\subsection{Overview of Proofs of Results with the Debiasing Matrix $W$}
{Now we proceed to the results involving debiasing using the \emph{optimal} weights matrix $\boldsymbol{W}$ obtained from Alg.~\ref{alg:design_W}. Our main result is Theorem~\ref{th:distribution_beta_delta_opt}. The proofs of these results largely follow the same approach as that for $\boldsymbol{W} = \boldsymbol{A}$ (i.e. Theorem~\ref{th:dist_debiased_beta_delta_A}). However there is one major point of departure---due to differences in properties of the weights matrix $\boldsymbol{W}$ designed from Alg.~\ref{alg:design_W} (these properties are given in Lemma~\ref{le:W_mat_prop}), as compared to the case where $\boldsymbol{W} = \boldsymbol{A}$ (given in Lemma~\ref{le:Rad_mat_prop}). First, for $\boldsymbol{\hat{\beta}_W}$, we express its deviation from $\boldsymbol{\beta^*}$ using the given measurement model. This decomposition consists of three terms: (i) a noise term involving $\boldsymbol{W}^{\top}\boldsymbol{\eta}$, (ii) a bias term involving the estimation error of $\boldsymbol{\hat{\beta}_{\lambda_1}}$, and (iii) an bias term involving the estimation error of $\boldsymbol{\hat{\delta}_{\lambda_2}}$. Theorem~\ref{th:2_3term_beta_decompose_W} ensures that the second and third terms are asymptotically negligible, decaying as $o_P(1/\sqrt{n})$. The leading term is $\frac{1}{\sqrt{n}}\boldsymbol{w}_{.j}^{\top}\boldsymbol{\eta}$, which is Gaussian with mean zero and variance $\Sigma_{\beta_{jj}}$, and which converges to $\sigma^2$ in probability, as shown in Theorem~\ref{th:dist_conv_W}. This helps establish the distributional property of $\boldsymbol{\hat{\beta}_W}$ from Theorem~\ref{th:distribution_beta_delta_opt} in \eqref{eq:beta_marginal_dist}.
For the deviation of $\boldsymbol{\hat{\delta}_W}$ from $\boldsymbol{\delta^*}$, a similar decomposition is applied. The second and third terms of this decomposition are shown to be asymptotically negligible, that is they are $o_P\left(\frac{p\sqrt{1-n/p}}{n}\right)$ whereas the standard deviation of the first term is $O_P\left(\frac{p\sqrt{1-n/p}}{n}\right)$. This is shown in Theorem~\ref{th:2_3term_beta_decompose_W}. The leading term, which involves a transformation of the Gaussian noise vector $\boldsymbol{\eta}$, remains Gaussian with mean zero and variance $\Sigma_{\delta_{ii}}$. Importantly in Theorem~\ref{th:dist_conv_W}, we show that $\frac{n^2}{p^2(1-n/p)} \Sigma_{\delta_{ii}}$ converges to $\sigma^2$ in probability, ensuring that the normalized estimator $\frac{\hat{\delta}_{Wi} - \delta^*_i}{\sqrt{\Sigma_{\delta_{ii}}}}$ converges to a standard normal distribution as shown in Theorem~\ref{th:distribution_beta_delta_opt}. This completes the proof of \eqref{eq:delta_marginal_dist}.}

\subsection{Proof of Theorem \ref{th:2_3term_beta_decompose_W}}
\noindent \textbf{Proof of \eqref{eq:bias_beta_1}:}
Using Result (4) of Lemma~\ref{le:mat_Op_properties}, we have
\begin{eqnarray}\label{eq:l1_decompose_W}
  \left\|\sqrt{n}\left(\boldsymbol{I_p}-\frac{1}{n}\boldsymbol{W^{\top}}\boldsymbol{A}\right)  (\boldsymbol{\beta^*-\hat{\beta}_{\lambda_1}})\right\|_{\infty} 
  \leq \sqrt{n} \|\boldsymbol{W^{\top}A}/n-\boldsymbol{I_p}\|_{\infty} \|\boldsymbol{\beta^*}-\boldsymbol{\hat{\beta}_{\lambda_1}}\|_1.
\end{eqnarray}
Using Result (2) of Lemma \ref{le:W_mat_prop}, 
 Result (1) of Theorem~\ref{th:upper_bound_robustLasso} and Result (5) of Lemma \ref{le:mat_Op_properties}, we have
\begin{equation}\label{eq:rate_B1_orig_W}
    \left\|\sqrt{n}\left(\boldsymbol{I_p}-\frac{1}{n}\boldsymbol{W^{\top}}\boldsymbol{A}\right)  (\boldsymbol{\beta^*-\hat{\beta}_{\lambda_1}})\right\|_{\infty} = O_P\left((s+r)\frac{\log p}{\sqrt{n}} \right).
\end{equation}
If $n$ is $\omega[((s+r)\log p)^2]$, we have:
\begin{equation}\label{eq:rate_B1_W}
    \left\|\sqrt{n}\left(\boldsymbol{I_p}-\frac{1}{n}\boldsymbol{W^{\top}}\boldsymbol{A}\right)  (\boldsymbol{\beta^*-\hat{\beta}_{\lambda_1}})\right\|_{\infty} = o_P(1).
\end{equation}
\newline
\textbf{Proof of \eqref{eq:bias_beta_2}:}
Using Result (4) of Lemma~\ref{le:mat_Op_properties}, we have
\begin{equation*}
    \left\|\frac{1}{\sqrt{n}}\boldsymbol{W^{\top}}(\boldsymbol{\delta^*}-\boldsymbol{\hat{\delta}_{\lambda_2}})\right\|_{\infty}  \leq  \left|\frac{1}{\sqrt{n}}\boldsymbol{W^{\top}}\right|_{\infty} \|\boldsymbol{\delta^*}-\boldsymbol{\hat{\delta}_{\lambda_2}}\|_1.
\end{equation*}
Using Result (3) of Lemma \ref{le:W_mat_prop}, Result (2) of  Theorem~\ref{th:upper_bound_robustLasso} and Result (5) of Lemma \ref{le:mat_Op_properties}, we have
\begin{equation}\label{eq:rate_B2_orig_W}
    \left\|\frac{1}{\sqrt{n}}\boldsymbol{W^{\top}}(\boldsymbol{\delta^*}-\boldsymbol{\hat{\delta}_{\lambda_2}})\right\|_{\infty} = O_P\left(\frac{r\sqrt{\log n}}{n^{3/2}}\right) =o_P(1).
\end{equation}
 \textbf{Proof of \eqref{eq:bias_delta_1}:}
    Using Result (4) of Lemma~\ref{le:mat_Op_properties}, we have
    \begin{eqnarray}\label{eq:bound_D1_W}
  \left\|\frac{n}{p\sqrt{1-n/p}}\left(\boldsymbol{I_n}-\frac{1}{n}\boldsymbol{A}\boldsymbol{W}^{\top}\right) \boldsymbol{A} (\boldsymbol{\beta^*-\hat{\beta}_{\lambda_1}})\right\|_{\infty} \leq \frac{n}{\sqrt{1-n/p}} \times \left|\frac1p\left(\boldsymbol{I_n}-\frac{1}{n}\boldsymbol{A}\boldsymbol{W}^{\top}\right) \boldsymbol{A}\right|_{\infty} \|\boldsymbol{\beta^*}-\boldsymbol{\hat{\beta}_{\lambda_1}}\|_1.
\end{eqnarray}
Using Result (4) of Lemma \ref{le:W_mat_prop}, Result (1) of Theorem \ref{th:upper_bound_robustLasso} and Result (5) of Lemma \ref{le:mat_Op_properties}, we have
{\begin{eqnarray}\label{eq:rate_D1_orig_W}
    \nonumber \left\|\frac{n}{p\sqrt{1-n/p}}\left(\boldsymbol{I_n}-\frac{1}{n}\boldsymbol{A}\boldsymbol{W}^{\top}\right) \boldsymbol{A} (\boldsymbol{\beta^*-\hat{\beta}_{\lambda_1}})\right\|_{\infty} &\leq& O_P\left(\frac{(s+r)n}{\sqrt{1-n/p}}\left(\sqrt{\frac{\log(pn)}{pn}}+\frac{1}{n}\right)\sqrt{\frac{\log p}{n}}\right) \\ &=& O_P\left(\frac{(s+r)}{\sqrt{1-n/p}}\sqrt{\frac{\log(np)\log(p)}{p}}+\frac{(s+r)}{\sqrt{1-n/p}}\sqrt{\frac{\log p}{n}}\right) .
\end{eqnarray}
Since $n$ is $o(p)$ and $n$ is $\omega[((s+r)\log p)^2]$}, we have
\begin{equation}\label{eq:rate_D1_W}
    \left\|\frac{n}{p\sqrt{1-n/p}}\left(\boldsymbol{I_n}-\frac{1}{n}\boldsymbol{A}\boldsymbol{W}^{\top}\right) \boldsymbol{A} (\boldsymbol{\beta^*-\hat{\beta}_{\lambda_1}})\right\|_{\infty} =o_P(1).
\end{equation}
\textbf{Proof of \eqref{eq:bias_delta_2}:}
Using Result (4) of Lemma~\ref{le:mat_Op_properties}, we have
\begin{eqnarray}\label{eq:bound_D2_W}
 \left\|\frac{n}{p\sqrt{1-n/p}}\frac{1}{n}\boldsymbol{AW}^{\top}
\big(\boldsymbol{\delta^{*}}-\boldsymbol{\hat{\delta}_{\lambda_2}}\big)\right\|_\infty \leq \frac{1}{\sqrt{1-n/p}} \times \left|\frac{1}{p}\boldsymbol{AW}^{\top}\right|_{\infty} \left\|\big(\boldsymbol{\delta^{*}}-\boldsymbol{\hat{\delta}_{\lambda_2}}\big)\right\|_1
\end{eqnarray}
Using Result (5) of Lemma \ref{le:W_mat_prop}, Result (2) of Theorem \ref{th:upper_bound_robustLasso} and Result (5) of Lemma \ref{le:mat_Op_properties}, we have
{\begin{eqnarray}\label{eq:rate_D2_orig_W}
   \nonumber \left\|\frac{n}{p\sqrt{1-n/p}}\frac{1}{n}\boldsymbol{AW}^{\top}
\big(\boldsymbol{\delta^{*}}-\boldsymbol{\hat{\delta}_{\lambda_2}}\big)\right\|_\infty &=& O_P\left(\frac{r}{\sqrt{1-n/p}}\left(\sqrt{\frac{n\log(np)}{p}}+1\right)\frac{\sqrt{\log n}}{{n^{3/2}}}\right) \\ &=& O_P\left(\frac{r}{\sqrt{1-n/p}}\sqrt{\frac{\log(np)}{p}}\sqrt{\frac{{\log n}}{{n}}}+\frac{r}{\sqrt{1-n/p}}\frac{\sqrt{\log n}}{{n^{3/2}}}\right) .
\end{eqnarray}}
Since $n$ is $o(p)$, we have 
\begin{equation}\label{eq:rate_D2_W}
    \left\|\frac{n}{p\sqrt{1-n/p}}\frac{1}{n}\boldsymbol{AW}^{\top}
\big(\boldsymbol{\delta^{*}}-\boldsymbol{\hat{\delta}_{\lambda_2}}\big)\right\|_\infty =o_P(1).
\end{equation}
This completes the proof.
\hfill{$\blacksquare$}

\subsection{Proof of Theorem~\ref{th:dist_conv_W}}
\noindent\textbf{Result(1):} 
Recall that $\boldsymbol{W}$ is the output of Alg.~\ref{alg:design_W}, $\boldsymbol{\Sigma_{\beta}}=\frac{\sigma^2}{n}\boldsymbol{W^{\top}W}$, $\boldsymbol{\Sigma_{\delta}}=\sigma^2\left(\boldsymbol{I_n}-\frac{1}{n}\boldsymbol{AW}^{\top}\right)\left(\boldsymbol{I_n}-\frac{1}{n}\boldsymbol{AW}^{\top}\right)^{\top}$ and $\mu_1=2\sqrt{\frac{2\log p}{n}}$.
We will derive the lower bound of $\Sigma_{\beta_{jj}}$ for all $j \in [p]$. Note that, $\Sigma_{\beta_{jj}} =\frac{\sigma^2}{n}\boldsymbol{w_{.j}^{\top}w_{.j}}$. For all $j \in [p]$, from \eqref{eq:bound_cons_1} of result (2) of Lemma \ref{le:W_mat_prop}, for any feasible $\boldsymbol{W}$ with probability at least $1-\left(\dfrac{2}{p^2}+\dfrac{2}{n^2}+\dfrac{3}{2np}\right)$, we have
\begin{equation*}
    1-\frac{1}{n}\boldsymbol{a_{.j}^\top w_{.j}} \leq \mu_1 \ \implies \ 1-\mu_1 \leq \frac{1}{n}\boldsymbol{a_{.j}^\top w_{.j}}.
\end{equation*}
For any feasible $\boldsymbol{W}$ of Alg.\ref{alg:design_W}, we have for any $c>0$,
\begin{eqnarray*}
    \frac{1}{n}\boldsymbol{w_{.j}^{\top}w_{.j}} &\geq& \frac{1}{n}\boldsymbol{w_{.j}^{\top}w_{.j}} + c(1-\mu_1) - c\frac{1}{n}\left(\boldsymbol{a_{.j}^\top w_{.j}}\right) \geq \underset{\boldsymbol{w_{.j}} \in \mathbb{R}^n}{\min} \left\{\frac{1}{n}\boldsymbol{w_{.j}^{\top}w_{.j}} + c(1-\mu_1) - c\frac{1}{n}\left(\boldsymbol{a_{.j}^\top w_{.j}}\right)\right\} \\ &=& \underset{\boldsymbol{w_{.j}} \in \mathbb{R}^n}{\min} \left\{\frac{1}{n}\left(\boldsymbol{w_{.j}} - c\boldsymbol{a_{.j}}/2\right)^{\top}\left(\boldsymbol{w_{.j}} - c\boldsymbol{a_{.j}}/2\right)\right\} + c(1-\mu_1) -\frac{c^2}{4}\frac{\boldsymbol{a_{.j}^{\top}a_{.j}}}{n}  \geq c(1-\mu_1) -\frac{c^2}{4}\frac{\boldsymbol{a_{.j}}^{\top}\boldsymbol{a_{.j}}}{n} \\ &=& c(1-\mu_1)- \dfrac{c^2}{4}.
\end{eqnarray*}
We obtain the last inequality by putting $\boldsymbol{w_{.j}} = c\boldsymbol{a_{.j}}/2$ which makes the square term 0. The rightmost equality is because $\boldsymbol{a_{.j}}^{\top}\boldsymbol{a_{.j}} = n$. The lower bound on the RHS is maximized for $c=2(1-\mu_1)$. Plugging in this value of $c$, we obtain the following with probability atleast $1-\left(\frac{2}{p^2}+\frac{2}{n^2}+\frac{1}{2np}\right)$: 
\begin{equation*}
    \frac{1}{n}\boldsymbol{w_{.j}^{\top}w_{.j}}  \geq  (1-\mu_1)^2.
\end{equation*}
Hence, from the above equation
and \eqref{eq:joint_feasibility}, we obtain the lower bound on $\Sigma_{\beta_{jj}}$ for any $j \in [p]$ as follows: 
\begin{equation}\label{eq:lower_bnd_beta}
    P\left(\Sigma_{\beta_{jj}} \geq \sigma^2(1-\mu_1)^2\right) \geq 1-\left(\frac{2}{p^2}+\frac{2}{n^2}+\frac{1}{2np}\right).
\end{equation}
Furthermore from Result (1) of Lemma.~\ref{le:W_mat_prop}, we have
\begin{equation}\label{eq:bound_cons_4}
    P\left(\boldsymbol{w_{.j}^{\top}w_{.j}}/n \leq 1 \ \forall j \in [p]\right) = 1.
\end{equation}
We use \eqref{eq:bound_cons_4} to get, for any $j \in [p]$, $P(\boldsymbol{w_{.j}^{\top}w_{.j}}/n \leq 1) =1$.
As $\Sigma_{\beta_{jj}} =\sigma^2\frac{\boldsymbol{w_{.j}^{\top} w_{.j}}}{n}$, we have for any $j \in [p]$:
\begin{equation}\label{eq:beta_var_uppr_bnd}
    P\left(\Sigma_{\beta_{jj}} \leq \sigma^2\right) = 1.
\end{equation}
Using \eqref{eq:beta_var_uppr_bnd} with \eqref{eq:lower_bnd_beta}, we obtain for any $j \in [p]$, 
\begin{equation}
\label{eq:sigma_beta_jj_bound}
    P\left(\sigma^2(1-\mu_1)^2\leq \Sigma_{\beta_{jj}} \leq \sigma^2 \right) \geq 1-\left(\frac{2}{p^2}+\frac{2}{n^2}+\frac{1}{2np}\right).
\end{equation}
Now under the assumption $n$ is $\omega[((s+r)\log p)^2]$, $\mu_1 \to 0$. Hence, we have, $\Sigma_{\beta_{jj}} \overset{P}{\to} \sigma^2$. This completes the proof of Result (1).
\newline
\textbf{Result (2):}
Recall that $\mu_3=2\sqrt{\frac{2\log(n)}{p}}$.
Now in order to obtain the upper and lower bounds for $\Sigma_{\delta_{ii}}$ for any $i \in [n]$, we use Result (6) of Lemma.~\ref{le:W_mat_prop}.
We have,
\begin{equation}\label{eq:bound_cons_3}
    P\left(\frac{n}{p\sqrt{1-\frac{n}{p}}}\left|\left(\frac{\boldsymbol{AW^{\top}}}{n}-\frac{p}{n}\boldsymbol{I_n}\right)\right|_{\infty} \leq \frac{1}{\sqrt{1-n/p}}\mu_3\right) \geq 1-\left(\frac{2}{p^2}+\frac{2}{n^2}+\frac{1}{2np}\right).
\end{equation}
We have for $ i \in [n]$, 
\begin{eqnarray}
   \nonumber \frac{n^2}{p^2(1-\frac{n}{p})}\Sigma_{\delta_{ii}}&=&\sigma^2\frac{n^2}{p^2(1-\frac{n}{p})}\left(1-\frac{2}{n}\boldsymbol{a_{i.}}\boldsymbol{w_{i.}}^{\top}
     +\frac{1}{n^2}\boldsymbol{a_{i.}W^{\top}W}\boldsymbol{a_{i.}}^{\top}\right)\\ &=&\sigma^2\left(\frac{n}{p\sqrt{({1-\frac{n}{p}})}}\left(1-\frac{\boldsymbol{a_{i.}w_{i.}^{\top}}}{n}\right)\right)^2 +\sigma^2\underset{k \ne i}{\sum_{k=1}^n} \left(\frac{n}{p\sqrt{({1-\frac{n}{p}})}}\left(\frac{\boldsymbol{a_{i.}w_{k.}^{\top}}}{n}\right)\right)^2. \label{eq:Sigma_delta_ii_decomp}
\end{eqnarray}
Let $\boldsymbol{V}=\left(\frac{\boldsymbol{AW}^{\top}}{n}-\frac{p}{n}\boldsymbol{I_n}\right)$. Note that, for $i \in [n]$, 
\begin{equation}\label{eq:vii_var_delta}
v_{ii}=\frac{\boldsymbol{a_{i.}w_{i.}}^{\top}}{n}-\frac{p}{n}=\left(\frac{\boldsymbol{a_{i.}w_{i.}}^{\top}}{n}-1\right)+\left(1-\frac{p}{n}\right).
\end{equation}
Hence from \eqref{eq:vii_var_delta}, we have
\begin{eqnarray}
   \nonumber \frac{n}{p\sqrt{1-n/p}}v_{ii} &=& \frac{n}{p\sqrt{1-n/p}}\left(\frac{\boldsymbol{a_{i.}w_{i.}}^{\top}}{n}-1\right) + \frac{1}{\sqrt{1-n/p}}\left(\frac{n}{p}-1\right)\\ \nonumber &=& \frac{n}{p\sqrt{1-n/p}}\left(\frac{\boldsymbol{a_{i.}w_{i.}}^{\top}}{n}-1\right) -  \frac{1}{\sqrt{1-n/p}}\left(1-\frac{n}{p}\right)\\ &=& \frac{n}{p\sqrt{1-n/p}}\left(\frac{\boldsymbol{a_{i.}w_{i.}}^{\top}}{n}-1\right) -\sqrt{1-\frac{n}{p}}. \label{eq:vii_var_delta_form}
\end{eqnarray}
We have, from \eqref{eq:bound_cons_3}, 
\begin{equation*}
    P\left(\frac{n}{p\sqrt{1-n/p}}v_{ii} \geq -\frac{1}{\sqrt{1-n/p}}\mu_3 \right) \geq 1-\left(\frac{2}{p^2}+\frac{2}{n^2}+\frac{1}{2np}\right). 
\end{equation*}
Therefore using \eqref{eq:vii_var_delta_form}, we have
\begin{eqnarray}
  \nonumber  P\left(\frac{n}{p\sqrt{1-n/p}}\left(\frac{\boldsymbol{a_{i.}w_{i.}}^{\top}}{n}-1\right) -\sqrt{1-\frac{n}{p}} \geq -\frac{1}{\sqrt{1-n/p}}\mu_3 \right) &\geq& 1-\left(\frac{2}{p^2}+\frac{2}{n^2}+\frac{1}{2np}\right) \\ \implies P\left(\frac{n}{p\sqrt{1-n/p}}\left(\frac{\boldsymbol{a_{i.}w_{i.}}^{\top}}{n}-1\right) \geq \sqrt{1-\frac{n}{p}}-\frac{1}{\sqrt{1-n/p}}\mu_3 \right) &\geq& 1-\left(\frac{2}{p^2}+\frac{2}{n^2}+\frac{1}{2np}\right).\label{eq:lower_bnd_aw}
\end{eqnarray}
Using \eqref{eq:lower_bnd_aw} in \eqref{eq:Sigma_delta_ii_decomp} yields the following inequality with probability at least $1-\left(\frac{2}{p^2}+\frac{2}{n^2}+\frac{1}{2np}\right)$,
\begin{eqnarray*}
    \frac{n^2}{p^2(1-\frac{n}{p})}\Sigma_{\delta_{ii}} &=& \sigma^2\left(\frac{n}{p\sqrt{({1-\frac{n}{p}})}}\left(\frac{\boldsymbol{a_{i.}w_{i.}^{\top}}}{n}-1\right)\right)^2 + \sigma^2\sum_{k=1,k\ne i}^n \left(\frac{n}{p\sqrt{1-n/p}}\frac{\boldsymbol{a_{i.}w_{k.}^{\top}}}{n}\right)^2 \\ &\geq& \sigma^2\left(\frac{n}{p\sqrt{({1-\frac{n}{p}})}}\left(\frac{\boldsymbol{a_{i.}w_{i.}^{\top}}}{n}-1\right)\right)^2 \geq \sigma^2\left(\sqrt{1-\frac{n}{p}}-\frac{1}{\sqrt{1-n/p}}\mu_3\right)^2.
\end{eqnarray*}
Therefore the lower bound on $\Sigma_{\delta_{ii}}$ is as follows: 
\begin{equation}\label{eq:delta_var_lwr_bnd}
    P\left(\frac{n^2}{p^2(1-\frac{n}{p})}\Sigma_{\delta_{ii}} \geq \sigma^2\left(\sqrt{1-\frac{n}{p}}-\frac{1}{\sqrt{1-n/p}}\mu_3\right)^2\right) \geq  1-\left(\frac{2}{p^2}+\frac{2}{n^2}+\frac{1}{2np}\right).
\end{equation}
We need to now derive an upper bound on $\Sigma_{\delta_{ii}}$. By the same argument as before, we have from \eqref{eq:bound_cons_3}
     \begin{eqnarray*}
        P\left(\frac{n}{p\sqrt{1-n/p}}v_{ii} \leq \frac{1}{\sqrt{1-n/p}}\mu_3 \right) &\geq& 1-\left(\frac{2}{p^2}+\frac{2}{n^2}+\frac{1}{2np}\right), \\     \implies P\left(\frac{n}{p\sqrt{1-n/p}}\left(\frac{\boldsymbol{a_{i.}w_{i.}}^{\top}}{n}-1\right) -\sqrt{1-\frac{n}{p}} \leq \frac{1}{\sqrt{1-n/p}}\mu_3 \right) &\geq& 1-\left(\frac{2}{p^2}+\frac{2}{n^2}+\frac{1}{2np}\right), \\ \implies P\left(\frac{n}{p\sqrt{1-n/p}}\left(\frac{\boldsymbol{a_{i.}w_{i.}}^{\top}}{n}-1\right) \leq \sqrt{1-\frac{n}{p}}+\frac{1}{\sqrt{1-n/p}}\mu_3 \right) &\geq& 1-\left(\frac{2}{p^2}+\frac{2}{n^2}+\frac{1}{2np}\right).
\end{eqnarray*}
 Again for $i \in [n] , k \in [n], k \ne i$, $v_{ik}=\frac{\boldsymbol{a_{i.}w_{k.}^{\top}}}{n}$. We have from 
\eqref{eq:bound_cons_3}, 
\begin{equation*}
    P\left(\frac{n}{p\sqrt{1-n/p}}\left|\frac{\boldsymbol{a_{i.}w_{k.}^{\top}}}{n}\right| \leq \frac{1}{\sqrt{1-n/p}}\mu_3 \right) \geq 1-\left(\frac{2}{p^2}+\frac{2}{n^2}+\frac{1}{2np}\right). 
\end{equation*}
     Now from \eqref{eq:Sigma_delta_ii_decomp}, we have 
\begin{eqnarray*}
    \frac{n^2}{p^2(1-\frac{n}{p})}\Sigma_{\delta_{ii}} &=& \sigma^2\left(\frac{n}{p\sqrt{({1-\frac{n}{p}})}}\left(\frac{\boldsymbol{a_{i.}w_{i.}^{\top}}}{n}-1\right)\right)^2 + \sigma^2\sum_{k=1,k\ne i}^n \left(\frac{n}{p\sqrt{1-n/p}}\frac{\boldsymbol{a_{i.}w_{k.}^{\top}}}{n}\right)^2\\  &\leq& \sigma^2\left(\sqrt{1-\frac{n}{p}}+\frac{1}{\sqrt{1-n/p}}\mu_3\right)^2+\frac{\sigma^2(n-1)\mu_3^2}{1-n/p}.
\end{eqnarray*}
The last inequality holds with probability at least $ 1-2\left(\frac{2}{p^2}+\frac{2}{n^2}+\frac{1}{2np}\right)$.
Hence, we have
\begin{equation}\label{eq:delta_var_uppr_bnd}
    P\left(\frac{n^2}{p^2(1-\frac{n}{p})}\Sigma_{\delta_{ii}} \leq \sigma^2\left(\sqrt{1-\frac{n}{p}}+\frac{1}{\sqrt{1-n/p}}\mu_3\right)^2+\frac{\sigma^2(n-1)\mu_3^2}{1-n/p}\right) \geq  1-2\left(\frac{2}{p^2}+\frac{2}{n^2}+\frac{1}{2np}\right).
\end{equation}
Using \eqref{eq:delta_var_uppr_bnd} with \eqref{eq:delta_var_lwr_bnd}, we obtain the following using the union bound, for all $i \in [n]$,
\begin{eqnarray*}
    P\left(\sigma^2\left(\sqrt{1-\frac{n}{p}}-\frac{1}{\sqrt{1-n/p}}\mu_3\right)^2 \leq \frac{n^2}{p^2(1-\frac{n}{p})}\Sigma_{\delta_{ii}} \leq \sigma^2\left(\sqrt{1-\frac{n}{p}}+\frac{1}{\sqrt{1-n/p}}\mu_3\right)^2+\frac{\sigma^2(n-1)\mu_3^2}{1-n/p}\right) \\ \geq 1-3\left(\frac{2}{p^2}+\frac{2}{n^2}+\frac{1}{2np}\right).
\end{eqnarray*}
Therefore, under the assumption $n \log n$ is $o(p)$, we have, $(n-1)\mu_3^2 = (n-1)\frac{\log n}{p} \to 0$ and $\left(\sqrt{1-\frac{n}{p}}+\frac{1}{\sqrt{1-n/p}}\mu_3\right)^2 \to 1$. Hence, we have $\frac{n^2}{p^2(1-\frac{n}{p})}\Sigma_{\delta_{ii}} \overset{P}{\to} \sigma^2$.
This completes the proof.
\hfill{$\blacksquare$}

\subsection{Proof of Theorem~\ref{th:distribution_beta_delta_opt}}Let $\boldsymbol{W}$ be {the output of Alg.~\ref{alg:design_W}}. Using the definition of $\boldsymbol{\hat{\beta}_W}$ from \eqref{eq:deb_beta_est} and the measurement model from \eqref{eq:fm_delta}, we have
\begin{eqnarray}
\boldsymbol{\hat{\beta}_W}-\boldsymbol{\beta^*}&=&\frac{1}{n}\boldsymbol{W}^{\top}\boldsymbol{{\eta}}- \left(\boldsymbol{I_p}-\frac{1}{n}\boldsymbol{W^{\top}}\boldsymbol{A}\right)  (\boldsymbol{\hat{\beta}_{\lambda_1}}-\boldsymbol{\beta^*})+
\frac{1}{n}\boldsymbol{W}^{\top}
\left(\boldsymbol{\delta^{*}}-\boldsymbol{\hat{\delta}_{\lambda_2}}\right).\label{eq:beta_d_decompose_W}
\end{eqnarray}
Using Results (1) and (2) of Theorem~\ref{th:2_3term_beta_decompose_W}, the second and third term on the RHS of \eqref{eq:beta_d_decompose_W} are $o_P(1/\sqrt{n})$. Recall that $\boldsymbol{\Sigma_\beta}=\frac{\sigma^2}{n}\boldsymbol{W^\top W} $.
Therefore, we have \begin{eqnarray}\label{beta_marginal_W}
\frac{\sqrt{n}(\hat{\beta}_{Wj}-\beta^*_j)}{\sqrt{\Sigma_{\beta_{jj}}}}&=&\frac{\frac{1}{\sqrt{n}}\boldsymbol{w}_{. j}^{\top}\boldsymbol{{\eta}}}{\sqrt{\Sigma_{\beta_{jj}}}}+ o_P\left(1/\sqrt{\Sigma_{\beta_{jj}}}\right),
\end{eqnarray}
where $\boldsymbol{w}_{.j}$ denotes the $j^{\text{th}}$ column of matrix $\boldsymbol{W}$. As $\boldsymbol{\eta}$ is Gaussian, the first term on the RHS of \eqref{beta_marginal_W} is a Gaussian random variable with mean $0$ and variance $1$. Using Result (1) of Theorem~\ref{th:dist_conv_W}, $\Sigma_{\beta_{jj}}$ converges to $\sigma^2$ in probability. This completes the proof of Result (1).   

Using \eqref{eq:deb_delta_est} and the measurement model \eqref{eq:fm_delta}, we have 
\begin{eqnarray}
 \boldsymbol{\hat{\delta}_W}-\boldsymbol{\delta^*}=
\left(\boldsymbol{I_n}-\frac{1}{n}\boldsymbol{A}\boldsymbol{W}^{\top}\right)
\boldsymbol{{\eta}}+
\left(\boldsymbol{I_n}-\frac{1}{n}\boldsymbol{A}\boldsymbol{W}^{\top}\right) \boldsymbol{A} (\boldsymbol{\beta^*-\hat{\beta}_{\lambda_1}})-
\frac{1}{n}\boldsymbol{AW}^{\top}
\big(\boldsymbol{\delta^{*}}-\boldsymbol{\hat{\delta}_{\lambda_2}}\big). \label{eq:delta_d_decompose_W}
\end{eqnarray}
Using Results (3) and (4) of Theorem~\ref{th:2_3term_beta_decompose_W}, the second and third term on the RHS of \eqref{eq:delta_d_decompose} are both $o_P\left(\frac{p\sqrt{1-n/p}}{n}\right)$. Recall from \eqref{eq:Sig_delta}, that $\boldsymbol{\Sigma_{\delta}}=\sigma^2\left(\boldsymbol{I_n}-\frac{1}{n}\boldsymbol{A}\boldsymbol{W}^{\top}\right) \left(\boldsymbol{I_n}-\frac{1}{n}\boldsymbol{A}\boldsymbol{W}^{\top}\right)^{\top}$. Therefore, we have  
\begin{eqnarray}
 \frac{\left({\hat{\delta}_{Wi}}-{\delta^*_i}\right)}{ \sqrt{\Sigma_{\delta_{ii}}}}=
\frac{\left(\boldsymbol{I_n}-\frac{1}{n}\boldsymbol{AW}^{\top}\right)^\top_{i.}
\boldsymbol{{\eta}}}{ \sqrt{\Sigma_{\delta_{ii}}}}+o_P\left(\frac{1}{\sqrt{\Sigma_{\delta_{ii}}}}\frac{p\sqrt{1-n/p}}{n}\right).\label{delta_W_dist_W}
\end{eqnarray}
As $\boldsymbol{\eta}$ is Gaussian, the first term on the RHS of \eqref{delta_W_dist_W} is a Gaussian random variable with mean $0$ and variance $1$.
Using Result (2) of Theorem~\ref{th:dist_conv_W}, $\frac{n^2}{p^2(1-n/p)} \Sigma_{\delta_{ii}}$ converges to $\sigma^2$ in probability so that $\left(\frac{1}{\sqrt{\Sigma_{\delta_{ii}}}}\frac{p\sqrt{1-n/p}}{n}\right) = 1$. This completes the proof of result (2).   
\hfill{$\blacksquare$}

\begin{lemma}\label{le:W_mat_prop}
    Let $\boldsymbol{A}$ be a $n \times p$ Rademacher matrix and $\boldsymbol{W}$ be the corresponding output of Alg.~\ref{alg:design_W}. If $n$ is $o(p)$, we have the following results:
    \begin{enumerate}[(1)]
        \item $P\left(\boldsymbol{w_{.j}^{\top}w_{.j}}/n \leq 1 \ \forall j \in [p]\right) = 1$.
        \item $\left|\frac{1}{n}\boldsymbol{W^{\top}}\boldsymbol{A} -\boldsymbol{I_p}\right|_{\infty}$ =  $O_P\left(\sqrt{\frac{\log(p)}{n}}\right)$.
        \item {$\left|\frac{1}{\sqrt{n}}\boldsymbol{W^{\top}}\right|_{\infty}=O\left(1\right)$.}
        \item {$\left|\frac{1}{p}\left(\frac{1}{n}\boldsymbol{A}\boldsymbol{W}^{\top}-\boldsymbol{I_n}\right)\boldsymbol{A}\right|_{\infty}$= $O_P\left(\sqrt{\frac{\log(pn)}{pn}}+\frac{1}{n}\right)$.}
        \item {$\left|\frac{1}{p}\boldsymbol{AW}^{\top}\right|_{\infty} = O_P\left(\sqrt{\frac{n\log(np)}{p}}+1\right)$.}
        \item $\frac{n}{p\sqrt{1-\frac{n}{p}}}\left|\left(\frac{\boldsymbol{AW^{\top}}}{n}-\frac{p}{n}\boldsymbol{I_n}\right)\right|_{\infty} = O_P\left(\frac{1}{\sqrt{1-n/p}}\sqrt{\frac{\log(n)}{p}}\right)$.
    \end{enumerate}
\end{lemma}
\textbf{Proof of Lemma~\ref{le:W_mat_prop}:}\\
In order to prove these results, we will first show that the intersection set of matrices ($\boldsymbol{W}$) that satisfy all four constraints of Alg.~\ref{alg:design_W} is non-null with probability at least $1-\left(\frac{2}{p^2}+\frac{2}{n^2}+\frac{1}{2np}\right)$. In particular, we show that the matrix $\boldsymbol{W}=\boldsymbol{A}$ also satisfies all four constraints. Let us first define the following sets:\\$G_{1}(n,p)=\left\{\boldsymbol{A} \in R^{n \times p}: |\boldsymbol{A^{\top}A}/n-\boldsymbol{I_p}|_{\infty} \leq \mu_1 \right\}$, $G_{2}(n,p)= \left\{ \boldsymbol{A} \in R^{n \times p}:\left|\frac{1}{p} (\boldsymbol{I_n}-\boldsymbol{AA^{\top}}/n)\boldsymbol{A}\right| \leq \mu_2\right\} $, \\ $ G_{3}(n,p) =\left\{ \boldsymbol{A} \in R^{n \times p}:\left|\left(\frac{\boldsymbol{AA^{\top}}}{p}-\boldsymbol{I_n}\right)\right|_{\infty} \leq \mu_3 \right\}$, $G_{4}(n,p)=\{\boldsymbol{A} \in R^{n \times p}: \boldsymbol{a_{.j}^{\top}a_{.j}}/n \leq 1 \ \forall \ j \in [p]\}$
, where, {$\mu_1=2\sqrt{\frac{2\log(p)}{n}}$, $\mu_2=2\sqrt{\frac{\log({2np})}{np}}+\frac{1}{n}$ and $\mu_3=\frac{2}{\sqrt{1-n/p}}\sqrt{\frac{2\log(n)}{p}}$.} Note that, here $\boldsymbol{A}$ is a $n \times p$ Rademacher matrix. We will now state the probabilities of the aforementioned sets.
 From \eqref{eq:V_bnd_max_p1} of Lemma~\ref{le:Rad_mat_prop}, we have 
 \begin{equation}\label{eq:G_1_prob}
      P\left(\boldsymbol{A} \in G_{1}(n,p)\right) \geq 1-\frac{2}{p^2}
 \end{equation}
 Again from \eqref{eq:bound_cons_2} of Lemma \ref{le:Rad_mat_prop}, we have
 \begin{equation}\label{eq:G_2_prob}
     P\left(\boldsymbol{A} \in G_{2}(n,p)\right) \geq 1-\frac{1}{2np}
 \end{equation}
 Similarly from \eqref{eq:bound_cons_3_cond} of Lemma~\ref{le:Rad_mat_prop}, we have 
\begin{equation}\label{eq:G_3_prob}
    P\left(\boldsymbol{A} \in G_{3}(n,p)\right) \geq 1-\frac{2}{n^2}
\end{equation}
Lastly, since $\boldsymbol{A}$ is Rademacher, $P(\boldsymbol{a_{.j}^{\top}a_{.j}}/n \leq 1 \ \forall \ j \in [p] ) = P\left(\cap_{j=1}^p \boldsymbol{a_{.j}^{\top}a_{.j}}/n \leq 1 \right)=\prod_{j=1}^p P(\boldsymbol{a_{.j}^{\top}a_{.j}}/n \leq 1) = 1^p=1$. Therefore, we have,
\begin{equation}\label{eq:G_4_prob}
    P\left(\boldsymbol{A} \in G_{4}(n,p) \right) = 1
\end{equation}
Note that, $P\left(\boldsymbol{A}\in \{\cap_{k=1}^4 G_k(n,p)\}^c\right)=P\left(\boldsymbol{A}\in \{\cup_{k=1}^4 (G_k(n,p))^c\}\right) \leq \sum_{k=1}^4 P\left(\boldsymbol{A}\in (G_k(n,p))^c\right) \leq \left(\frac{2}{p^2}+\frac{2}{n^2}+\frac{1}{2np}\right)$. Therefore, we obtain, $P\left(\boldsymbol{A}\in \{\cap_{k=1}^4 G_k(n,p)\}^c\right)=1-P\left(\boldsymbol{A}\in \{\cap_{k=1}^4 G_k(n,p)\}\right) \leq \left(\frac{2}{p^2}+\frac{2}{n^2}+\frac{1}{2np}\right)$ . Hence, 
\begin{equation}\label{eq:feasibility_Identity}
        P\left(\boldsymbol{A}\in \{\cap_{k=1}^4 G_{k}(n,p)\}\right) \geq 1-\left(\frac{2}{p^2}+\frac{2}{n^2}+\frac{1}{2np}\right).
    \end{equation}
    {Therefore, $\boldsymbol{A}$ satisfies the constraints of Alg.~\ref{alg:design_W} with high probability. This implies that there exists $\boldsymbol{W^*}$ that satisfies the constraints.}
    {Let \\
    \begin{multline*}
        E(n,p)=\Bigg\{\boldsymbol{A}: \exists \boldsymbol{W^*} \ \text{s.t.}\ |\boldsymbol{{W^*}^{\top}A}/n-\boldsymbol{I_p}|_{\infty} \leq \mu_1,\ \left|\frac{1}{p} (\boldsymbol{I_n}-\boldsymbol{A{W^*}^{\top}}/n)\boldsymbol{A}\right| \leq \mu_2,\\ \left|\left(\frac{\boldsymbol{A{W^*}^{\top}}}{p}-\boldsymbol{I_n}\right)\right|_{\infty} \leq \mu_3, \boldsymbol{{w^*}_{.j}^{\top} {w^*}_{.j}}/n \leq 1 \ \forall \ j \in [p] \Bigg\}.\end{multline*} Hence, we have
\begin{equation}\label{eq:joint_feasibility}
    P(\boldsymbol{A}\in E(n,p)) \geq P(\boldsymbol{A}\in \cap_{k=1}^{4}G_k(n,p)) \geq 1-\left(\frac{2}{p^2}+\frac{2}{n^2}+\frac{1}{2np}\right)
\end{equation}
    Given that the set of feasible solutions is non-null, we can say that the optimal solution of Alg.~\ref{alg:design_W} denoted by $\boldsymbol{W}$ satisfies the constraints of Alg.~\ref{alg:design_W} with probability $1$.
    \newline
\noindent\textbf{Result (1):}
{
Recall that the event that there exists a matrix $\boldsymbol{A}$ satisfying constraints $\mathsf{C0}$--$\mathsf{C3}$ is $ E(n,p)$.
We have 
\begin{eqnarray}
   \nonumber P\left(\boldsymbol{w_{.j}^{\top}w_{.j}}/n \leq 1 \ \forall j \in [p]\right) &=& P\left(\boldsymbol{w_{.j}^{\top}w_{.j}}/n \leq 1 \ \forall j \in [p] \bigg | \boldsymbol{A}\in E(n,p)\right) P\left(\boldsymbol{A}\in E(n,p)\right) \\ &+& P\left(\boldsymbol{w_{.j}^{\top}w_{.j}}/n \leq 1 \ \forall j \in [p] \bigg | \boldsymbol{A}\in E(n,p)^c\right) P\left(\boldsymbol{A}\in E(n,p)^c\right)  \label{eq:wj_int_break}
\end{eqnarray}
If there exists a feasible solution to $\mathsf{C0}$--$\mathsf{C3}$ then $\boldsymbol{w}_{.j}^\top \boldsymbol{w}_{.j}/n\leq 1$. Therefore,} we have
\begin{equation}\label{eq:cond_A_feas}
    P\left(\boldsymbol{w_{.j}^{\top}w_{.j}}/n \leq 1 \ \forall j \in [p] \bigg | \boldsymbol{A}\in E(n,p)\right)=1.
\end{equation}
 Now, we have from Alg.~\ref{alg:design_W}, if the constraints of the optimisation problem are not satisfied, then we choose $\boldsymbol{W}=\boldsymbol{A}$ as the output.
This event is given by $\boldsymbol{A}\in E(n,p)^c$. Now, we know that for Rademacher matrix $\boldsymbol{A}$, $\boldsymbol{a_{.j}^{\top}a_{.j}}/n = 1$ with probability 1. Therefore, we have
\begin{equation}\label{eq:cond_A_not_feas}
    P\left(\boldsymbol{w_{.j}^{\top}w_{.j}}/n \leq 1 \ \forall j \in [p] \bigg | \boldsymbol{A}\in E(n,p)^c\right)=1.
\end{equation}
Therefore, we have from \eqref{eq:cond_A_feas},\eqref{eq:cond_A_not_feas} and \eqref{eq:wj_int_break},
\begin{equation}\label{eq:col_Wj_prob}
    P\left(\boldsymbol{w_{.j}^{\top}w_{.j}}/n \leq 1 \ \forall j \in [p]\right)=  P\left(\boldsymbol{A}\in E(n,p)\right)+ P\left(\boldsymbol{A}\in E(n,p)^c\right)=1.
\end{equation}
\newline    
\noindent \textbf{Result (2):}
Recall that $\mu_1=2\sqrt{\frac{2\log(p)}{n}}$. Note that we have for any two events $F_1,F_2$, $P(F_1)=P(F_1 \cap F_2)+P(F_1 \cap F_2^c) \leq P(F_1 \cap F_2) + P(F_2^c)$.
 Therefore, we have,
\begin{eqnarray*}
    P\left(|\boldsymbol{W^{\top}A}/n-\boldsymbol{I_p}|_{\infty} \geq \mu_1\right) &\leq&  P\left(\{|\boldsymbol{W^{\top}A}/n-\boldsymbol{I_p}|_{\infty} \geq \mu_1\}\cap E(n,p)\right) + P\left(\{E(n,p)\}^c\right) \\
    &\leq& P\left(\{|\boldsymbol{W^{\top}A}/n-\boldsymbol{I_p}|_{\infty} \geq \mu_1\}\cap E(n,p)\right) + \left(\frac{2}{p^2}+\frac{2}{n^2}+\frac{1}{2np}\right)
\end{eqnarray*}
The last inequality comes from \eqref{eq:joint_feasibility}. Since $\boldsymbol{W}$ is a feasible solution, given $\boldsymbol{A} \in E(n,p)$, it will satisfy the second constraint of Alg.~\ref{alg:design_W} with probability 1. This means that
\begin{equation*}
    P\left(\{|\boldsymbol{W^{\top}A}/n-\boldsymbol{I_p}|_{\infty} \geq \mu_1\}\cap E(n,p)\right) = 0.
\end{equation*}
Therefore we have,
\begin{equation}\label{eq:bound_cons_1}
    P\left(|\boldsymbol{W^{\top}A}/n-\boldsymbol{I_p}|_{\infty} \leq \mu_1\right) \geq 1-\left(\frac{2}{p^2}+\frac{2}{n^2}+\frac{1}{2np}\right).
\end{equation}
Since $\mu_1=2\sqrt{\frac{2\log(p)}{n}}$, we have, $|\boldsymbol{W^{\top}A}/n-\boldsymbol{I_p}|_{\infty}= O_P(\sqrt{\log(p)/n})$.
\newline
\textbf{Result (3):}
From \eqref{eq:col_Wj_prob}, we have that, for each $j \in [p]$, $\|\boldsymbol{w_{.j}}\|_2 \leq \sqrt{n}$ with probability $1$.
Note that for any vector $\boldsymbol{x}$, $\|\boldsymbol{x}\|_{\infty} \leq \|\boldsymbol{x}\|_2$, we have for every $j \in [p]$, $\|\boldsymbol{w_{.j}}\|_\infty \leq \sqrt{n}$ with probability $1$.
Since $|\boldsymbol{W}^\top|_{\infty} \leq \underset{j \in [p]}{\max}\|\boldsymbol{w_{.j}}\|_\infty \leq \sqrt{n}$ with probability 1.
Therefore, we have
\begin{equation}\label{eq:W^T_bnd_inf}
    \left|\frac{1}{\sqrt{n}}\boldsymbol{W}^\top\right|_{\infty}=O(1).
\end{equation}}
\newline
\textbf{Result (4):}
{Recall that $\mu_2=\frac{1}{n}+2\sqrt{\frac{\log({2np})}{np}}$.} Therefore, we have
\begin{eqnarray*}
    P\left(\left|\frac1p\left(\boldsymbol{I_n}-\frac{1}{n}\boldsymbol{A}\boldsymbol{W}^{\top}\right) \boldsymbol{A}\right|_{\infty} \geq \mu_2\right) &\leq&  P\left(\left\{\left|\frac1p\left(\boldsymbol{I_n}-\frac{1}{n}\boldsymbol{A}\boldsymbol{W}^{\top}\right) \boldsymbol{A}\right|_{\infty} \geq \mu_2\right\}\cap E(n,p)\right) + P\left(\{E(n,p)\}^c\right) \\
    &\leq& P\left(\left\{\left|\frac1p\left(\boldsymbol{I_n}-\frac{1}{n}\boldsymbol{A}\boldsymbol{W}^{\top}\right) \boldsymbol{A}\right|_{\infty} \geq \mu_2\right\}\cap E(n,p)\right) + \left(\frac{2}{p^2}+\frac{2}{n^2}+\frac{1}{2np}\right)
\end{eqnarray*}
The last inequality comes from \eqref{eq:joint_feasibility}. Note that, since $\boldsymbol{W}$ is a feasible solution, given $\boldsymbol{A} \in E(n,p)$, it will satisfy the third constraint of Alg.~\ref{alg:design_W} with probability 1. This implies
\begin{equation*}
    P\left(\left\{\left|\frac1p\left(\boldsymbol{I_n}-\frac{1}{n}\boldsymbol{A}\boldsymbol{W}^{\top}\right) \boldsymbol{A}\right|_{\infty}\geq \mu_2\right\}\cap E(n,p)\right) = 0.
\end{equation*}
Therefore, we have
\begin{equation}
    P\left(\left|\frac1p\left(\boldsymbol{I_n}-\frac{1}{n}\boldsymbol{A}\boldsymbol{W}^{\top}\right) \boldsymbol{A}\right|_{\infty} \leq \mu_2\right) \geq 1-\left(\frac{2}{p^2}+\frac{2}{n^2}+\frac{1}{2np}\right).
\end{equation}
Hence, $\left|\frac{1}{p}\left(\frac{1}{n}\boldsymbol{A}\boldsymbol{W}^{\top}-\boldsymbol{I_n}\right)\boldsymbol{A}\right|_{\infty}$= $O_P\left(\sqrt{\frac{\log(pn)}{pn}}+1/n\right)$.
\newline
\textbf{Result (5):}
Recall that from Eqn.\eqref{eq:joint_feasibility}, we have with probability at least $1-\left(\frac{2}{p^2}+\frac{2}{n^2}+\frac{1}{2np}\right)$, 
\begin{equation*}
    \left|\frac{1}{p} (\boldsymbol{AW^{\top}}/n-\boldsymbol{I_n})\boldsymbol{A}\right|_{\infty} \leq \mu_2.
\end{equation*}
Applying triangle inequality, we have with probability atleast $1-\left(\frac{2}{p^2}+\frac{2}{n^2}+\frac{1}{2np}\right)$,
\begin{equation}\label{eq:bnd_tr_cons_2}
    \left|\frac{1}{np}\boldsymbol{AW}^{\top}\boldsymbol{A}\right|_{\infty} \leq \left|\frac{1}{p} (\boldsymbol{AW^{\top}}/n-\boldsymbol{I_n})\boldsymbol{A}\right|_{\infty} + \frac{1}{p}|\boldsymbol{A}|_{\infty} \leq \mu_2 + 1/p.
\end{equation}
We now present some useful results about the norms being used in this proof. Let $\boldsymbol{X}$ be a $p \times p$ matrix and $\boldsymbol{Y}$ be a $p \times n$ matrix .
Recall the following definitions from \cite{Todd1977},
\begin{align*}
\|\boldsymbol{Y}\|_{\infty\rightarrow\infty}  \triangleq \max_{\boldsymbol{x} \in \mathbb{R}^{n}\setminus\{\boldsymbol{0}\}}\frac{\|\boldsymbol{Yx}\|_{\infty}}{\|\boldsymbol{x}\|_{\infty}}\quad\textup{and} \ \|\boldsymbol{Y}\|_{2\rightarrow2}  \triangleq \max_{\boldsymbol{x} \in \mathbb{R}^{n}\setminus\{\boldsymbol{0}\}}\frac{\|\boldsymbol{Yx}\|_{2}}{\|\boldsymbol{x}\|_{2}}=\sigma_{max}(\boldsymbol{Y}).
\end{align*}
 Note that $|\boldsymbol{XY}|_{\infty} \leq \|\boldsymbol{X}\|_{\infty\rightarrow\infty}|\boldsymbol{Y}|_{\infty}$\footnote{This is because $|\boldsymbol{XY}|_{\infty} = \max_{i} \|\boldsymbol{X Y_{.i}}\|_{\infty}
\leq \max_{i} \|\boldsymbol{X}\|_{\infty \rightarrow \infty} \|\boldsymbol{Y_{.i}}\|_{\infty}$ 
(by the definition of the induced norm)
$= \|\boldsymbol{X}\|_{\infty \rightarrow \infty} \max_{i}\|\boldsymbol{Y_{.i}}\|_{\infty} = \|\boldsymbol{X}\|_{\infty \rightarrow \infty} |\boldsymbol{Y}|_{\infty}$.}. Since $\frac{1}{\sqrt{p}}\|\boldsymbol{x}\|_2\le\|\boldsymbol{x}\|_{\infty}$  {and $ \|\boldsymbol{Y}^\top\boldsymbol{x}\|_{\infty} \leq \|\boldsymbol{Y}^\top\boldsymbol{x}\|_2$ for all $\boldsymbol{x}\in\mathbb{R}^{p}$ , we have} 
{\begin{equation}
    \label{eq:normineq}    \|\boldsymbol{Y^\top}\|_{\infty\rightarrow\infty} \leq \sqrt{p}\|\boldsymbol{Y}^\top\|_{2\rightarrow 2}.
\end{equation}}
Then by using \eqref{eq:normineq}, we have 
\begin{eqnarray}\label{eq:inv_ineq}
|\boldsymbol{XY}|_{\infty}=|\boldsymbol{Y}^{\top}\boldsymbol{X}^{\top}|_{\infty} &\leq \|\boldsymbol{Y}^{\top}\|_{\infty\rightarrow\infty}|\boldsymbol{X}^{\top}|_{\infty} \leq \sqrt{p}\|\boldsymbol{Y}^{\top}\|_{2\rightarrow 2}|\boldsymbol{X}^{\top}|_{\infty}
     = \sqrt{p}\sigma_{max}(\boldsymbol{Y})|\boldsymbol{X}|_{\infty}.
\end{eqnarray}
Substituting $\boldsymbol{X} =\frac{1}{np}\boldsymbol{A}\boldsymbol{W}^\top\boldsymbol{A}$ and $\boldsymbol{Y=A^{\dagger}}\triangleq \boldsymbol{A}^{\top}(\boldsymbol{AA}^{\top})^{-1}$, the Moore-Penrose pseudo-inverse of $\boldsymbol{A}$, in \eqref{eq:inv_ineq}, we obtain:
\begin{equation}\label{eq:Bnd_first_term}
\left|\frac{1}{np}\boldsymbol{AW^{\top}}\right|_{\infty} =\left|\frac{1}{np}\boldsymbol{AW}^{\top}\boldsymbol{AA}^{\dagger}\right|_{\infty}\leq \frac{\sqrt{p}}{np}|\boldsymbol{AW}^{\top}\boldsymbol{A}|_{\infty}\|\boldsymbol{A^{\dagger}}\|_{2\rightarrow2}.
\end{equation}
We now derive the upper bound for the second factor of \eqref{eq:Bnd_first_term}. We have,
\begin{equation}
\|\boldsymbol{A^{\dagger}}\|_{2\rightarrow2}=\sigma_{max}(\boldsymbol{A^{\dagger}})=\frac{1}{\sigma_{min}(\boldsymbol{A})}=\frac{1}{\sigma_{min}(\boldsymbol{A}^{\top})}.
\end{equation}
Note that, for an arbitrary $\epsilon_1>0$, using Theorem 1.1. from \cite{Rudelson2009} for the mean-zero sub-Gaussian random matrix $\boldsymbol{A}$, we have the following 
\begin{equation}\label{eq:min_sing_A}
    P(\sigma_{min}(\boldsymbol{A}^{\top})\leq \epsilon_1(\sqrt{p}-\sqrt{n-1})) \leq (c_{6}\epsilon_1)^{p-n+1}+(c_5)^{p}.
\end{equation}
 where $c_6 > 0$ and $c_5 \in (0,1)$ are constants dependent on the sub-Gaussian norm of the entries of $\boldsymbol{A}^{\top}$. Let for some small constant $\psi \in (0,1)$ $\epsilon_1 c_6\triangleq \psi$, we have   
\begin{equation}\label{eq:psuedo_sing}
    P\left(\sigma_{max}(\boldsymbol{A^{\dagger}}) \leq \frac{c_6}{\psi(\sqrt{p}-\sqrt{n-1})}\right) \geq 1-((\psi)^{p-n+1}+(c_5)^{p}).
\end{equation}
 we have
\begin{eqnarray} 
  P\left(\sigma_{max}(\boldsymbol{A^{\dagger}}) \leq \frac{c_6}{\psi\sqrt{p}\left(1-\frac{\sqrt{n-1}}{\sqrt{p}}\right)}\right) \geq 1-((\psi)^{p-n+1}+(c_5)^{p}). \label{eq:Sig_max}
\end{eqnarray} 
Using Eqns.~\eqref{eq:Sig_max} and \eqref{eq:bnd_tr_cons_2}, we have
{\begin{eqnarray*}
    P\left[\sqrt{p}\frac{1}{np}|\boldsymbol{AW}^{\top}\boldsymbol{A}|_{\infty}|\boldsymbol{A^{\dagger}}|_{2\rightarrow2} \leq \left(\mu_2+\frac{1}{p}\right)\frac{c_6}{\psi(1-\sqrt{n/p})}\right] \geq 1-\left(\left(\frac{2}{p^2}+\frac{2}{n^2}+\frac{1}{2np}\right)+(\psi)^{p-n+1}+(c_5)^{p}\right).
\end{eqnarray*}
Therefore we have by Bonferroni's inequality,
\begin{equation*}
    P\left(\left|\frac{1}{np}\boldsymbol{AW}^{\top}\right|_{\infty} \leq \left(\mu_2+\frac{1}{p}\right)\frac{c_6}{\psi(1-\sqrt{n/p})}\right) \geq 1-\left(\left(\frac{2}{p^2}+\frac{2}{n^2}+\frac{1}{2np}\right)+(\psi)^{p-n+1}+(c_5)^{p}\right).
\end{equation*}}
Under the condition $n=o(p)$ as $n,p \to \infty$, the probability $1-\left(\frac{2}{p^2}+\frac{2}{n^2}+\frac{1}{2np}+(\psi)^{p-n+1}+(c_5)^{p}\right) \to 1$. Therefore, we have,
{\begin{equation}\label{eq:Mat_D3_bnd}
   \left|\frac{1}{p}\boldsymbol{AW}^{\top}\right|_{\infty} = O_P\left(n\left(\sqrt{\frac{\log(np)}{np}}+\frac{1}{n}+\frac{1}{p}\right)\right)=O_P\left(\sqrt{\frac{n\log(np)}{p}}+1+\frac{n}{p}\right)=O_P\left(\sqrt{\frac{n\log(np)}{p}}+1\right).
\end{equation}}
\newline
\textbf{Result (6):}
Recall that $\mu_3=2\sqrt{\frac{2\log(n)}{p}}$.
We have from \eqref{eq:bound_cons_3_cond} of Lemma \ref{le:Rad_mat_prop},
\begin{eqnarray*}
   P\left(\left|\left(\frac{\boldsymbol{AW^{\top}}}{p}-\boldsymbol{I_n}\right)\right|_{\infty} \geq \mu_3\right) &=& P\left(\frac{n}{p\sqrt{1-\frac{n}{p}}}\left|\left(\frac{\boldsymbol{AW^{\top}}}{n}-\frac{p}{n}\boldsymbol{I_n}\right)\right|_{\infty} \geq \frac{1}{\sqrt{1-n/p}}\mu_3\right)\\ &\leq&  P\left(\left\{\frac{n}{p\sqrt{1-\frac{n}{p}}}\left|\left(\frac{\boldsymbol{AW^{\top}}}{n}-\frac{p}{n}\boldsymbol{I_n}\right)\right|_{\infty} \geq \frac{1}{\sqrt{1-n/p}}\mu_3\right\}\cap E(n,p)\right) + P\left(\{E(n,p)\}^c\right) \\
    &\leq& P\left(\left\{\frac{n}{p\sqrt{1-\frac{n}{p}}}\left|\left(\frac{\boldsymbol{AW^{\top}}}{n}-\frac{p}{n}\boldsymbol{I_n}\right)\right|_{\infty} \geq \frac{1}{\sqrt{1-n/p}}\mu_3\right\}\cap E(n,p)\right) + \left(\frac{2}{p^2}+\frac{2}{n^2}+\frac{1}{2np}\right)
\end{eqnarray*}
The last inequality comes from \eqref{eq:joint_feasibility}. Note that, since $\boldsymbol{W}$ is a feasible solution, given $\boldsymbol{A} \in E_3(n,p)$, it will satisfy the fourth constraint of Alg.~\ref{alg:design_W} with probability 1. This implies that:
\begin{equation*}
    P\left(\left\{\frac{n}{p\sqrt{1-\frac{n}{p}}}\left|\left(\frac{\boldsymbol{AW^{\top}}}{n}-\frac{p}{n}\boldsymbol{I_n}\right)\right|_{\infty} \geq \frac{1}{\sqrt{1-n/p}}\mu_3\right\}\cap E(n,p)\right) = 0,
\end{equation*}
which yields:
\begin{equation}
    P\left(\frac{n}{p\sqrt{1-\frac{n}{p}}}\left|\left(\frac{\boldsymbol{AW^{\top}}}{n}-\frac{p}{n}\boldsymbol{I_n}\right)\right|_{\infty} \leq \frac{1}{\sqrt{1-n/p}}\mu_3\right) \geq 1-\left(\frac{2}{p^2}+\frac{2}{n^2}+\frac{1}{2np}\right).
\end{equation}
Since, $1-\left(\frac{2}{p^2}+\frac{2}{n^2}+\frac{1}{2np}\right)$ goes to $0$ as $n,p \to \infty$, the proof is completed.
\hfill{$\blacksquare$}

\section{Lemma on properties of Rademacher \texorpdfstring{$\boldsymbol{A}$}{A}}\label{sec:App_A3}
\begin{lemma}\label{le:Rad_mat_prop}
    Let $\boldsymbol{A}$ be a $n\times p$ random Rademacher matrix. Then the following holds:
    \begin{enumerate}
        \item $\left|\frac{1}{n}\boldsymbol{A^{\top}}\boldsymbol{A} -\boldsymbol{I_p}\right|_{\infty}$ = $O_P\left(\sqrt{\frac{\log(p)}{n}}\right)$.
        \item $\left|\frac{1}{p}\left(\frac{1}{n}\boldsymbol{A}\boldsymbol{A}^{\top}-\boldsymbol{I_n}\right)\boldsymbol{A}\right|_{\infty}=O_P\left(\sqrt{\frac{\log(pn)}{pn}}+\frac{1}{n}\right)$.
        \item If $n<p$, then $\frac{n}{p\sqrt{1-\frac{n}{p}}}\left|\left(\frac{\boldsymbol{AA^{\top}}}{n}-\frac{p}{n}\boldsymbol{I_n}\right)\right|_{\infty} = O_P\left(\frac{1}{\sqrt{1-n/p}}\sqrt{\frac{\log(n)}{p}}\right)$. 
    \end{enumerate}
\end{lemma}
\textbf{Proof of Lemma \ref{le:Rad_mat_prop}, [Result (1)]:}
 Let $\boldsymbol{V} \triangleq \boldsymbol{A^{\top}A}/n-\boldsymbol{I_p}$. Note that elements of $\boldsymbol{V}$ matrix satisfies the following:
 \begin{equation} \label{eq:v_result1}
  v_{lj}=   \begin{cases}
  \sum_{k=1}^n \frac{(a_{kl})^2}{n}-1=0& \mbox{if } j=l, \ l\in[p]\\
  \sum_{k=1}^n \frac{a_{kl}a_{kj}}{n} & \mbox{if } j\ne l, \ j,l\in[p]
 \end{cases}
 \end{equation}
 Therefore, we now consider off-diagonal elements of $\boldsymbol{V}$ (i.e., $l\ne j$) for the bound.  
Each summand of $v_{lj}$ is uniformly bounded  $-\frac{1}{n}\leq \frac{a_{kl}a_{kj}}{n} \leq \frac{1}{n}$ since the elements of $\boldsymbol{A}$ are $\pm 1$. Note that  $E\left[\frac{a_{kl}a_{kj}}{n}\right]=0$  $\forall k \in [n], \ l\ne j\in[p]$. Furthermore, for $l\ne j\in[p]$, each of the summands of $v_{lj}$ are independent since the elements of $\boldsymbol{A}$ are independent. Therefore, using Hoeffding's Inequality (see Lemma 1 of \cite{Lugosi}) for $t>0$, 
\begin{eqnarray*}
    P(|v_{lj}|\geq t) = P\left(\left|\sum_{k=1}^n \frac{a_{kl}a_{kj}}{n}\right|\geq t\right)\leq 2e^{-\frac{nt^2}{2}}, \  l\ne j\in[p].
\end{eqnarray*}
Therefore we have 
\begin{eqnarray}
    P\left(\max_{l \ne j \in [n]}|v_{lj}|\geq t\right) &=&P\left(\cup_{l \ne j \in [n]}|v_{lj}|\geq t\right) \leq \sum_{l \ne j \in [n]} P(|v_{lj}|\geq t)  \leq  2p(p-1)e^{-\frac{nt^2}{2}} < 2p^2e^{-\frac{nt^2}{2}}. \label{eq:max_R1}
\end{eqnarray}
 Putting  $t=2\sqrt{\frac{2\log p}{n}}$ in \eqref{eq:max_R1}, we obtain:
\begin{eqnarray}\label{eq:V_bnd_max_p1}
P\left(\max_{l \ne j \in [n]}|v_{lj}| \geq 2\sqrt{\frac{2\log p}{n}}\right) \leq 2p^2 e^{-4\log(p)} =2p^{-2}.
\end{eqnarray}
Thus, we have:
\begin{equation}\label{eq:V_bound_r1}
|\boldsymbol{V}|_{\infty}=O_P\left(\sqrt{\frac{\log p}{n}}\right).
\end{equation}
This completes the proof of Result (1).

\noindent \textbf{Result (2):} 
Note that, 
\begin{equation}\label{eq:R2_mat_equiv}
    \frac1p\left(\frac{1}{n}\boldsymbol{A}\boldsymbol{A}^{\top}-\boldsymbol{I_n}\right) \boldsymbol{A}=\frac1p\left(\frac{1}{n}\boldsymbol{A}\boldsymbol{A}^{\top}\boldsymbol{A}-\boldsymbol{A}\right) \triangleq \boldsymbol{V}.
\end{equation}
Fix $i \in [n]$ and $j \in [p]$, and consider 
\[ v_{ij}  = -\frac{a_{ij}}{p} + \frac{1}{np}\sum_{l=1}^{p}\sum_{k=1}^{n}a_{il}a_{kl}a_{kj}\]
By splitting the sum over $l$ into the terms where $l\neq j$ and the term where $l=j$, and simplifying by using the fact that $a_{kj}^2=1$ for all $k,j$, we obtain
\begin{align*}
    v_{ij}
    & = -\frac{a_{ij}}{p} + \frac{1}{np}\left(\sum_{\substack{l=1\\l\neq j}}^{p}\sum_{k=1}^{n}a_{il}a_{kl}a_{kj}\right) + \frac{1}{np}\left(\sum_{k=1}^{n}a_{ij}a_{kj}^2\right)\\
    & = -\frac{a_{ij}}{p} + \frac{1}{np}\left(\sum_{\substack{l=1\\l\neq j}}^{p}\sum_{k=1}^{n}a_{il}a_{kl}a_{kj}\right) + \frac{a_{ij}}{p}\frac{1}{n}\left(\sum_{k=1}^{n}1\right)\\
    & = \frac{1}{np}\left(\sum_{\substack{l=1\\l\neq j}}^{p}\sum_{k=1}^{n}a_{il}a_{kl}a_{kj}\right).
    \end{align*}
    Next we split the sum over $k$ into the terms where $k\neq i$ and the term where $k=i$ to obtain
    \begin{equation} v_{ij}  = \frac{1}{np}\left(\sum_{\substack{l=1\\l\neq j}}^{p}\sum_{k=1}^{n}a_{il}a_{kl}a_{kj}\right)
    = \frac{1}{np}\left(\sum_{\substack{l=1\\l\neq j}}^{p}\sum_{\substack{k=1\\k\neq i}}^{n}a_{il}a_{kl}a_{kj}\right) + \frac{1}{np}\sum_{\substack{l=1\\l\neq j}}^{p}a_{il}^2a_{ij}
    = \frac{1}{np}\left(\sum_{\substack{l=1\\l\neq j}}^{p}\sum_{\substack{k=1\\k\neq i}}^{n}a_{il}a_{kl}a_{kj}\right) + \frac{p-1}{np}a_{ij}.
    \end{equation}
If we condition on $\boldsymbol{a}_{i.}$ and $\boldsymbol{a}_{.j}$,  the $(n-1)(p-1)$ random variables $\frac{1}{np}a_{il}a_{kl}a_{kj}$ for $k\in [n]\setminus\{i\}$ and $l\in [p]\setminus \{j\}$ are independent, have mean zero, and are bounded between $-\frac{1}{np}$ and $\frac{1}{np}$. Therefore, by Hoeffding's inequality, for $t>0$, we have
\begin{equation}\label{eq:uncond_prob_r2}
        P\left(\left.\left|\frac{1}{np}\sum_{\substack{l=1\\l\neq j}}^{p}\sum_{\substack{k=1\\k\neq i}}^{n}a_{il}a_{kl}a_{kj}\right|\geq t\;\right|\;\boldsymbol{a}_{i.},\boldsymbol{a}_{.j}\right)\leq 2e^{-\frac{t^2/2}{(n-1)(p-1)/(n^2p^2)}} \leq 2e^{-\frac{n^2p^2t^2}{2(n-1)(p-1)}}\leq 2e^{-npt^2/2}.
    \end{equation}
    Since the RHS of \eqref{eq:uncond_prob_r2} is independent of $\boldsymbol{a}_{i,}$ and $\boldsymbol{a}_{.j}$ the bound also holds on the unconditional probability, i.e., we have 
    \begin{equation}\label{eq:cond_prob_r2}
    P\left(\left|\frac{1}{np}\sum_{\substack{l=1\\l\neq j}}^{p}\sum_{\substack{k=1\\k\neq i}}^{n}a_{il}a_{kl}a_{kj}\right|\geq t\right) \leq 2e^{-\frac{n^2p^2t^2}{2(n-1)(p-1)}}\leq 2e^{-npt^2/2}.
    \end{equation}
    Since $a_{ij}$ is Rademacher, $\frac{p-1}{np}|a_{ij}|< \frac{1}{n}$ with probability $1$. Choosing $t = 2\sqrt{\frac{\log(2np)}{np}}$ and using the triangle inequality, we have from \eqref{eq:cond_prob_r2}, 
    \begin{align*} 
    P\left(|v_{ij}| \geq \frac{1}{n}+2\sqrt{\frac{\log(2np)}{np}}\right) &\leq P\left(\left|\frac{1}{np}\sum_{\substack{l=1\\l\neq j}}^{p}\sum_{\substack{k=1\\k\neq i}}^{n}a_{il}a_{kl}a_{kj}\right| + \frac{p-1}{np}|a_{ij}|\geq 2\sqrt{\frac{\log(2np)}{np}} + \frac{1}{n}\right)\\
    &\leq P\left(\left|\frac{1}{np}\sum_{\substack{l=1\\l\neq j}}^{p}\sum_{\substack{k=1\\k\neq i}}^{n}a_{il}a_{kl}a_{kj}\right|\geq 2\sqrt{\frac{\log(2np)}{np}}\right) \leq \frac{1}{2n^2p^2}.\end{align*}
    Then, by a union bound over $np$ events,
    \begin{equation}\label{eq:bound_cons_2}
    P\left(\max_{i,j}|v_{ij}| \geq \frac{1}{n} + 2\sqrt{\frac{\log(2np)}{np}}\right) \leq np\,\frac{1}{2n^2p^2} = \frac{1}{2np}.
    \end{equation}
    This completes the proof of Result (2).    
\newline
\noindent\textbf{Result (3):}  
Reversing the roles of $n$ and $p$ in result (1), \eqref{eq:v_result1} and \eqref{eq:V_bnd_max_p1} of this lemma, we have 
\begin{equation}
    P\left(\left|\frac{\boldsymbol{AA^{\top}}}{p}-\boldsymbol{I_n}\right|_{\infty}
 \leq 2\sqrt{\frac{2\log(n)}{p}}\right) \geq 1-\frac{2}{n^2}.
\end{equation}
% Now, multiplying by $\frac{1}{\sqrt{1-n/p}}$, we get
%\begin{equation}\label{eq:bound_cons_3_cond_cent}
 %   P\left(\frac{n}{p\sqrt{1-\frac{n}{p}}}\left|\left(\frac{\boldsymbol{AA^{\top}}}{n}-\frac{p}{n}\boldsymbol{I_n}\right)\right|_{\infty}\leq \frac{2}{\sqrt{1-n/p}}\sqrt{\frac{2\log(n)}{p}}\right) \geq  1-\frac{2}{n^2}.
  %  \end{equation}
This completes the proof.
\hfill{$\blacksquare$}
\section{{Results for the centered bounded pooling matrix $\boldsymbol{A}$}}\label{Sec:App_A4}
{We now establish the EREC property of the centered bounded pooling matrix $\boldsymbol{A}$, which  leads to performance bounds for the robust \textsc{Lasso} estimator for this matrix model (see Remark~4 of Theorem~\ref{th:upper_bound_robustLasso}).
\begin{lemma}\label{le:EREC_bound_mat} 
Let $\boldsymbol{A}$ be an $n \times p$ matrix with i.i.d. entries obtained from a distribution with mean $0$ and variance $1$ and defined on a bounded domain $[-h,h]$. There exists positive constants $C_1(h),C_2(h),c_3,c_4$ such that if $n \geq C_1(h) s \log p$ and $r \leq \textrm{min}\{C_2(h) \frac{n}{ \log n},\frac{s\log p}{\log n}\}$ then
     \begin{equation*}
        P\left(\forall \ (\boldsymbol{h_{\beta}},\boldsymbol{h_{\delta}}) \in \mathcal{C}\left(\mathcal{S},\mathcal{R},\lambda\right), \ \frac{1}{\sqrt{n}} \|\boldsymbol{Ah_{\beta}}+\sqrt{n}\boldsymbol{h_{\delta}}\|_2 \geq \frac{1}{16}(\|\boldsymbol{h_{\beta}}\|_2+\|\boldsymbol{h_{\delta}}\|_2)\right) \geq 1-c_3 \exp{\{-c_4n\}},
    \end{equation*}
      where $\lambda := \sqrt{\frac{\log n}{\log p}}$ and $\mathcal{C}$ as in \eqref{eq:Coneconstraint}.
\hfill{$\blacksquare$}
\end{lemma}
\textbf{Proof of Lemma \ref{le:EREC_bound_mat}:}
The proof follows along similar lines as that of Lemma \ref{le:Ext_RE} except the adaptation of the properties given in \eqref{eq:lwr_bnd_Rade} and \eqref{eq:max_sing_val} for the random centered bounded matrix $\boldsymbol{A}$. Recall that the elements of $\boldsymbol{A}$ are i.i.d. with mean $0$ and variance $1$ and defined on a bounded domain $[-h,h]$
We use the following results to obtain the equivalent of \eqref{eq:lwr_bnd_Rade} and \eqref{eq:max_sing_val} for this matrix:
\begin{enumerate}
    \item In Lemma~1 of \cite{LiRaskutti}, we set $\bar{D}$ as identity matrix, since %we are concerned with signals that 
    the chosen signal $\boldsymbol{\beta^*}$ is sparse in the canonical basis. We show in Lemma \ref{le:isotropic} that the distribution of the elements of $\boldsymbol{A}$ is isotropic (Defn. 2.2 of \cite{LiRaskutti}) which is used to obtain the elements of $\boldsymbol{A}$. 
    Therefore using Lemma 1 of \cite{LiRaskutti}, there exist positive constants $c_2=c_2(h),c'_3,c'_4$, such that with probability
at least $1-c'_3 \exp{\{-c'_4n\}}$:
\begin{equation}\label{eq:lwr_bnd_cent_Ber}
    \frac{1}{\sqrt{n}}\|\boldsymbol{Ah_{\beta}}\|_2 \geq \frac{\|\boldsymbol{h_{\beta}}\|_2}{4}-c_2\sqrt{\frac{\log p}{n}}\|\boldsymbol{h_{\beta}}\|_1 \ \forall \ \boldsymbol{h_{\beta}} \in \mathbb{R}^p. 
\end{equation} 
\item From Proposition 2.4 of \cite{Rudelson2010}, for a $s \times r'$ dimensional sub-Gaussian matrix $\boldsymbol{A_{\mathcal{R}_i \mathcal{S}_j}}$, there exists a constant $c_1>0$ such that, for any $\tau'>0$, with probability at least $1-2\exp{\{-n\tau'^2\}}$ we have 
\begin{equation}\label{eq:max_sing_val_cent_ber}
   \frac{1}{\sqrt{n}} \|\boldsymbol{A_{\mathcal{R}_i \mathcal{S}_j}}\|_2 = \frac{1}{\sqrt{n}} \sigma_{max}({\boldsymbol{A_{\mathcal{R}_i \mathcal{S}_j}}}) \leq c_1 \left(\sqrt{\frac{s}{n}}+\sqrt{\frac{r'}{n}}+\tau'\right).
\end{equation}
Since a centered bounded matrix is sub-Gaussian, \eqref{eq:max_sing_val_cent_ber} holds with $c_1$ being a function of $h$.
\end{enumerate}
The rest of the proof follows exactly the same as Lemma \ref{le:Ext_RE}. This completes the proof. \hfill{$\blacksquare$}

In the upcoming lemma, we extend the matrix properties from Lemma~\ref{le:Rad_mat_prop} for Rademacher matrices to centered bounded matrices. We show that for centered bounded matrices, the rates of convergence for the infinity norm of the bias terms of both \eqref{eq:beta_d_unscaled} and \eqref{eq:delta_d_unscaled} with $\boldsymbol{W}=\boldsymbol{A}$ converge to $0$ at the same rate as that for Rademacher matrices. This implies that there $\boldsymbol{W} = \boldsymbol{A}$ is a feasible solution for the optimization problem in Alg.~\ref{alg:design_W_bnd}. This ensures that Theorems ~\ref{th:2_3term_beta_decompose_W}, \ref{th:dist_conv_W} and \ref{th:distribution_beta_delta_opt} follow directly for the centered bounded matrix model.

\begin{lemma}\label{le:Bnd_mat_prop}
    Let $\boldsymbol{A}$ be an $n \times p$ matrix with i.i.d. entries obtained from a distribution with mean $0$ and variance $1$, and defined on the domain $[-h,h]$ with $ h>0$. Then the following are true:
    \begin{enumerate}
        \item $\left|\frac{1}{n}\boldsymbol{A^{\top}}\boldsymbol{A} -\boldsymbol{I_p}\right|_{\infty}$ = $O_P\left(\sqrt{\frac{\log(p)}{n}}\right)$.
         \item $\left|\frac{1}{p}\left(\frac{1}{n}\boldsymbol{A}\boldsymbol{A}^{\top}-\boldsymbol{I_n}\right)\boldsymbol{A}\right|_{\infty}=O_P\left(\sqrt{\frac{\log(2pn)}{pn}}+\frac{1}{n}\right)$.
        \item If $n<p$, then $\frac{n}{p\sqrt{1-\frac{n}{p}}}\left|\left(\frac{\boldsymbol{AA^{\top}}}{n}-\frac{p}{n}\boldsymbol{I_n}\right)\right|_{\infty} = O_P\left(\frac{1}{\sqrt{1-n/p}}\sqrt{\frac{\log(n)}{p}}\right)$. 
        \item $P\left(\underset{\forall j \in [p]}{\max} \frac{\boldsymbol{a_{.j}^\top a_{.j}}}{n} \geq 1+h^2\sqrt{\frac{\log p}{n}}\right) \leq \frac{1}{p}.$
    \end{enumerate}
\end{lemma}
\textbf{Proof of Lemma \ref{le:Bnd_mat_prop}, 
 [Result 1]:}
Recall that the elements of $\boldsymbol{A}$ is drawn i.i.d. from distribution with mean $0$ and variance $1$ and defined on a bounded domain $[-h,h]$ with $h>0$. 
Let $\boldsymbol{V} \triangleq \frac{1}{n} \boldsymbol{A^{\top}A}-\boldsymbol{I_p}$. Note that the elements of $\boldsymbol{V}$ satisfy the following:
 \begin{equation} \label{eq:v_result_cent}
  v_{lj}=   \begin{cases}
  \sum_{k=1}^n \frac{(a_{kl})^2}{n}-1& \mbox{if } j=l, \ l\in[p]\\
  \sum_{k=1}^n \frac{a_{kl}a_{kj}}{n} & \mbox{if } j\ne l, \ j,l\in[p]
 \end{cases}
 \end{equation}
We first consider the diagonal elements of $\boldsymbol{V}$. For $j=l,l \in [p]$, each summand of $v_{lj}$ is independent of each other and $E\left[ (a_{kl})^2-1\right]=0$ since $E(a^2_{kl}) = 1$. Furthermore, $v_{lj}$ is uniformly bounded by $-\frac{1}{n} \leq  \frac{1}{n}\left((a_{kl})^2-1\right) \leq \frac{1}{n}h^2-\frac{1}{n}$. Note that, $\frac{1}{n}h^2-\frac{1}{n}-\left(-\frac{1}{n}\right)=\frac{1}{n}h^2$. Therefore, using Hoeffding's inequality for $t>0$,
\begin{eqnarray*}
    P(|v_{lj}|\geq t) = P\left(\left|\sum_{k=1}^n \frac{(a_{kl})^2}{n}-1\right| \geq t\right) \leq 2e^{-\frac{2nt^2}{h^4}}.
\end{eqnarray*}
Therefore, using union bound, we have,
\begin{equation}\label{eq:union_bnd_R1_diag}
    P\left(\underset{l=j,l\in [p]}{\max}|v_{lj}| \geq t\right) \leq 2pe^{-\frac{2nt^2}{h^4}}.
\end{equation}
Taking $t=h^2\sqrt{\frac{2\log p}{n}}$ in \eqref{eq:union_bnd_R1_diag}, we have,
\begin{eqnarray}\label{eq:diag_bnd_R1}
    P\left(\underset{l=j,l\in [p]}{\max}|v_{lj}| \geq h^2\sqrt{\frac{2\log p}{n}}\right) \leq 2pe^{-4\log p} =\frac{2}{p^3}.
\end{eqnarray}

Now, we go on to analyze the off-diagonal terms of $\boldsymbol{V}$. Each summand of $v_{lj}$ is uniformly bounded as $-\frac{h^2}{n}\leq \frac{a_{kl}a_{kj}}{n} \leq \frac{h^2}{n}$. 
Note that  $E\left[\frac{a_{kl}a_{kj}}{n}\right]=0$  $\forall k \in [n], \ l\ne j\in[p]$. Furthermore, for $l\ne j\in[p]$, each of the summands of $v_{lj}$ are independent since the elements of $\boldsymbol{A}$ are independent. Therefore, using Hoeffding's Inequality  for $t>0$, 
\begin{equation*}
    P(|v_{lj}|\geq t) \leq 2e^{-\frac{nt^2}{2h^4}}.
\end{equation*}
 Using union bound we have,
\begin{equation}\label{eq:union_bnd_off_diag_R1}
    P\left(\max_{l \ne j \in [n]}|v_{lj}|\geq t\right)\leq 2p^2 e^{-\frac{nt^2}{2h^4}}.
\end{equation}
Setting $t =  2h^2\sqrt{\frac{2\log p}{n}}$ in \eqref{eq:union_bnd_off_diag_R1}, we have,
\begin{equation}\label{eq:off_diag_bnd_R1}
     P\left(\max_{l \ne j \in [n]}|v_{lj}|\geq 2h^2 \sqrt{\frac{2\log p}{n}}\right) \leq \frac{2}{p^2}.
\end{equation}
Since $h>0$, $h^2 \leq 2h^2$, we have,
\begin{equation}\label{eq:diag_bnd_R1_fin}
    P\left(\underset{l=j,l\in [p]}{\max}|v_{lj}|\leq 2h^2 \sqrt{\frac{2\log p}{n}}\right) \geq  P\left(\underset{l=j,l\in [p]}{\max}|v_{lj}| \leq h^2\sqrt{\frac{2\log p}{n}}\right) \geq 1- \frac{2}{p^2}.
\end{equation}
Therefore using union bound on \eqref{eq:off_diag_bnd_R1} and \eqref{eq:diag_bnd_R1_fin}, we have,
\begin{equation}\label{eq:max_bnd_R1}
    P\left(\underset{j,l\in [p]}{\max}|v_{lj}|\leq 2h^2 \sqrt{\frac{2\log p}{n}}\right) \geq 1- \frac{2}{p^3}-\frac{2}{p^2}.
\end{equation}
This completes the proof.

\noindent \textbf{Result (2):}
In this part, we define an auxiliary matrix $\boldsymbol{V}$ as follows: 
\begin{equation}\label{eq:R2_mat_equiv_cent}
     \boldsymbol{V} \triangleq \frac1p\left(\frac{1}{n}\boldsymbol{A}\boldsymbol{A}^{\top}-\boldsymbol{I_n}\right) \boldsymbol{A}=\frac1p\left(\frac{1}{n}\boldsymbol{A}\boldsymbol{A}^{\top}\boldsymbol{A}-\boldsymbol{A}\right).
\end{equation}
Fix $i \in [n]$ and $j \in [p]$, and consider 
\[ v_{ij}  = -\frac{a_{ij}}{p} + \frac{1}{np}\sum_{l=1}^{p}\sum_{k=1}^{n}a_{il}a_{kl}a_{kj}\]
By splitting the sum over $l$ into the terms where $l\neq j$ and the term where $l=j$, we obtain
\begin{align*}
    v_{ij}
    & = -\frac{a_{ij}}{p} + \frac{1}{np}\left(\sum_{\substack{l=1\\l\neq j}}^{p}\sum_{k=1}^{n}a_{il}a_{kl}a_{kj}\right) + \frac{a_{ij}}{np}\sum_{k=1}^{n}\left(a_{kj}^2-1\right)+\frac{a_{ij}}{p}\frac{1}{n}\left(\sum_{k=1}^{n}1\right)\\
    & = \frac{1}{np}\left(\sum_{\substack{l=1\\l\neq j}}^{p}\sum_{k=1}^{n}a_{il}a_{kl}a_{kj}\right) + \frac{a_{ij}}{np}\sum_{k=1}^{n}\left(a_{kj}^2-1\right).
    \end{align*}
    Next we split the first term of $v_{lj}$ over $k$ into the terms where $k\neq i$ and the term where $k=i$ to obtain
    \begin{eqnarray*}
     \frac{1}{np}\left(\sum_{\substack{l=1\\l\neq j}}^{p}\sum_{k=1}^{n}a_{il}a_{kl}a_{kj}\right)
    &=& \frac{1}{np}\left(\sum_{\substack{l=1\\l\neq j}}^{p}\sum_{\substack{k=1\\k\neq i}}^{n}a_{il}a_{kl}a_{kj}\right) + \frac{1}{np}\sum_{\substack{l=1\\l\neq j}}^{p}a_{il}^2a_{ij}
    \\ &=& \frac{1}{np}\left(\sum_{\substack{l=1\\l\neq j}}^{p}\sum_{\substack{k=1\\k\neq i}}^{n}a_{il}a_{kl}a_{kj}\right) + \frac{a_{ij}}{np}\left(\sum_{\substack{l=1\\l\neq j}}^{p}a_{il}^2-1\right) + \frac{p-1}{np}a_{ij}.
    \end{eqnarray*} 
    Therefore, we have for all $i\in [n], j \in [p]$,
    \begin{equation}\label{eq:v_ij_struct}
        v_{ij}= \frac{1}{np}\left(\sum_{\substack{l=1\\l\neq j}}^{p}\sum_{\substack{k=1\\k\neq i}}^{n}a_{il}a_{kl}a_{kj}\right)+ \frac{a_{ij}}{np}\left(\sum_{k=1}^{n}a_{kj}^2-1\right) + \frac{a_{ij}}{np}\left(\sum_{\substack{l=1\\l\neq j}}^{p}a_{il}^2-1\right) +  \frac{p-1}{np}a_{ij}.
    \end{equation}
    We will now obtain the tail bounds for each of the four structures on RHS of \eqref{eq:v_ij_struct}.
If we condition on $\boldsymbol{a}_{i.}$ and $\boldsymbol{a}_{.j}$,  the $(n-1)(p-1)$ random variables $\frac{1}{np}a_{il}a_{kl}a_{kj}$ for $k\in [n]\setminus\{i\}$ and $l\in [p]\setminus \{j\}$ are independent, have mean zero, and are bounded between $-\frac{h^3}{np}$ and $\frac{h^3}{np}$. 

Therefore, by Hoeffding's inequality, for $t>0$, we have
\begin{eqnarray}\label{eq:uncond_prob_r2_theta}
        P\left(\left.\left|\frac{1}{np}\sum_{\substack{l=1\\l\neq j}}^{p}\sum_{\substack{k=1\\k\neq i}}^{n}a_{il}a_{kl}a_{kj}\right|\geq t\;\right|\;\boldsymbol{a}_{i.},\boldsymbol{a}_{.j}\right) \leq  2\exp \left(-\frac{2t^2}{(n-1)(p-1)4h^6\frac{1}{n^2p^2}}\right)\nonumber \\  \leq 2e^{-\frac{npt^2}{2h^6}}.
    \end{eqnarray}
    Since the RHS of \eqref{eq:uncond_prob_r2_theta} is independent of $\boldsymbol{a}_{i,.}$ and $\boldsymbol{a}_{.j}$, the bound also holds for the unconditional probability, i.e., we have 
    \begin{equation}\label{eq:cond_prob_r2_theta}
    P\left(\left|\frac{1}{np}\sum_{\substack{l=1\\l\neq j}}^{p}\sum_{\substack{k=1\\k\neq i}}^{n}a_{il}a_{kl}a_{kj}\right|\geq t\right) \leq 2e^{-\frac{npt^2}{2h^6}}.
    \end{equation}
Taking $t=2h^3\sqrt{\frac{\log 2np}{np}}$ in \eqref{eq:cond_prob_r2_theta} and taking union bound over all $i\in [n], j\in [p]$, we have,
\begin{equation}\label{eq:union_cond_prob_r2_theta}
    P\left(\underset{i \in [n], j\in [p]}{\max}\left|\frac{1}{np}\sum_{\substack{l=1\\l\neq j}}^{p}\sum_{\substack{k=1\\k\neq i}}^{n}a_{il}a_{kl}a_{kj}\right|\geq 2h^3\sqrt{\frac{\log 2np}{np}}\right) \leq \frac{1}{2n^2p^2}.
    \end{equation}
Now, we analyse the tail bounds for the second and third terms of the RHS of \eqref{eq:v_ij_struct}.
Since $|a_{ij}| \leq h$, we have $ \left|\frac{a_{ij}}{np}\left(\sum_{k=1}^{n}a_{kj}^2-1\right)\right| \leq \frac{h}{p} \left|\sum_{k=1}^n \frac{(a_{kj})^2}{n}-1\right|$. From \eqref{eq:max_bnd_R1}, we have,
$$P\left(\underset{j\in [p]}{\max}\left|\sum_{k=1}^n \frac{(a_{kj})^2}{n}-1\right|\leq 2h^2 \sqrt{\frac{2\log p}{n}}\right) \geq 1- \frac{2}{p^3}-\frac{2}{p^2}.$$
Therefore, we get,
\begin{equation}\label{eq:bnd_2nd_term_R2}
    P\left(\underset{i\in [n],j\in [p]}{\max}\left|\frac{a_{ij}}{p}\sum_{k=1}^n \frac{(a_{kl})^2}{n}-1\right|\leq 2h^3\frac{1}{p} \sqrt{\frac{2\log p}{n}}\right) \geq 1- \frac{2}{p^3}-\frac{2}{p^2}.
\end{equation}
Similarly for the third term in the RHS of \eqref{eq:v_ij_struct}, we reverse the roles of $n$ and $p$ in \eqref{eq:max_bnd_R1} to obtain the following:
\begin{equation}\label{eq:bnd_3nd_term_R2}
    P\left(\underset{i\in [n],j\in [p]}{\max}\left|\frac{a_{ij}}{np}\left(\sum_{\substack{l=1\\l\neq j}}^{p}a_{il}^2-1\right)\right|\leq 2h^3\frac{1}{n} \sqrt{\frac{2\log n}{p}}\right) \geq 1- \frac{2}{n^3}-\frac{2}{n^2}.
\end{equation}
Lastly, we have $\frac{p-1}{np}|a_{ij}| \leq \frac{h}{n}$ for all $i\in [n], j\in [p]$. Combining this with \eqref{eq:union_cond_prob_r2_theta}, \eqref{eq:bnd_2nd_term_R2} and \eqref{eq:bnd_3nd_term_R2}, and using the union bound and triangle inequality, we obtain:
\begin{eqnarray}\label{eq:max_bnd_vij_theta}
   \nonumber P\Bigg(\underset{i\in [n],j\in [p]}{\max} |v_{ij}| \leq 2h^3\sqrt{\frac{\log 2np}{np}}+2h^3\left(\frac{1}{p}\sqrt{\frac{2\log p}{n}}+\frac{1}{p}\sqrt{\frac{2\log n}{p}}\right) +\frac{h}{n}\Bigg) \geq 1-\frac{1}{2n^2p^2}-\frac{2}{n^3}-\frac{2}{n^2}-\frac{2}{p^3}-\frac{2}{p^2}.
\end{eqnarray}
Since $h^3\sqrt{\frac{\log 2np}{np}} \geq 2h^3\frac{1}{p} \sqrt{\frac{2\log p}{n}}$ and $h^3\sqrt{2\frac{\log 2n^2p^2}{np}} \geq 2h^3\frac{1}{n} \sqrt{\frac{2\log n}{p}}$ for $n,p>4$, \eqref{eq:max_bnd_vij_theta} becomes
\begin{equation}\label{eq:max_bnd_vij_theta_fin}
    P\left(\underset{i\in [n],j\in [p]}{\max} |v_{ij}| \leq 4h^3\sqrt{\frac{\log 2np}{np}}+\frac{h}{n}\right)\geq 1-\frac{1}{2n^2p^2}-\frac{2}{n^3}-\frac{2}{n^2}-\frac{2}{p^3}-\frac{2}{p^2}.
\end{equation}
This completes the proof.

\noindent\textbf{Result (3):}
Reversing the roles of $n$ and $p$ in result (1), \eqref{eq:max_bnd_R1} of this lemma, we have 
\begin{equation}\label{eq:bound_cons_3_cond}
    P\left(\left|\frac{\boldsymbol{AA^{\top}}}{p}-\boldsymbol{I_n}\right|_{\infty}
 \leq 2h^2\sqrt{\frac{2\log(n)}{p}}\right) \geq 1-\frac{2}{n^2}-\frac{2}{n^3}.
\end{equation}
 %Now, multiplying by $\frac{1}{\sqrt{1-n/p}}$, we get
%\begin{equation}
 %   P\left(\frac{n}{p\sqrt{1-\frac{n}{p}}}\left|\left(\frac{\boldsymbol{AA^{\top}}}{n}-\frac{p}{n}\boldsymbol{I_n}\right)\right|_{\infty}\leq 2h^2\frac{1}{\sqrt{1-n/p}}\sqrt{\frac{2\log(n)}{p}}\right) \geq  1-\frac{2}{n^3}-\frac{1}{n^2}.
  %  \end{equation}
This completes the proof of Result (3).

\noindent\textbf{Result (4):}
From \eqref{eq:union_bnd_R1_diag} using Hoeffdings' inequality, we have for $t>0$, 
\begin{equation*}
    P\left(\sum_{i=1}^n \frac{a_{ij}^2}{n}-1 \geq t\right) \leq e^{-\frac{2nt^2}{h^4}}
\end{equation*}
Taking $t=h^2\sqrt{\frac{\log p}{n}}$, we have,
\begin{equation}\label{eq:aij_bound_one_side}
    P\left(\sum_{i=1}^n \frac{a_{ij}^2}{n} \geq 1+h^2\sqrt{\frac{\log p}{n}}\right) \leq \frac{1}{p^2}.
\end{equation}
Using union bound of probability on \eqref{eq:aij_bound_one_side} over all $j \in [p]$, we have,
\begin{equation}\label{eq:aij_bound_one_side_union}
    P\left(\underset{\forall j \in [p]}{\max} \frac{\boldsymbol{a_{.j}^\top a_{.j}}}{n} \geq 1+h^2\sqrt{\frac{\log p}{n}}\right) \leq p\frac{1}{p^2} =\frac{1}{p}.
\end{equation}
This completes the proof.
\hfill{$\blacksquare$}

Using Lemma \ref{le:Bnd_mat_prop}, we now show in the following lemma, that the properties of the matrix $\boldsymbol{W}$ obtained from Alg.~\ref{alg:design_W_bnd} are very similar to those for the matrix $\boldsymbol{W}$ obtained from Alg.~\ref{alg:design_W} (compare with Lemma \ref{le:W_mat_prop}). %We restate the lemma for $\boldsymbol{W}$ obtained from Alg.\ref{alg:design_W_bnd} and state the respective changes in the proof for the sake of completeness.

\begin{lemma}\label{le:W_mat_prop_bnd}
    Let $\boldsymbol{A}$ be an $n \times p$ matrix with i.i.d. entries obtained from a distribution with mean $0$ and variance $1$ and defined on a bounded domain  $[-h,h], h>0$ and $\boldsymbol{W}$ be the corresponding output of Alg.~\ref{alg:design_W_bnd}. If $n$ is $o(p)$, we have the following results:
    \begin{enumerate}[(1)]
        \item $P\left(\boldsymbol{w_{.j}^{\top}w_{.j}}/n \leq 1+h^2\sqrt{\frac{\log p}{n}} \ \forall j \in [p]\right) = 1-1/p$.
        \item $\left|\frac{1}{n}\boldsymbol{W^{\top}}\boldsymbol{A} -\boldsymbol{I_p}\right|_{\infty}$ =  $O_P\left(\sqrt{\frac{\log(p)}{n}}\right)$.
        \item {$\left|\frac{1}{\sqrt{n}}\boldsymbol{W^{\top}}\right|_{\infty}=O\left(1\right)$.}
        \item {$\left|\frac{1}{p}\left(\frac{1}{n}\boldsymbol{A}\boldsymbol{W}^{\top}-\boldsymbol{I_n}\right)\boldsymbol{A}\right|_{\infty}$= $O_P\left(\sqrt{\frac{\log(pn)}{pn}}+\frac{1}{n}\right)$.}
        \item {$\left|\frac{1}{p}\boldsymbol{AW}^{\top}\right|_{\infty} = O_P\left(\sqrt{\frac{n\log(np)}{p}}+1\right)$.}
        \item $\frac{n}{p\sqrt{1-\frac{n}{p}}}\left|\left(\frac{\boldsymbol{AW}^{\top}}{n}-\frac{p}{n}\boldsymbol{I_n}\right)\right|_{\infty} = O_P\left(\frac{1}{\sqrt{1-n/p}}\sqrt{\frac{\log(n)}{p}}\right)$.
    \end{enumerate}
\end{lemma}
\noindent\textbf{Proof of Lemma \ref{le:W_mat_prop_bnd}:}
The proof is very similar to that of Lemma \ref{le:W_mat_prop}. The only change is the definition of $G_{4}(n,p)$ which now becomes $G_{4}(n,p):=\{\boldsymbol{A} \in \mathbb{R}^{n \times p}: \boldsymbol{a_{.j}^{\top}a_{.j}}/n \leq 1+h^2\sqrt{\frac{\log p}{n}} \ \forall \ j \in [p]\}$ and the constraints of the problem involve parameters $\mu_1(h),\mu_2(h)$ and $\mu_3(h)$ defined as in Alg.~\ref{alg:design_W_bnd}. From Result (4) of Lemma \ref{le:Bnd_mat_prop}, we have  $\underset{\forall j \in [p]}{\max}\boldsymbol{a_{.j}^{\top}a_{.j}}/n \leq 1+h^2\sqrt{\frac{\log p}{n}}$ with probability at least $1-\frac{1}{p}$. Hence, the result \eqref{eq:joint_feasibility} holds.

\textbf{Result [1]:}
The set of $n \times p$ matrices ($\boldsymbol{W}$)  satisfying constraints $\mathsf{C0}$--$\mathsf{C3}$ is denoted by $ E(n,p)$.
We have 
\begin{eqnarray}
   \nonumber P\left(\underset{j \in [p]}{\max} \ \boldsymbol{w_{.j}^{\top}w_{.j}}/n \leq 1+h^2\sqrt{\frac{\log p}{n}}\right) &=& P\left(\underset{j \in [p]}{\max} \ \boldsymbol{w_{.j}^{\top}w_{.j}}/n \leq 1+h^2\sqrt{\frac{\log p}{n}}  \bigg | \boldsymbol{A}\in E(n,p)\right) P\left(\boldsymbol{A}\in E(n,p)\right) \\ &+& P\left(\underset{j \in [p]}{\max} \ \boldsymbol{w_{.j}^{\top}w_{.j}}/n \leq 1+h^2\sqrt{\frac{\log p}{n}} \bigg | \boldsymbol{A}\in E(n,p)^c\right) P\left(\boldsymbol{A}\in E(n,p)^c\right)  \label{eq:wj_int_break_bnd}
\end{eqnarray}
If the optimization problem in Alg.~\ref{alg:design_W_bnd} is feasible, then $\underset{\forall j \in [p]}{\max}\boldsymbol{w}_{.j}^\top \boldsymbol{w}_{.j}/n\leq 1+h^2\sqrt{\frac{\log p}{n}}$. Therefore, we have
\begin{equation}\label{eq:cond_A_feas_bnd}
    P\left(\underset{j \in [p]}{\max} \ \boldsymbol{w_{.j}^{\top}w_{.j}}/n \leq 1+h^2\sqrt{\frac{\log p}{n}} \bigg | \boldsymbol{A}\in E(n,p)\right)\geq 1-\frac{1}{p}.
\end{equation}
 As per Alg.~\ref{alg:design_W_bnd}, if the constraints of the optimization problem are not satisfied, then we choose $\boldsymbol{W}:=\boldsymbol{A}$ as the output.
This event is given by $\boldsymbol{A}\in E(n,p)^c$. We have from Result (4) of Lemma \ref{le:Bnd_mat_prop}, $\underset{\forall j \in [p]}{\max}\boldsymbol{a_{.j}^{\top}a_{.j}}/n \leq 1+h^2\sqrt{\frac{\log p}{n}}$ with probability atleast $1-\frac{1}{p}$. Therefore, we have
\begin{equation}\label{eq:cond_A_not_feas_bnd}
    P\left(\underset{j \in [p]}{\max} \ \boldsymbol{w_{.j}^{\top}w_{.j}}/n \leq 1+h^2\sqrt{\frac{\log p}{n}}  \bigg | \boldsymbol{A}\in E(n,p)^c\right)\geq 1-\frac{1}{p}.
\end{equation}
Therefore, we have from \eqref{eq:cond_A_feas_bnd},\eqref{eq:cond_A_not_feas_bnd} and \eqref{eq:wj_int_break_bnd},
\begin{equation}\label{eq:col_Wj_prob_bnd}
    P\left(\underset{\forall j \in [p]}{\max}\frac1n\boldsymbol{w_{.j}^{\top}w_{.j}} \leq 1+h^2\sqrt{\frac{\log p}{n}}\right)=  \left(1-\frac{1}{p}\right)P\left(\boldsymbol{A}\in E(n,p)\right)+ P\left(\boldsymbol{A}\in E(n,p)^c\right)\geq 1-\frac{1}{p}.
\end{equation}

\textbf{Result [3]:}
From \eqref{eq:col_Wj_prob_bnd}, for each $j \in [p]$, we have $\|\boldsymbol{w_{.j}}\|_2 \leq \sqrt{n}\left(1+h^2\sqrt{\frac{\log p}{n}}\right)$ with probability $1-1/p$.
For any vector $\boldsymbol{x}$, $\|\boldsymbol{x}\|_{\infty} \leq \|\boldsymbol{x}\|_2$. Hence, for every $j \in [p]$, we have $\|\boldsymbol{w_{.j}}\|_\infty \leq \sqrt{n}\left(1+h^2\sqrt{\frac{\log p}{n}}\right)$ with probability $1-1/p$.
Since $|\boldsymbol{W}^\top|_{\infty} \leq \underset{j \in [p]}{\max}\|\boldsymbol{w_{.j}}\|_\infty \leq \sqrt{n}\left(1+h^2\sqrt{\frac{\log p}{n}}\right)$ with probability atleast $1-\frac{1}{p}$.
Since $h$ is a constant,
\begin{equation}\label{eq:W^T_bnd_inf_bnd}
    \left|\frac{1}{\sqrt{n}}\boldsymbol{W}^\top\right|_{\infty}=O(1).
\end{equation}

The proofs of Results (2), (4), (5) and (6) follow exactly in the same manner as for Lemma~\ref{le:W_mat_prop}, except for the changes that were stated in the beginning of this proof. This completes the proof. \hfill{$\blacksquare$}

We now state the lemma for the convergence of marginal variances of the debiased \textsc{Lasso} estimators for the centered bounded sensing matrix $\boldsymbol{A}$. 

\begin{lemma}\label{le:dist_conv_W_bnd}
    Let $\boldsymbol{A}$ be an $n \times p$ matrix with i.i.d. entries obtained from a distribution with mean $0$ and variance $1$ defined on bounded domain $[-h,h]$ with $h>0$. Suppose  $\boldsymbol{W}$ is obtained from Alg.~\ref{alg:design_W_bnd} and $\boldsymbol{\Sigma_{\beta}}$ and $\boldsymbol{\Sigma_{\delta}}$ are defined as in \eqref{eq:Sig_beta} and \eqref{eq:Sig_delta}, respectively. If $n \log n$ is $o(p)$ and $n$ is $\omega[((s+r)\log p)^2]$, as $n,p \to \infty$, we have the following:
    \begin{enumerate}[(1)]
        \item For $j \in [p]$,
        \begin{equation}
             \Sigma_{\beta_{jj}} \overset{P}{\to} \sigma^2
        \end{equation}
        \item For $i \in [n]$,
        \begin{equation}
            \frac{n^2}{p^2(1-n/p)}\Sigma_{\delta_{ii}} \overset{P}{\to} \sigma^2. 
        \end{equation}
        \hfill $\blacksquare$
    \end{enumerate}
\end{lemma}
\textbf{Proof of Lemma \ref{le:dist_conv_W_bnd}, Result (1):}
Note that, $\Sigma_{\beta_{jj}} =\frac{\sigma^2}{n}\boldsymbol{w_{.j}^{\top}w_{.j}}$. For all $j \in [p]$, from  result (2) of Lemma \ref{le:W_mat_prop_bnd}, for any feasible $\boldsymbol{W}$ with probability at least $1-\left(\dfrac{2}{p^2}+\dfrac{2}{n^2}+\dfrac{3}{2np}\right)$, we have
\begin{equation*}
    1-\frac{1}{n}\boldsymbol{a_{.j}^\top w_{.j}} \leq \mu_1 \ \implies \ 1-\mu_1 \leq \frac{1}{n}\boldsymbol{a_{.j}^\top w_{.j}}.
\end{equation*}
For any feasible $\boldsymbol{W}$ of Alg.\ref{alg:design_W_bnd}, we have for any $c>0$,
\begin{eqnarray*}
    \frac{1}{n}\boldsymbol{w_{.j}^{\top}w_{.j}} &\geq& \frac{1}{n}\boldsymbol{w_{.j}^{\top}w_{.j}} + c(1-\mu_1) - \frac{c}{n}\boldsymbol{a_{.j}^\top w_{.j}} \geq \underset{\boldsymbol{w_{.j}} \in \mathbb{R}^n}{\min} \left\{\frac{1}{n}\boldsymbol{w_{.j}^{\top}w_{.j}} + c(1-\mu_1) -  \frac{c}{n}\boldsymbol{a_{.j}^\top w_{.j}}\right\} \\ &=& \underset{\boldsymbol{w_{.j}} \in \mathbb{R}^n}{\min} \left\{\frac{1}{n}\left(\boldsymbol{w_{.j}} - c\boldsymbol{a_{.j}}/2\right)^{\top}\left(\boldsymbol{w_{.j}} - c\boldsymbol{a_{.j}}/2\right)\right\} + c(1-\mu_1) -\frac{c^2}{4}\frac{\boldsymbol{a_{.j}^{\top}a_{.j}}}{n}  \geq c(1-\mu_1) -\frac{c^2}{4}\frac{\boldsymbol{a_{.j}}^{\top}\boldsymbol{a_{.j}}}{n} \\ &\geq& c(1-\mu_1)- \dfrac{c^2\left(1+h^2\sqrt{\frac{\log p}{n}}\right)}{4}.
\end{eqnarray*}
We obtain the last inequality by putting $\boldsymbol{w_{.j}} = c\boldsymbol{a_{.j}}/2$ which makes the square term 0. The rightmost equality is because $\boldsymbol{a_{.j}}^{\top}\boldsymbol{a_{.j}} \leq n\left(1+h^2\sqrt{\frac{\log p}{n}}\right)$ from Result (4) of Lemma \ref{le:Bnd_mat_prop}. 
%Now, using Popoviciu's Inequality for variances we have for a random variable $X$ and defined on a bounded domain  $u\leq X \leq v$, $Var(X) \leq (v-u)^2/4$. This implies $1 \leq h^2$. Therefore, the lower bound becomes $ch^2(1-\mu_1)- \dfrac{c^2h^6}{4} \geq ch^2(1-\mu_1)- \dfrac{c^2h^2}{4} = h^2(c(1-\mu_1)-c^2/4) $. 
This lower bound is maximized for $c=\frac{2(1-\mu_1)}{\left(1+h^2\sqrt{\frac{\log p}{n}}\right)}$. Plugging in this value of $c$, we obtain the following with probability at least $1-\left(\frac{2}{p^2}+\frac{2}{n^2}+\frac{1}{2n^2p^2}+\frac{2}{n^3}+\frac{2}{p^3}\right)$: 
\begin{equation*}
    \frac{1}{n}\boldsymbol{w_{.j}^{\top}w_{.j}}  \geq  (1-\mu_1)^2\left(1+h^2\sqrt{\frac{\log p}{n}}\right).
\end{equation*}
Hence, from the above equation
and \eqref{eq:joint_feasibility}, we obtain the lower bound on $\Sigma_{\beta_{jj}}$ for any $j \in [p]$ as follows: 
\begin{equation}\label{eq:lower_bnd_beta_bnd}
    P\left(\Sigma_{\beta_{jj}} \geq \frac{\sigma^2(1-\mu_1)^2}{1+h^2\sqrt{\frac{\log p}{n}}}\right) \geq 1-\left(\frac{2}{p^2}+\frac{2}{n^2}+\frac{1}{2n^2p^2}+\frac{2}{n^3}+\frac{2}{p^3}\right).
\end{equation}
Furthermore from Result (1) of Lemma.~\ref{le:W_mat_prop_bnd}, we have
\begin{equation}\label{eq:bound_cons_4_bnd}
    P\left(\frac1n\boldsymbol{w_{.j}^{\top}w_{.j}} \leq \left(1+h^2\sqrt{\frac{\log p}{n}}\right) \ \forall j \in [p]\right) \geq 1-\frac{1}{p}.
\end{equation}
As $\Sigma_{\beta_{jj}} =\sigma^2\frac{\boldsymbol{w_{.j}^{\top} w_{.j}}}{n}$, we have from \eqref{eq:bound_cons_4_bnd} for any $j \in [p]$:
\begin{equation}\label{eq:beta_var_uppr_bnd_bound}
    P\left(\Sigma_{\beta_{jj}} \leq \sigma^2\left(1+h^2\sqrt{\frac{\log p}{n}}\right)\right) \geq 1-\frac{1}{p}.
\end{equation}
Using \eqref{eq:beta_var_uppr_bnd_bound} with \eqref{eq:lower_bnd_beta_bnd}, we obtain for any $j \in [p]$, 
\begin{equation}
\label{eq:sigma_beta_jj_bound_bnd}
    P\left(\frac{\sigma^2(1-\mu_1)^2}{1+h^2\sqrt{\frac{\log p}{n}}}\leq \Sigma_{\beta_{jj}} \leq \sigma^2\left(1+h^2\sqrt{\frac{\log p}{n}}\right) \right) \geq 1-\left(\frac{2}{p^2}+\frac{2}{n^2}+\frac{1}{2n^2p^2}+\frac{2}{n^3}+\frac{2}{p^3}+\frac{1}{p}\right).
\end{equation}
Now under the assumption $n$ is $\omega[((s+r)\log p)^2]$, $\mu_1 \to 0$ and $\left(1+h^2\sqrt{\frac{\log p}{n}}\right) \to 1$. Hence, we have, $\Sigma_{\beta_{jj}} \overset{P}{\to} \sigma^2$. 
This completes the proof of Result (1).

The proof of Result (2) is exactly the same as in Lemma \ref{th:dist_conv_W}. This completes the proof. \hfill{${\blacksquare}$} }

\section{Some auxiliary useful lemmas}
  \begin{lemma} \label{le:mat_Op_properties}
    Let $\boldsymbol{U}$ and $\boldsymbol{V}$ be two $n \times p$ random matrices. Let $\vartheta \in \mathbb{R}$ and $\boldsymbol{w} \in \mathbb{R}^{n}$. Then,
    \begin{enumerate}
        \item $|\vartheta \boldsymbol{U}|_{\infty} = |\vartheta| |\boldsymbol{U}|_{\infty}$.
        \item $|\boldsymbol{U}+\boldsymbol{V}|_{\infty} \leq |\boldsymbol{U}|_{\infty}+|\boldsymbol{V}|_{\infty}$.
        \item If $|\boldsymbol{U}|_{\infty}=O_P(h_1(n,p))$ and $|\boldsymbol{V}|_{\infty}=O_P(h_2(n,p))$, then $|\boldsymbol{U}+\boldsymbol{V}|_{\infty}\leq O_P(\max\{h_1(n,p),h_2(n,p)\})$.
        \item $\|\boldsymbol{w^{\top}V}\|_{\infty} \leq |\boldsymbol{V}|_{\infty} \|\boldsymbol{w}\|_1$.
        \item If $|\boldsymbol{V}|_{\infty}=O_P(h_1(n,p))$ and $\|\boldsymbol{w}\|_1 = O_P(h_{\boldsymbol{w}}(n,p))$, then $\|\boldsymbol{w^{\top}V}\|_{\infty} \leq O_P(h_1(n,p)h_w(n,p))$. $\blacksquare$
        \end{enumerate}
\end{lemma}
\textbf{Proof:}
\newline
\noindent\textbf{Result (1):} We have, $|\vartheta \boldsymbol{U}|_{\infty}= \underset{i \in [n], j \in [p]}{\max} |\vartheta u_{ij}| = \underset{i \in [n], j \in [p]}{\max} |\vartheta| |u_{ij}| = |\vartheta| |\boldsymbol{U}|_{\infty}$.

\noindent\textbf{Result (2):} We have, $|\boldsymbol{U}+\boldsymbol{V}|_{\infty} = \underset{i \in [n], j \in [p]}{\max} |u_{ij}+v_{ij}| \leq \underset{i \in [n], j \in [p]}{\max} \{|u_{ij}|+|v_{ij}|\} \leq \underset{i \in [n], j \in [p]}{\max}|u_{ij}| +\underset{i \in [n], j \in [p]}{\max}|v_{ij}| = |\boldsymbol{U}|_\infty+|\boldsymbol{V}|_\infty$.

\noindent\textbf{Result (3):} Given $|\boldsymbol{U}|_{\infty}=O_P(h_1(n,p))$ and $|\boldsymbol{V}|_{\infty}=O_P(h_2(n,p))$. From Part (2), we have, \[|\boldsymbol{U}+\boldsymbol{V}|_{\infty} \leq |\boldsymbol{U}|_{\infty}+|\boldsymbol{V}|_{\infty} \leq O_P(h_1(n,p))+O_P(h_1(n,p)) = O_P(h_1(n,p)+h_2(n,p)) \leq O_P(\max\{h_1(n,p),h_2(n,p)\}).\]

\noindent\textbf{Result (4):} For any $j \in [p]$, we have \[|(\boldsymbol{w^{\top}V})_j| = |\sum_{i=1}^n {v}_{ij} w_i| \leq \sum_{i=1}^n |{v}_{ij}| |w_i| \leq \sum_{i=1}^n \max_{j \in [p]} |{v}_{ij}| |w_i| \leq \max_{i \in [n], j \in [p]} |{v}_{ij}| \sum_{i=1}^n |w_i| = |\boldsymbol{V}|_{\infty} \|\boldsymbol{w}\|_1.\]  

\noindent\textbf{Result (5):} We have from Part (4), $\|\boldsymbol{w^{\top}V}\|_{\infty} \leq |\boldsymbol{V}|_{\infty} \|\boldsymbol{w}\|_1$ = $O_P(h_1(n,p)) O_P(h_{\boldsymbol{w}}(n,p))$ = $O_P(h_1(n,p)h_w(n,p))$.  $\blacksquare$ 

\begin{lemma}\label{le:ma_Gaussian}
    Let $X_i$, for $i=1,2,\ldots,k$, be Gaussian random variables with mean $0$ and variance $\sigma^2$. Then, we have
    \begin{equation}
        P\left[\max_{i\in[k]} |X_i|\ge 2\sigma \sqrt{\log(k)}\right] \leq 1/k.
    \end{equation}
\end{lemma}
Note that Lemma \ref{le:ma_Gaussian} does not require independence. For a proof see, e.g., \cite{Boucheron}.

{\begin{lemma}\label{le:isotropic}
    Let $\boldsymbol{a}$ be an $p \times 1$ vector with i.i.d. entries obtained from a distribution with mean $0$ and variance $1$ and defined on a bounded domain  $[-h,h]$. Then for every $\boldsymbol{x} \in \mathbb{R}^p$, the following results hold:
    \begin{enumerate}
        \item $\mathbb{E}|\boldsymbol{a^\top x}|^2=\|\boldsymbol{x}\|_2^2$.
        \item For some constant $c>0$, $\|\boldsymbol{a^\top x}\|_{\psi_2} \leq c \|\boldsymbol{x}\|_2$, where $\|\cdot\|_{\psi_2}$ is the Orlicz sub-Gaussian norm.
    \end{enumerate}
\end{lemma}
\textbf{Proof of Lemma \ref{le:isotropic}, Result[1]:}
Since the $a_i$'s are independent with zero mean, we have,
\begin{equation}\label{eq:exp_iso}
    \mathbb{E} \left[ \left( \sum_{i=1}^p x_i a_i \right)^2 \right] = \sum_{i=1}^p x_i^2 \mathbb{E}[a_i^2] + \sum_{i \neq j} x_i x_j \mathbb{E}[a_i a_j].
\end{equation}
$\mathbb{E}[a_i a_j]=0$ due to independence and zero mean. Hence, the cross product terms on the RHS of \eqref{eq:exp_iso} vanishes. Furthermore, since $\mathbb{E}[a_{ij}]^2=1$, we have, $$\mathbb{E}|\boldsymbol{a^\top x}|^2=\|\boldsymbol{x}\|_2^2.$$

\noindent\textbf{Result [2]:}
 Let $v_i := x_i a_i$. Then $v_i$ is a zero-mean random variable bounded in $[-h|x_i|, h|x_i|]$ since $a_i \in [-h, h]$. By Hoeffding's Lemma, for all $\lambda \in \mathbb{R}$, we have
\[
\mathbb{E}[e^{\lambda v_i}] \leq \exp\left( \frac{\lambda^2 K^2 x_i^2}{2} \right).
\]
Since the $v_i$'s are independent, the moment generating function of $S = \sum_{i=1}^p v_i$ satisfies,
\[
\mathbb{E}[e^{\lambda S}] = \prod_{i=1}^p \mathbb{E}[e^{\lambda v_i}] \leq \exp\left( \frac{\lambda^2 h^2 \|\boldsymbol{x}\|_2^2}{2} \right).
\]
This is the MGF of a sub-Gaussian random variable with parameter $ K \|\boldsymbol{x}\|_2$. From the equivalence of the Result 1 and 4 of Proposition 2.5.2 of \cite{Vershynin2018}, the Orlicz norm satisfies,
\[
\|S\|_{\psi_2} \leq \sqrt{2} h \|\boldsymbol{x}\|_2.
\]
This completes the proof. \hfill{$\blacksquare$}}

\section{Supplemental: Experimental Results with Signals Generated Using Different Priors}
In this section, we perform the same experiments as in Sec. IV-D, IV-E, IV-F of the main paper, but with the non-zero elements of the signal $\boldsymbol{\beta^*}$ obtained from the distribution given in \cite{Buchan2020}. Referring to the empirical distribution of the CT (cycle threshold) values for the RT-PCR test from \cite{Buchan2020}, we converted the empirical distribution (probability mass function) of CT values to the empirical distribution of the viral loads using the relationship $Q=2e^{40-CT}$, where $CT$ and $Q$ denote the cycle threshold and viral load respectively. We sampled the non-zero elements of $\boldsymbol{\beta^*}$ from the distribution of $Q$. We compared the sensitvity, specificity and RRMSE for the estimates of the \textsc{Odrlt}, \textsc{Drlt} and \textsc{Rl} estimators for signals generated in this manner. The experiments in this section essentially show that our algorithms work well in terms of good sensitivity, specificity and RRMSE even for signals $\boldsymbol{\beta^*}$ with non-zero elements drawn from a distribution that is different from uniform. 

\subsection{Sensitivity and Specificity for estimates of $\boldsymbol{\delta^*}$}
We performed experiments to study sensitivity and specificity \textsc{Rl}, \textsc{Drlt} and \textsc{Odrlt} for the estimates of $\boldsymbol{\delta^*}$. 
In experimental setup \textsf{EA}, we varied $f_{adv} \in \{0.01,0.03,\ldots,0.09\}$ with fixed values $n = 400, f_{sp} = 0.01, f_{\sigma} = 0.05$. In \textsf{EB}, we varied $n$ from 200 to 450 in steps of 50 with $f_{adv} = 0.01, f_{sp} = 0.01, f_{\sigma} = 0.05$. In \textsf{EC}, we varied $f_{\sigma} \in \{0.01,0.03,\ldots,0.09\}$ with $n = 400, f_{adv} = 0.01, f_{sp} = 0.1$. In  \textsf{ED}, we varied $f_{sp} \in \{0.01,0.03,\ldots,0.09\}$ with $n = 400, f_{adv} = 0.01, f_{\sigma} = 0.05$. The experiments were run 25 times across different noise instances in $\boldsymbol{\eta}$, for the same signal $\boldsymbol{\beta^*}$.
The sensitivity and specificity was computed the same way as described in Sec. IV-D of the main paper. In Fig.~\ref{fig:sens_spec_signal_delta}, we see that the \textsc{Odrlt} performs the best.

\begin{figure*}
\centering
    \includegraphics[height=1.95in]{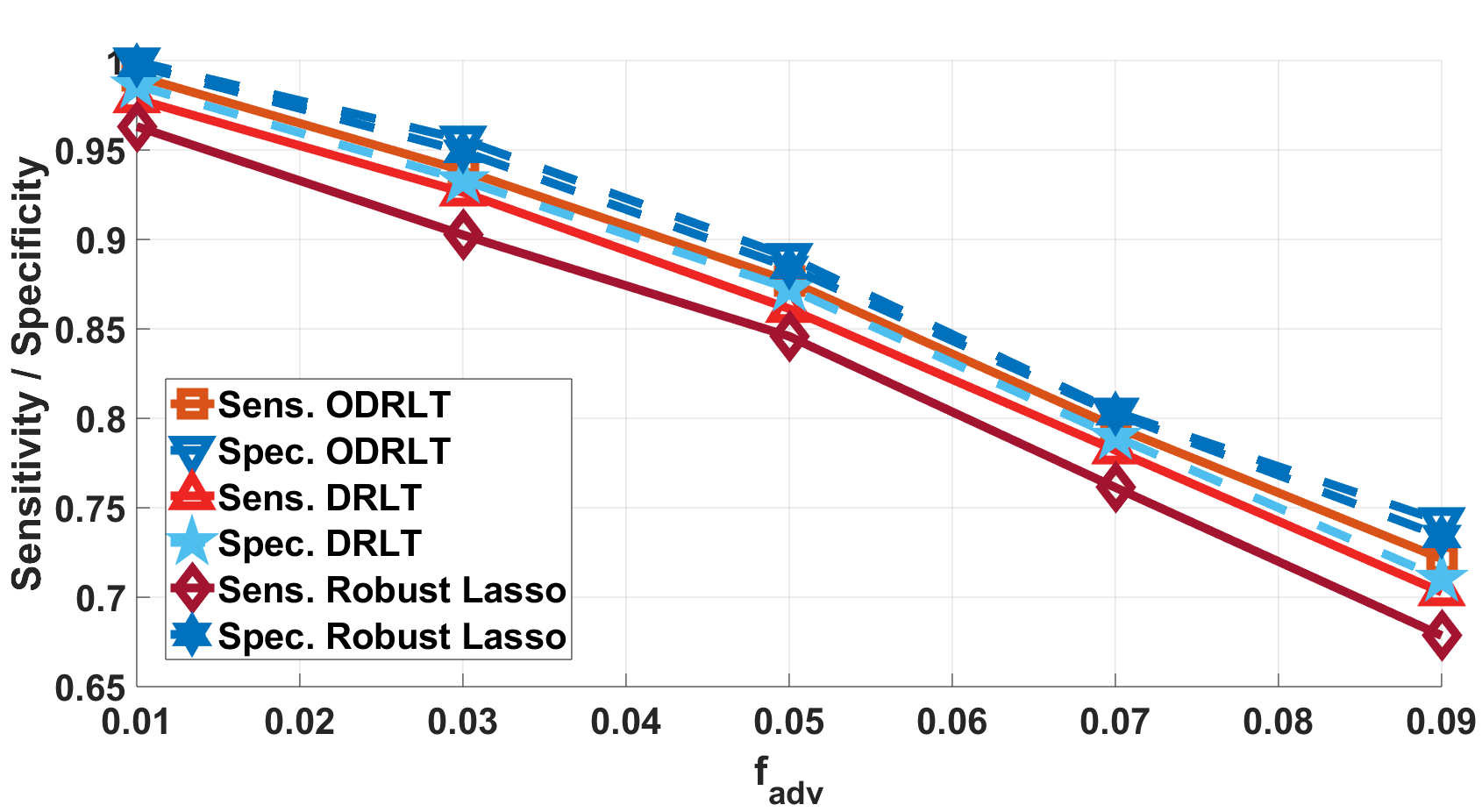}
     \includegraphics[height=1.95in]{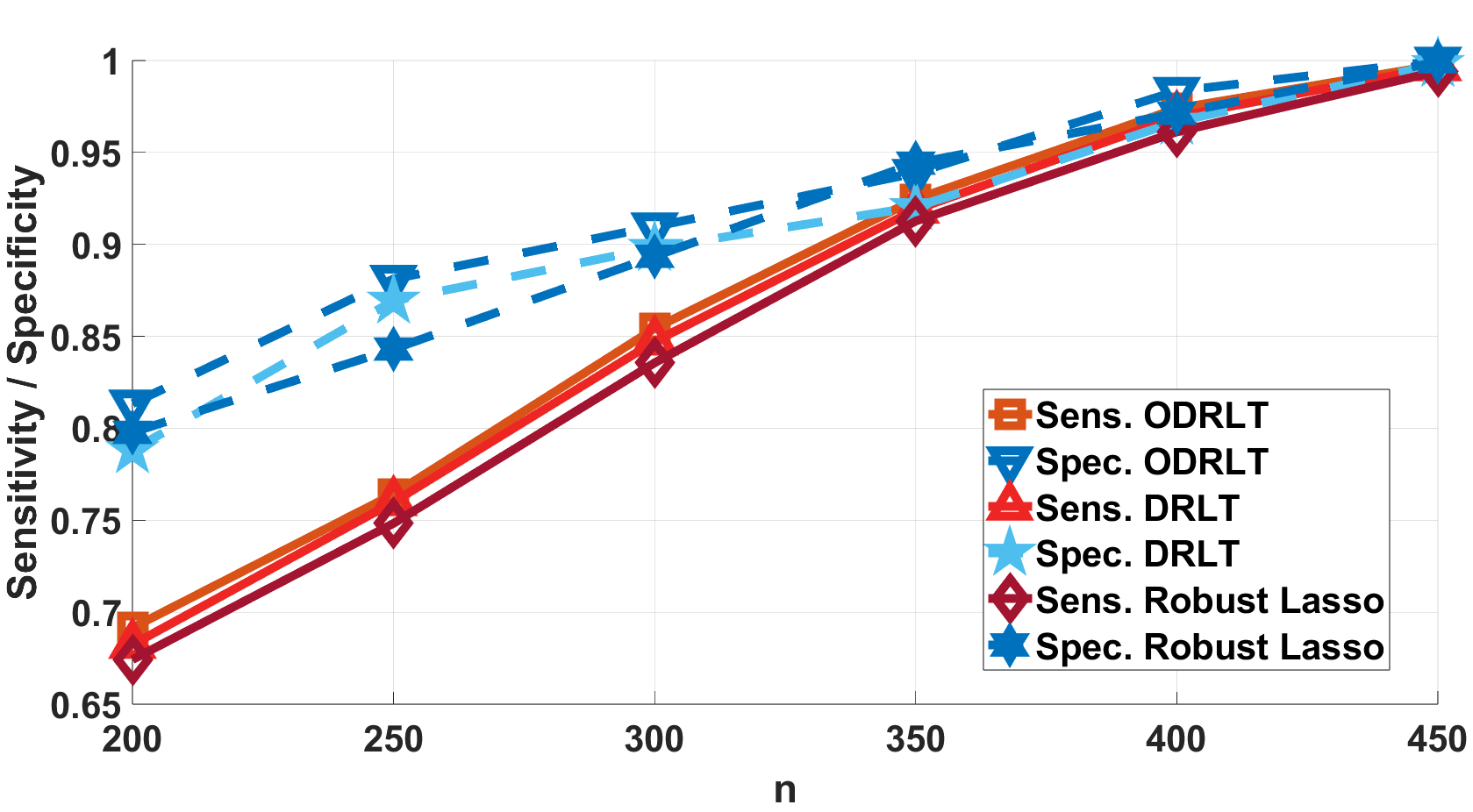}\\
    \includegraphics[height=1.95in]{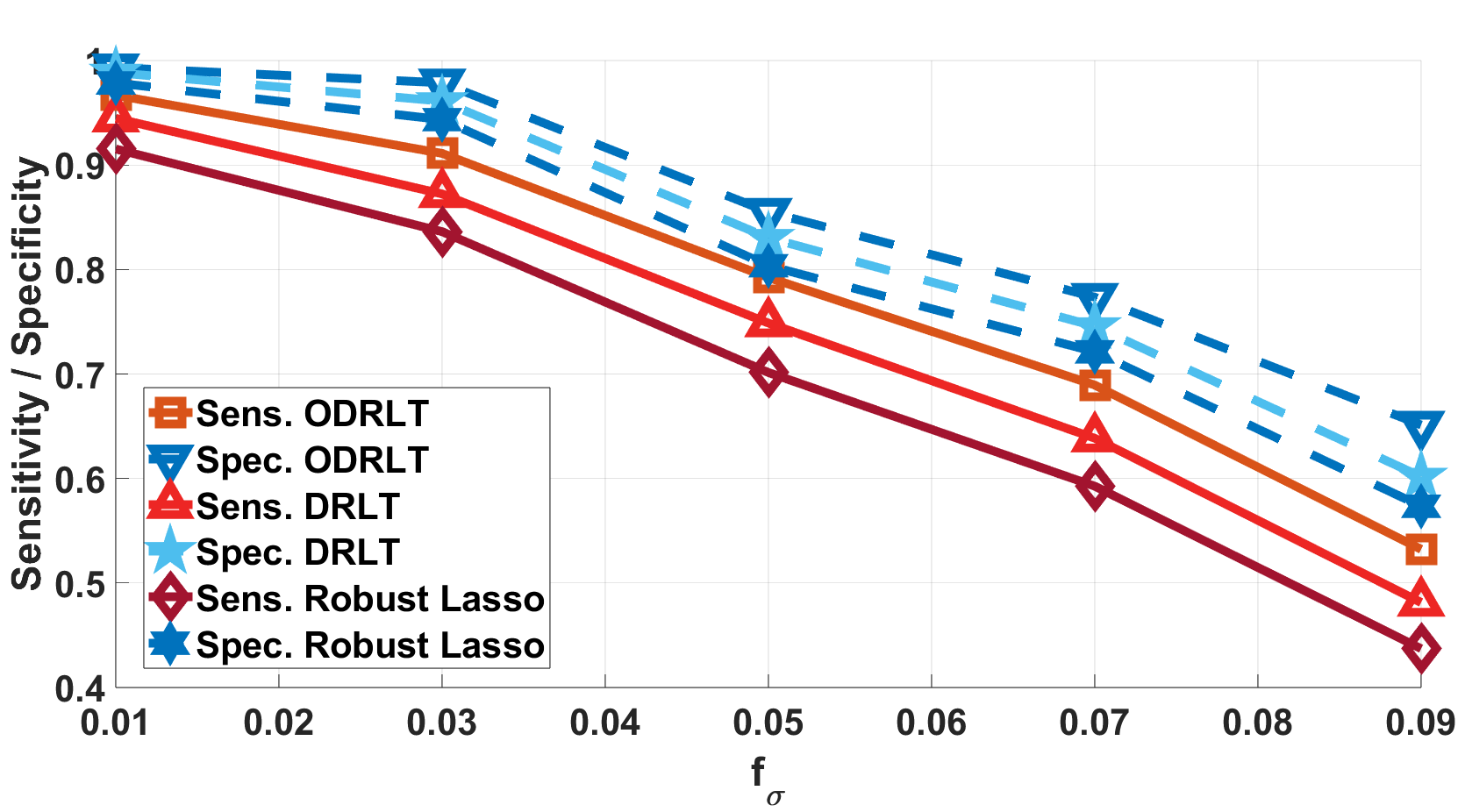}      
    \includegraphics[height=1.95in]{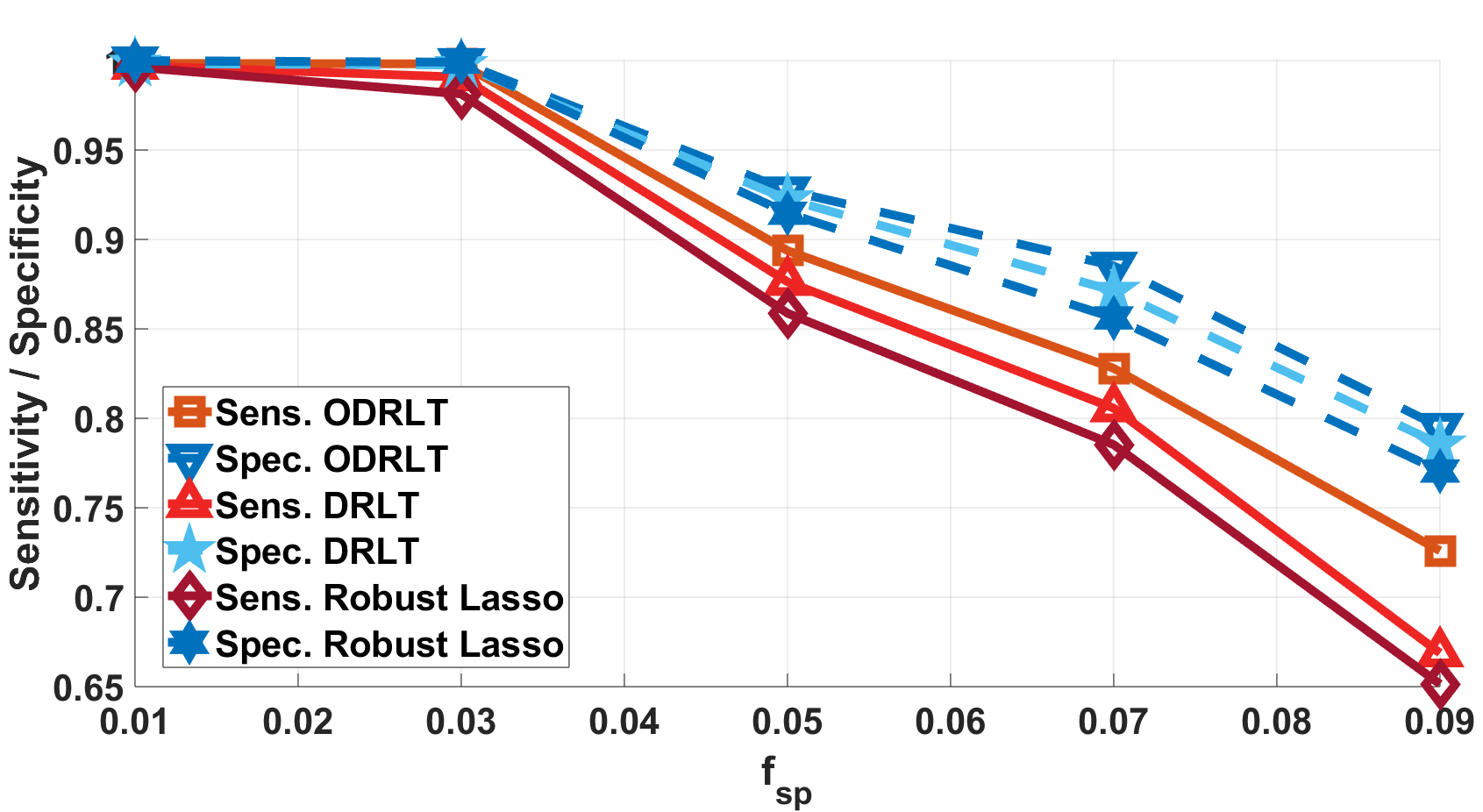}
    \caption{Average Sensitivity and Specificity plots (over 25 independent noise runs) for detecting measurements containing MMEs (i.e. detecting non-zero values of $\boldsymbol{\delta^*}$) using \textsc{Odrlt}. The experimental parameters are $p = 500, f_{\sigma}=0.05, f_{adv}=0.01,f_{sp}=0.1,n = 400$. Left to right, top to bottom: results for experiments \textsf{EA}, \textsf{EB}, \textsf{EC}, \textsf{ED}.}   
\label{fig:sens_spec_signal_delta}    
\end{figure*}

\subsection{Sensitivity and Specificity of $\boldsymbol{\beta^*}$}
In this set of experimental results, we examined the effectiveness of \textsc{Rl}, \textsc{Drlt} and \textsc{Odrlt} to detect defective samples using estimates of $\boldsymbol{\beta^*}
$ in the presence of bit-flips in $\boldsymbol{A}$ induced as per adversarial MMEs. 
We examined the variation in sensitivity and specificity with regard to change in the following parameters, keeping all other parameters fixed.
For the bit-flips experiment i.e., (\textsf{EA}), $f_{adv}$ was varied in $\{0.01,0.03,\ldots,0.09\}$ with $n=400,f_{sp}=0.01,f_{\sigma}=0.05$. For the measurements experiment (\textsf{EB}), $n$ was varied over $\{200,150,\ldots,450\}$ with $f_{sp} = 0.01 , f_{adv} = 0.01, f_{\sigma} = 0.05$. For the noise experiment (i.e., (\textsf{EC}), we varied $f_{\sigma}$ in $\{0.01,0.03,\ldots,0.09\}$ with $n=400,f_{sp}=0.1,f_{adv}=0.01$. For the sparsity experiment (i.e., (\textsf{ED}), $f_{sp}$ was varied in $\{0.01,0.03,\ldots,0.09\}$ with $n = 400, f_{adv} = 0.01, f_{\sigma} = 0.05$. Sensitivity and Specificity is calculated based on the technique described in Sec. IV-E of the main paper. The experiments were run 25 times across different noise instances in $\boldsymbol{\eta}$, for the same signal $\boldsymbol{\beta^*}$. In Fig.~\ref{fig:Sens_spec_beta_signal}, we see the same patter as that for the Sensitivity and Specificity for $\boldsymbol{\delta^*}$. The Hypothesis test using \textsc{Odrlt} for $\boldsymbol{\beta^*}$ performs the best in detecting the infected samples in $\boldsymbol{\beta^*}$ followed by \textsc{Drlt} and then \textsc{Rl}.

\begin{figure*}
   \centering
    \includegraphics[height=1.95in]{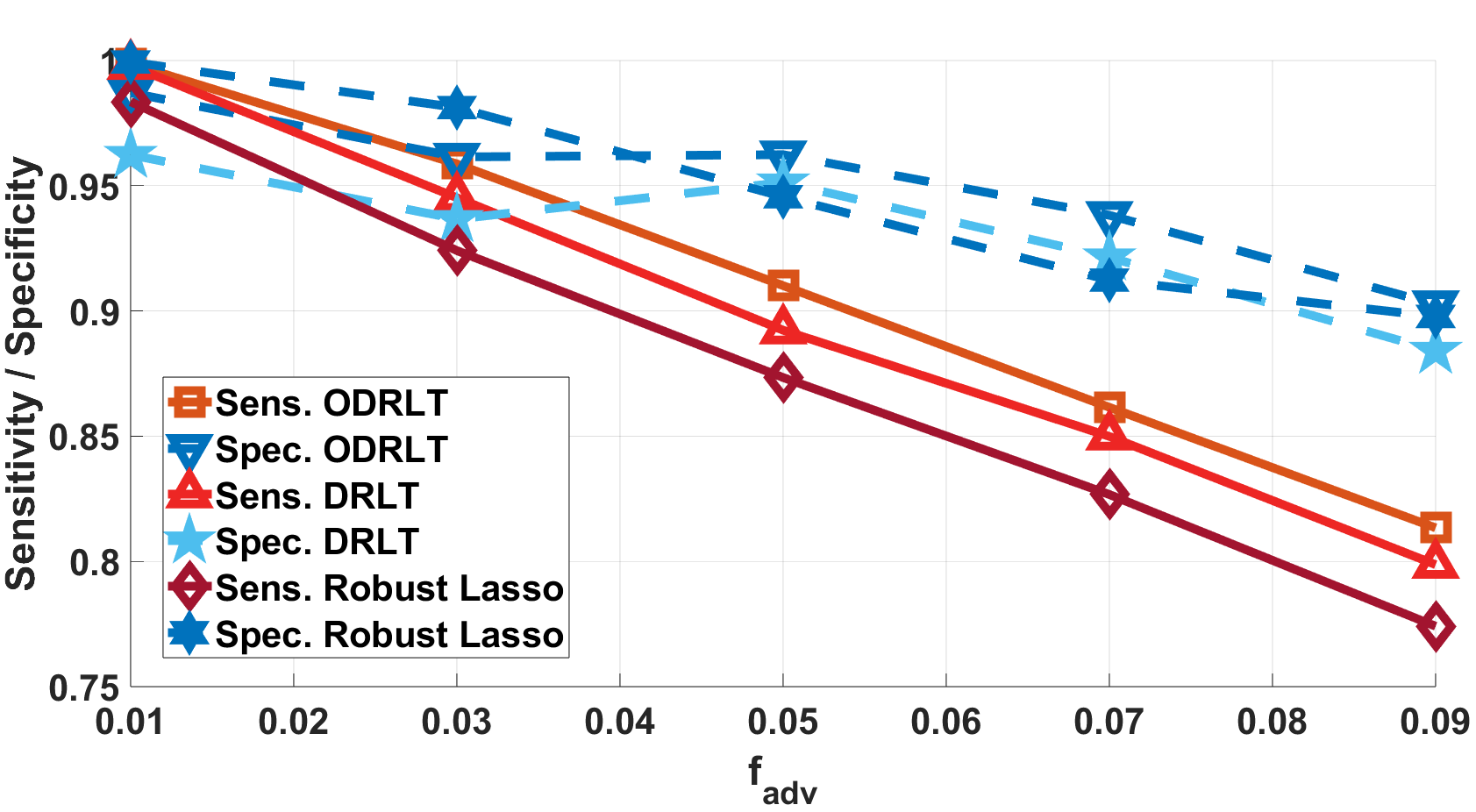}
     \includegraphics[height=1.95in]{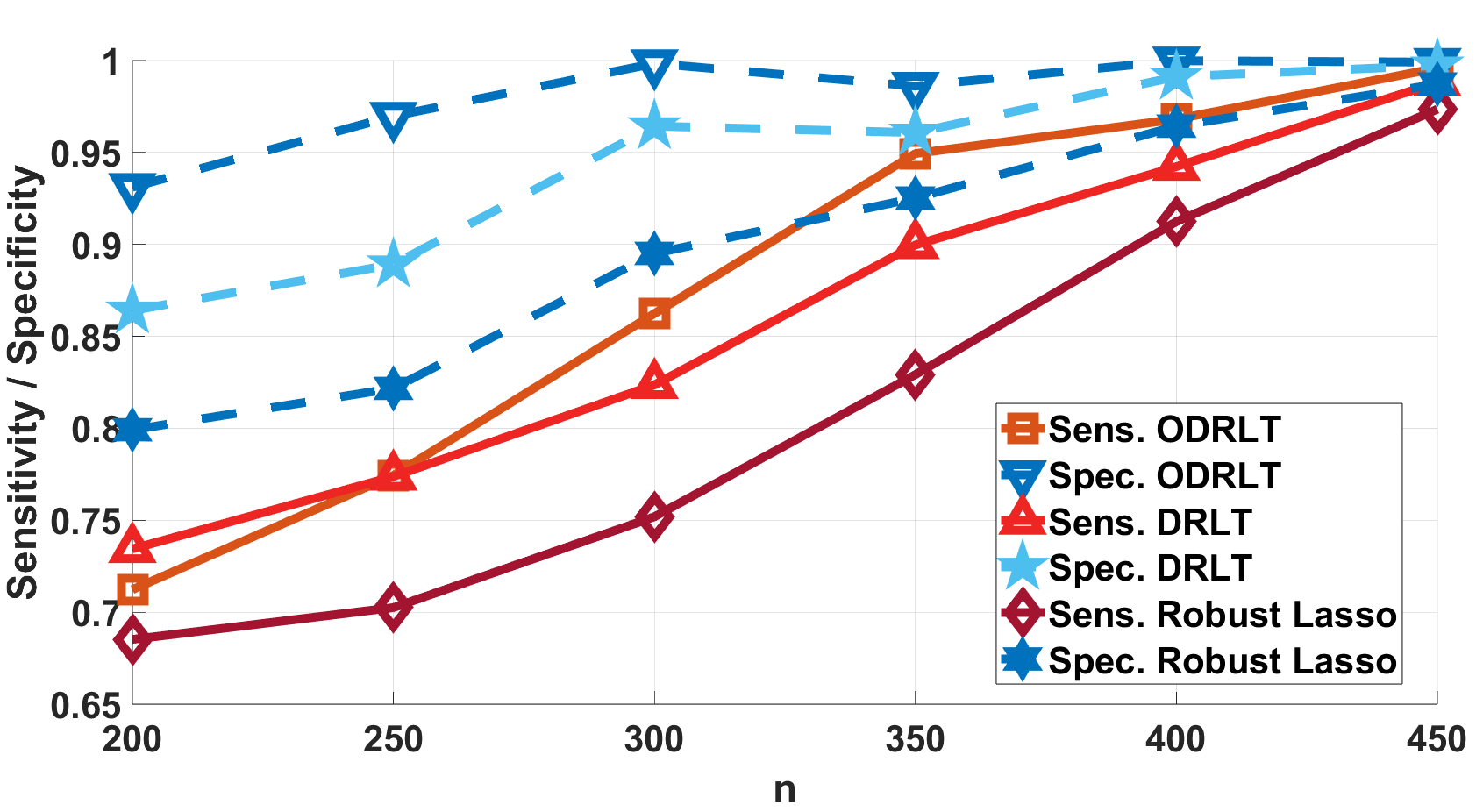}\\
    \includegraphics[height=1.95in]{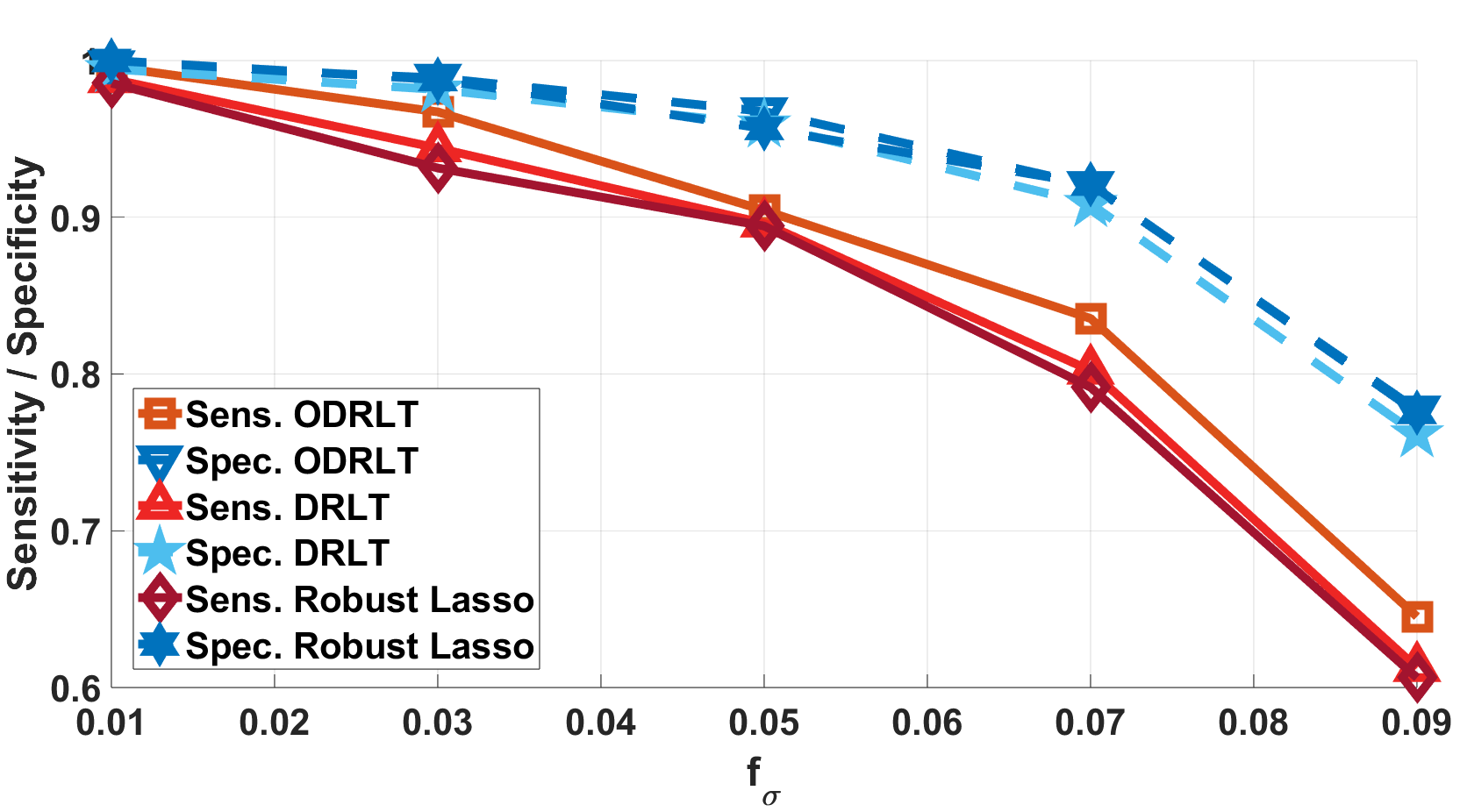}      
    \includegraphics[height=1.95in]{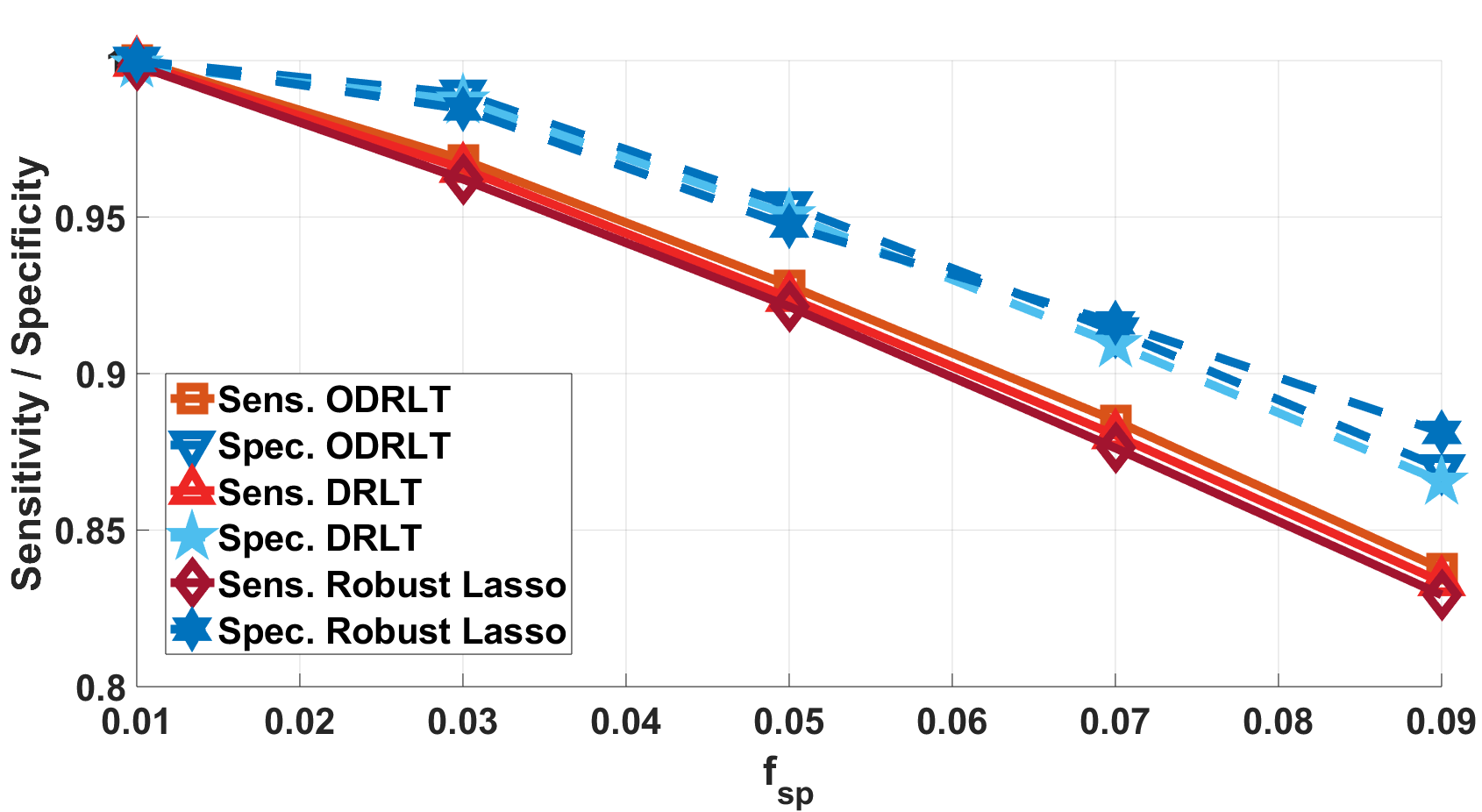}
    \caption{Average Sensitivity and Specificity plots (over 25 independent noise runs) plots for detecting defective samples (i.e., non-zero values of $\boldsymbol{\beta^*}$) using
    \textsc{Odrlt}. Left to right, top to bottom: results for experiments (\textsf{EA}), (\textsf{EB}), (\textsf{EC}), (\textsf{ED}). The experimental parameters are $p = 500, f_{\sigma}=0.05, f_{adv}=0.01,f_{sp}=0.1,n = 400$. } 
    \label{fig:Sens_spec_beta_signal}   
\end{figure*}

\subsection{RRMSE Comparison}
We computed the RRMSE of \textsc{Rl}, \textsc{Drlt} and \textsc{Odrlt} estimates in the same way as described in Sec. IV-F of the main paper. We examined the variation in RRMSE with regard to change in the following parameters, keeping all other parameters fixed.
For the bit-flips experiment i.e., (\textsf{EA}), $f_{adv}$ was varied in $\{0.01,0.03,\ldots,0.09\}$ with $n=400,f_{sp}=0.01,f_{\sigma}=0.05$. For the measurements experiment (\textsf{EB}), $n$ was varied over $\{200,150,\ldots,450\}$ with $f_{sp} = 0.01 , f_{adv} = 0.01, f_{\sigma} = 0.05$. For the noise experiment (i.e., (\textsf{EC}), we varied $f_{\sigma}$ in $\{0.01,0.03,\ldots,0.09\}$ with $n=400,f_{sp}=0.1,f_{adv}=0.01$. For the sparsity experiment (i.e., (\textsf{ED}), $f_{sp}$ was varied in $\{0.01,0.03,\ldots,0.09\}$ with $n = 400, f_{adv} = 0.01, f_{\sigma} = 0.05$. The experiments were run 25 times across different noise instances in $\boldsymbol{\eta}$, for the same signal $\boldsymbol{\beta^*}$. We see that in Fig.~\ref{fig:beta_RMSE}, the RRMSE for \textsc{Odrlt} is the best followed by \textsc{Drlt} and then \textsc{Rl}.
\begin{figure*}
\centering
    \includegraphics[scale=0.20]{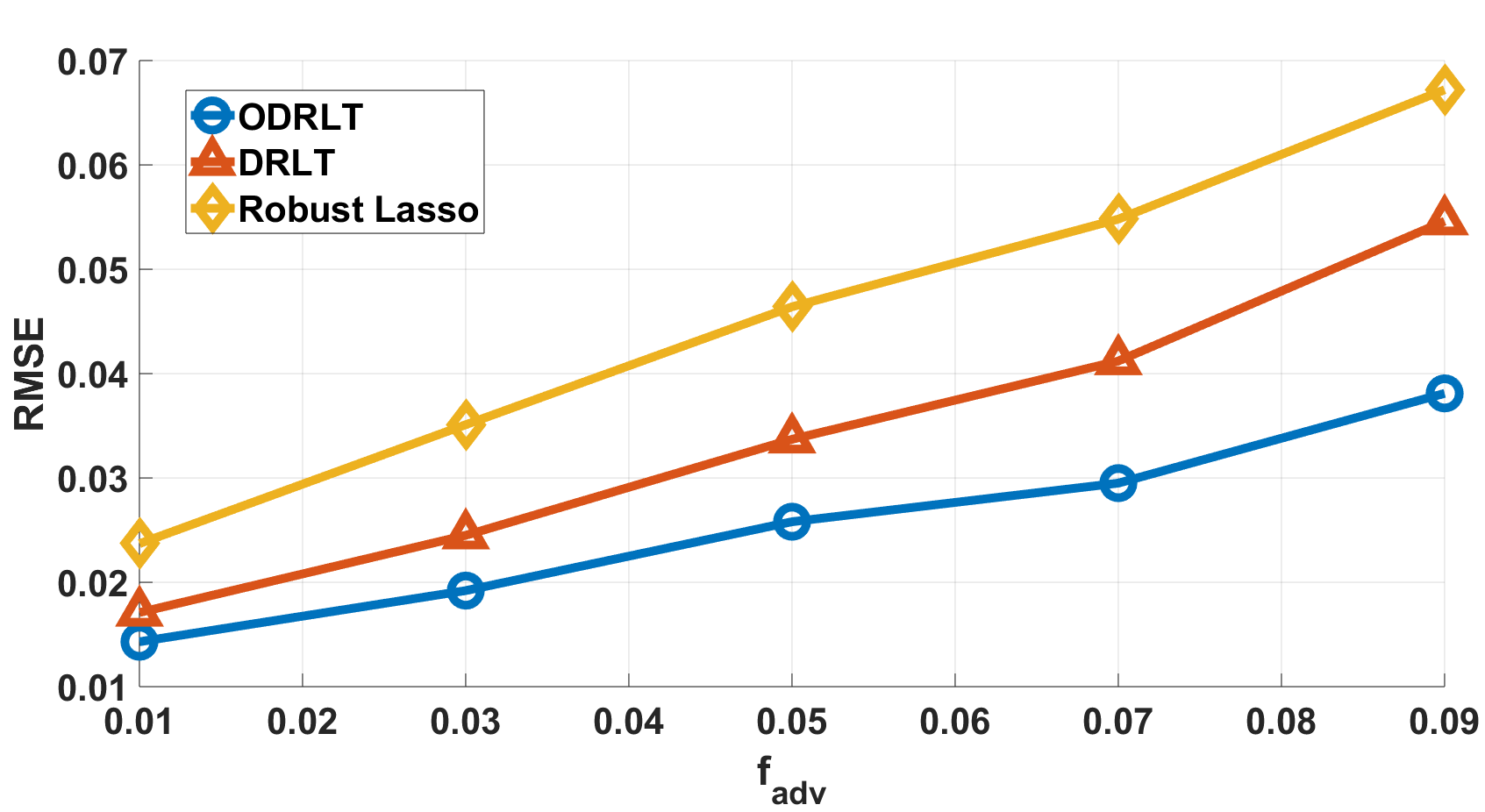}
    \includegraphics[scale=0.2]{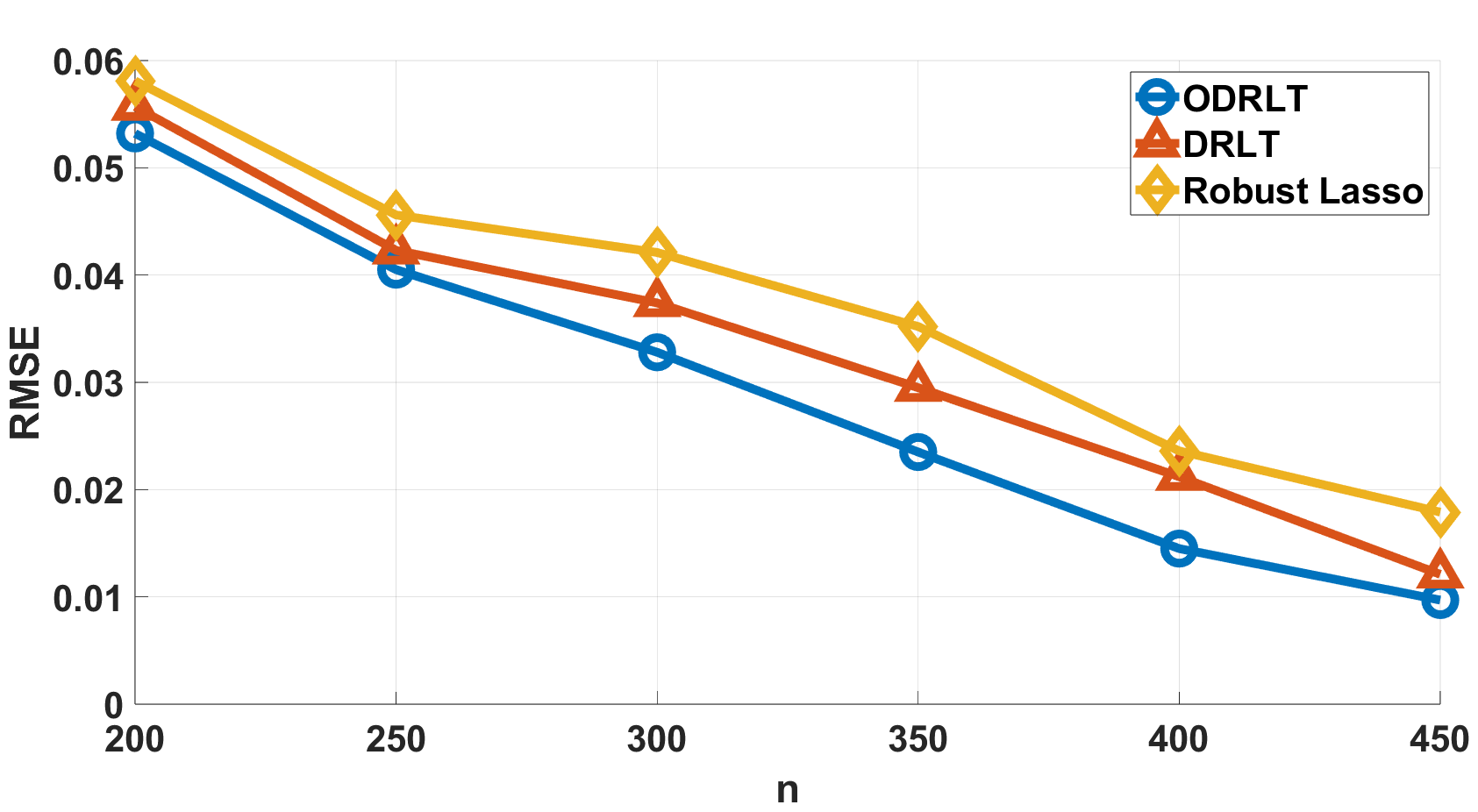}\\
    \includegraphics[scale=0.20]{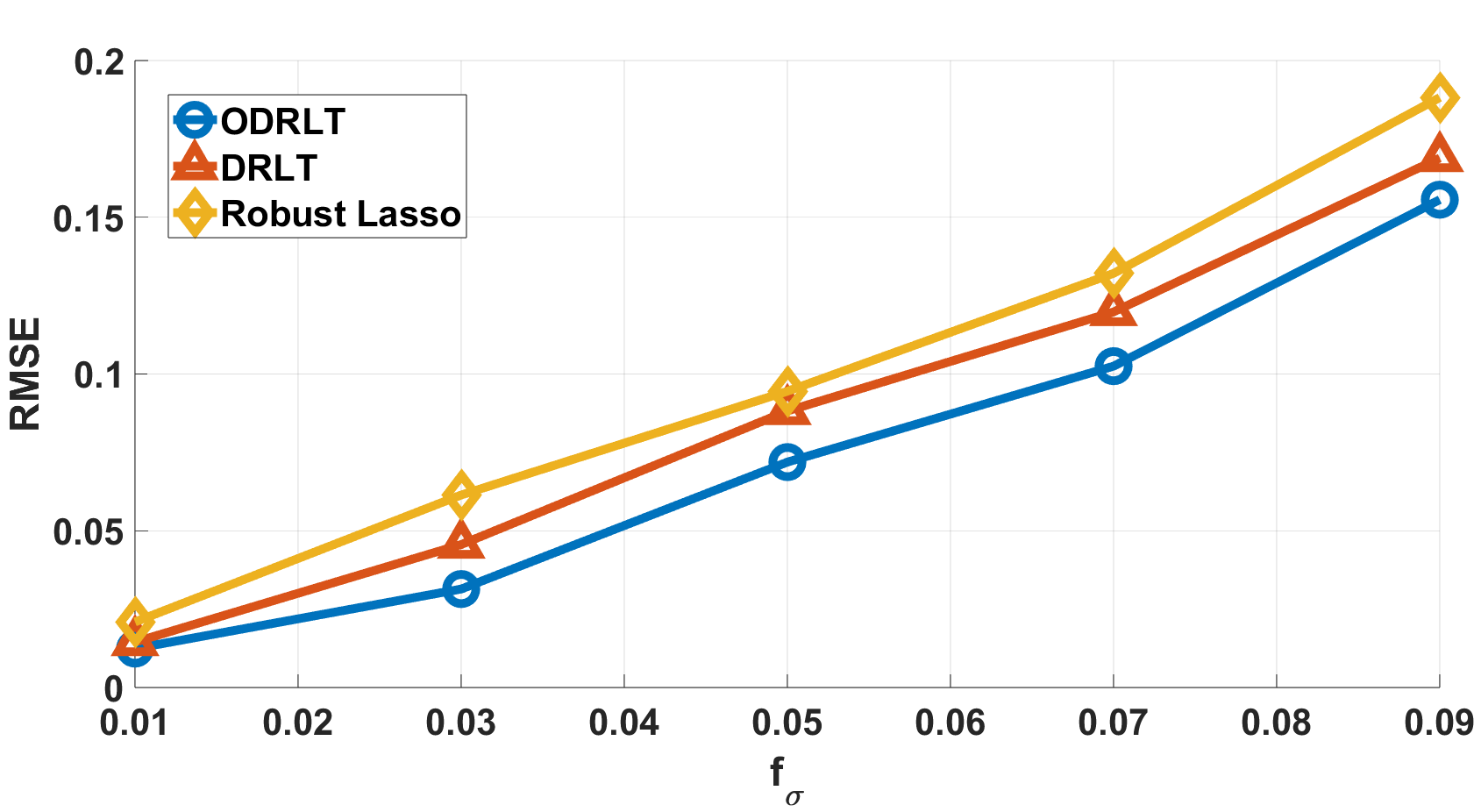}
    \includegraphics[scale=0.20]{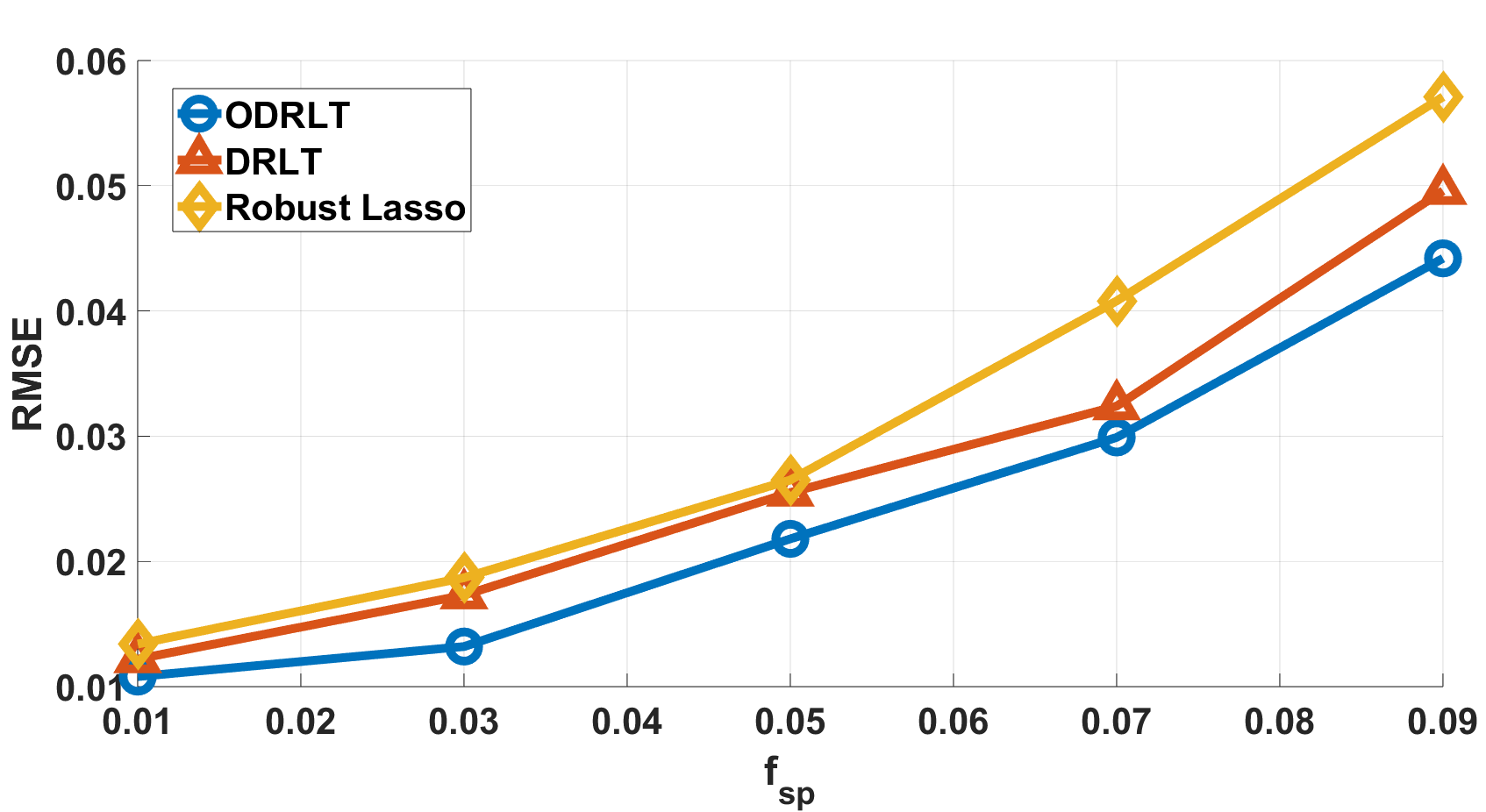}
    \caption{Average RRMSE comparison (over 25 independent noise runs) using  \textsc{Odrlt} for the viral loads $\boldsymbol{\beta^*}$ w.r.t. variation in the following parameters keeping others fixed: bit-flip proportions $f_{adv}$ as in setup (\textsf{EA}) (topleft), measurements $n$ (top right) as in setup (\textsf{EB}), noise level $f_{\sigma}$ as in setup (\textsf{EC}) (bottom left) and sparsity $f_{sp}$ as in setup (\textsf{ED}) (bottom right). The fixed parameters are dimension of $p = 500, f_{\sigma}=0.05,f_{adv}=0.01,f_{sp}=0.01,n=400$. }
    \label{fig:beta_RMSE}    
\end{figure*}

\section{Supplemental: Comparison with Different Sensing Matrices}
In this subsection, we compare the performance of \textsc{Odrlt} with different sensing matrices:
(\textit{i}) Centered Bernoulli($0.1$), (\textit{ii}) Centered Bernoulli($0.3$), (\textit{iii}) Centered Bernoulli($0.5$), and (\textit{iv}) Centered \textbf{Doubly-regular}. The elements of the sensing matrices in (\textit{i}), (\textit{ii}) and (\textit{iii}) have a distribution given in Section III-D of the main paper. Doubly-regular matrices, i.e. matrices with equal number of $1$'s and $0$'s in each row and column ($n/50$ ones per column and $p/50$ ones per row) with the locations of the $1$'s randomly chosen in each row, are a common model in group testing \cite{tan2022performance}. The centering for doubly-regular matrices was done as described in Section III-D of the main paper by choosing $\theta=1/50$. We compared the performances of these matrices in terms of RRMSE, Sensitivity and Specificity for the \textsc{Odrlt} estimates of $\boldsymbol{\beta^*}$ and $\boldsymbol{\delta^*}$.

For \textsc{Odrlt} estimates of $\boldsymbol{\delta^*}$ using all four types of sensing matrices, 
%In experimental setup \textsf{EA}, we varied $f_{adv} \in \{0.01,0.03,\ldots,0.09\}$ with fixed values $n = 400, f_{sp} = 0.01, f_{\sigma} = 0.05$. In \textsf{EB}, we varied $n$ from 200 to 450 in steps of 50 with $f_{adv} = 0.01, f_{sp} = 0.01, f_{\sigma} = 0.05$. In \textsf{EC}, we varied $f_{\sigma} \in \{0.01,0.03,\ldots,0.09\}$ with $n = 400, f_{adv} = 0.01, f_{sp} = 0.1$. In  \textsf{ED}, we varied $f_{sp} \in \{0.01,0.03,\ldots,0.09\}$ with $n = 400, f_{adv} = 0.01, f_{\sigma} = 0.05$. The experiments were run 25 times across different noise instances in $\boldsymbol{\eta}$, for the same signal $\boldsymbol{\beta^*}$.
the experimental setups \textsf{EA}, \textsf{EB}, \textsf{EC} and \textsf{ED} were chosen as described earlier. In Fig.~\ref{fig:sens_spec_matrix_delta}, we see that the \textsc{Odrlt} for $\boldsymbol{\delta^*}$ for doubly regular designs performs the best, followed by Centered Bernoulli($0.5$) matrices,  Centered Bernoulli($0.3$) and lastly Centered Bernoulli($0.1$).
\begin{figure*}
\centering
    \includegraphics[height=1.95in]{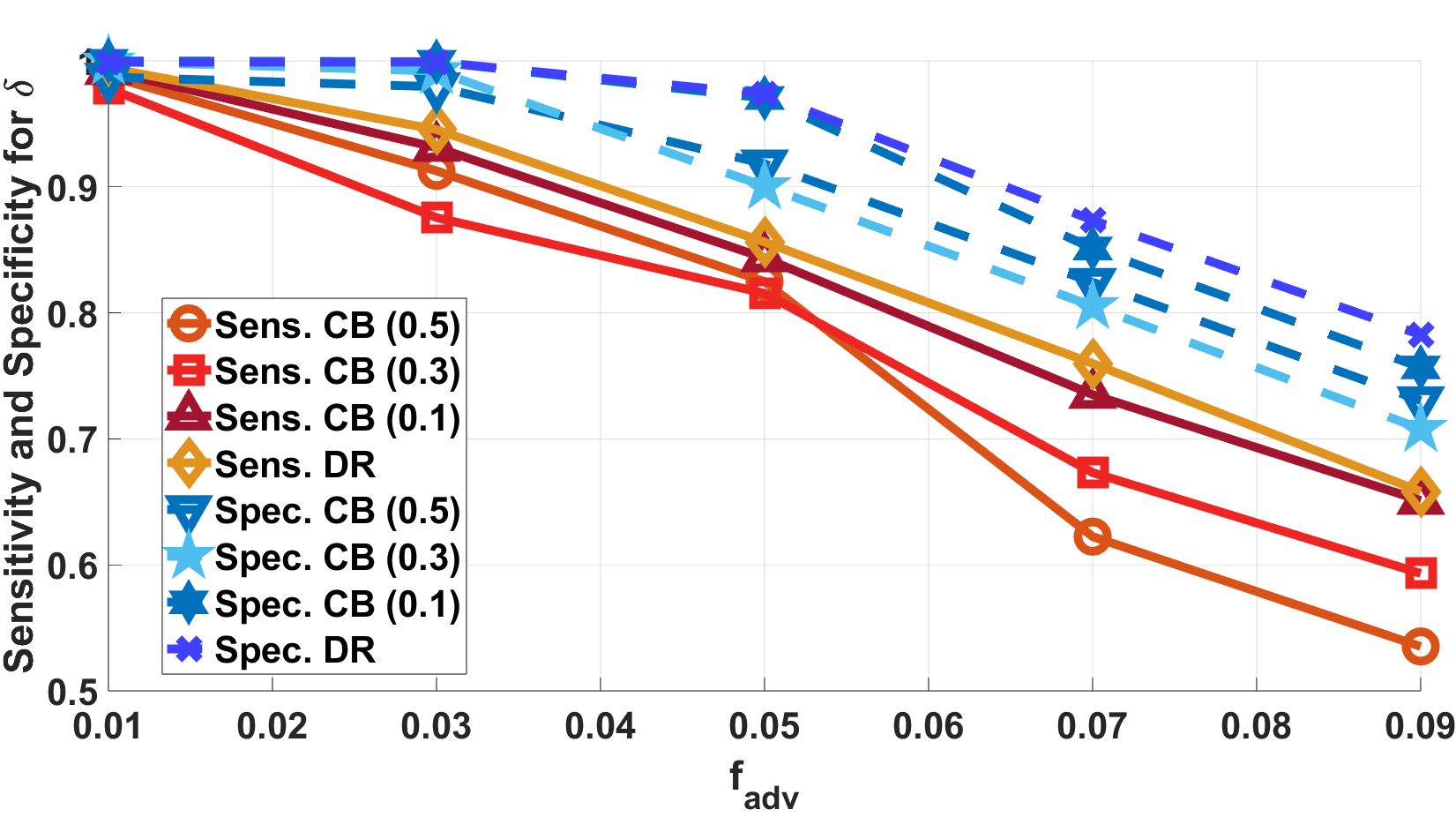}      
    \includegraphics[height=1.95in]{sens_spec_mat_n_delta.png}\\
    \includegraphics[height=1.95in]{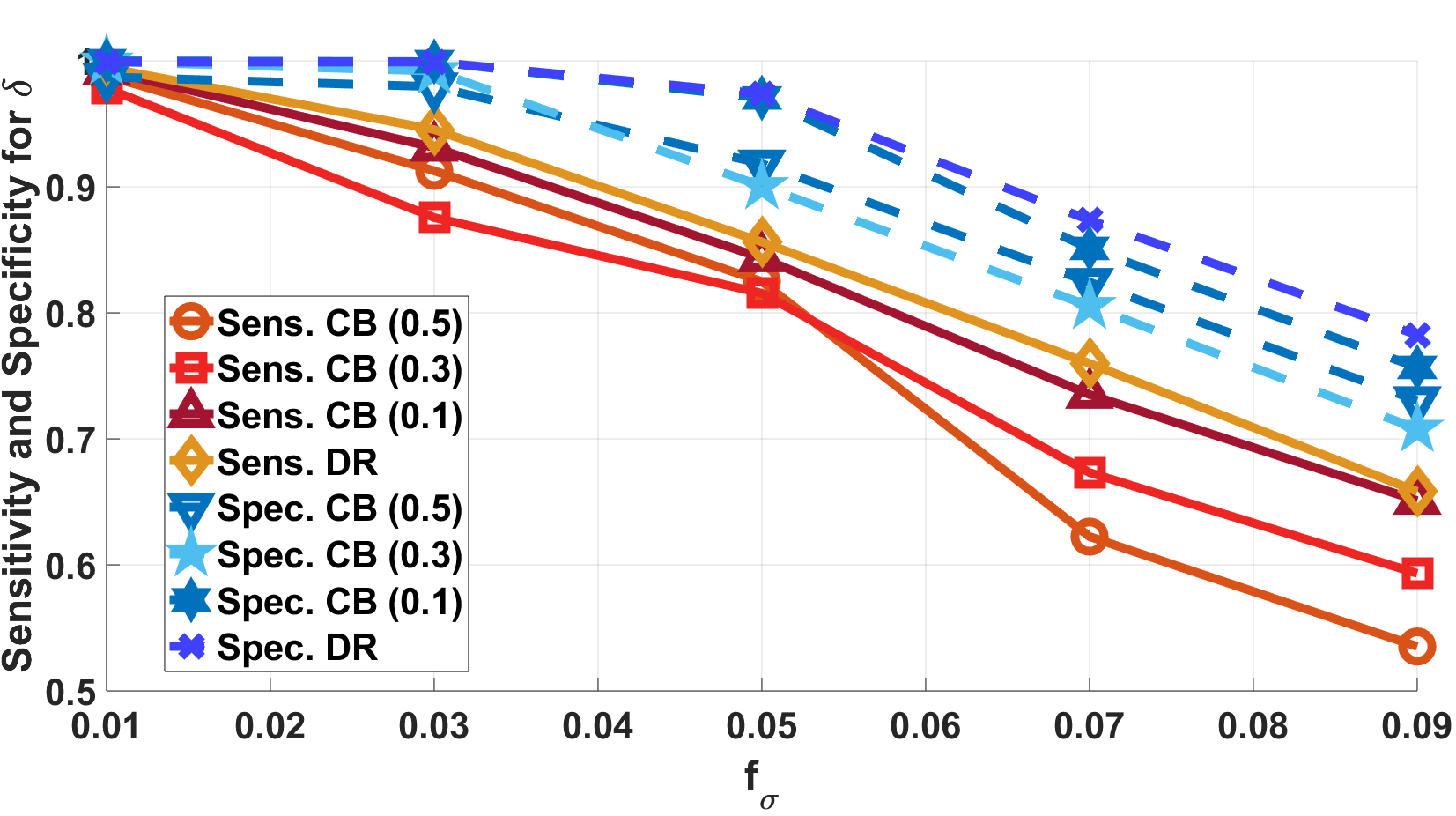}      
    \includegraphics[height=1.95in]{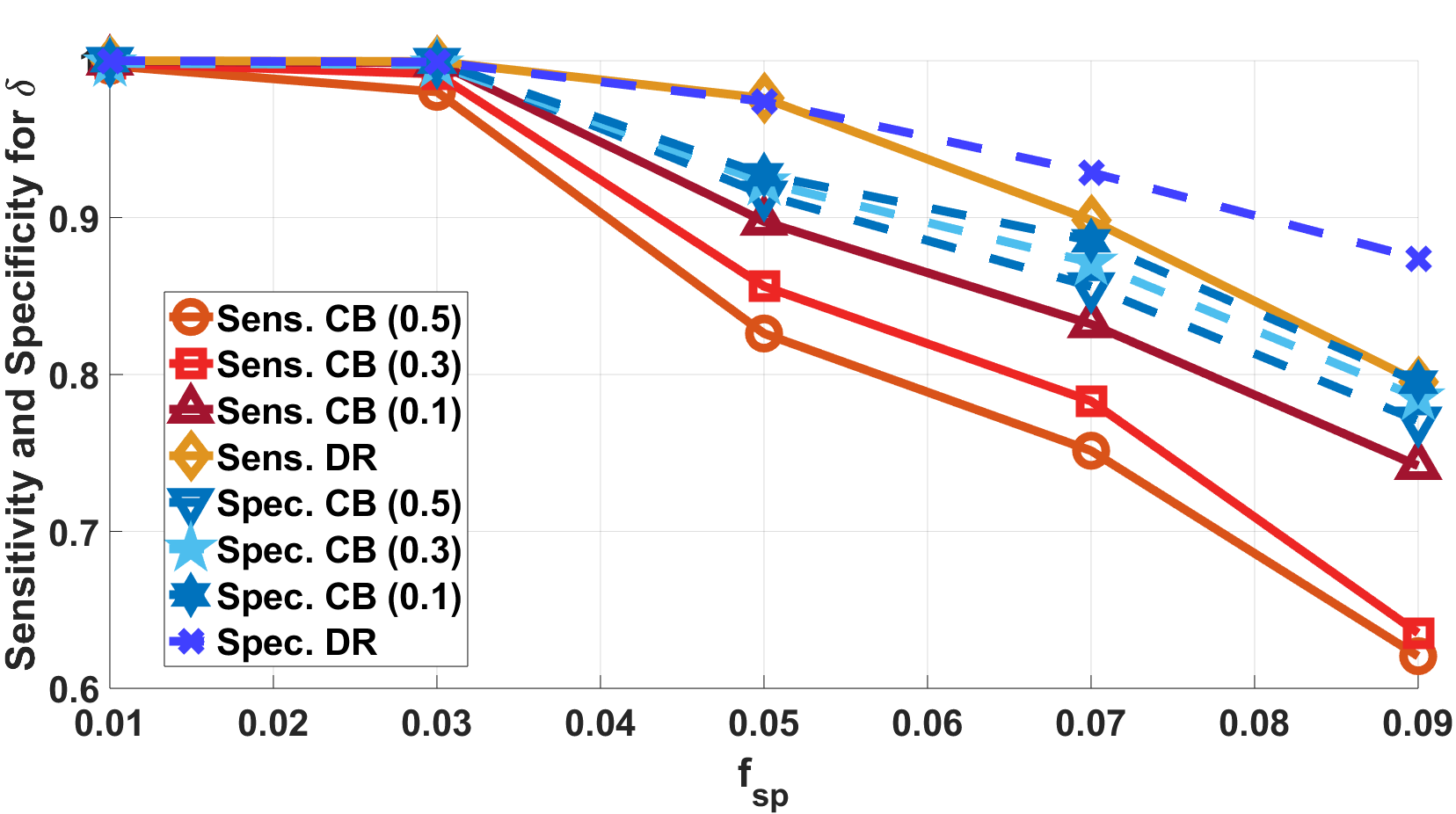}
    \caption{Average Sensitivity and Specificity plots (over 25 independent noise runs keeping $\boldsymbol{\beta^*}$, $\boldsymbol{A}$ and $\boldsymbol{\delta^*}$ fixed) for detecting measurements containing MMEs (i.e. detecting non-zero values of $\boldsymbol{\delta^*}$) using  \textsc{Odrlt} for Centered Bernoulli (0.5), Centered Bernoulli(0.3), Centered Bernoulli(0.1) and centered Doubly Regular sensing matrix. The experimental parameters are $p = 500, f_{\sigma}=0.05, f_{adv}=0.01,f_{sp}=0.1,n = 400$. Left to right, top to bottom: results for experiments \textsf{EA}, \textsf{EB}, \textsf{EC}, \textsf{ED}.}   
    \label{fig:sens_spec_matrix_delta}    
\end{figure*}
For the \textsc{Odrlt} for $\boldsymbol{\beta^*}
$, we again used experiment setups \textsf{EA}, \textsf{EB}, \textsf{EC} and \textsf{ED}. In Fig.~\ref{fig:Sens_spec_beta_matrix}, we observe that \textsc{Odrlt} for $\boldsymbol{\beta^*}$ produces the highest sensitivity and specificity for Centered Bernoulli($0.5$), followed by doubly-regular, Centered Bernoulli($0.3$) and lastly Centered Bernoulli($0.1$). Similar trends are observed for the RRMSE, as shown in Fig.~\ref{fig:matrix_RMSE}.
%For the bit-flips experiment i.e., (\textsf{EA}), $f_{adv}$ was varied in $\{0.01,0.03,\ldots,0.09\}$ with $n=400,f_{sp}=0.01,f_{\sigma}=0.05$. For the measurements experiment (\textsf{EB}), $n$ was varied over $\{200,150,\ldots,450\}$ with $f_{sp} = 0.01 , f_{adv} = 0.01, f_{\sigma} = 0.05$. For the noise experiment (i.e., (\textsf{EC}), we varied $f_{\sigma}$ in $\{0.01,0.03,\ldots,0.09\}$ with $n=400,f_{sp}=0.1,f_{adv}=0.01$. For the sparsity experiment (i.e., (\textsf{ED}), $f_{sp}$ was varied in $\{0.01,0.03,\ldots,0.09\}$ with $n = 400, f_{adv} = 0.01, f_{\sigma} = 0.05$. Sensitivity and Specificity is calculated based on the technique described in Sec.\ref{subsec:exp_beta_hypothesis}. The experiments were run 25 times across different noise instances in $\boldsymbol{\eta}$, for the same signal $\boldsymbol{\beta^*}$.

\begin{figure*}
   \centering
    \includegraphics[height=1.9in]{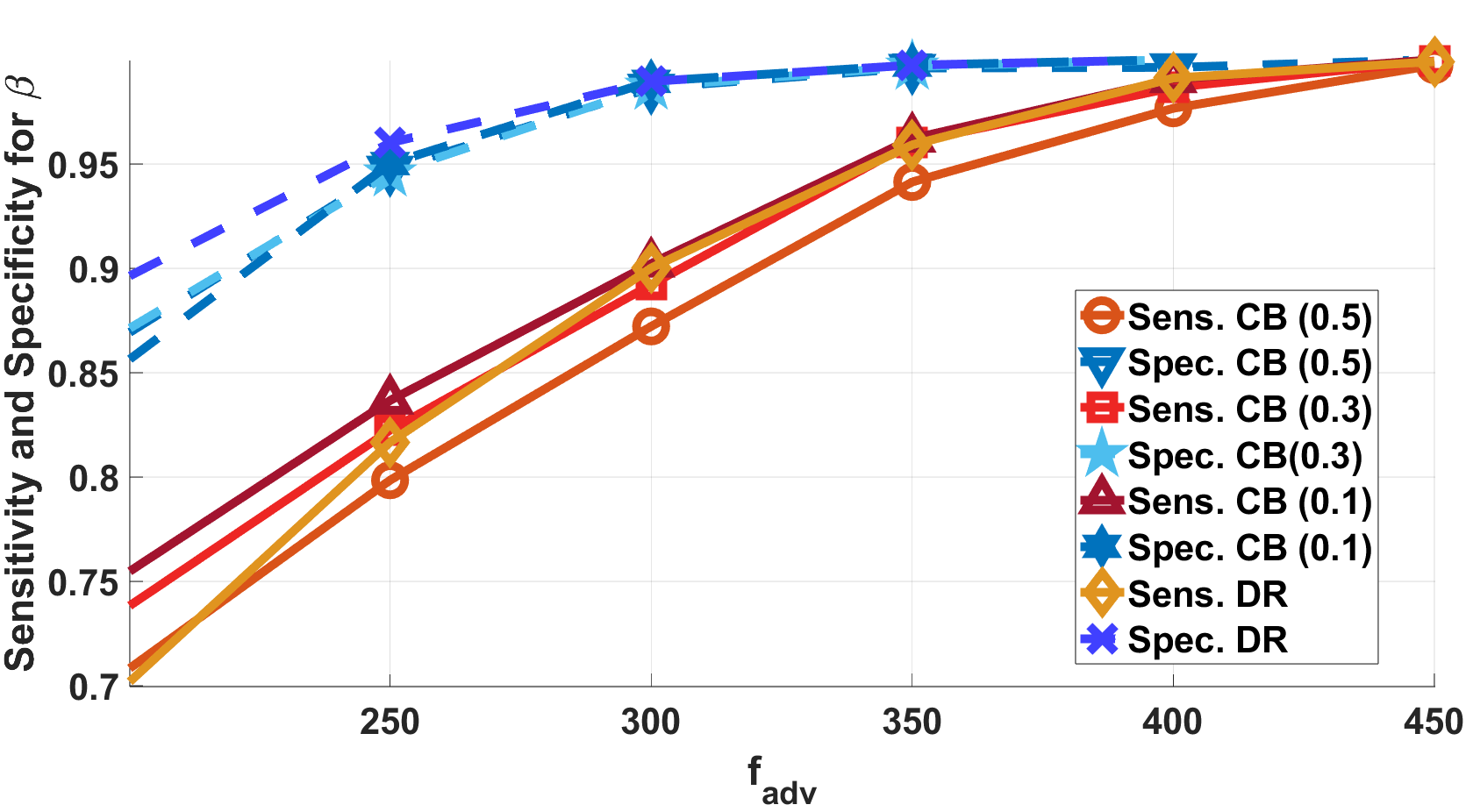}      
    \includegraphics[height=1.9in]{sens_spec_mat_n.png}\\
    \includegraphics[height=1.9in]{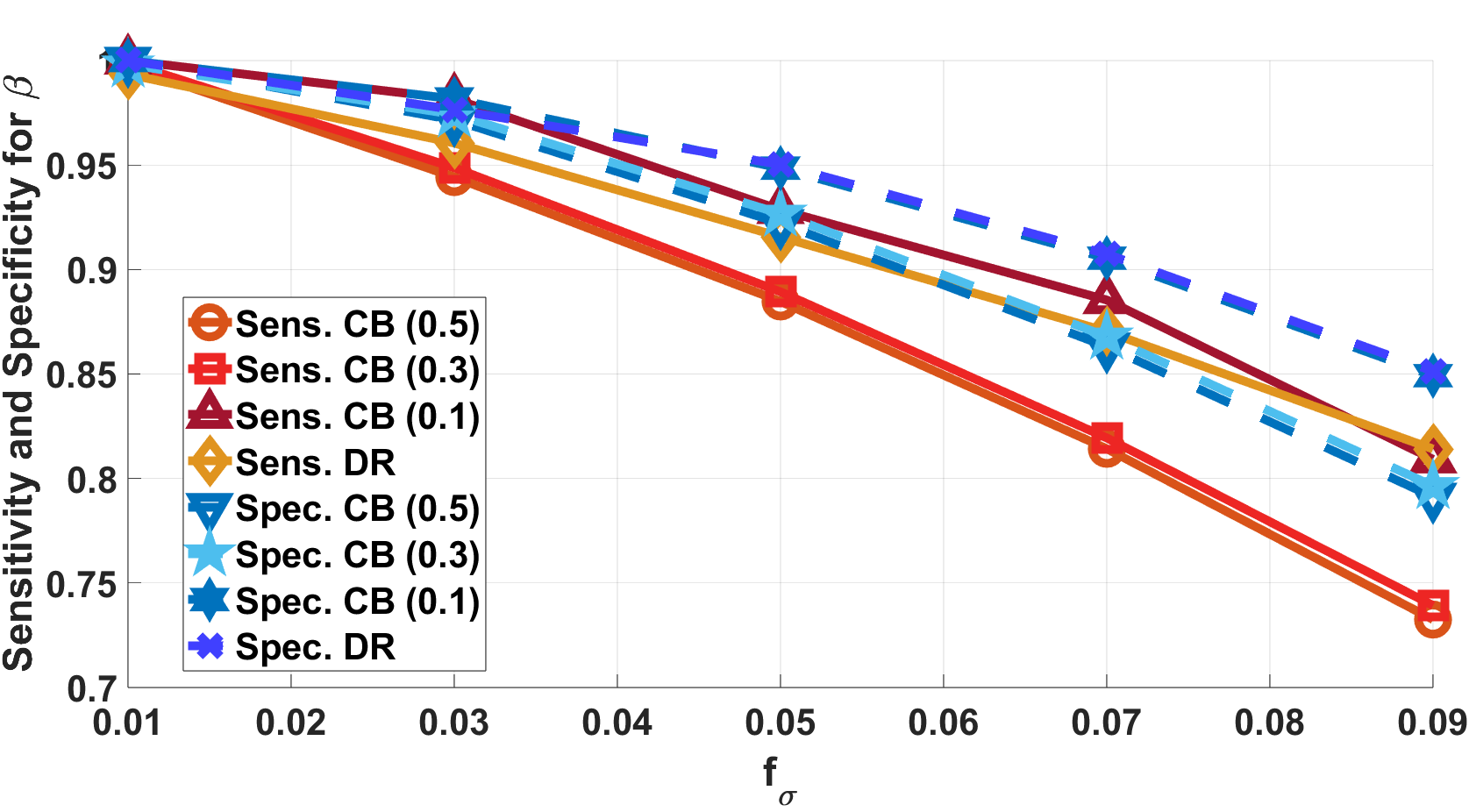}      
    \includegraphics[height=1.9in]{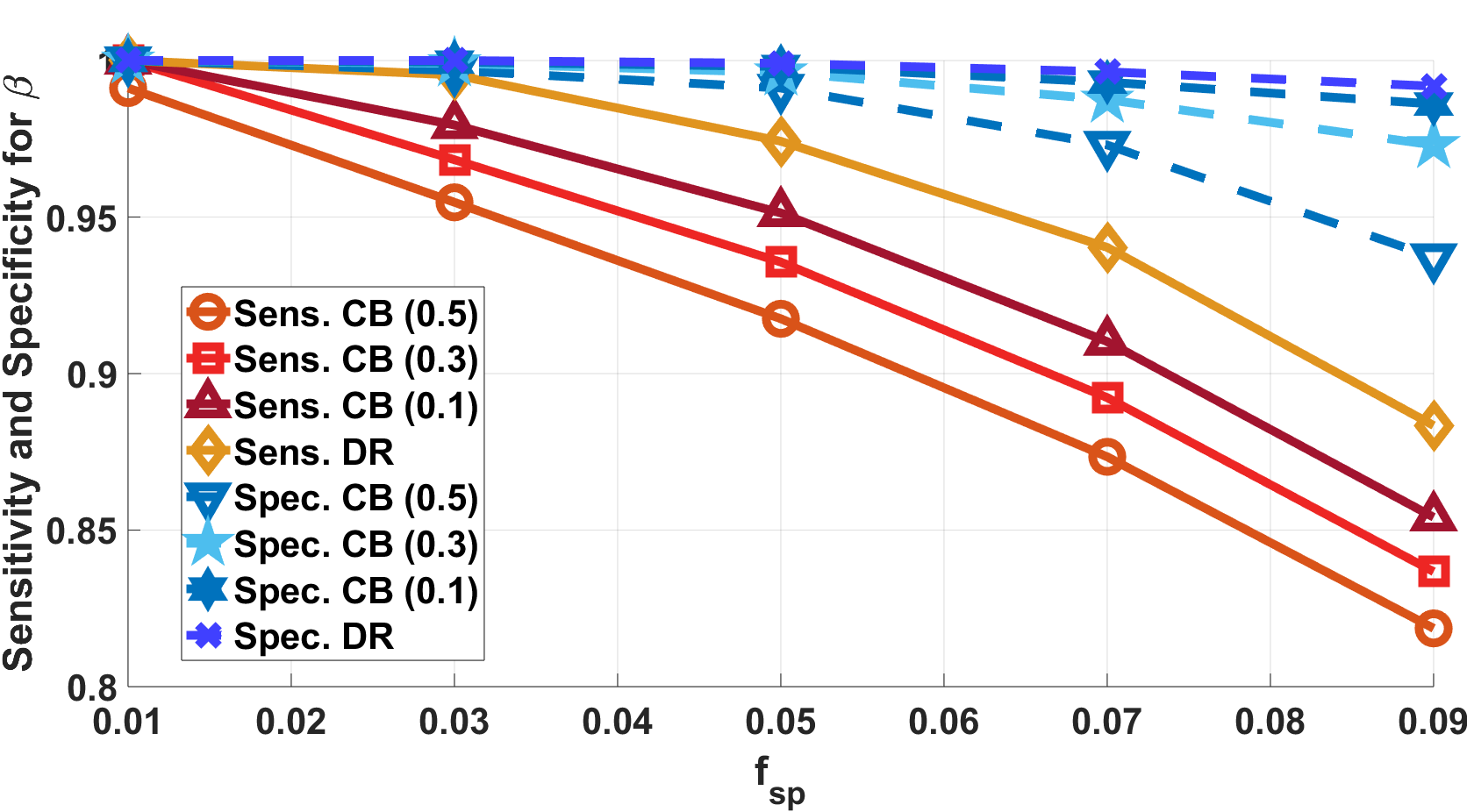}
    \caption{Average Sensitivity and Specificity plots (over 25 independent noise runs keeping $\boldsymbol{\beta^*}$, $\boldsymbol{A}$ and $\boldsymbol{\delta^*}$ fixed) for detecting defective samples (i.e., non-zero values of $\boldsymbol{\beta^*}$) using  \textsc{Odrlt} for Centered Bernoulli (0.5), Centered Bernoulli(0.3), Centered Bernoulli(0.1) and centered Doubly Regular sensing matrix. Left to right, top to bottom: results for experiments (\textsf{EA}), (\textsf{EB}), (\textsf{EC}), (\textsf{ED}). The experimental parameters are $p = 500, f_{\sigma}=0.05, f_{adv}=0.01,f_{sp}=0.1,n = 400$. }  
    \label{fig:Sens_spec_beta_matrix}    
\end{figure*}

%We computed the RRMSE of the \textsc{Odrlt} estimates for all four sensing matrices the same way as described in Sec.\ref{sec:rmse comparison}. We examined the variation in RRMSE with regard to change in the following parameters, keeping all other parameters fixed. For the bit-flips experiment i.e., (\textsf{EA}), $f_{adv}$ was varied in $\{0.01,0.03,\ldots,0.09\}$ with $n=400,f_{sp}=0.01,f_{\sigma}=0.05$. For the measurements experiment (\textsf{EB}), $n$ was varied over $\{200,150,\ldots,450\}$ with $f_{sp} = 0.01 , f_{adv} = 0.01, f_{\sigma} = 0.05$. For the noise experiment (i.e., (\textsf{EC}), we varied $f_{\sigma}$ in $\{0.01,0.03,\ldots,0.09\}$ with $n=400,f_{sp}=0.1,f_{adv}=0.01$. For the sparsity experiment (i.e., (\textsf{ED}), $f_{sp}$ was varied in $\{0.01,0.03,\ldots,0.09\}$ with $n = 400, f_{adv} = 0.01, f_{\sigma} = 0.05$. Sensitivity and Specificity is calculated based on the technique described in Sec.\ref{subsec:exp_beta_hypothesis}. The experiments were run 25 times across different noise instances in $\boldsymbol{\eta}$, for the same signal $\boldsymbol{\beta^*}$.
\begin{figure*}
\centering
    \includegraphics[scale=0.19]{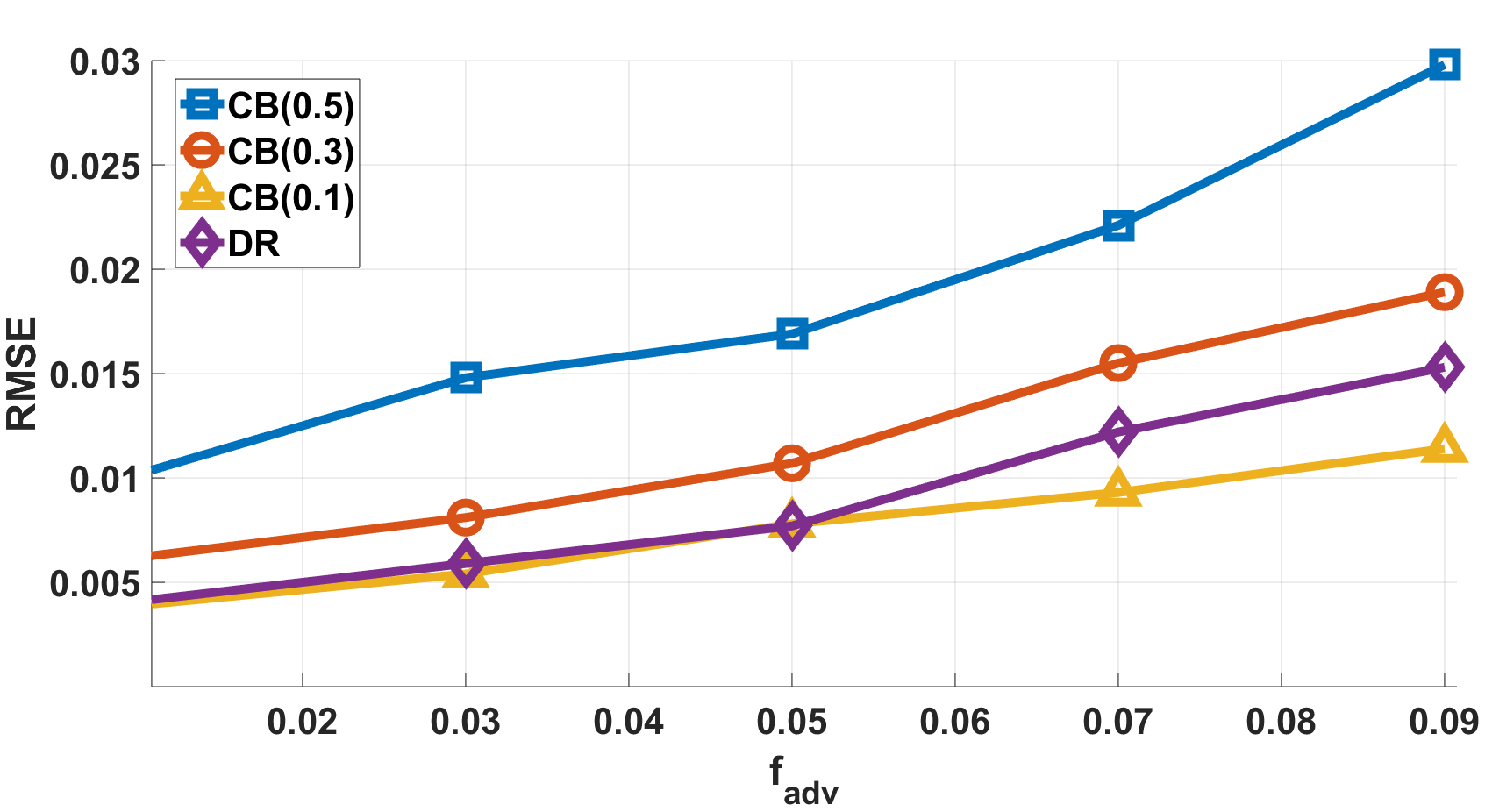}
    \includegraphics[scale=0.19]{rmse_mat_n.png}\\
    \includegraphics[scale=0.19]{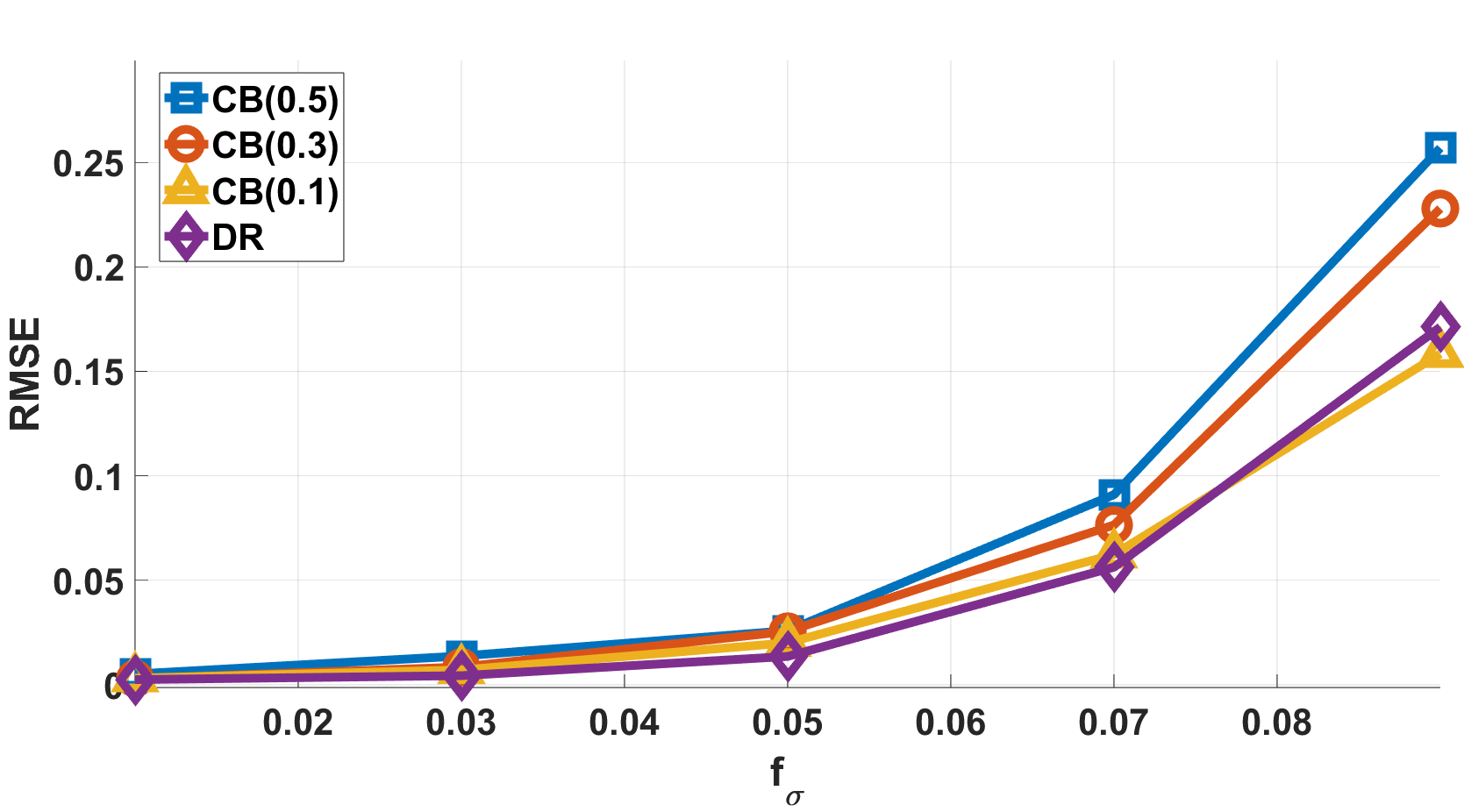}
    \includegraphics[scale=0.19]{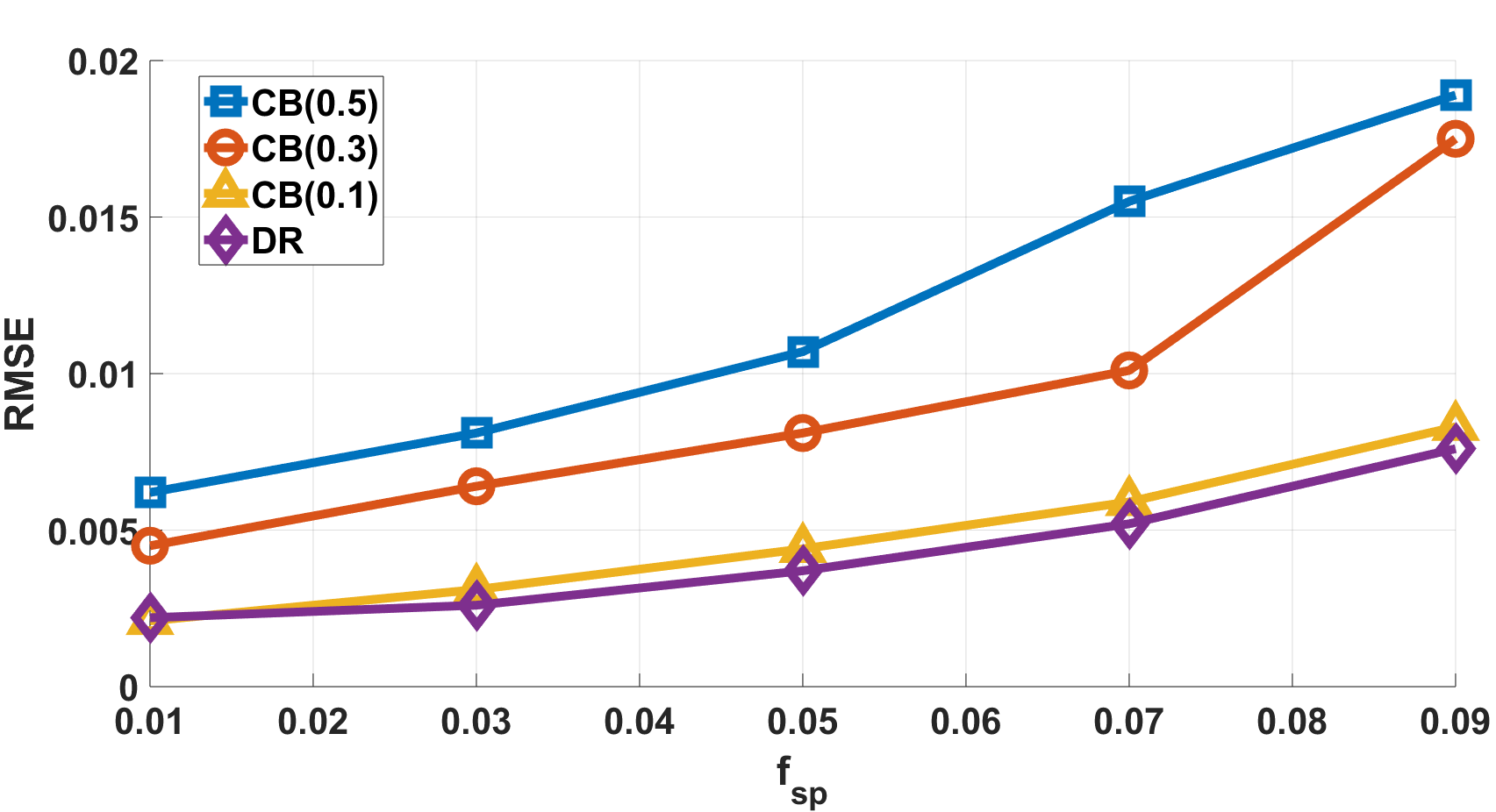}
    \caption{Average RRMSE comparison (over 25 independent noise runs keeping $\boldsymbol{\beta^*}$, $\boldsymbol{A}$ and $\boldsymbol{\delta^*}$ fixed) using  \textsc{Odrlt} for Centered Bernoulli (0.5), Centered Bernoulli(0.3), Centered Bernoulli(0.1) and centered Doubly Regular sensing matrix w.r.t. variation in the following parameters keeping others fixed: bit-flip proportions $f_{adv}$ as in setup (\textsf{EA}) (topleft), measurements $n$ (top right) as in setup (\textsf{EB}), noise level $f_{\sigma}$ as in setup (\textsf{EC}) (bottom left) and sparsity $f_{sp}$ as in setup (\textsf{ED}) (bottom right). The fixed parameters are dimension of $p = 500, f_{\sigma}=0.05,f_{adv}=0.01,f_{sp}=0.01,n=400$. }
    \label{fig:matrix_RMSE}    
\end{figure*}

\section{Supplemental: Comparison with Different Noise Models}
{In this set of experiments, we compared the performance of \textsc{Odrlt} in the presence of additive noise $\boldsymbol{\eta}$ obtained from three different distributions in addition to $\mathcal{N}(0,\sigma^2)$): (\textit{i}) Bounded Uniform$[-\sqrt{3}\sigma,\sqrt{3}\sigma]$, and (\textit{ii}) a Generalized Gaussian (GG) distribution with shape parameter $1.5$ and scale $\sigma^2$ with the following probability density function $$f(x) = \frac{3}{4 \sigma \, \Gamma\left( \frac{2}{3} \right)} \exp\left( -\left| \frac{x}{\sigma} \right|^{1.5} \right)
.
$$}

For the different noise models and different experimental setups, we observe the sensitivity and specificity of the \textsc{ODrlt} estimates of $\boldsymbol{\delta^*}$ and $\boldsymbol{\beta^*}$ in Fig.~\ref{fig:sens_spec_noise_delta}and~\ref{fig:Sens_spec_beta_noise} respectively. We present the RRMSE for the estimates of $\boldsymbol{\beta^*}$ in Fig.~\ref{fig:noise_RMSE}. We observe that the performance under Gaussian and Uniform noise is approximately the same with symmetric Beta being slightly worse. Since the Generalised Gaussian is sub-Gaussian with heavier tails than standard Gaussian, it does perform slightly worse. However, for large $n$, small $f_{sp}, f_{adv}$ and $f_{\sigma}$, the performance of \textsc{Odrlt} in the presence of Generalised Gaussian noise is good.

%\subsubsection{\textbf{Sensitivity and Specificity of} $\boldsymbol{\delta^*}$} We performed experiments to study sensitivity and specificity \textsc{Odrlt} for $\boldsymbol{\delta^*}$ of all $3$ type of additive noise models. In experimental setup \textsf{EA}, we varied $f_{adv} \in \{0.01,0.03,\ldots,0.09\}$ with fixed values $n = 400, f_{sp} = 0.01, f_{\sigma} = 0.05$. In \textsf{EB}, we varied $n$ from 200 to 450 in steps of 50 with $f_{adv} = 0.01, f_{sp} = 0.01, f_{\sigma} = 0.05$. In \textsf{EC}, we varied $f_{\sigma} \in \{0.01,0.03,\ldots,0.09\}$ with $n = 400, f_{adv} = 0.01, f_{sp} = 0.1$. In  \textsf{ED}, we varied $f_{sp} \in \{0.01,0.03,\ldots,0.09\}$ with $n = 400, f_{adv} = 0.01, f_{\sigma} = 0.05$. The experiments were run 25 times across different noise instances in $\boldsymbol{\eta}$, for the same signal $\boldsymbol{\beta^*}$. The sensitivity and specificity was computed the same way as described in Sec.\ref{sec:sensitivity_delta}.

\begin{figure*}
\centering
\includegraphics[height=1.9in]{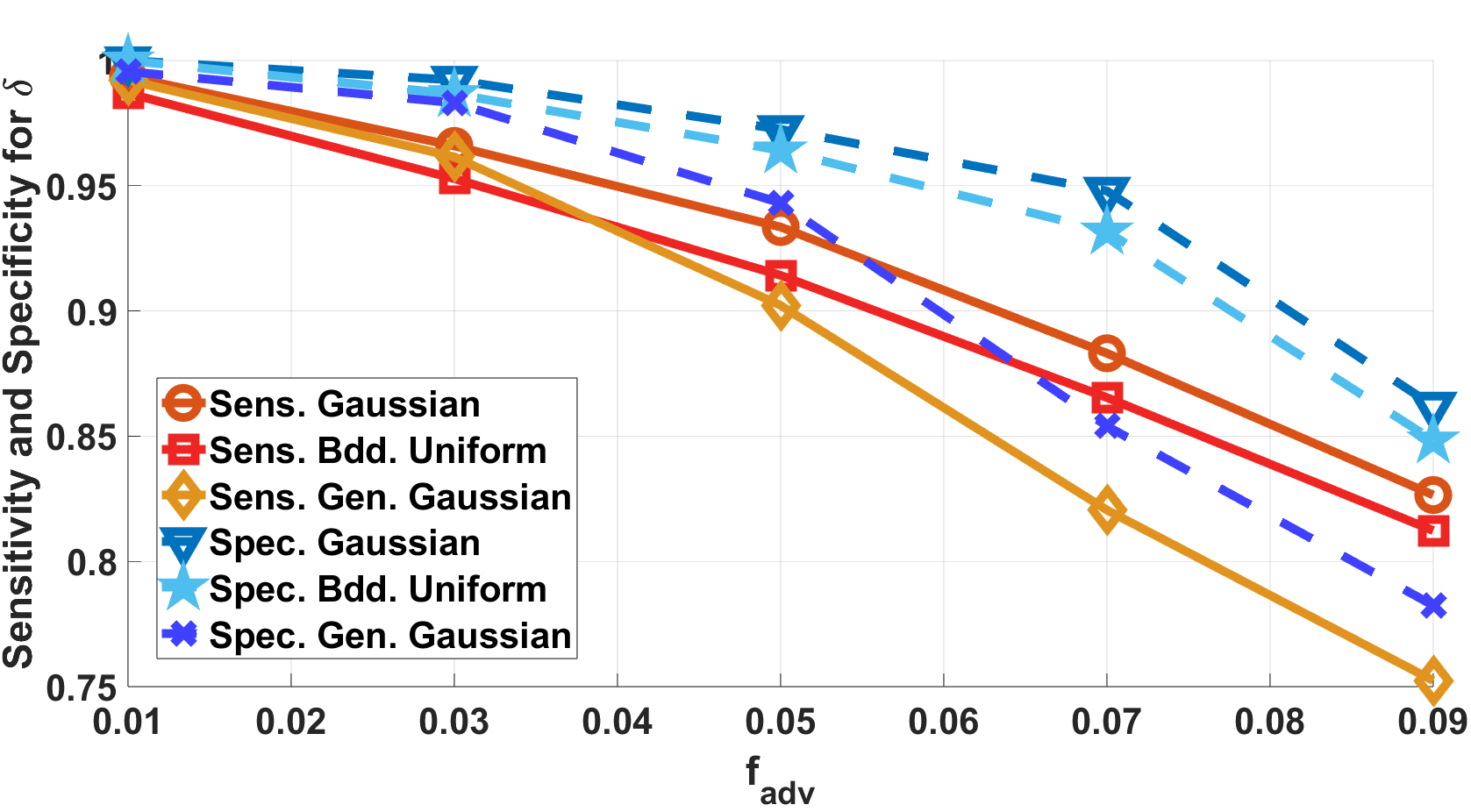}      
    \includegraphics[height=1.9in]{sens_spec_noise_n_delta.png}\\
    \includegraphics[height=1.9in]{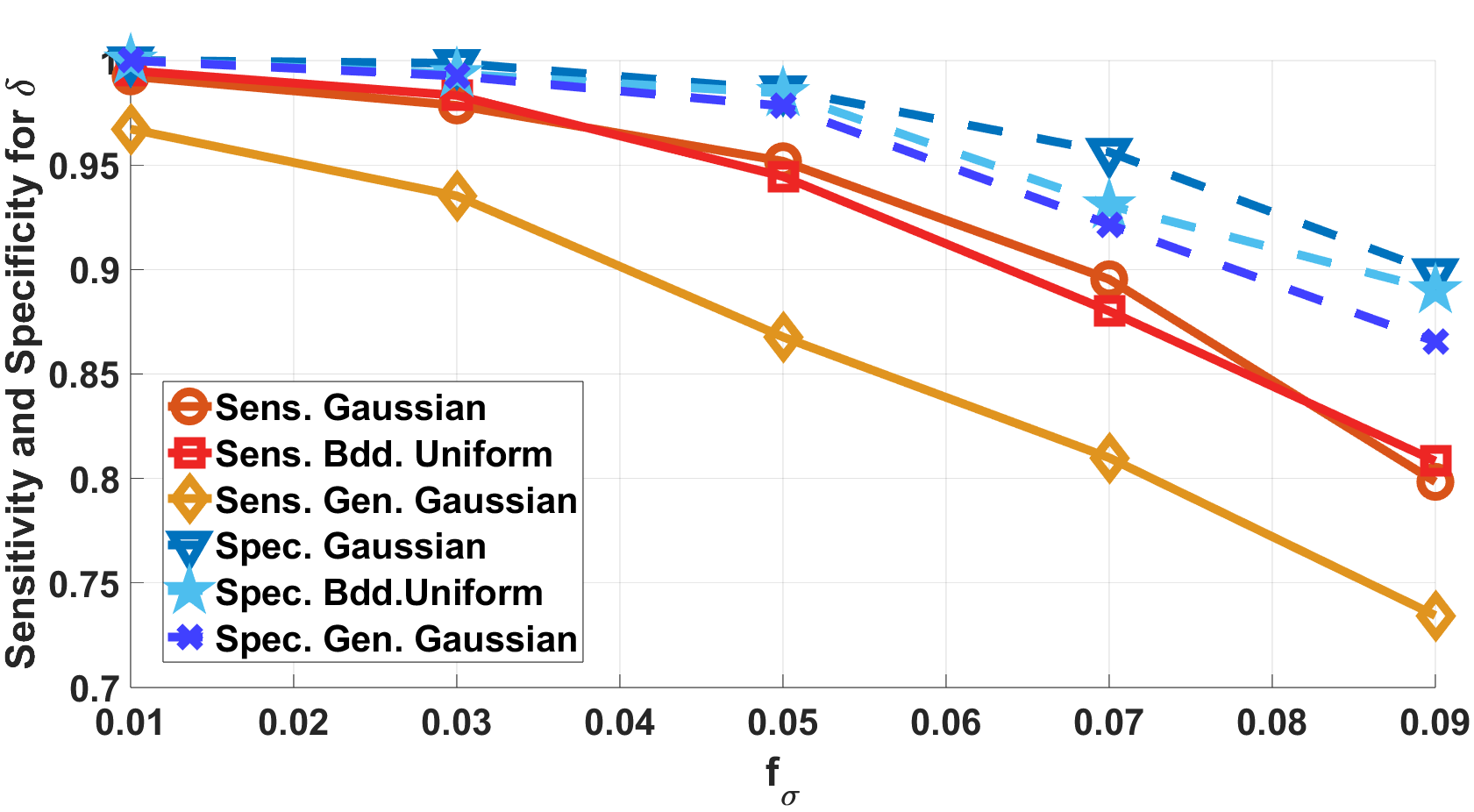}      
    \includegraphics[height=1.9in]{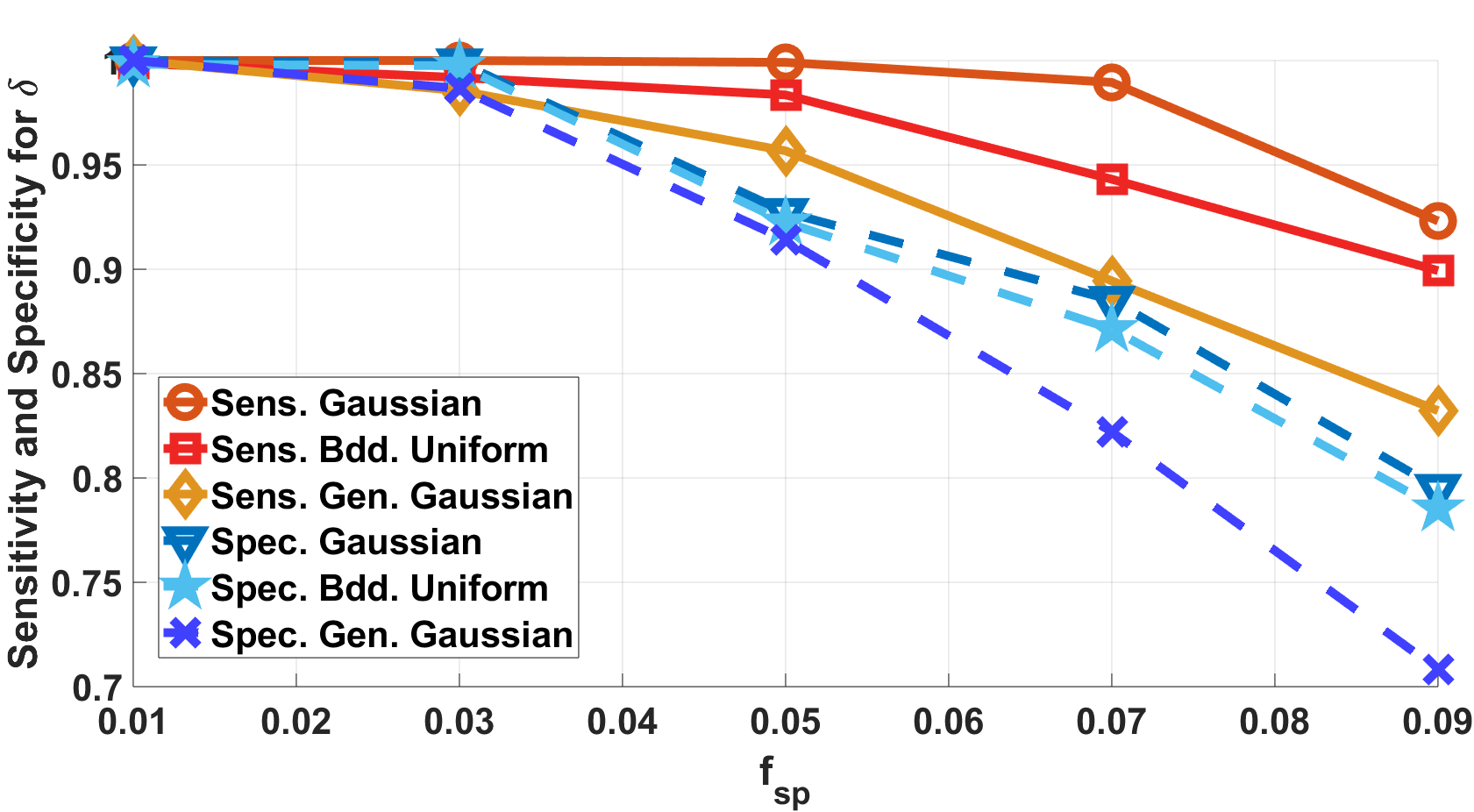}
    \caption{Average Sensitivity and Specificity plots (over 25 independent noise runs keeping $\boldsymbol{\beta^*}$, $\boldsymbol{A}$ and $\boldsymbol{\delta^*}$ fixed) for detecting measurements containing MMEs (i.e. detecting non-zero values of $\boldsymbol{\delta^*}$) using  \textsc{ODrlt} for additive noise models $N(0,\sigma^2)$, Empricial \textsc{ODrlt} for Bounded uniform $[-\sqrt{3}\sigma,\sqrt{3}\sigma]$ and Generalised Gaussian with shape $3/2$. The experimental parameters are $p = 500, f_{\sigma}=0.05, f_{adv}=0.01,f_{sp}=0.1,n = 400$. Left to right, top to bottom: results for experiments \textsf{EA}, \textsf{EB}, \textsf{EC}, \textsf{ED}.}   
    \label{fig:sens_spec_noise_delta}    
\end{figure*}
%\subsection{\textsc{Odrlt} for }
%In this set of experimental results, we examined the effectiveness of \textsc{Odrlt} to detect defective samples in $\boldsymbol{\beta^*}  in the presence of bit-flips in $\boldsymbol{A}$ induced as per adversarial MMEs for all $3$ type of additive noise models. We examined the variation in sensitivity and specificity with regard to change in the following parameters, keeping all other parameters fixed. For the bit-flips experiment i.e., (\textsf{EA}), $f_{adv}$ was varied in $\{0.01,0.03,\ldots,0.09\}$ with $n=400,f_{sp}=0.01,f_{\sigma}=0.05$. For the measurements experiment (\textsf{EB}), $n$ was varied over $\{200,150,\ldots,450\}$ with $f_{sp} = 0.01 , f_{adv} = 0.01, f_{\sigma} = 0.05$. For the noise experiment (i.e., (\textsf{EC}), we varied $f_{\sigma}$ in $\{0.01,0.03,\ldots,0.09\}$ with $n=400,f_{sp}=0.1,f_{adv}=0.01$. For the sparsity experiment (i.e., (\textsf{ED}), $f_{sp}$ was varied in $\{0.01,0.03,\ldots,0.09\}$ with $n = 400, f_{adv} = 0.01, f_{\sigma} = 0.05$. Sensitivity and Specificity is calculated based on the technique described in Sec.\ref{subsec:exp_beta_hypothesis}. The experiments were run 25 times across different noise instances in $\boldsymbol{\eta}$, for the same signal $\boldsymbol{\beta^*}$.

\begin{figure*}
   \centering
   \includegraphics[height=1.9in]{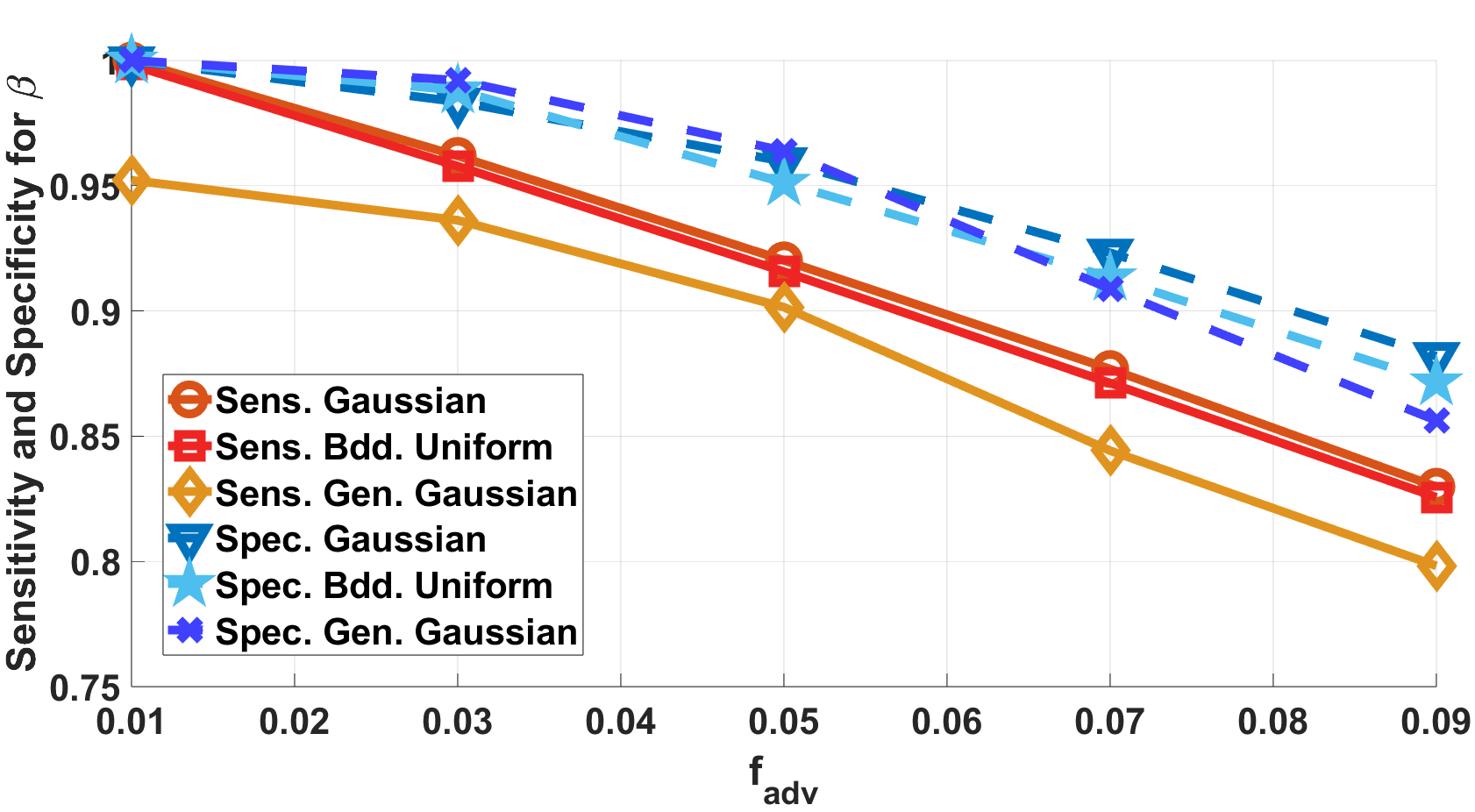}      
    \includegraphics[height=1.9in]{sens_spec_noise_n_beta.png}\\
    \includegraphics[height=1.9in]{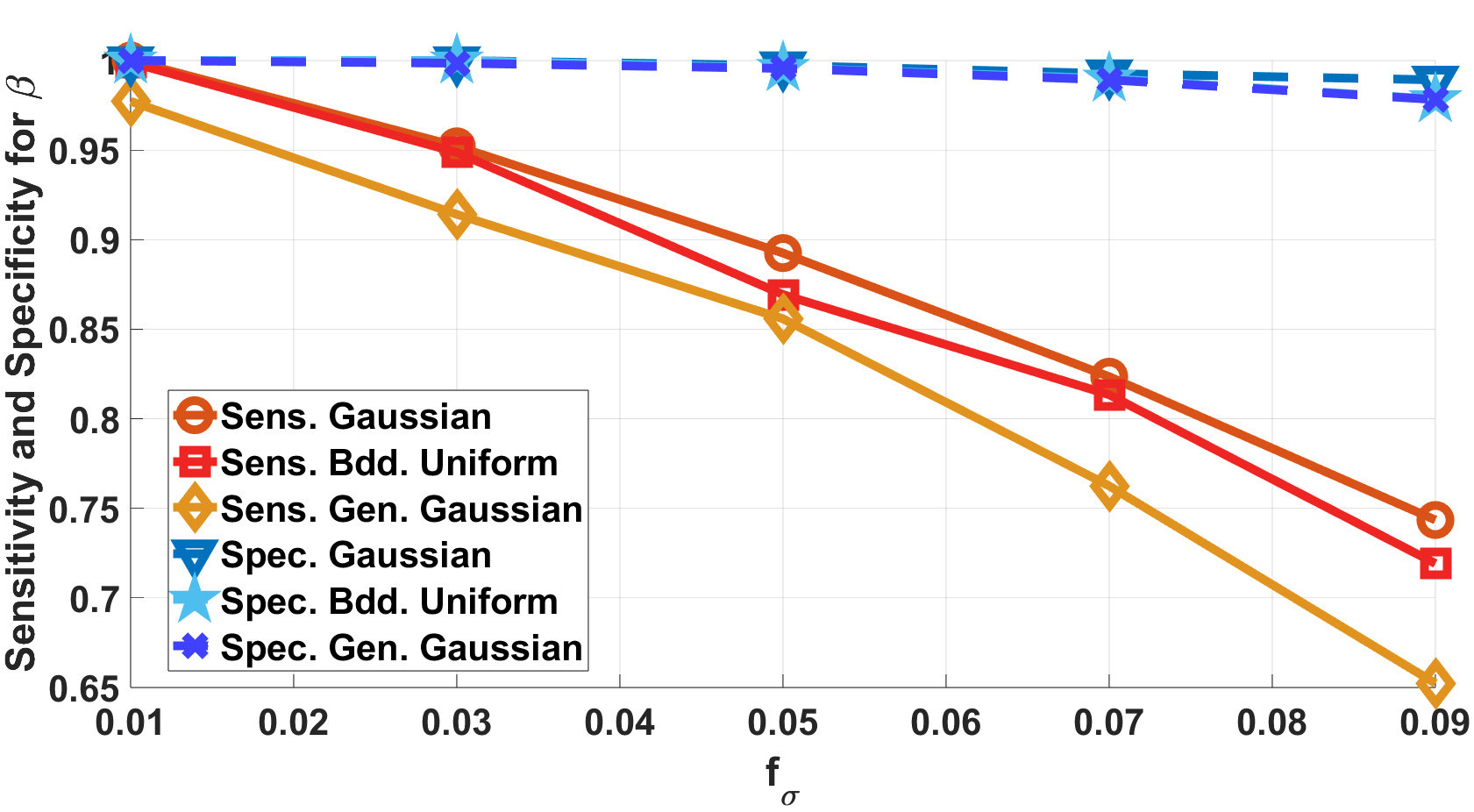}      
    \includegraphics[height=1.9in]{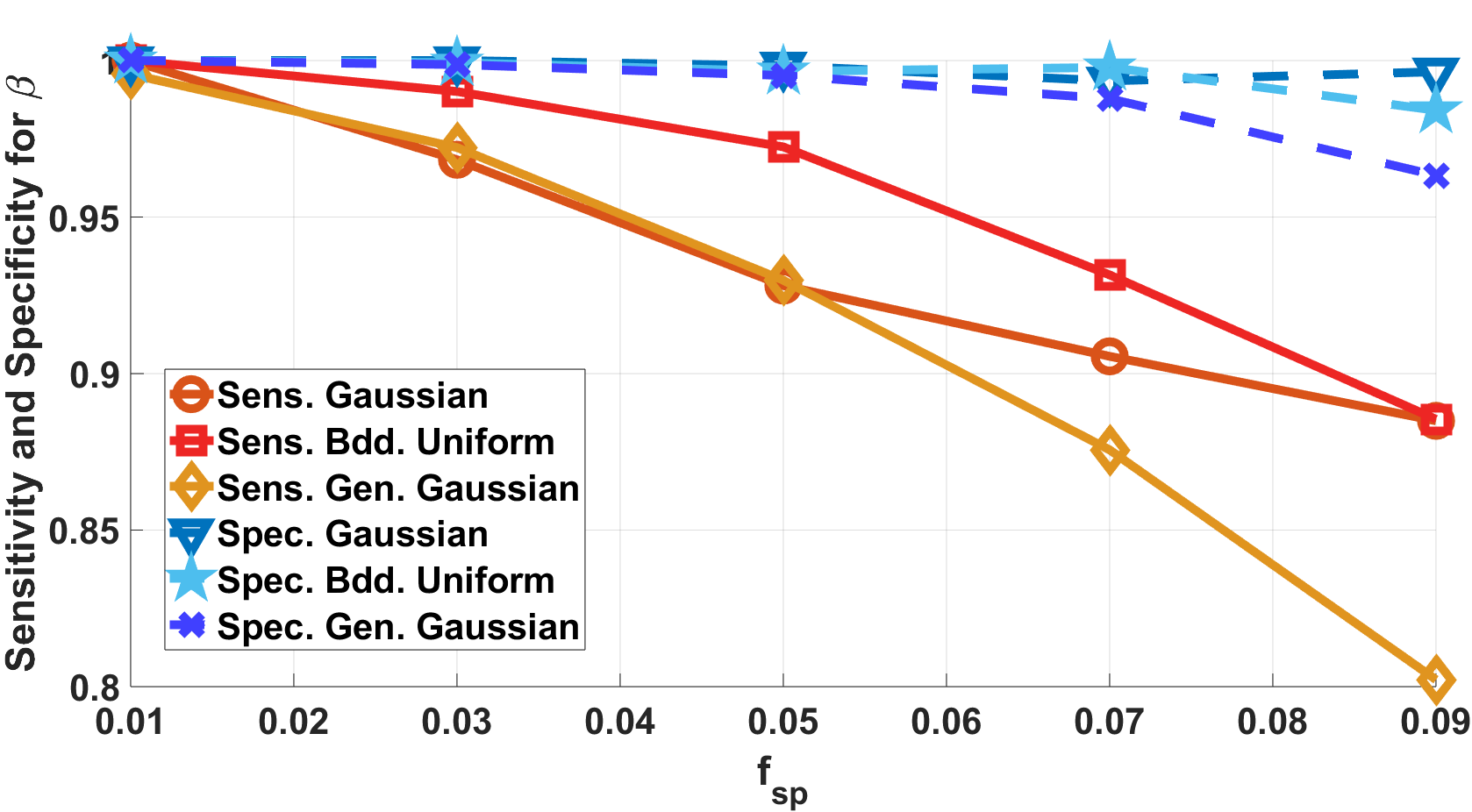}
    \caption{Average Sensitivity and Specificity plots(over 25 independent noise runs keeping $\boldsymbol{\beta^*}$, $\boldsymbol{A}$ and $\boldsymbol{\delta^*}$ fixed) for detecting defective samples (i.e., non-zero values of $\boldsymbol{\beta^*}$) using  \textsc{ODrlt} for additive noise models $N(0,\sigma^2)$, Empricial \textsc{ODrlt} for Bounded uniform $[-\sqrt{3}\sigma,\sqrt{3}\sigma]$ and Generalised Gaussian with shape $3/2$. Left to right, top to bottom: results for experiments (\textsf{EA}), (\textsf{EB}), (\textsf{EC}), (\textsf{ED}). The experimental parameters are $p = 500, f_{\sigma}=0.05, f_{adv}=0.01,f_{sp}=0.1,n = 400$. }  
    \label{fig:Sens_spec_beta_noise}    
\end{figure*}

%We computed the RRMSE of the \textsc{Odrl} estimates for all three additive noise models the same way as described in Sec.\ref{sec:rmse comparison}. We examined the variation in RRMSE with regard to change in the following parameters, keeping all other parameters fixed. For the bit-flips experiment i.e., (\textsf{EA}), $f_{adv}$ was varied in $\{0.01,0.03,\ldots,0.09\}$ with $n=400,f_{sp}=0.01,f_{\sigma}=0.05$. For the measurements experiment (\textsf{EB}), $n$ was varied over $\{200,150,\ldots,450\}$ with $f_{sp} = 0.01 , f_{adv} = 0.01, f_{\sigma} = 0.05$. For the noise experiment (i.e., (\textsf{EC}), we varied $f_{\sigma}$ in $\{0.01,0.03,\ldots,0.09\}$ with $n=400,f_{sp}=0.1,f_{adv}=0.01$. For the sparsity experiment (i.e., (\textsf{ED}), $f_{sp}$ was varied in $\{0.01,0.03,\ldots,0.09\}$ with $n = 400, f_{adv} = 0.01, f_{\sigma} = 0.05$. Sensitivity and Specificity is calculated based on the technique described in Sec.\ref{subsec:exp_beta_hypothesis}. The experiments were run 25 times across different noise instances in $\boldsymbol{\eta}$, for the same signal $\boldsymbol{\beta^*}$.
\begin{figure*}
\centering
\includegraphics[scale=0.19]{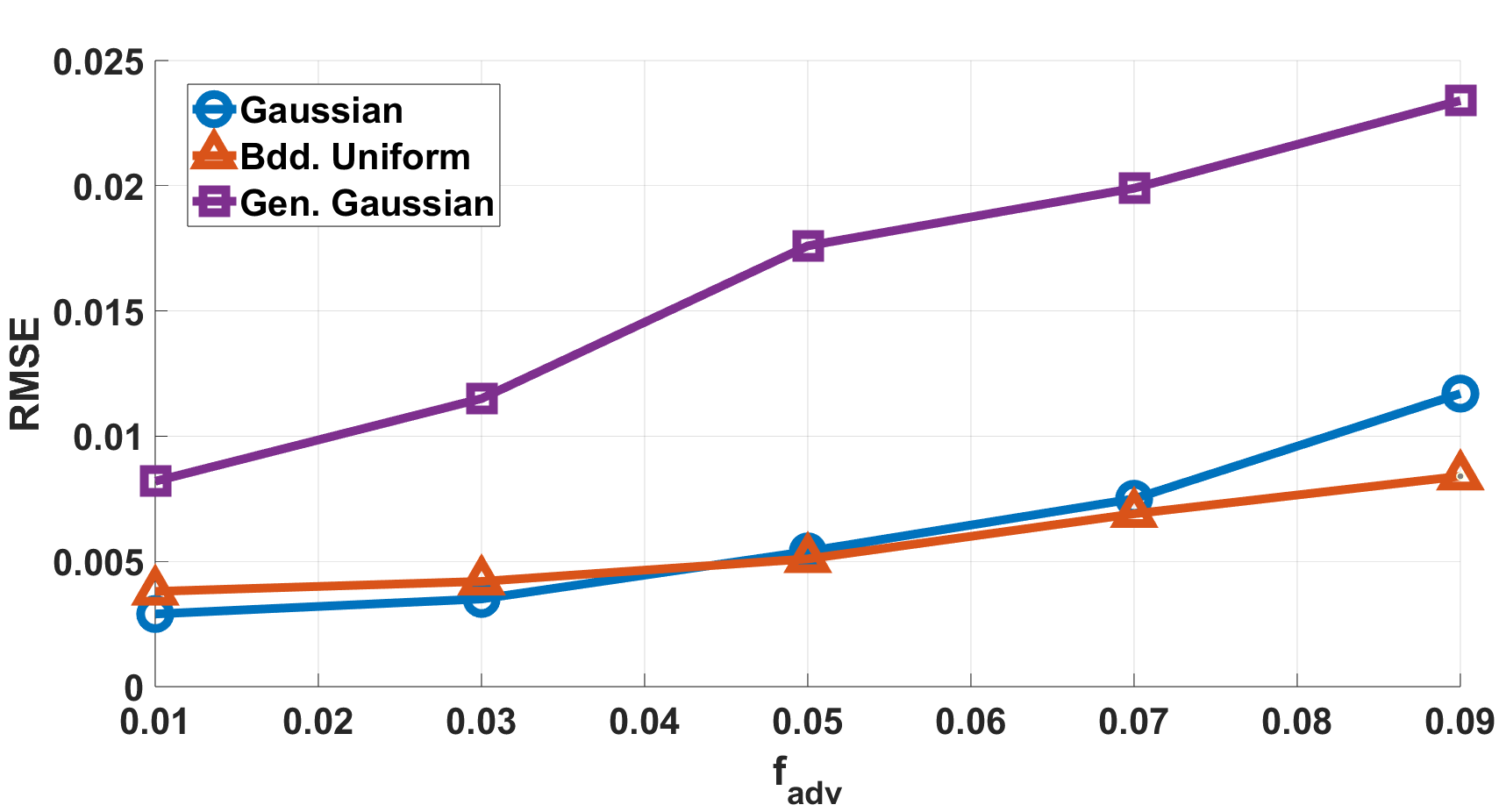}
    \includegraphics[scale=0.19]{rmse_noise_n.png}\\
    \includegraphics[scale=0.19]{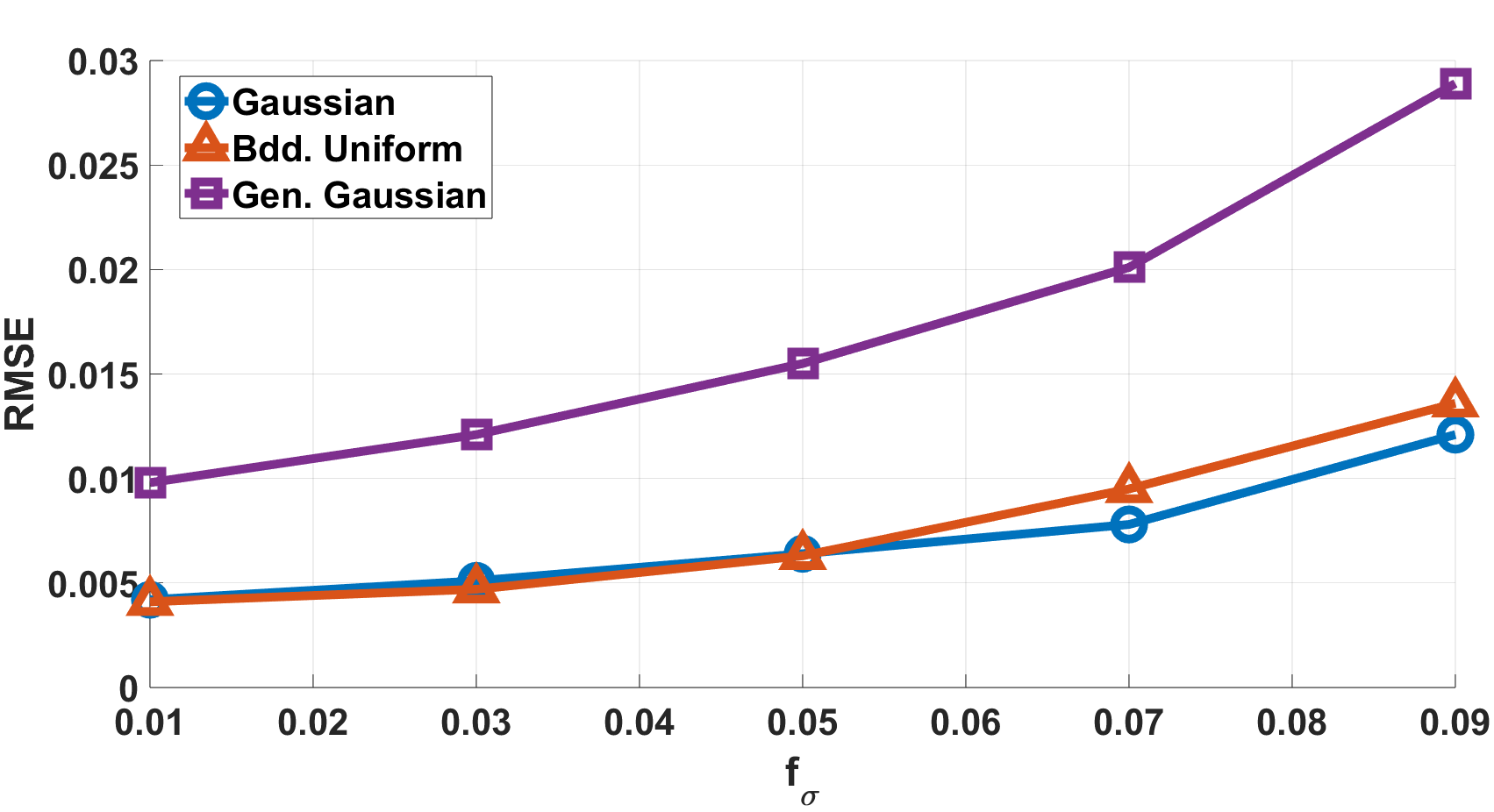}
    \includegraphics[scale=0.19]{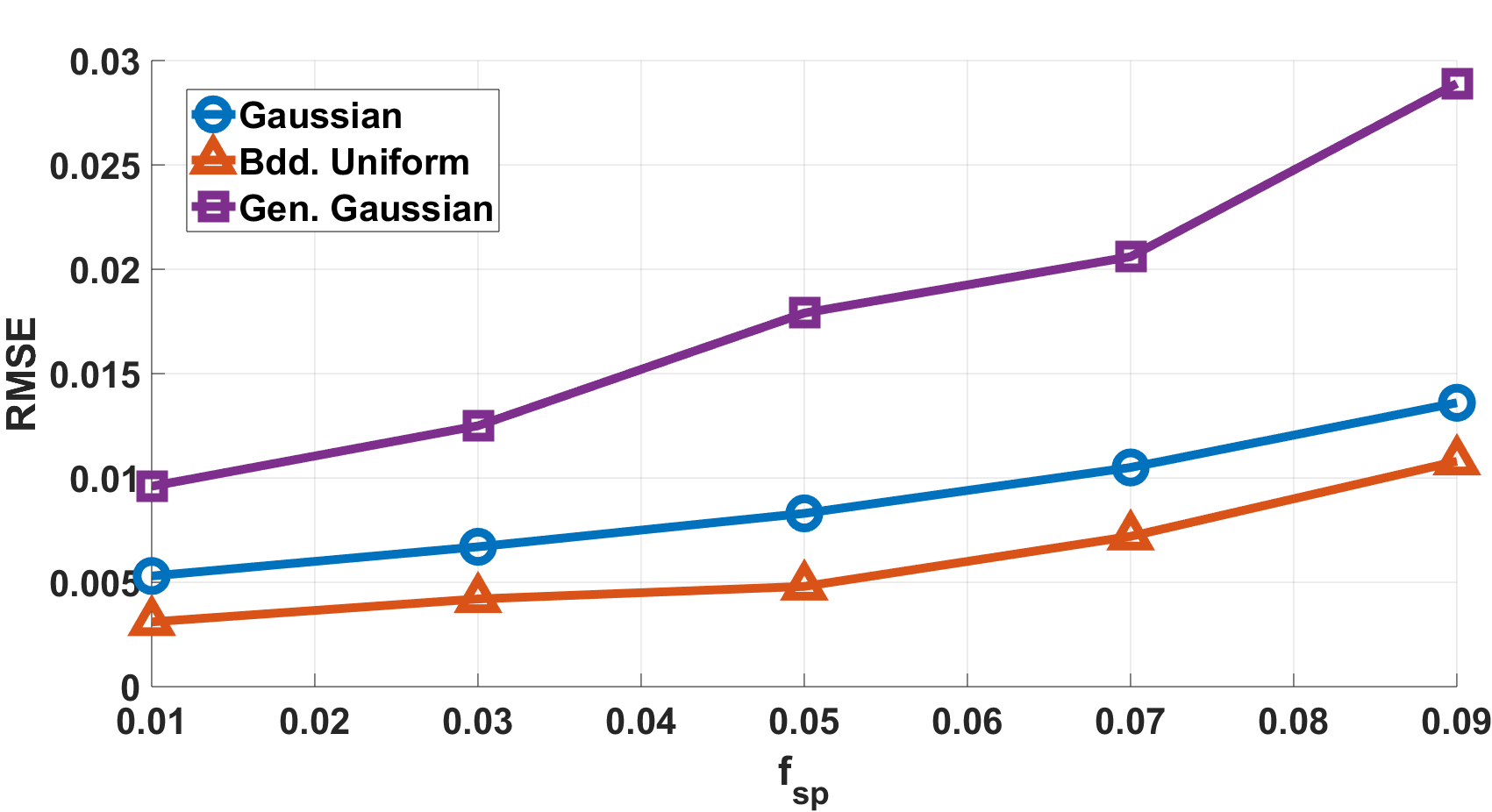}
    \caption{Average RRMSE comparison (over 25 independent noise runs keeping $\boldsymbol{\beta^*}$, $\boldsymbol{A}$ and $\boldsymbol{\delta^*}$ fixed) using  \textsc{ODrlt} for additive noise models $N(0,\sigma^2)$, Empricial \textsc{ODrlt} for Bounded uniform $[-\sqrt{3}\sigma,\sqrt{3}\sigma]$ and Generalised Gaussian with shape $3/2$ w.r.t. variation in the following parameters keeping others fixed: bit-flip proportions $f_{adv}$ as in setup (\textsf{EA}) (topleft), measurements $n$ (top right) as in setup (\textsf{EB}), noise level $f_{\sigma}$ as in setup (\textsf{EC}) (bottom left) and sparsity $f_{sp}$ as in setup (\textsf{ED}) (bottom right). The fixed parameters are dimension of $p = 500, f_{\sigma}=0.05,f_{adv}=0.01,f_{sp}=0.01,n=400$. }
    \label{fig:noise_RMSE}    
\end{figure*}

\bibliographystyle{plain}
\bibliography{arxiv_v2}
\end{document}